\documentclass[natbib]{pretty-preprint}
\usepackage[section]{placeins} 
\usepackage{booktabs}
\usepackage{multirow}
\usepackage{graphicx}
\usepackage{booktabs}
\usepackage{multirow}
\usepackage{graphicx}
\usepackage{amssymb}
\usepackage{wrapfig}

\usepackage{booktabs}
\usepackage{multirow}
\usepackage{graphicx}
\usepackage[table]{xcolor}
\usepackage{longtable}
\usepackage{pifont}
\usepackage{fontawesome5}
\usepackage{tikz}
\usepackage{xurl} 
\usetikzlibrary{arrows.meta,positioning,fit,backgrounds}

\title{GameAnything: From Any Inspirations to Playable Game Worlds}
\runningtitle{GameAnything}

\author[1,*]{Wenyi Wu}
\author[1,*]{Siqi Liu}
\author[1,*]{Yuru Feng}
\author[1]{Zixuan Song}
\author[1]{Aayush Salvi}
\author[1]{Sibo Zhu}
\author[2,\textdagger]{Kun Zhou}
\author[1,2]{Biwei Huang}
\affiliation[1]{University of California, San Diego}
\affiliation[2]{Aether AI}
\contribution[*]{Equal contribution}
\contribution[\textdagger]{Corresponding author}

\date{November 21, 2026}
\correspondence{\email{franciskunzhou@gmail.com}}
\metadata[Project]{\href{https://huggingface.co/spaces/GameAnything/GameAnything}{huggingface.co/spaces/GameAnything/GameAnything}}
\metadata[Code]{\href{https://github.com/AetherLabsAI/GameAnything}{github.com/AetherLabsAI/GameAnything}}
\metadata[Benchmark]{\href{https://github.com/AetherLabsAI/GameAnything-Bench}{github.com/AetherLabsAI/GameAnything-Bench}}
\metadata[Early Access]{\href{https://huggingface.co/spaces/GameAnything/GameAnything-Trial}{huggingface.co/spaces/GameAnything/GameAnything-Trial}}
\metadata[Contribute]{\href{https://huggingface.co/spaces/GameAnything/gameanything-submit}{huggingface.co/spaces/GameAnything/gameanything-submit}}

\abstract{%
AI-driven game development holds the potential to transform diverse human inspirations into interactive experiences. However, existing systems and benchmarks primarily focus on explicit development specifications, with limited support for adapting multimodal references and evaluating actual gameplay experiences. We introduce \textbf{GameAnything}, an open-source agentic platform that transforms multimodal inspirations into playable extensions of existing games through iterative development. GameAnything integrates coding agents with an independent playtesting critic, persistent development state, and reusable experience memory to support continuous refinement. We further present \textbf{GameAnything-Bench}, a benchmark spanning diverse reference modalities, game creation objectives, and gaming environments, with exploratory gameplay testing to assess both functional correctness and gameplay quality. Evaluations of nine coding agent configurations reveal that current agents can adapt reference content but still struggle with playable interactions and precise modifications. Moreover, multimodal references improve content fidelity, while iterative refinement yields substantial but varying improvements across agents. These findings highlight both the potential and remaining challenges of inspiration-driven game development.
}


\begin{document}
\maketitle

\section{Introduction}
\label{sec:introduction}

\begin{figure}[p]
    \centering
    \includegraphics[width=\textwidth]{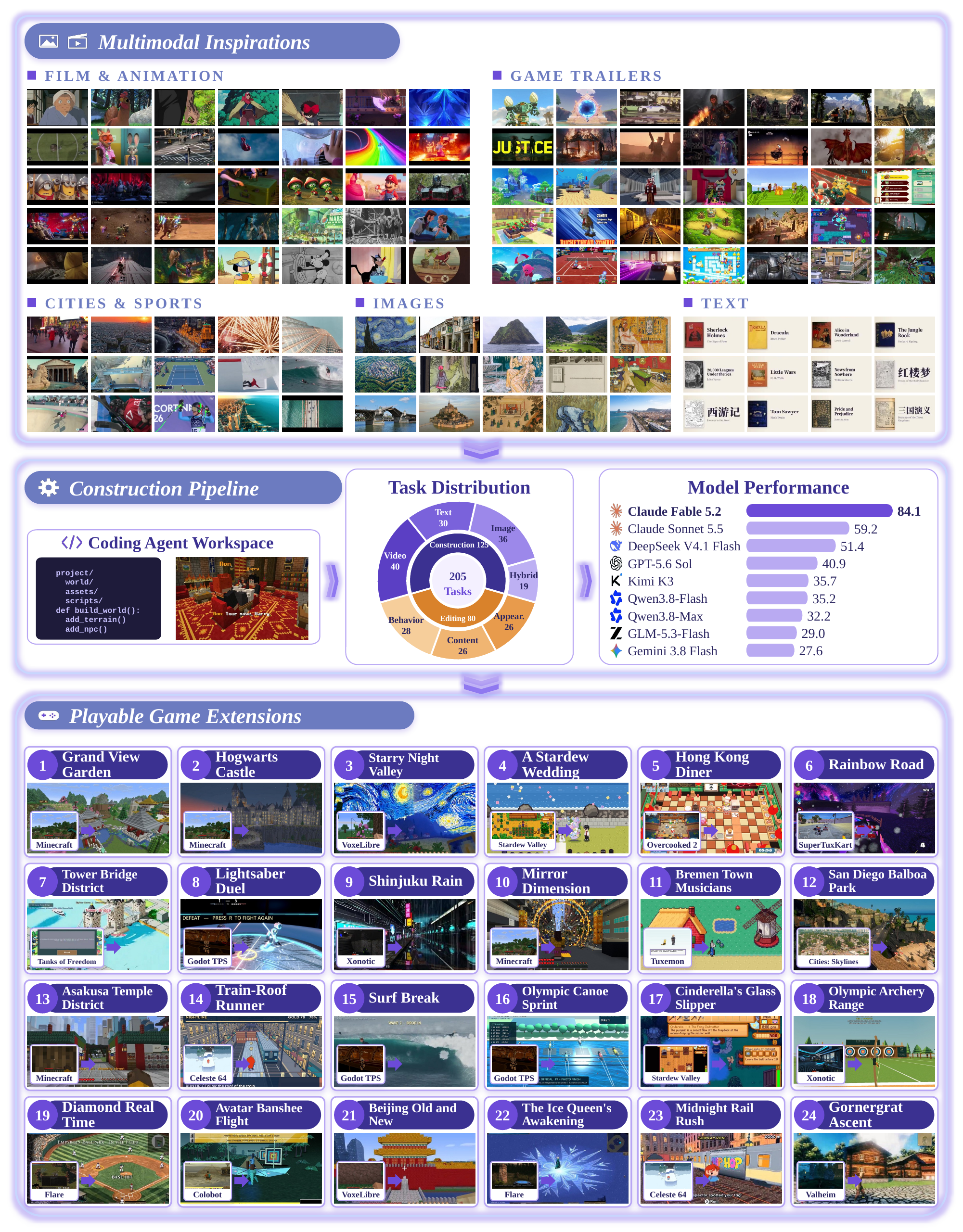}
\caption{\textbf{GameAnything: From Any Inspiration to Playable Game Worlds.} GameAnything transforms diverse inspirations, including films, game trailers, real-world footage, images, and literature, into playable game extensions. Middle: illustrates the agentic construction pipeline alongside benchmark statistics and model performance. Bottom: showcases representative game extensions across diverse gaming environments.}

    \label{fig:gameanything_teaser}
\end{figure}

AI-driven game generation holds the potential to make game development more accessible and efficient, enabling a broader range of creators to transform their ideas into interactive experiences. Recent advances in large language models have further enabled coding agents to generate playable games and modify existing game projects from natural language instructions~\cite{voyager}~\cite{mariogpt}. However, game creation often begins with inspiration from novels, photographs, films, gameplay videos, or personal experiences rather than explicit specifications. To this end, agents must identify transferable mechanisms, visual characteristics, entities, and relationships from reference content, and integrate them into coherent game worlds, gameplay mechanics, and narratives while preserving the functionality of the original game. For example, adapting a tennis match into an existing game requires translating its rules and interactions into playable mechanics. Achieving this requires both accurate understanding of diverse reference materials and reliable game development capabilities.


Existing game development systems and evaluation benchmarks primarily focus on game generation from textual specifications or design documents~\cite{luo2026gamecraftbench,dong2026cegamebench,chen2026swegame}, feature implementation and repair in existing projects~\cite{chi2026gamedevbench,chen2026swegame}, and game modding from natural language requests~\cite{ku2026igmbench}. However, these systems and benchmarks have not systematically addressed how diverse external references can be understood and adapted into playable game content, including their underlying mechanisms, visual characteristics, and relationships. Moreover, existing evaluation benchmarks largely rely on scripted action replay and runtime probes to verify predefined behaviors, providing limited insight into actual gameplay experiences from a player's perspective. More comprehensive game development systems and evaluation benchmarks are therefore needed to support and assess inspiration-driven content adaptation and its coherent integration into playable games, better reflecting real-world game development needs and player experiences.

To address these limitations, we introduce \textbf{GameAnything}, an open-source agentic platform that transforms diverse human inspirations into playable extensions of existing games through autonomous and iterative development. GameAnything establishes a continuous development loop in which coding agents interpret reference content and implement game modifications, while an independent critic autonomously playtests the resulting games to identify unmet requirements, functional failures, and regressions. These findings are incorporated into a persistent development state to guide subsequent revisions, with sparse human feedback steering high-level creative decisions. Beyond individual tasks, GameAnything consolidates verified development experience into reusable memory, allowing subsequent development cycles to benefit from previous experience. Inspired by the experience-guided improvement mechanism of RSIAgent~\citep{zhu2026rsiagent}, this design supports recursive self-improvement across game versions and tasks. With modular interfaces for different coding agents and game environments, GameAnything further serves as a general evaluation framework for systematically assessing inspiration-driven game development, development reliability, and gameplay quality.

We further introduce \textbf{GameAnything-Bench}, a benchmark for systematically evaluating how coding agents understand multimodal references and transform their content into playable extensions of existing games. The benchmark organizes tasks into three categories based on their primary creation objectives: \textit{Scene \& World}, \textit{Mechanics \& Behavior}, and \textit{Narrative \& Event}, covering diverse references including text, images, videos, and their combinations. To support reproducible evaluation across diverse gaming environments, GameAnything-Bench includes a core track built on open-source games and a separate commercial game track for broader assessment. For evaluation, the benchmark combines runtime probes, scripted action replay, and exploratory gameplay testing to assess both functional correctness and gameplay quality. Unlike evaluations limited to predefined test trajectories, this approach enables the discovery of unexpected failures and gameplay inconsistencies, providing a more comprehensive assessment of game development capabilities from a player's perspective.


We evaluate nine coding agent configurations on GameAnything-Bench and uncover three key findings. First, current agents can realize reference-grounded content and generally preserve existing games, but struggle to implement fully playable interactions and precise edits. 
Second, multimodal references substantially improve content and visual fidelity, while agent capabilities vary sharply. Claude Fable 5.2~\cite{anthropic2026fable52} leads Construction by 24.9 points, whereas DeepSeek V4.1 Flash~\cite{deepseek2026v41flash} offers the strongest open-weight performance at substantially lower cost. Most agents perform best on image references, but Kimi K3~\cite{kimiteam2026kimik3} and Qwen3.8-Max~\cite{alibaba2026qwen38max} exhibit relative strengths on hybrid and video inputs, respectively. 
Third, iterative refinement reveals distinct improvement trajectories across agents. Qwen3.8-Flash~\cite{alibaba2026qwen38flash} gains only 0.8 Construction points after one round but 14.8 after three, while refined DeepSeek V4.1 Flash nearly matches native Claude Sonnet 5.5~\cite{anthropic2026sonnet55} at comparable development cost.
{\let\FloatBarrier\relax \section{Related Work}}
\label{sec:related-work}

\definecolor{checkgreen}{RGB}{25,145,55}
\definecolor{crossred}{RGB}{210,40,40}

\providecommand{\tabyes}{%
  \textcolor{checkgreen}{\ding{51}}}
\providecommand{\tabno}{%
  \textcolor{crossred}{\ding{55}}}

\providecommand{\txticon}{\faIcon{file-alt}}
\providecommand{\imgicon}{\faIcon{image}}
\providecommand{\vidicon}{\faIcon{video}}

\newcommand{\modbox}[1]{%
  \begingroup
  \setlength{\fboxsep}{2pt}%
  \colorbox{black!6}{\strut\,#1\,}%
  \endgroup
}

\begin{table*}[t]
\centering
\caption{\textbf{Comparison with existing game development benchmarks.}}
\label{tab:benchmark_comparison}
\begingroup
\small
\setlength{\tabcolsep}{2.5pt}
\renewcommand{\arraystretch}{1.08}
\begin{tabular*}{\textwidth}{@{\extracolsep{\fill}}l c cc l cccc@{}}
\toprule
\multirow{2}{*}{\textbf{Benchmark}}
& \textbf{Inputs}
& \multicolumn{3}{c}{\textbf{Development}}
& \multicolumn{4}{c}{\textbf{Evaluation}} \\
\cmidrule(lr){2-2}\cmidrule(lr){3-5}\cmidrule(l){6-9}
& \textbf{Modality}
& \textbf{Constr.}
& \textbf{Editing}
& \shortstack{\textbf{Target}\\\textbf{Environments}}
& \shortstack{\textbf{Pres.}\\\textbf{Checks}}
& \shortstack{\textbf{Runtime}\\\textbf{Checks}}
& \shortstack{\textbf{Action}\\\textbf{Replay}}
& \shortstack{\textbf{Agent}\\\textbf{Playtest}} \\
\midrule
V-GameGym & T & \tabyes & \tabno & Pygame & \tabno & \tabno & \tabno & \tabno \\
OpenGame-Bench & T & \tabyes & \tabno & Browser & \tabno & \tabno & \tabno & \tabno \\
GameDevBench & T & \tabno & \tabyes & Godot projects & \tabno & \tabyes & \tabno & \tabno \\
PlaytestArena & T & \tabyes & \tabno & Browser & \tabno & \tabno & \tabno & \tabyes \\
GameCraft-Bench & T & \tabyes & \tabno & Godot & \tabno & \tabno & \tabyes & \tabno \\
SWE-Game & T, V & \tabyes & \tabyes & 41 reference games & \tabyes & \tabyes & \tabyes & \tabno \\
IGMBench & T & \tabno & \tabyes & 2 existing games & \tabyes & \tabyes & \tabno & \tabno \\
CEGame-Bench & T & \tabyes & \tabno & 5 engines & \tabno & \tabyes & \tabyes & \tabno \\
Movie2Game-Bench & V & \tabyes & \tabno & Browser & \tabno & \tabno & \tabno & \tabyes \\
\midrule
\rowcolor{black!5}
\textbf{GameAnything} & \textbf{T, I, V, H} & \tabyes & \tabyes & \textbf{14 existing games} & \tabyes & \tabyes & \tabyes & \tabyes \\
\bottomrule
\end{tabular*}
\par\vspace{4pt}
\begin{minipage}{\textwidth}\footnotesize
\textit{Notes.} T, I, V, H denote text, image, video, and hybrid inputs. Checkmarks indicate explicit support, and Pres. denotes Preservation Checks. GameAnything covers nine open-source and five commercial games. Other entries identify engines or reference projects where applicable.
\end{minipage}
\endgroup
\end{table*}
\label{sec:related}

\paragraph{Benchmarking game generation.}
Recent benchmarks increasingly focus on end-to-end game generation, which requires coding agents to build complete playable games from scratch. V-GameGym~\cite{zhang2026vgamegym} curates Pygame tasks from real repositories and scores generated code, screenshots, and gameplay videos with LLM/MLLM judges. OpenGame-Bench~\cite{jiang2026opengame} evaluates browser game generation by grading rendered visuals with a VLM. PlaytestArena~\cite{huang2026playtestarena} instead pairs browser-game prompts with behavioral rubrics verified by an autonomous GUI playtesting agent. GameCraft-Bench~\cite{luo2026gamecraftbench} validates complete Godot projects by replaying agent-submitted interaction traces. CEGame-Bench~\cite{dong2026cegamebench} likewise builds new games and verifies them with runtime probes and action replay. Movie2Game-Bench~\cite{movie2gamebench} replaces the text description with a full film and, alongside general game quality and agent playtesting, scores how much film-specific content survives into player interaction, leaving the game genre and design to the agent. In contrast, GameAnything-Bench moves beyond blank-canvas generation to evaluate cross-world transfer and preservation-aware editing, challenging agents to adapt multimodal references into existing games' native mechanics and iteratively modify them without breaking intact content.

\paragraph{Game editing.}
More closely related are benchmarks grounded in existing game repositories. GameDevBench~\cite{chi2026gamedevbench} turns Godot tutorials into feature-implementation tasks and verifies them with engine-side tests. SWE-Game~\cite{chen2026swegame} combines construction tasks, which build a game from a brief or a design document, with skeleton-completion, repair, and porting tasks on executable reference games, and scores both fault restoration and preservation of unaffected behavior. IGMBench~\cite{ku2026igmbench} frames Minecraft and Terraria modding as world editing at increasing intervention depth, graded by executable state, behavioral, and regression checks. In these benchmarks, the target change is specified by a text request or by a reference version of the same game. GameAnything-Bench instead requires agents to transfer content from external multimodal references into a different existing game as a new mod. It covers both construction and editing, and evaluates them by combining preservation checks, runtime probes, action replay, and agent playtesting (Table~\ref{tab:benchmark_comparison}).

\paragraph{Simulated environment generation.}
RobotAgentBench~\citep{mao2026robotagentbench} uses 50 simulated household layouts to evaluate task discovery, manipulation, and recovery from failures. In game development, most benchmarks target a single engine or modding framework. V-GameGym~\cite{zhang2026vgamegym} targets Pygame, while OpenGame-Bench~\cite{jiang2026opengame} and PlaytestArena~\cite{huang2026playtestarena} target browser games. GameDevBench~\cite{chi2026gamedevbench}, GameCraft-Bench~\cite{luo2026gamecraftbench}, and SWE-Game~\cite{chen2026swegame} are built on Godot projects, with Unity appearing only as a porting target. IGMBench~\cite{ku2026igmbench} reaches commercial titles but covers only Minecraft and Terraria through their modding frameworks. GameAnything-Bench instead decouples what to build from where to build it, treating the reference and the host game as independent variables. It thereby tests whether agents can generalize across heterogeneous games and modding conventions.




\section{Preliminaries}
\label{sec:preliminaries}

\paragraph{Task Formulation.}
We study how coding agents transfer ideas and experiences from multimodal references into existing games, and further develop them into new playable worlds, mechanics, and narratives.
Given multimodal references $X$, a user instruction $u$, and a target game $G$, a coding agent $\mathcal{A}$ develops a game modification $M$.
The modification includes code, assets, and configurations needed to extend the target game.
We formulate this process as
\begin{equation}
M \sim \mathcal{A}(\cdot \mid X, u, G),
\qquad
G' = \operatorname{Apply}(G, M),
\label{eq:task}
\end{equation}
where $G'$ denotes the game with the new content installed.
Unlike game generation from scratch, the agent must understand and work with the existing game's completed and comples projects, and conduct further development in it without destroying original functions.

\paragraph{Playable Game Extension.}
Rather than simply reproducing the references, agents must identify and transfer their underlying mechanisms, visual characteristics, and relationships, then build upon them to create new playable content in the target game.
Executing the modified game $G'$ produces a gameplay trace
\begin{equation}
\tau = (o_0, a_0, o_1, \ldots, a_{T-1}, o_T),
\label{eq:trace}
\end{equation}
where $o_t$ denotes a game observation and $a_t$ denotes a player action.
A successful extension should preserve the intended characteristics of the references, realize them through functional gameplay, and preserve the core functionality of the original game.
Since the same creative intent may admit different valid implementations, success is determined by observable gameplay experience rather than a single hard-coded implementation.

\section{GameAnything}
\label{sec:method}

\begin{figure}[t]
    \centering
    \includegraphics[width=\textwidth]{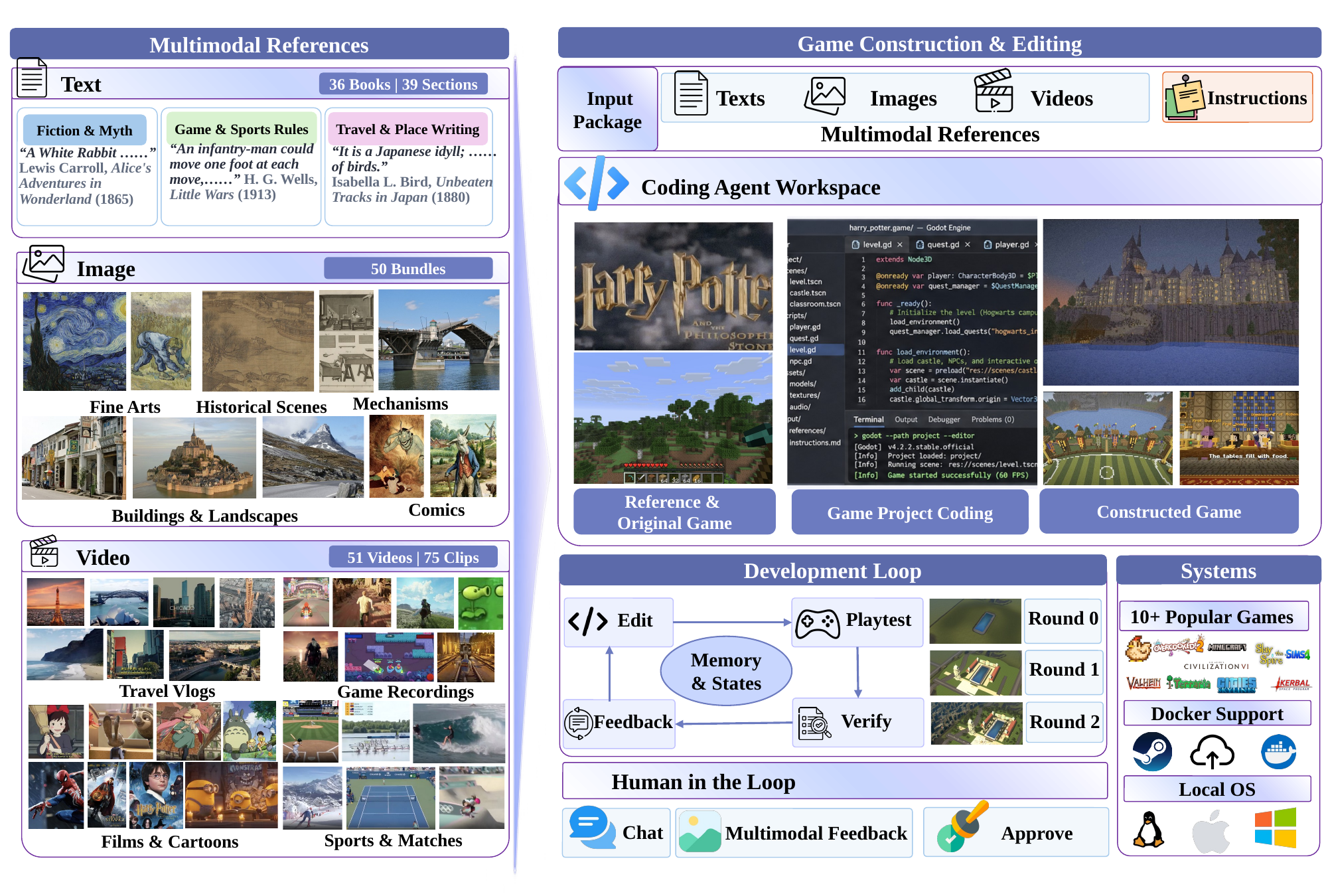}
    \caption{\textbf{Overview of GameAnything.}
    \textit{Left:} the multimodal reference pool.
    \textit{Top right:} each task gives a coding agent an input package of multimodal references and instructions; the agent works in the original game's project and modding interfaces to build or edit playable content.
    \textit{Bottom right:} the agent development loop with iterative refinement, and supported games and systems.}
    \label{fig:bench_overview}
\end{figure}

GameAnything extends coding agents from one-time generation to iterative game development grounded in gameplay evidence.
Given multimodal references $R$, a user instruction $I$, and a game project $P_t$, the framework iteratively updates the project:
\begin{equation}
\Delta_t \sim \pi_{\mathrm{builder}}
(\cdot \mid R, I, P_t, S_t, \mathcal{M}),
\qquad
P_{t+1} = \operatorname{Apply}(P_t, \Delta_t).
\label{eq:ga_update}
\end{equation}
Here, $S_t$ maintains verified development progress, while $\mathcal{M}$ stores reusable experience from previous tasks.
The framework combines two complementary processes: autonomous playtesting and verification within each task, and experience accumulation across tasks.

\subsection{Evidence-Grounded Iterative Development}
\label{sec:iterative_development}
\paragraph{Persistent Development State.}
GameAnything maintains a structured development state $S_t$ that tracks task requirements, completion status, supporting evidence, and unresolved issues:
\begin{equation}
S_t = (\mathcal{Q}, \boldsymbol{\sigma}_t,
       \mathcal{E}_t, \mathcal{K}_t),
\label{eq:dev_state}
\end{equation}
where $\mathcal{Q}$ denotes task requirements, $\boldsymbol{\sigma}_t$ records their status, $\mathcal{E}_t$ stores supporting evidence, and $\mathcal{K}_t$ maintains a recheck list.
Each requirement is tracked as \textit{todo}, \textit{partial}, \textit{done}, \textit{broken}, or \textit{regressed}.
The builder records development progress, while an independent playtester verifies completion through gameplay evidence.
Previously satisfied requirements that fail in subsequent rounds are marked as regressions and added to the recheck list.
The state carries verified progress and unresolved issues between rounds, much like the verifier-backed checkpoints in StructAgent~\citep{wu2026structagent}.

\paragraph{Autonomous Playtesting and Verification.}
At each round, the builder implements or revises the game according to the current development state.
It then launches the updated project $P_{t+1}$ and performs gameplay interactions to collect an observation trace $\tau_t$.
An independent playtester examines the resulting evidence against the user instruction and reference materials:
\begin{equation}
\tau_t \sim \pi_{\mathrm{play}}(
\cdot \mid P_{t+1}, I, S_t),
\qquad
v_t \sim \pi_{\mathrm{verify}}(
\cdot \mid R, I, P_{t+1}, S_t, \tau_t).
\label{eq:playtest}
\end{equation}
Here, $v_t$ identifies verified requirements, remaining issues, and regressions.
The verification results update the development state:
\begin{equation}
S_{t+1} = \operatorname{Update}(S_t, \tau_t, v_t).
\label{eq:state_update}
\end{equation}
The updated state guides the next revision, and the process continues until the requirements are verified or the development budget is exhausted.
This grounds iterative improvement in actual gameplay rather than code inspection alone.
Appendix~\ref{app:iter_case_studies} shows this process on three projects.

\subsection{Cross-Task Experience Memory}
\label{sec:experience_memory}

\paragraph{Experience Extraction and Consolidation.}
HyMEM~\citep{zhu2026hymem} organizes computer-use trajectories in a graph for retrieval across tasks. In GameAnything, each reusable lesson comes from a completed development cycle.
Let $\mathcal{H}_j$ denote the development history of task $j$, including encountered problems, corrective actions, and verification evidence.
The framework extracts candidate lessons and consolidates them into a persistent memory:
\begin{equation}
\mathcal{L}_j = \operatorname{Extract}(\mathcal{H}_j),
\qquad
\mathcal{M}_{j+1}
= \operatorname{Consolidate}(\mathcal{M}_j,\mathcal{L}_j).
\label{eq:memory}
\end{equation}
Each lesson records the observed issue, its cause, the corrective action, and supporting evidence.
We organize these lessons into game-specific knowledge and general development practices.
A new lesson enters the memory as a candidate; similar lessons are merged rather than duplicated, and a candidate that a later task reproduces is promoted to a confirmed lesson.

\paragraph{Budgeted Experience Retrieval.}
Before each refinement round, GameAnything retrieves relevant experiences based on the target game, the agent role, and the current development state (the user request and the latest critic review).
Rather than including the entire memory history, it selects applicable lessons under a fixed context budget:
\begin{equation}
\mathcal{M}_j^{\mathrm{ret}}
=
\operatorname{Retrieve}(
\mathcal{M}_j \mid G, \rho, S_t),
\qquad
\operatorname{Length}(\mathcal{M}_j^{\mathrm{ret}})
\leq B,
\label{eq:retrieve}
\end{equation}
where $G$ is the target game, $\rho$ denotes the agent role, $S_t$ is the current development state, and $B$ is the retrieval budget.
Confirmed lessons are selected first, then candidates ranked by relevance to $S_t$.
The retrieved experiences guide the next revision; for every round we log which lessons were retrieved and, after the round, which of them the agent applied.
As more tasks are completed, newly validated lessons become available for future development, helping agents avoid repeated failures.

\subsection{Interactive Platform and Extensibility}
\label{sec:platform}

\paragraph{User-Guided Game Creation.}
GameAnything provides a local web interface that allows users to transform personal memories, stories, images, and videos into playable content without writing code.
Users provide multimodal references and creative goals, while the platform presents interactive cards to clarify important design choices, such as characters, locations, visual styles, and gameplay mechanics.
After each development cycle, users can review recorded gameplay and provide feedback on specific video frames or image regions.
The platform converts this feedback into actionable development requirements and verifies subsequent modifications through corresponding gameplay views.
This enables users to guide creative decisions and progressively refine their games through natural interaction; Appendix~\ref{app:ui} describes the interface and workflow in detail.

\paragraph{Extensible Game and Agent Integration.}
GameAnything supports heterogeneous game engines and modding ecosystems through modular game adapters that provide interfaces for building, installing, launching, and interacting with games.
New games can be integrated through standardized configurations, adapters, and validation tests.
The platform also supports multiple coding agents, including Claude Code~\citep{anthropic2026claudecode}, Codex~\citep{openai2026codexcli}, Gemini CLI~\citep{google2026geminicli}, and OpenCode~\citep{anomaly2026opencode}, allowing users to work with their preferred agent configurations.
Development can run locally or in isolated server environments.
User materials are managed locally, external model access requires authorization, and game saves are backed up before modifications are installed.
Together, these capabilities provide an accessible and extensible platform for creating and continuously refining content in existing games.

\section{GameAnything-Bench}
\label{sec:bench}

\begin{figure}[t]
    \centering
    \includegraphics[width=\textwidth]{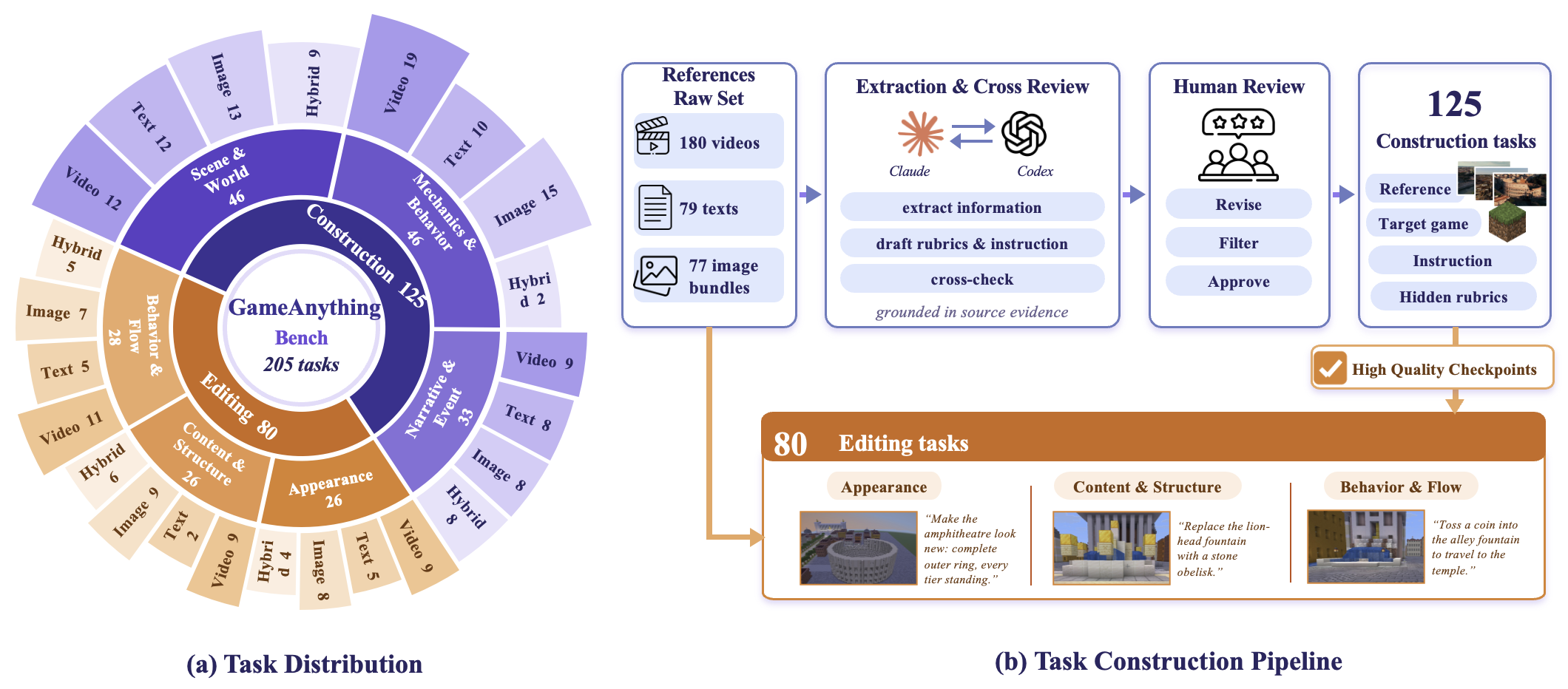}
\caption{\textbf{Overview of GameAnything-Bench.} The left panel shows the distribution of 205 tasks across Construction and Editing, task categories, and input modalities. The right panel illustrates the benchmark construction pipeline, where multimodal references undergo information extraction, model cross-review, and human validation to produce 125 Construction tasks. Validated Construction checkpoints further support 80 Editing tasks covering appearance, content and structure, and behavior and flow.}
    \label{fig:bench_pipeline}
\end{figure}

\subsection{Task Design}
\label{sec:taskdesign}

\paragraph{Task Taxonomy}

We organize GameAnything-Bench around three fundamental dimensions of reference-to-game adaptation objective: \textit{Scene \& World Adaptation} transfers the visual and spatial structure of places, including objects, landmarks, layouts, and environmental styles.
\textit{Mechanics \& Behavior Adaptation} distills underlying rules, motion dynamics, and interaction patterns into executable gameplay.
\textit{Narrative \& Event Adaptation} grounds character relationships and causal event sequences into interactive experiences.
Crucially, tasks require visual and interactive grounding rather than surface-level visual replication: agents must abstract transferable principles and re-express them through the target game's native affordances during construction, while executing targeted modifications without disrupting unspecified game systems during editing. 
\paragraph{Input Modalities and Task Settings.}
In real-world game development, creative ideas often draw on multimodal references, while user requests involve both creating and modifying game content.
To reflect this diversity, we consider four input modalities and two task settings.
For input modalities, we include text, image, video, and hybrid references.
Different modalities provide complementary information about visual appearance, mechanisms, and narratives, while hybrid tasks require agents to integrate information across modalities into coherent playable content.
For task settings, \textit{Construction} requires agents to create new playable content in an existing game based on the provided references.
\textit{Editing} requires agents to modify existing playable content according to specific user requests while preserving unrelated content and functionality.
Both settings evaluate the ability to translate multimodal inspirations into playable game content.

\paragraph{Target Game Selection.}
We select target games based on diversity, adaptation feasibility, and reproducibility.
The selected games span different genres and interaction mechanisms, allowing us to evaluate adaptation across heterogeneous environments.
When pairing references with target games, we ensure that each game can meaningfully realize the reference content through its mechanics and modding interfaces.
We also avoid source--target pairs that allow direct replication without substantive adaptation.
Our core benchmark uses open-source games to enable reproducible execution and evaluation, while a separate commercial-game track assesses generalization to commercial games. Further details on game selection and reference-to-game matching criteria are provided in Appendix~\ref{app:game_selection}.

\subsection{Benchmark Construction}
\label{sec:benchmark-construction}

\paragraph{Reference Collection and Annotation.}
We curate diverse multimodal references, including fictions, photographs, manga, film clips, sports footage, gameplay recordings, and travel videos, etc.
For video references, we first remove duplicates and extract timestamped clips, keyframes, and transcripts. Then we perform two independent annotation passes to identify transferable content, including mechanisms, entities and their properties, and spatial, temporal, and causal relations. Each annotation is grounded in evidence from the original reference.
Based on these annotations, we manually select representative references, assign task categories, and match them to target games, considering adaptation potential, diversity, and evaluation feasibility.
We extend the same annotation scheme to image and text references, grounding annotations in specific image regions or text passages. More details about reference sources and annotation process can be found in Appendix~\ref{app:reference_collection}.

\paragraph{Task Construction.}
\label{sec:task_construction}
Each task consists of reference materials, a target game, an agent-visible instruction, and hidden evaluation rubrics.
Editing tasks additionally provide a validated game checkpoint containing existing playable content.
To align task instructions with evaluation criteria, we adopt a rubric-first construction process.
We define rubrics for content, functionality, visual presentation, and preservation of the host game's character, with criteria tailored to Construction and Editing.
Each rubric item includes supporting evidence and scoring anchors.
We first calibrate the authoring guidelines through multiple rounds of model generation and human feedback on pilot tasks.
For large-scale construction, all rubrics undergo independent model cross-review and multiple rounds of human revision before finalization.
We then derive instructions by grouping rubrics into broader requirements, and additionally include explicit user requests to \textit{preserve, specify, and modify} content.
Further details are provided in Appendix~\ref{app:task_construction}.



\paragraph{Quality Control and Benchmark Statistics.}
We apply several quality checks to ensure task validity and evaluation reliability.
We verify that rubric items are grounded in reference evidence or explicit requirements, exclude incidental details that do not define the intended experience, and review every task before finalization.
We also validate game capability probes on unmodified and deliberately corrupted game instances to check executability and host functionality.
Further details are provided in Appendix~\ref{app:task_construction} and Appendix~\ref{app:game_capability}.

GameAnything-Bench contains 205 tasks across 14 target games, comprising 125 Construction and 80 Editing tasks.
The open-source core includes 77 Construction tasks across nine games and 64 Editing tasks on six of them, while the commercial track includes 48 Construction tasks across five games and 16 Editing tasks across two of them.
Construction covers text, image, video, and hybrid references, while Editing spans appearance, content and structure, and behavior and flow modifications.
Full distributions are provided in Tables~\ref{tab:app_task_distribution} and~\ref{tab:app_edit_distribution}.

\subsection{Evaluation Protocol}
\label{sec:evaluation-protocol}

\paragraph{Evaluation Dimensions.}
Successful game development requires both creating new playable content from multimodal references and precisely modifying existing content.
We therefore evaluate two complementary capabilities: \textit{Construction} and \textit{Editing}.
For Construction, \textit{Content} assesses whether the required entities, properties, structures, and relations are correctly realized,
\textit{Functions \& Playability} evaluates whether the created content is functional, interactive, and playable,
and \textit{Visual \& Presentation} measures visual fidelity, spatial appearance, and overall presentation quality.
For Editing, \textit{Edit Success} assesses whether the requested change is correctly realized in the game, and \textit{Preservation} whether content and functions that the edit should not touch remain as they were in the initial checkpoint.
Both settings share \textit{Host Fidelity}, which assesses whether the result is built on and played through the target game, using its world, avatar and controls, systems, and look, rather than a replacement that merely runs inside it.


\paragraph{Evidence Grounded Agentic Evaluation.}
Creative game development allows multiple valid implementations, making fixed tests and static visual evaluation insufficient to fully assess interactive correctness.
We therefore combine rubric based evaluation with gameplay replay and independent agentic playtesting.
Automated probes act as a gate that checks whether the original game still runs and remains present after modification.
Gameplay replay executes action traces submitted by the development agent to verify the implemented features.
Independent agentic playtesting further explores the game beyond these traces, assessing whether the content can be discovered and experienced through normal gameplay.
Visual evaluation compares captured screenshots against reference grounded rubrics.
For Editing, we additionally compare the modified game with its initial checkpoint to verify targeted changes and preservation of unrelated content.
Further details are provided in Appendix~\ref{app:agentic_eval}; agreement with human judgments and the contribution of each evidence source are analyzed in Appendices~\ref{app:human_agreement} and~\ref{app:evidence_ablation}.

\paragraph{Scoring Rubrics.}
Each rubric item is scored as 0, 0.5, or 1, and scores are aggregated within each evaluation dimension. Construction Overall combines content, functionality, visual quality, and host fidelity, while Editing Overall prioritizes edit success with penalties for unintended changes and loss of host fidelity. Submissions that fail the executability gate receive zero. Detailed scoring rules are provided in Appendix~\ref{app:eval_details}, and the stability of the scores across repeated evaluations in Appendix~\ref{app:eval_stability}.

\section{Experiments}
\label{sec:experiments}

\setcounter{topnumber}{1}
\begin{table*}[t]
\centering
\caption{
\textbf{Native coding-agent performance on GameAnything-Bench}, ordered by Construction Overall.
}
\label{tab:main_results}

\begingroup
\setlength{\tabcolsep}{4.5pt}
\renewcommand{\arraystretch}{1.25}
\definecolor{ovcol}{HTML}{EEF1F6}

\resizebox{\textwidth}{!}{%
\begin{tabular}{@{}ll cccc>{\columncolor{ovcol}}cc ccc>{\columncolor{ovcol}}cc@{}}
\toprule

\multirow{2}{*}{\textbf{Model}}
& \multirow{2}{*}{\textbf{Harness}}
& \multicolumn{6}{c}{\textbf{Construction}}
& \multicolumn{5}{c}{\textbf{Editing}} \\

\cmidrule(lr){3-8}
\cmidrule(lr){9-13}

&
& Content
& Func.
& Vis.
& Host
& Overall $\uparrow$
& \$Cost $\downarrow$
& Edit Succ.
& Pres.
& Host
& Overall $\uparrow$
& \$Cost $\downarrow$ \\

\midrule
Claude Fable 5.2 & Claude Code
& 85.9 & 82.9 & 80.1 & 94.4 & \textbf{84.1} & 5.96
& 87.4 & 98.7 & 95.0 & \textbf{86.5} & 1.63 \\

Claude Sonnet 5.5 & Claude Code
& 59.4 & 56.8 & 50.4 & 96.6 & \underline{59.2} & 1.84
& 74.4 & 95.4 & 91.2 & \underline{72.8} & 0.53 \\

DeepSeek V4.1 Flash & OpenCode
& 49.7 & 48.4 & 45.2 & 95.9 & 51.4 & \textbf{0.26}
& 71.4 & 95.6 & 94.4 & 69.9 & \textbf{0.11} \\

GPT-5.6 Sol & Codex CLI
& 39.2 & 34.7 & 38.5 & 82.4 & 40.9 & 2.55
& 74.3 & 95.0 & 94.4 & 72.5 & 2.13 \\

Kimi K3 & OpenCode
& 34.4 & 29.7 & 30.0 & 98.9 & 35.7 & 5.48
& 69.4 & 95.8 & 92.5 & 67.7 & 2.62 \\

Qwen3.8-Flash & OpenCode
& 31.9 & 27.2 & 32.2 & 92.5 & 35.2 & 0.35
& 62.8 & 94.5 & 95.0 & 61.7 & 0.15 \\

Qwen3.8-Max & OpenCode
& 29.6 & 22.4 & 29.9 & 92.7 & 32.2 & 3.33
& 69.8 & 94.7 & 96.2 & 68.5 & 1.65 \\

GLM-5.3-Flash & OpenCode
& 25.7 & 19.6 & 26.2 & 95.4 & 29.0 & 0.52
& 64.7 & 96.1 & 92.5 & 63.4 & 0.19 \\

Gemini 3.8 Flash & OpenCode
& 23.3 & 16.5 & 26.5 & 92.8 & 27.6 & 2.65
& 62.6 & 95.6 & 93.8 & 61.3 & 2.42 \\

\bottomrule
\end{tabular}%
}

\vspace{3pt}
\begin{minipage}{\textwidth}
\footnotesize
Func.: functions \& playability; Vis.: visual \& presentation; Host: host fidelity; Edit Succ.: edit success; Pres.: preservation of content the edit should not touch.
Scores are percentages; Cost is development cost in USD per task. Bold: best Overall and lowest Cost; underline: second-best Overall.
Per-game results are in Appendix~\ref{app:per_game_results}, refinement results in Table~\ref{tab:refinement_results}.
\end{minipage}

\endgroup
\end{table*}






\subsection{Experimental Setup}
\label{sec:experimental-setup}

\paragraph{Coding Agents.}
We evaluate nine coding-agent configurations using Claude Code, Codex CLI, and OpenCode. The backbone models are Claude Fable 5.2~\citep{anthropic2026fable52}, Claude Sonnet 5.5~\citep{anthropic2026sonnet55}, GPT-5.6 Sol~\citep{openai2026gpt56}, Gemini 3.8 Flash~\citep{google2026gemini38flash}, GLM-5.3-Flash~\citep{zai2026glm53flash,glm5team2026glm5}, Qwen3.8-Max~\citep{alibaba2026qwen38max}, Qwen3.8-Flash~\citep{alibaba2026qwen38flash}, Kimi K3~\citep{kimiteam2026kimik3}, and DeepSeek V4.1 Flash~\citep{deepseek2026v41flash}. Each model and its native harness is a complete agent system. We retain the default prompts, tools, and reasoning settings of each harness without model-specific tuning. All agents receive identical task instructions, reference materials, and game development interfaces, with the same initial state for each task. Each native agent performs one autonomous attempt with a development budget of 120 minutes. We further evaluate iterative refinement with one additional round for eight configurations and three rounds for three selected configurations. Each round provides critic feedback based on gameplay replay, followed by 60 minutes of revision by the same coding agent with lessons retrieved from experience memory. The first Construction round on ten of the fourteen games (98 of 125 tasks) was run before experience memory was added and used 40 minutes without lessons; all later rounds and all Editing rounds use the full setting. Detailed configurations and execution settings are provided in Appendix~\ref{app:agent_config}.

\paragraph{Evaluation.}
We evaluate all submissions independently of the development agents and refinement critics using a consistent evaluation protocol. Claude Sonnet 5.5 serves as the judge for replay verification, playtest verification, visual assessment, and adjudication. Each submission undergoes executability checks with three probes per game (a common stability probe and two game-specific checks), followed by action replay and independent agentic playtesting for executable submissions. Independent playtesting is limited to 40 actions and 30 minutes per submission. We evaluate each submission once for the main results and conduct three independent evaluations on 47 submissions to assess scoring stability. We report 95\% confidence intervals using 4{,}000 paired bootstrap resamples over tasks. Reported development costs exclude independent evaluation. Further details are provided in Appendix~\ref{app:eval_cost}.

\subsection{Main Results}
\label{sec:main-results}

\paragraph{Turning multimodal inspirations into playable content remains challenging.}
Table~\ref{tab:main_results} reveals substantial capability gaps among current coding agents. Claude Fable 5.2 achieves the highest Construction Overall of 84.1, while the second-best agent reaches only 59.2 and five of the nine agents score below 40. Moreover, across all evaluated agents, Content scores are higher than Functions \& Playability scores, suggesting that creating the requested content is easier than making it fully functional and playable. These findings highlight the difficulty of transforming multimodal references into meaningful, interactive experiences within existing games.

\paragraph{Editing failures arise primarily from incomplete changes.}
Coding agents generally preserve existing game content but struggle to implement the requested modifications accurately. As shown in Table~\ref{tab:main_results}, Preservation exceeds 94 across all nine agents, while Edit Success ranges from only 62.6 to 87.4. These results indicate that maintaining the original game is not the primary difficulty in editing. Instead, reliably realizing targeted changes remains a major challenge.

\paragraph{Iterative development can substantially improve initial game quality.}
Native coding agents leave considerable room for further improvement. As shown in Table~\ref{tab:refinement_results}, a single refinement round raises both Construction and Editing scores for all eight evaluated configurations; 11 of these 16 gains have a 95\% confidence interval that excludes zero (Table~\ref{tab:app_ci_refine}). Three rounds further increase Construction scores by 7.6 to 14.8 points for the selected agents. Notably, Qwen3.8-Flash gains only 0.8 points after the first round but 14.8 points after three rounds, suggesting that limited early progress does not necessarily imply diminishing potential for further development. We further examine the effectiveness and cost of iterative refinement in Section~\ref{sec:analysis_refinement}.

\section{Analysis}
\label{sec:analysis}

\subsection{Iterative Refinement and Efficiency}
\label{sec:analysis_refinement}

\begin{figure}[t]
    \centering
    \includegraphics[width=0.82\textwidth]{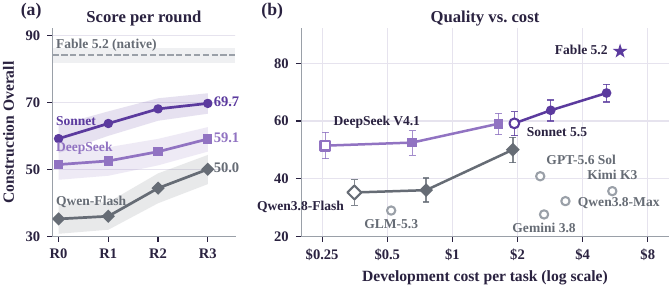}
    \caption{\textbf{Refinement quality and cost.} (a) Construction Overall across development rounds for three backbones, with native Claude Fable 5.2 as a reference. (b) Overall against cumulative development cost per task on a logarithmic scale. Shaded bands and error bars are 95\% bootstrap confidence intervals over tasks.}
    \label{fig:analysis_refinement}
\end{figure}

\looseness=-1
Small initial improvements can be followed by substantial gains in later refinement rounds. As shown in Figure~\ref{fig:analysis_refinement}, Qwen3.8-Flash and DeepSeek V4.1 Flash improve Construction by only 0.8 and 1.1 points after the first round, neither of which is statistically significant. After three rounds, however, their cumulative gains reach 14.8 and 7.6 points, both with confidence intervals excluding zero. Editing exhibits different improvement patterns (Table~\ref{tab:refinement_results}), with DeepSeek obtaining most of its gains in the first round while Qwen3.8-Flash continues improving in later rounds. These findings suggest that limited early progress does not necessarily indicate that further refinement has little value.

RSIGame~\citep{wu2026rsigame} demonstrates the benefits of autonomous game refinement while highlighting the importance of detecting improvement saturation and regression. Our results reveal a further tradeoff between continued refinement and starting with a stronger coding agent. DeepSeek V4.1 Flash reaches 59.1 Construction Overall after three rounds at \$1.72 per task, approaching native Claude Sonnet 5.5 at 59.2 and \$1.84. However, Qwen3.8-Flash reaches only 50.0 after three rounds at \$1.90, still below native DeepSeek at 51.4 and \$0.26. Thus, substantial improvements over an agent's initial performance do not necessarily translate into better cost efficiency than using a stronger agent directly. These comparisons concern development cost rather than matched execution time, and the observed gains reflect additional development time alongside critic feedback and experience memory.

\subsection{Multimodal Reference Understanding and Capability Diversity}
\label{sec:analysis_capability}

\begin{figure}[t]
    \centering
    \includegraphics[width=\textwidth]{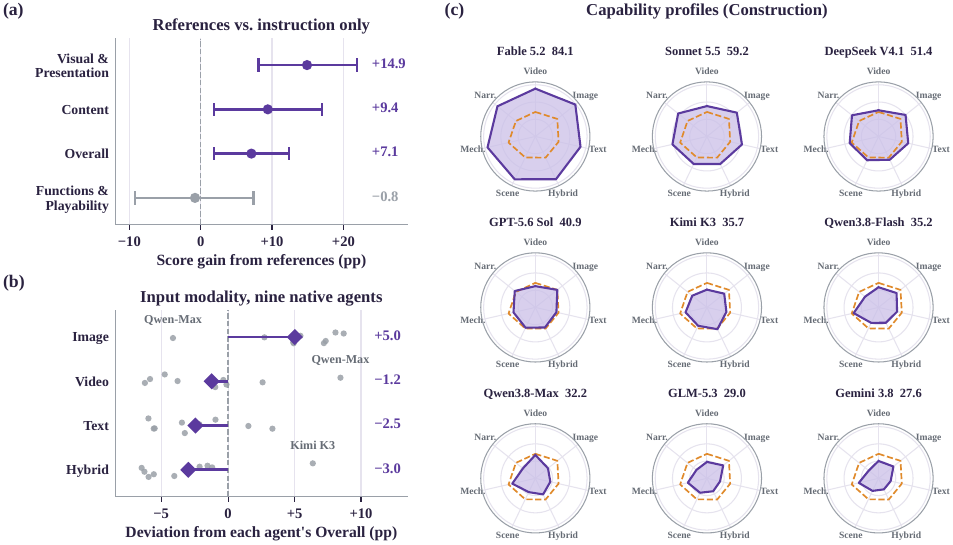}
    \caption{\textbf{Multimodal reference contributions and capability profiles.} (a) The score gain from providing references over instruction-only inputs for Claude Sonnet 5.5 on a 40-task subset, with paired-bootstrap 95\% confidence intervals. (b) Each native agent's deviation from its own Construction Overall across input modalities; diamonds indicate the mean over nine agents. (c) Construction Overall of each native agent across input modalities and task categories; the orange dashed curve denotes the mean over nine agents. Exact scores and task counts are provided in Table~\ref{tab:app_capability}.}
    \label{fig:analysis_capability}
\end{figure}

Multimodal references provide information that task instructions alone cannot capture. To examine their contribution, we remove the references from Construction tasks while keeping the instructions and evaluation unchanged (Appendix~\ref{app:reference_ablation}). As shown in Figure~\ref{fig:analysis_capability} (a), removing references reduces Claude Sonnet 5.5's Overall score by 7.1 points under matched harness settings. The decline is concentrated in Content and Visual \& Presentation, with the latter dropping by 14.9 points, while Functions \& Playability shows no clear change. This indicates that reference materials are particularly important for realizing the intended content and visual characteristics.

Different input modalities also reveal distinct agent capabilities. As shown in Figure~\ref{fig:analysis_capability} (b), image tasks achieve the highest mean Construction score across the nine agents and are the best-performing modality for seven of them. However, this pattern is not universal. Kimi K3 performs best on hybrid references, while Qwen3.8-Max achieves its highest score on video tasks. These differences reveal model-specific strengths that are obscured by aggregate performance and motivate evaluating coding agents across diverse reference modalities.

Agent performance also varies across task categories. As shown in Figure~\ref{fig:analysis_capability} (c), the Qwen models perform relatively better on Mechanics \& Behavior tasks, while Claude Sonnet 5.5 and DeepSeek V4.1 Flash achieve their highest scores on Narrative \& Event tasks. The exact scores in Table~\ref{tab:app_capability} further illustrate these differences across models and task categories.

\setcounter{topnumber}{2}
\subsection{Editing Failure Modes}
\label{sec:analysis_editing}

\begin{figure}[!t]
    \centering
    \includegraphics[width=0.86\textwidth]{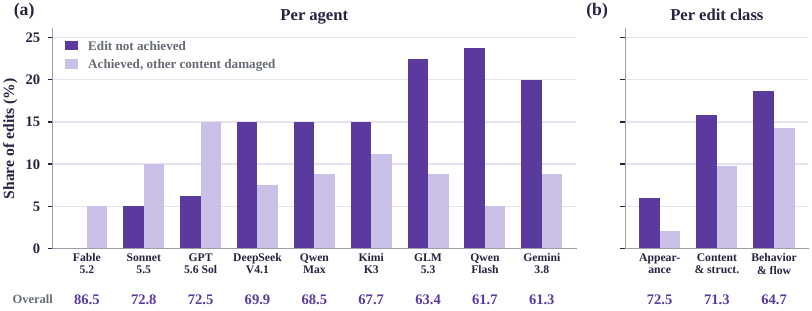}
    \caption{\textbf{Editing failure modes.} Fractions of edits that do not achieve the requested change and edits that achieve it but damage unrelated content. The left panel groups results by agent, and the right panel by edit class. The thresholds are Edit Success below 0.5 for missed edits and Edit Success at least 0.5 with Preservation below 0.9 for collateral damage.}
    \label{fig:analysis_editing}
\end{figure}

Behavioral edits are the most difficult to implement without disrupting existing gameplay. As shown in Figure~\ref{fig:analysis_editing}, Behavior \& Flow edits have the lowest Edit Success of 66.3, compared with 73.3 for Appearance, and a substantially higher rate of collateral damage of 14.3\% versus 2.1\%. This pattern is consistent across agents. Eight of the nine agents achieve lower Edit Success on behavioral edits, and all nine show lower Preservation. These findings highlight the difficulty of modifying gameplay behavior while keeping surrounding functionality intact.

Structural edits reveal a different failure pattern. Content \& Structure edits achieve nearly the same Edit Success as Appearance edits, with scores of 73.0 and 73.3, respectively. However, their Preservation drops from 98.7 to 94.5, indicating a greater risk of unintended changes to existing content. This distinction suggests that structural modifications can satisfy the requested change while compromising other parts of the game, revealing limitations that Edit Success alone cannot capture.

\subsection{Reliability of Agentic Evaluation}
\label{sec:eval_reliability}

\begin{figure}[!t]
    \centering
    \includegraphics[width=0.9\textwidth]{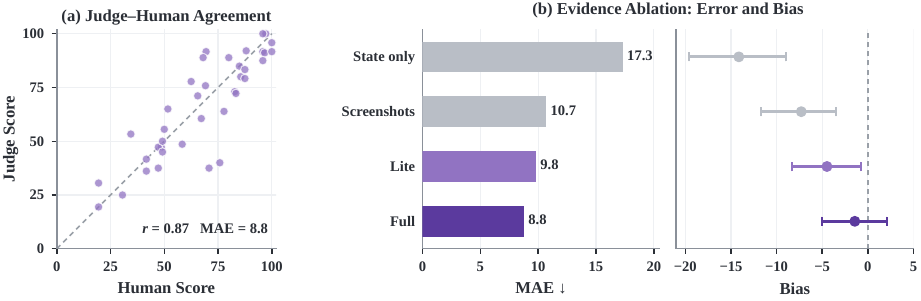}
    \caption{\textbf{Reliability of agentic evaluation.} The left panel compares the Full evaluator with human judgments on 40 submissions. The right panel reports mean absolute error and bias across evaluation settings, with 95\% bootstrap intervals for bias. Bias is the evaluator score minus the human score.}
    \label{fig:eval_human_agreement}
\end{figure}

Agentic evaluation closely aligns with human judgments and remains stable across repeated runs. As shown in Figure~\ref{fig:eval_human_agreement}, the Full evaluator achieves a Pearson correlation of $r=0.87$ on 40 human-labeled submissions and exactly reproduces the human ranking of six agent models. Compared with the Lite evaluator, independent playtesting and adjudication reduce scoring error and mitigate conservative scoring bias, while maintaining the same correlation with human judgments. Repeated Full evaluations on 47 submissions further demonstrate strong consistency, with an intraclass correlation of 0.93. Additional analyses of evaluator stability and playtesting budget, and of judges from other model families, are provided in Appendices~\ref{app:eval_stability} and~\ref{app:judge_families}.

\setcounter{topnumber}{2}
{\let\FloatBarrier\relax
\section{Conclusion}}
\label{sec:conclusion}

\looseness=-1
In this work, we introduced \textbf{GameAnything}, an open-source agentic platform for transforming diverse human inspirations into playable extensions of existing games through autonomous and iterative development. We further presented \textbf{GameAnything-Bench}, a comprehensive benchmark for evaluating multimodal reference understanding, game content adaptation, and gameplay quality across diverse game environments. Through evaluations of nine coding agent configurations, we found that current agents can effectively transfer reference content and preserve existing game functionality, but still face challenges in implementing playable interactions and precise modifications. Our results further demonstrate the benefits of multimodal references and iterative refinement, while revealing substantial differences in development capabilities and improvement trajectories across agents. Together, GameAnything and GameAnything-Bench provide a unified framework for developing and evaluating inspiration-driven game creation systems, highlighting opportunities for more reliable, adaptable, and autonomous game development.





\clearpage
\phantomsection
\addcontentsline{toc}{section}{References}
\bibliographystyle{plainnat}
\bibliography{reference}

\clearpage

\beginappendix
\setcounter{figure}{0}
\setcounter{table}{0}
\renewcommand{\thefigure}{A\arabic{figure}}
\renewcommand{\thetable}{A\arabic{table}}
\renewcommand{\topfraction}{0.9}\renewcommand{\bottomfraction}{0.8}\renewcommand{\textfraction}{0.07}\renewcommand{\floatpagefraction}{0.75}
\setcounter{topnumber}{3}\setcounter{bottomnumber}{2}\setcounter{totalnumber}{4}

\newcommand{\apxsec}[4]{%
  \noindent\hyperref[#1]{%
    \makebox[1.9em][l]{\textcolor{black!70}{\bfseries\ref*{#1}}}%
    \textbf{#2}}%
  \nobreak\hfill\nobreak\hyperref[#1]{\textbf{\pageref*{#1}}}\par
  \nobreak\vspace{-4pt}%
  \noindent\hspace{1.9em}\begin{minipage}[t]{\dimexpr\linewidth-1.9em\relax}
    \footnotesize\textcolor{black!45}{#3}\par
    #4
  \end{minipage}\par\vspace{1.5pt}}
\newcommand{\apxsub}[2]{%
  \nobreak\noindent\hyperref[#1]{\small\ref*{#1}\hspace{0.6em}#2}%
  \nobreak\hfill\nobreak\hyperref[#1]{\small\pageref*{#1}}\par}
\newcommand{\apxsubsub}[2]{%
  \nobreak\noindent\hspace{1.2em}\hyperref[#1]{\small\ref*{#1}\hspace{0.6em}#2}%
  \nobreak\hfill\nobreak\hyperref[#1]{\small\pageref*{#1}}\par}

\begingroup
\hypersetup{linkcolor=PrettyText, pdfborder={0 0 0}}
\setlength{\parindent}{0pt}
{\small\textcolor{black!60}{This appendix first invites community contributions, then details how the benchmark was built, how submissions are evaluated, the experimental setup, the reliability of the evaluation, and how GameAnything is used in practice.}}\par\vspace{4pt}

\apxsec{app:contribute}{Call for Contributions}
  {Why we need community help, two ways to join, what to submit, submission standards, and how contributors are credited.}
  {\apxsub{app:why_contribute}{Why Contribute}%
   \apxsub{app:two_ways}{Two Ways to Join}%
   \apxsub{app:what_submit}{What to Submit}%
   \apxsub{app:standards}{Submission Standards}%
   \apxsub{app:credit}{Review and Credit}}

\apxsec{app:bench_details}{Benchmark Construction Details}
  {Target games, reference collection and annotation, and how references become tasks and rubrics.}
  {\apxsub{app:game_selection}{Game Selection}%
   \apxsubsub{app:opensource_game_selection}{Open-source Games}%
   \apxsubsub{app:commercial_game_selection}{Commercial Games}%
   \apxsub{app:reference_collection}{Reference Collection and Annotation}%
   \apxsub{app:task_construction}{Task and Rubric Construction}%
   \apxsub{app:case_studies}{Task Construction Case Studies}}

\apxsec{app:eval_protocol}{Evaluation Protocol Details}
  {Executability gate, agent execution and independent evaluation, and scoring.}
  {\apxsub{app:game_capability}{Game Capability Probes and Executability Gate}%
   \apxsub{app:agentic_eval}{Agent Execution and Independent Evaluation}%
   \apxsub{app:eval_details}{Scoring Rules and Aggregation}}

\apxsec{app:exp_settings}{Experimental Setup Details}
  {Coding-agent configurations, execution settings, and evaluation and statistical settings.}
  {\apxsub{app:agent_config}{Agent Configurations and Execution Settings}%
   \apxsub{app:evaluation_cost}{Evaluation and Cost Settings}}

\apxsec{app:per_game_results}{Per-Game Results}
  {Construction Overall of every agent configuration, native and with +Ours refinement, on each of the 14 games.}{}

\apxsec{app:analysis_tables}{Additional Analysis Results}
  {Refinement results per round (R0 to R3), and scores by input modality and task category, by edit class, and across open-source and commercial games.}{}

\apxsec{app:ablations}{Additional Ablations and Robustness Analysis}
  {The effect of withholding the references and the variance across independent development runs.}
  {\apxsub{app:ablation_protocol}{Shared Task Subset and Protocol}%
   \apxsub{app:reference_ablation}{Reference Ablation}%
   \apxsub{app:dev_variance}{Variance across Development Runs}}

\apxsec{app:reliability_eval}{Evaluation Reliability Analysis}
  {Agreement with human labels, evidence ablation, stability and playtest budget, and judges from other model families.}
  {\apxsub{app:human_agreement}{Agreement with Human Judgments}%
   \apxsub{app:evidence_ablation}{Evidence Ablation}%
   \apxsub{app:eval_stability}{Evaluation Stability and Playtest Budget}%
   \apxsub{app:judge_families}{Judges from Other Model Families}}

\apxsec{app:ui}{GameAnything: User Interface and Workflow}
  {The web application, decision cards, the four modes, evidence-anchored review, and installation.}{}

\apxsec{app:iter_case_studies}{GameAnything: Iterative Development Case Studies}
  {Hogwarts and the Grand View Garden in Minecraft, and San Diego in Cities: Skylines, round by round.}{}
\endgroup
\clearpage

\section{Call for Contributions}
\label{app:contribute}
GameAnything is an open project, and we warmly welcome contributions from
players, game designers, modders, and researchers.

\subsection{Why Contribute}
\label{app:why_contribute}
Human inspiration is open-ended. No fixed team can cover every film, novel,
place, or play style that people want to bring into games, and no small group
of testers can tell us everything that works or breaks on the platform.
Community input helps in two ways: it makes GameAnything more useful for real
creators, and it makes GameAnything-Bench a broader and harder test of coding
agents.

\subsection{Two Ways to Join}
\label{app:two_ways}
The two links below serve different purposes and ask for different levels of
effort.

\begin{center}\small
\begin{tabular}{@{}lp{0.38\linewidth}p{0.38\linewidth}@{}}
\toprule
 & \href{https://huggingface.co/spaces/GameAnything/GameAnything-Trial}{\textbf{Try GameAnything}}
 & \href{https://huggingface.co/spaces/GameAnything/gameanything-submit}{\textbf{Contribute Tasks}} \\
\midrule
For       & Anyone who wants to use the platform & Anyone with a game idea and a reference for it \\
You share & The games you play, the mods you have made, and one idea you would like to build (about 3 minutes) & A reference, a target game, a request, and scoring points (about 10 minutes, no coding) \\
You get   & Free trial access, if selected & Credit in this paper (Appendix~\ref{app:credit}) \\
\bottomrule
\end{tabular}
\end{center}

For the trial, we look first at players who have spent hundreds of hours in
one game and at people who have made their own mods, maps, or levels. A link to
past work helps most. We contact selected testers by email and ask for honest
feedback in return.

\subsection{What to Submit}
\label{app:what_submit}
A task (a \emph{challenge} on the submission page) follows the format in
Section~\ref{sec:task_construction} and has the parts listed in
Table~\ref{tab:submit_parts}. The submission page shows three complete
examples.

\begin{table}[h]
\centering\small
\caption{\textbf{Parts of a submitted task.}}
\label{tab:submit_parts}
\begin{tabular}{@{}lp{0.72\linewidth}@{}}
\toprule
Part & What to provide \\
\midrule
Reference      & A video clip (30--60\,s works best), 3--6 images of one subject, a text passage of at least 300 words, or a mix of these \\
Target game    & One of the 14 games on the submission page, or a new PC game you propose that supports mods, maps, or scripts \\
Request        & What the agent should build and what the player can do with it (30--300 words), plus why the game fits the idea (10--100 words) \\
Scoring points & 2--6 points each for \emph{content} (what must exist), \emph{function} (what must work when played), and \emph{visual} (what it must look like) \\
Follow-up edit & Optional. One change to ask for after the build (10--80 words), used to create an Editing task \\
\bottomrule
\end{tabular}
\end{table}

\subsection{Submission Standards}
\label{app:standards}
Scoring points are the checklist the agent's work is graded on: each one is a
single sentence about something that must appear in the game. Each point must
meet the four criteria in Table~\ref{tab:point_criteria}. All examples come
from the beekeeping challenge on the submission page.

\begin{table}[h]
\centering\small
\caption{\textbf{Criteria for scoring points.}}
\label{tab:point_criteria}
\begin{tabular}{@{}lp{0.30\linewidth}p{0.50\linewidth}@{}}
\toprule
Criterion & Requirement & Example \tabularnewline
\midrule
Grounded  & \raggedright It appears in the reference, and you say where
          & \raggedright ``The hives are set out right beside flowering vegetation.'' (img 1, 3, 4) \tabularnewline
Checkable & \raggedright It can be seen or tried in the game, not a matter of taste
          & \raggedright ``The player keeps several hives side by side, each opened and harvested on its own.'' \tabularnewline
Defining  & \raggedright It helps identify the reference; generic details are left out
          & \raggedright ``Working an open hive: a frame of comb lifted out of the opened hive.'' \tabularnewline
Positive  & \raggedright It says what must be there, not what must be absent
          & \raggedright ``Harvested frames are processed in a separate extraction step into honey the player obtains.'' \tabularnewline
\bottomrule
\end{tabular}
\end{table}

Points such as ``The hives look realistic'' (not checkable) or ``Bees must not
fly through walls'' (not positive) are not accepted.

Beyond the points, the whole task must ask for something \textbf{playable},
not a reskin that only changes how existing content looks. If the reference is
your own work, in the public domain, or under a Creative Commons license,
upload it. If it is someone else's copyrighted work, give only a link and a
time range. We do not accept submissions with unclear sources, reskins with
nothing to play, gore, sexual or hateful content, private information about
real people, game files, cracks, or cheats. Duplicates and low-effort text are
also declined.

\subsection{Review and Credit}
\label{app:credit}
Our team reviews every submission and may ask for one revision. An accepted
submission becomes a full task: we write the final instruction for the agent
and refine the scoring points into its rubric. Only tasks that enter the
released benchmark count toward credit. With \textbf{five accepted tasks}, you
are thanked by name (or a nickname) in the Acknowledgements. With \textbf{a
complete game package plus 10 accepted tasks}, you are invited onto the author
list; a game package is a new game fully connected to our platform, built
together with our team. Credit is counted per person by email address, and the
same idea submitted twice counts once, for the first sender. Accepted tasks are
released under CC~BY~4.0 with the name you choose. Tasks built on copyrighted
references are kept private but are credited all the same. Credit appears in
the next version of this paper after your work is included. There is no
payment.

\section{Benchmark Construction Details}
\label{app:bench_details}

\subsection{Game Selection}
\label{app:game_selection}

\definecolor{GameRow}{HTML}{F6F7FB}
\definecolor{TagThreeD}{HTML}{DCE6F4}
\definecolor{TagTwoD}{HTML}{F4E6DC}
\newcommand{\gname}[2]{\textbf{#1}\newline{\scriptsize\color{PrettyMuted}#2}}
\newcommand{\gtag}[1]{\leavevmode{\footnotesize\color{PrettyNavy}#1}}   
\newcommand{\vtag}[1]{{\setlength{\fboxsep}{1.5pt}\ifnum#1=3 \colorbox{TagThreeD}{\scriptsize\strut\,\textbf{3D}\,}\else\colorbox{TagTwoD}{\scriptsize\strut\,\textbf{2D}\,}\fi}}
\newcommand{\mf}[1]{{\footnotesize\ttfamily #1}}
\newcommand{\gthumb}[1]{\raisebox{\dimexpr-\height+1.7ex\relax}{\includegraphics[width=2.3cm]{figures/games/#1.jpg}}}

\newlength{\pfw}\setlength{\pfw}{0.88\dimexpr(\textwidth-1.8cm-15pt)/4\relax}
\newcommand{\rf}[1]{\includegraphics[width=\pfw]{figures/games/play_ref/#1}}
\newcommand{\pf}[2]{\begin{tabular}[t]{@{}l@{}}\includegraphics[width=\pfw]{figures/games/play_annot/#1}\\[-1pt]{\scriptsize\sffamily #2}\end{tabular}}
\newcommand{\rlab}[2]{\parbox[c]{1.8cm}{\raggedright\textbf{\scriptsize #1}\\{\tiny #2}}}
\newcommand{\extlabel}{\multicolumn{7}{@{}l}{\rule{0pt}{13pt}\footnotesize\textbf{\color{PrettyNavy}Example extensions}} \\*}
\newcommand{\extrow}[1]{\multicolumn{7}{@{}c@{}}{\footnotesize\setlength{\tabcolsep}{1.5pt}\renewcommand{\arraystretch}{1.0}\begin{tabular}{@{}m{1.8cm}cccc@{}}#1\end{tabular}}}

\begin{table*}[!htbp]
\centering
\caption{Number of games at each selection stage, grouped by gameplay category. Counts combine open-source and commercial games. Selection procedures are detailed in Appendices~\ref{app:opensource_game_selection} and~\ref{app:commercial_game_selection}.}
\label{tab:game_selection_funnel}
\begingroup
\footnotesize
\setlength{\tabcolsep}{5pt}
\begin{tabular}{@{}lccccccc@{}}
\toprule
\textbf{Category}
& \shortstack{\textbf{Steam}\\\textbf{ref.}}
& \shortstack{\textbf{First}\\\textbf{round}}
& \shortstack{\textbf{Added}\\\textbf{later}}
& \shortstack{\textbf{Deploy}\\\textbf{tried}}
& \shortstack{\textbf{Ran \&}\\\textbf{recorded}}
& \shortstack{\textbf{Passed}\\\textbf{review}}
& \textbf{Final} \\
\midrule
Shooter & 5 & 6 & 3 & 8 & 7 & 3 & 1 \\
Action & 6 & 3 & 1 & 2 & 2 & 2 & 1 \\
Action RPG & 5 & 5 & 2 & 5 & 4 & 3 & 1 \\
Creature-collection RPG & 1 & 0 & 1 & 1 & 1 & 1 & 1 \\
Turn-based strategy & 4 & 9 & 1 & 8 & 8 & 2 & 1 \\
Real-time strategy & 5 & 3 & 2 & 3 & 3 & 2 & 1 \\
Building / management & 7 & 8 & 0 & 6 & 6 & 2 & 1 \\
Survival / sandbox & 13 & 13 & 4 & 15 & 15 & 8 & 4 \\
Platformer & 2 & 3 & 1 & 3 & 3 & 2 & 1 \\
Racing / simulation & 6 & 9 & 4 & 12 & 12 & 4 & 1 \\
Life sim / farming & 2 & 2 & 0 & 1 & 1 & 1 & 1 \\
Other & 3 & 9 & 1 & 3 & 3 & 0 & 0 \\
\midrule
\textbf{Total} & \textbf{59} & \textbf{70} & \textbf{20} & \textbf{67} & \textbf{65} & \textbf{30} & \textbf{14} \\
\bottomrule
\end{tabular}
\vspace{3pt}
\begin{minipage}{\textwidth}
\footnotesize
\textit{Notes.} Counts include both open-source and commercial game candidates. Minecraft was added outside the Steam reference set. The final games are listed in Tables~\ref{tab:final_games} and~\ref{tab:commercial_games}.
\end{minipage}
\endgroup
\end{table*}

\subsubsection{Open-source Games}
\label{app:opensource_game_selection}

\paragraph{Game Sources}
We first built a reference set of single-player Steam games with active modding communities, using Steam review counts, Nexus Mods listings, and Steam Workshop entries as indicators of player base and modding activity.
Within the categories reported in Table~\ref{tab:game_selection_funnel}, this set contains 59 games. The counts in this section follow Table~\ref{tab:game_selection_funnel} and include the commercial candidates of Appendix~\ref{app:commercial_game_selection}.
This set served only as a map of popular, moddable genres. We then searched for open-source counterparts genre by genre, matching on genre rather than on individual titles, through the Wikipedia list of open-source games, the \texttt{bobeff/open-source-games} list, \texttt{osgameclones}, GamingOnLinux, and general web search.
Within these categories, this first round produced 70 candidates.
Because many of these candidates had dated visuals, we ran a second search focused on high-quality 3D open-source games, using six parallel web searches totaling about 600 queries.
We also included games from our earlier internal tests.
In total, 20 games outside the first-round list entered the trial stage: 14 from the 3D searches, 2 tested earlier, 3 from an earlier 2D test, and Minecraft for the commercial track.

\paragraph{Selection Requirements}
Table~\ref{tab:game_selection_funnel} reports the number of games at each stage.
We screened first-round candidates on five requirements: the game runs on our Linux servers without root access, contains real playable content rather than an empty engine, offers a modding route such as scripts, data files, level editors, or plugins, has acceptable visuals, and has been maintained since October 2024.

\textit{Runtime trial.} Within these categories, we attempted to deploy 67 games, 57 open-source and the 10 shortlisted commercial games, in a Docker sandbox on a GPU server, without accounts or logins.
Using each game's own modding or scripting interface and without patching binaries, we made one or two small visible changes and recorded before-and-after gameplay.
Two games could not be deployed: Ryzom Core requires a database, web login, and many services, and the data for Arma: Cold War Assault is available only through a logged-in Steam account.

\textit{Manual quality review.} We then manually reviewed the recordings of the 65 running games and judged each game's overall quality, with visual quality as the primary consideration.
Visuals served as a minimum bar rather than a preference for a particular style: 2D and low-polygon games were retained when their presentation stayed clear and coherent, while games with the most dated presentation were removed.
We also considered overall game quality and complexity, favoring games with enough content and interacting systems to support substantive extension.
We also favored games that could show a scene at the scale of a playable character or world, and removed games dominated by menus, abstract views, or distant map cameras.
This step was a holistic judgment rather than a fixed scoring procedure.
It retained 30 games, 25 open-source and 5 commercial; two more were marked as uncertain and did not advance.

\textit{Development trial.} For each of the 25 open-source games, we built four video-adaptation tasks, one each from city, film, game trailer, and sports references, for 100 tasks in total.
Each task went through repeated rounds of development, independent playtesting, and independent review; Table~\ref{tab:final_games} shows two of the resulting extensions for each selected game.

\textit{Final selection.} We kept one game per gameplay type by manually comparing the final outputs of these tasks, asking whether the reference content could be clearly expressed in the game.
We removed vehicle simulators because their task outputs were weak, and social 3D sandboxes because their gameplay was too thin.
The final set requires that code and data can be downloaded without an account, the game can run and be recorded in our GPU sandbox, the game has been successfully modified, the player can take the role of the reference's protagonist, the game has the strongest visuals in its type, and the selected games cover different kinds of gameplay.

\paragraph{Final Game List}
Table~\ref{tab:final_games} lists the nine open-source games in the benchmark, in the same layout as Table~\ref{tab:commercial_games}.

\begingroup
\small
\setlength{\tabcolsep}{4pt}
\setlength{\LTleft}{0pt}\setlength{\LTright}{0pt}
\renewcommand{\arraystretch}{1.3}
\begin{longtable}{@{}p{2.3cm}>{\raggedright\arraybackslash}p{2.5cm}>{\raggedright\arraybackslash}p{1.95cm}>{\centering\arraybackslash}p{0.75cm}>{\raggedright\arraybackslash}p{3.55cm}>{\raggedright\arraybackslash}p{2.75cm}>{\raggedright\arraybackslash}p{0.85cm}@{}}
\caption{\textbf{Open-source target games.} Game information, original-game screenshots, and two example extensions per game from development trials. Reference frames are shown above the corresponding in-game results. Task counts refer to Construction tasks.}
\label{tab:final_games}\\
\toprule
\textbf{Key art} & \textbf{Game} & \textbf{Genre} & \textbf{View} & \textbf{What the player does} & \textbf{Modding surface} & \textbf{Tasks} \\
\midrule
\endfirsthead
\multicolumn{7}{@{}l}{\small\textit{Table~\ref{tab:final_games} continued}} \\
\toprule
\textbf{Key art} & \textbf{Game} & \textbf{Genre} & \textbf{View} & \textbf{What the player does} & \textbf{Modding surface} & \textbf{Tasks} \\
\midrule
\endhead
\bottomrule
\endlastfoot
\gthumb{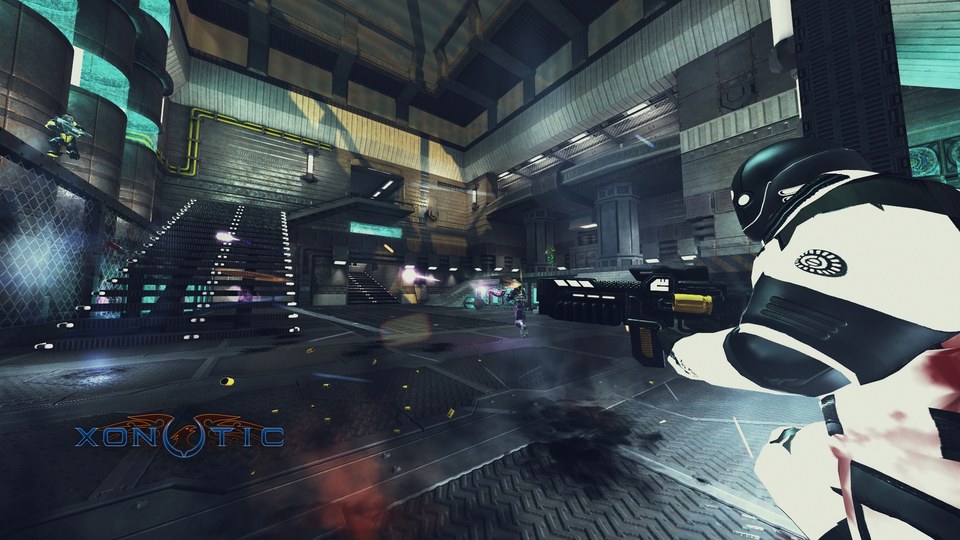} & \gname{Xonotic}{0.8.6 \textperiodcentered{} DarkPlaces \textperiodcentered{} GPL-3.0} & \gtag{Shooter} & \vtag{3} & Fights fast deathmatches against bots in arenas: moves, picks up weapons and items, frags opponents. & \mf{.cfg} configs, \mf{pk3} data packs, QuakeC & 7 \\*
\multicolumn{7}{@{}l@{}}{\rule{0pt}{12pt}\includegraphics[width=0.995\textwidth]{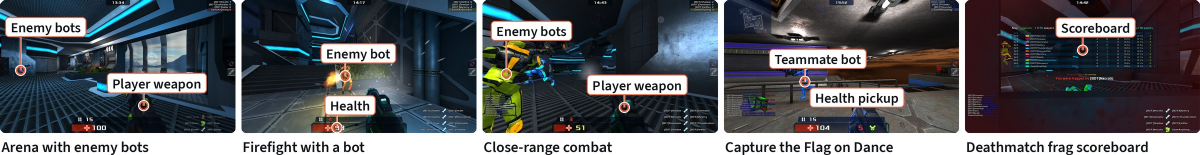}} \\[2pt]
\extlabel
\extrow{\multicolumn{5}{@{}l}{\scriptsize\textbf{City}: a Manhattan evening from dusk to the lit billboards} \\*
\rlab{Reference}{New York city footage} & \rf{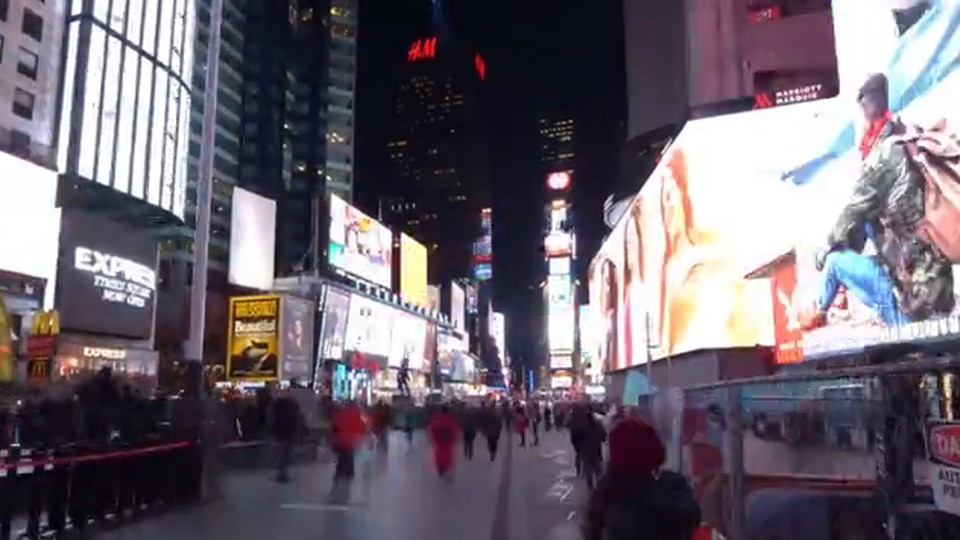} & \rf{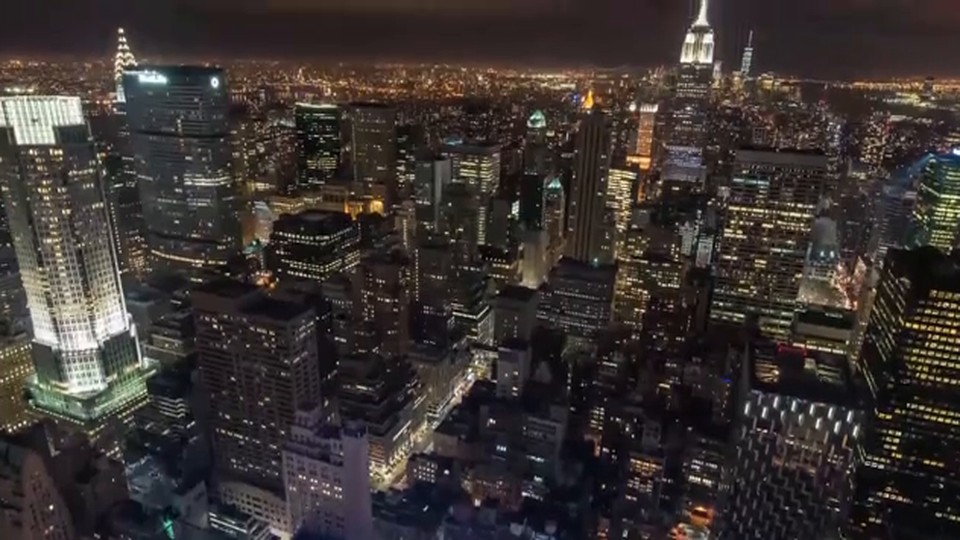} & \rf{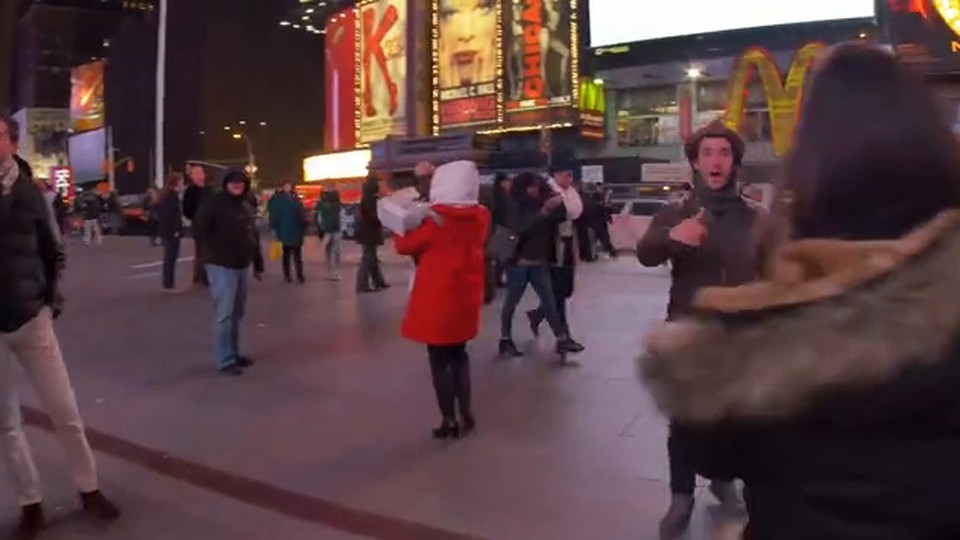} & \rf{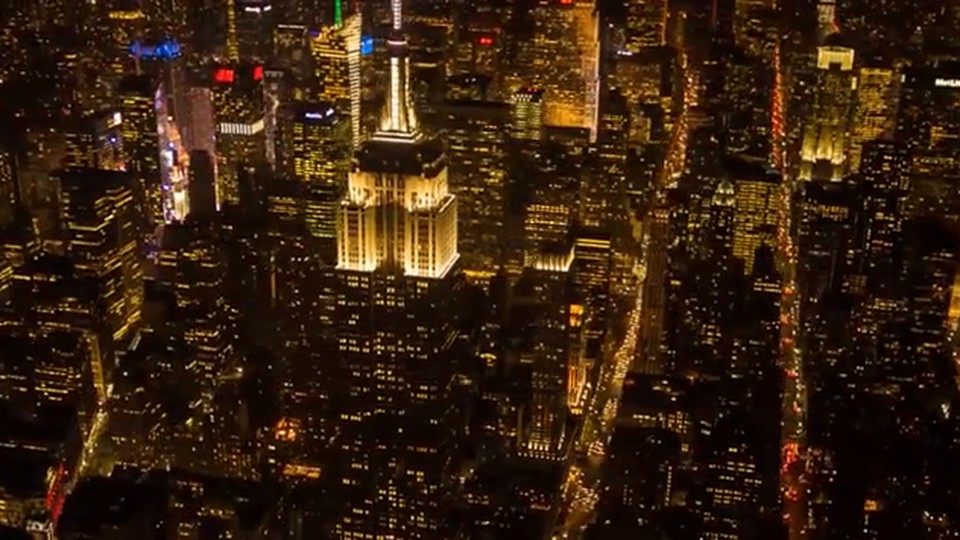} \\*
\rlab{In game}{} & \pf{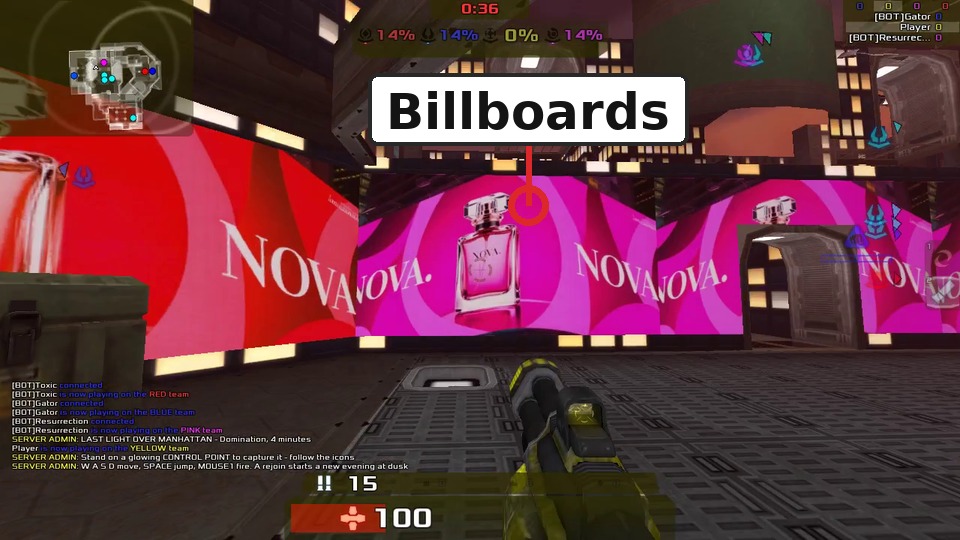}{Times Square billboards} & \pf{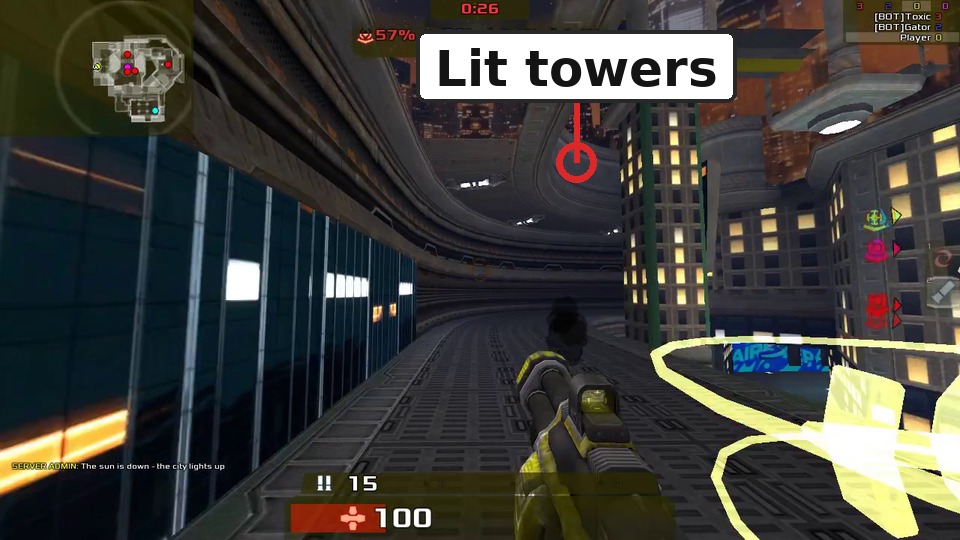}{Midtown towers at night} & \pf{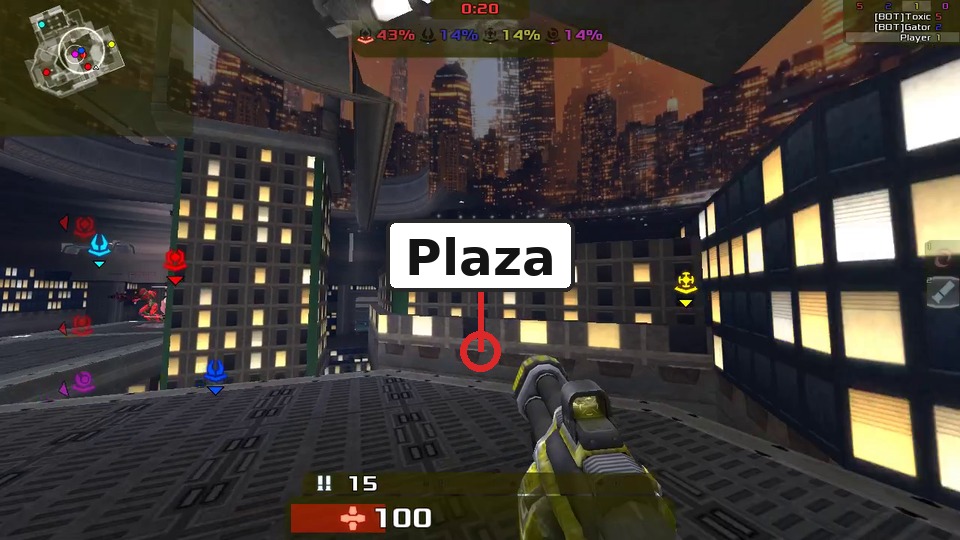}{Crowded plaza} & \pf{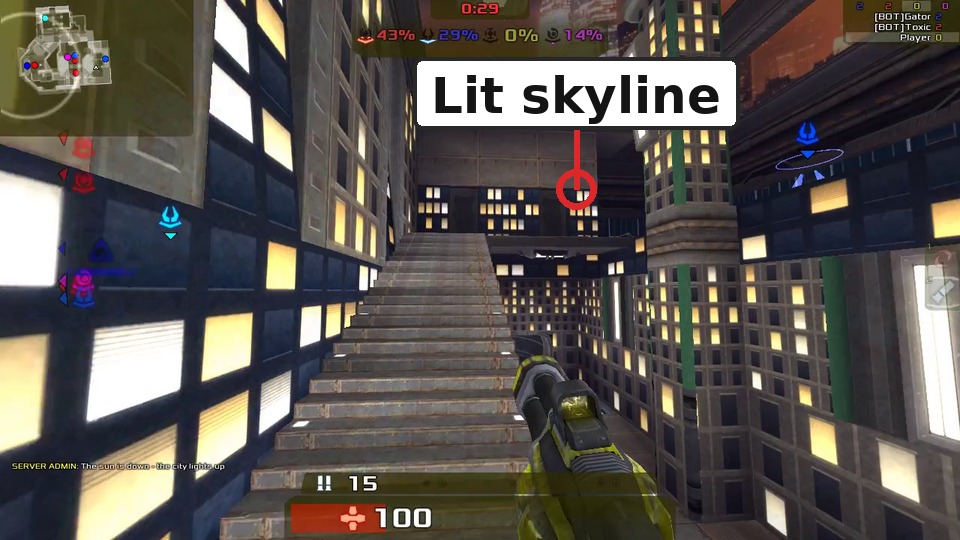}{Night city from above}} \\[2pt]
\extrow{\multicolumn{5}{@{}l}{\scriptsize\textbf{Film}: Mario drops into the Bob-omb hall and floats out on Peach's parasol} \\*
\rlab{Reference}{The Super Mario Bros. Movie} & \rf{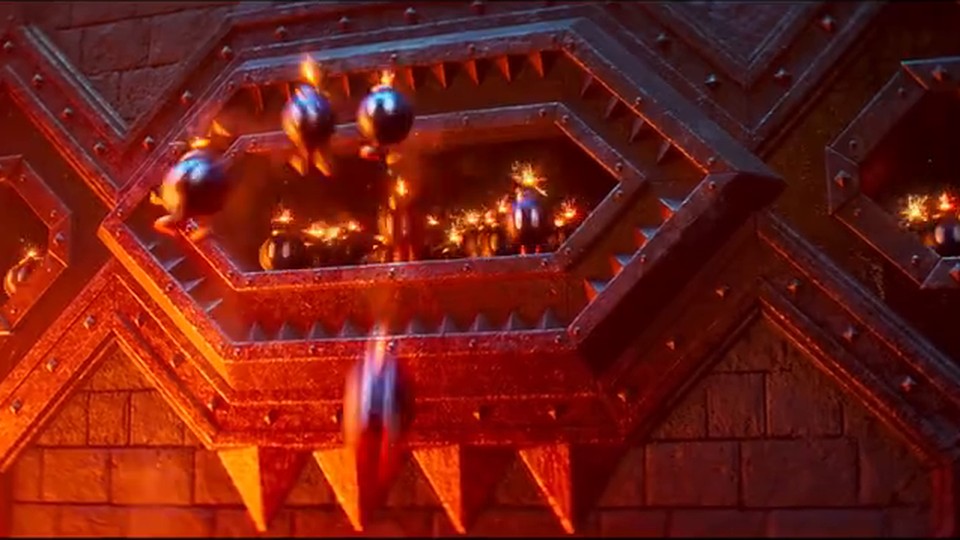} & \rf{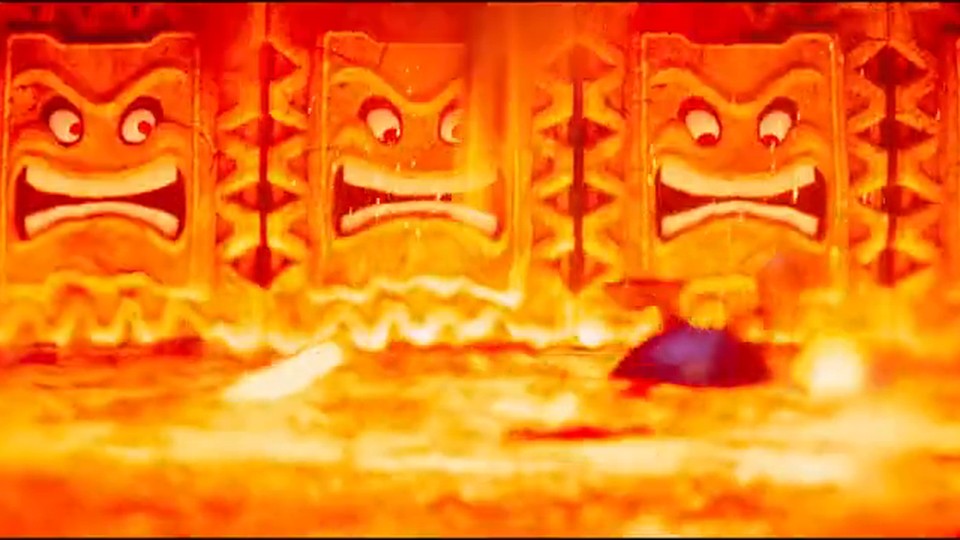} & \rf{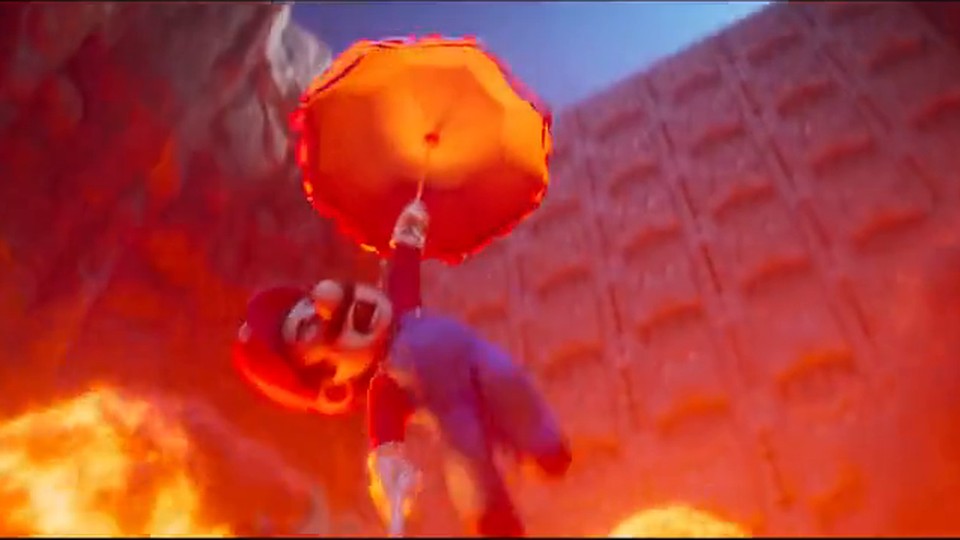} & \rf{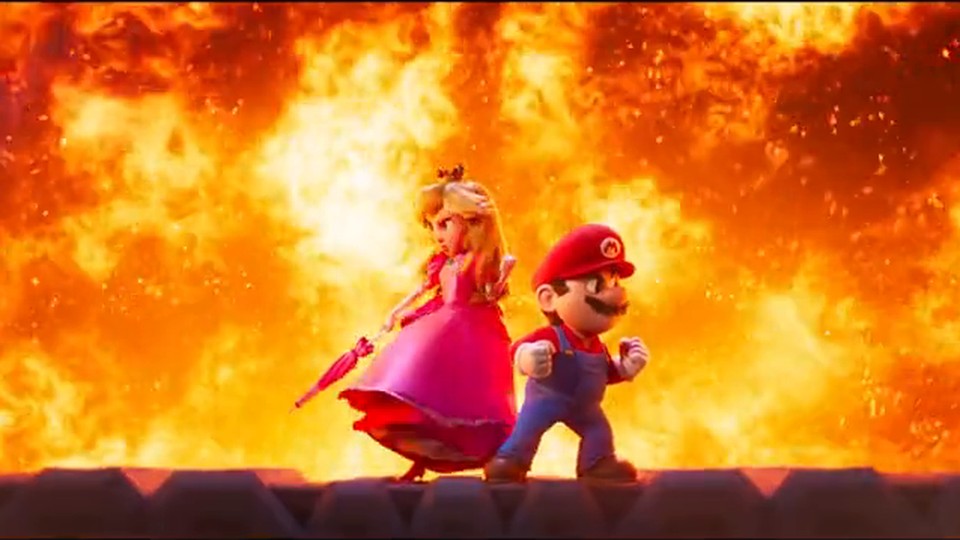} \\*
\rlab{In game}{} & \pf{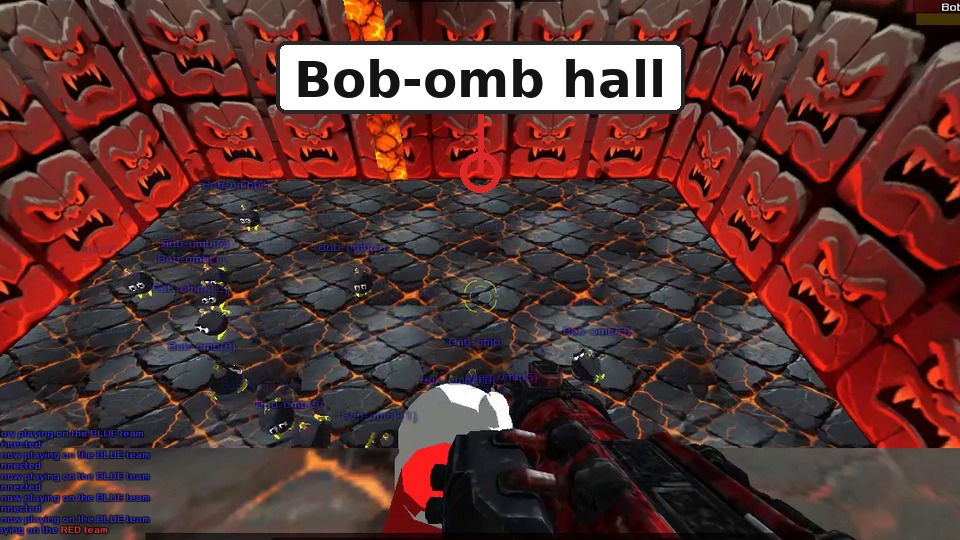}{Bob-omb hall} & \pf{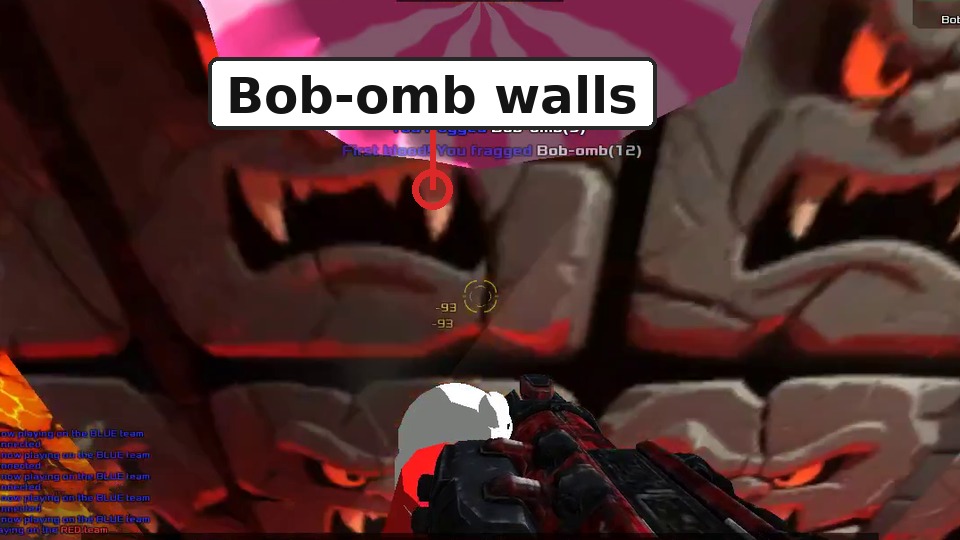}{Bob-omb faces} & \pf{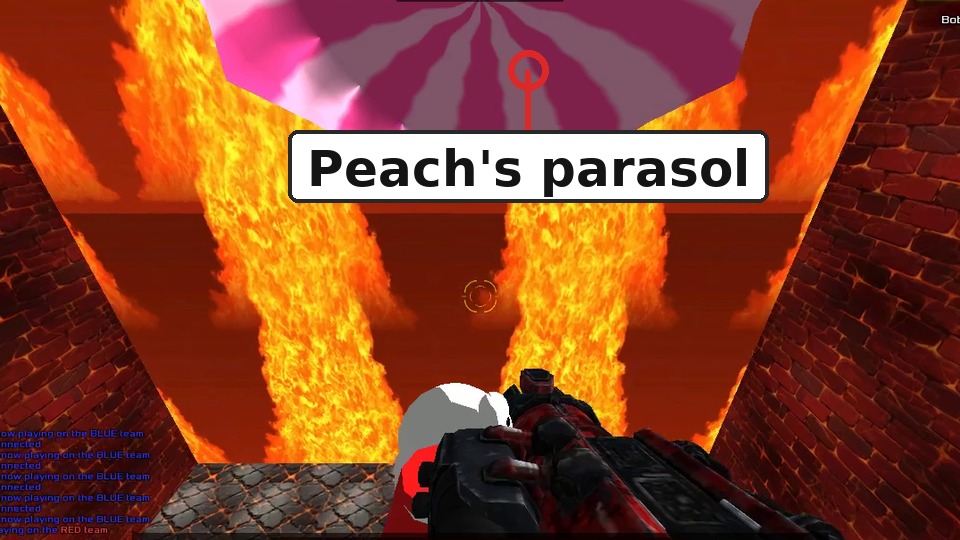}{Peach's parasol} & \pf{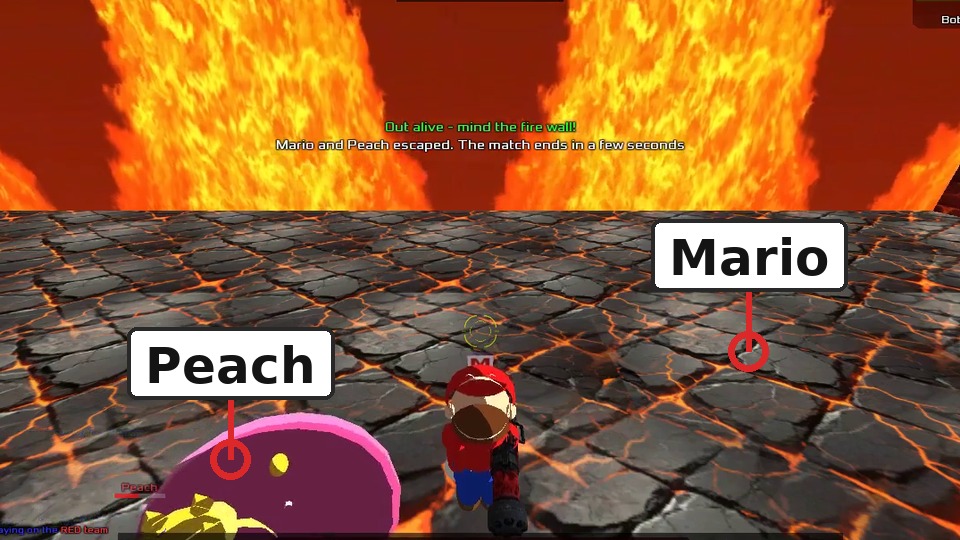}{Mario and Peach land}} \\[4pt]
\midrule
\gthumb{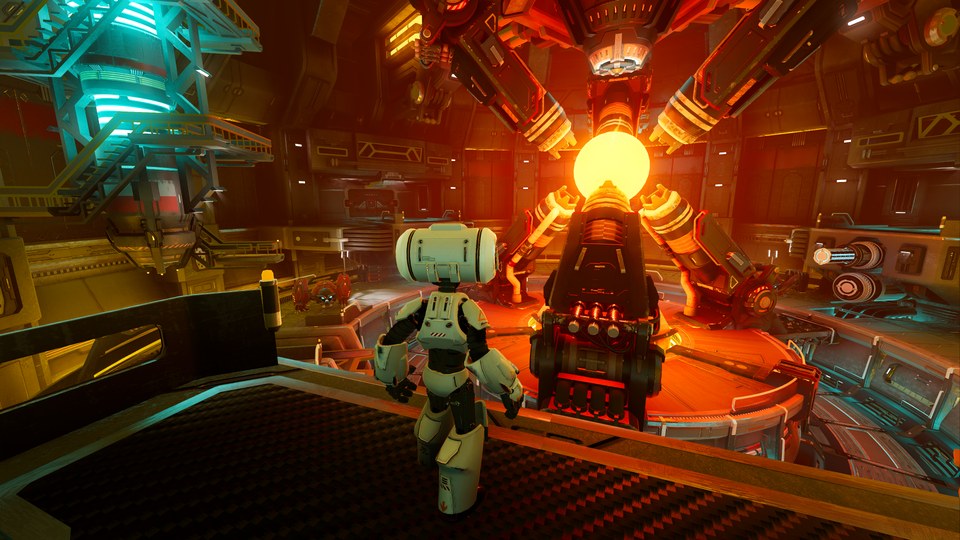} & \gname{Godot TPS Demo}{a82f154 \textperiodcentered{} Godot 4.7 \textperiodcentered{} MIT / CC-BY} & \gtag{Action} & \vtag{3} & Guides a robot through a sci-fi level, aiming and shooting at enemy drones and turrets. & Text scenes, resources, GDScript & 11 \\*
\multicolumn{7}{@{}l@{}}{\rule{0pt}{12pt}\includegraphics[width=0.995\textwidth]{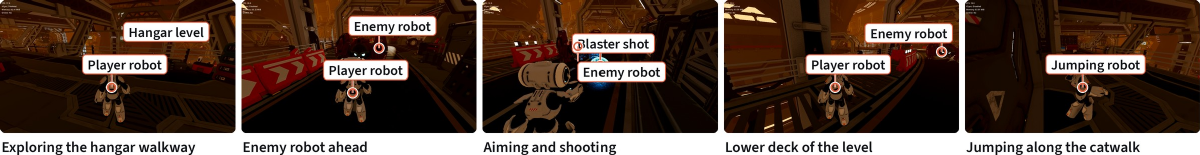}} \\[2pt]
\extlabel
\extrow{\multicolumn{5}{@{}l}{\scriptsize\textbf{Game trailer}: Kirby crosses the meadow to the Waddle Dees} \\*
\rlab{Reference}{Kirby trailer} & \rf{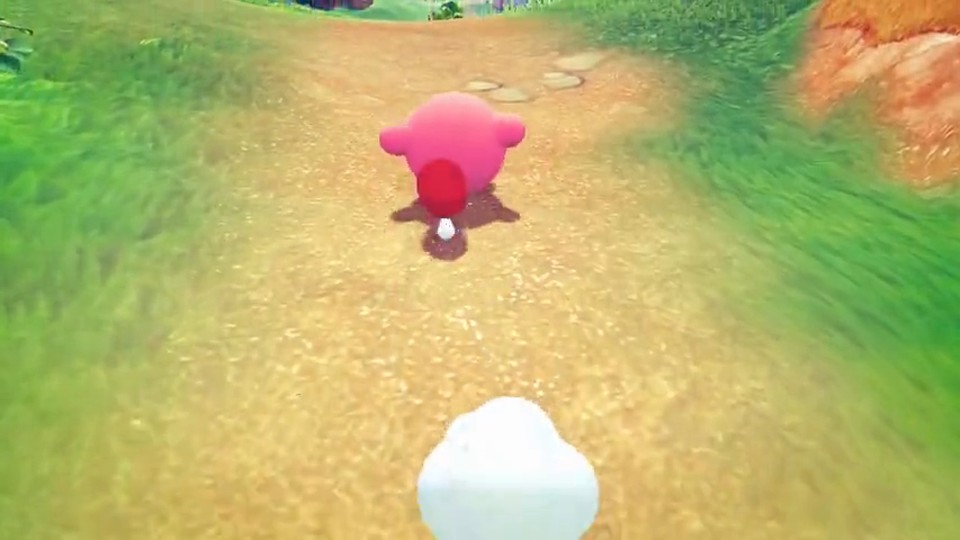} & \rf{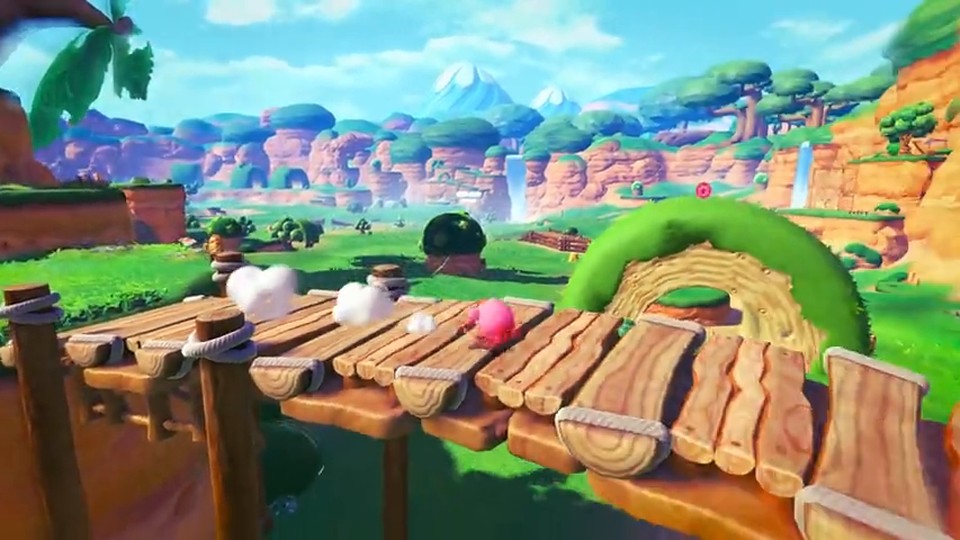} & \rf{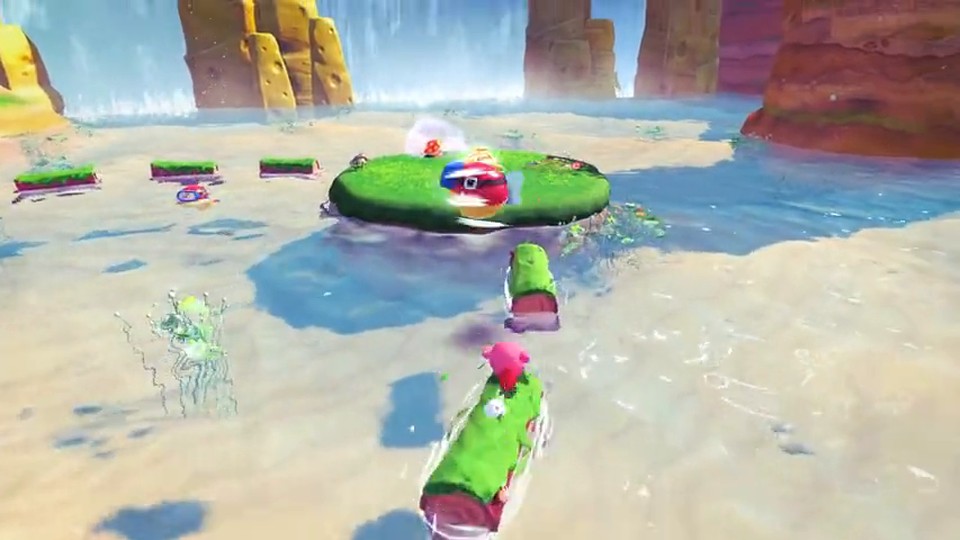} & \rf{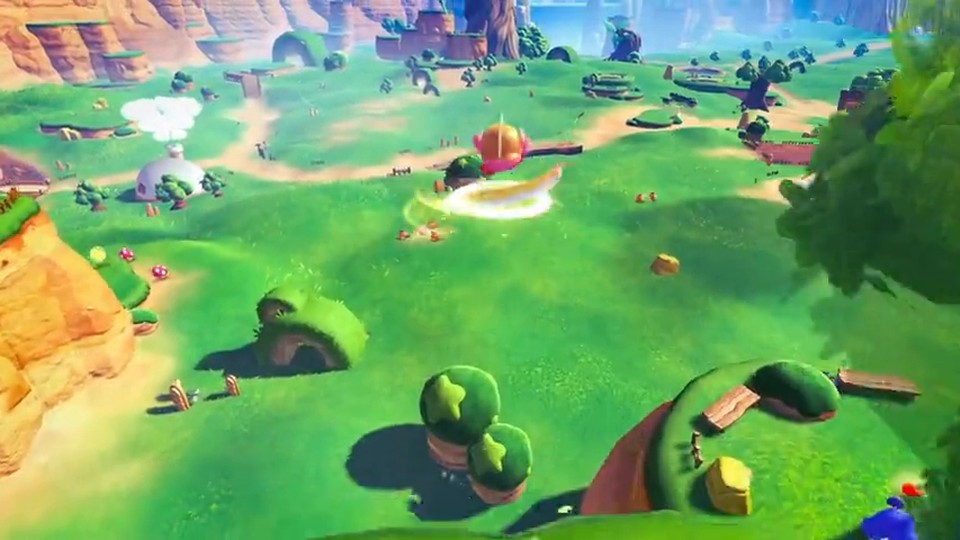} \\*
\rlab{In game}{} & \pf{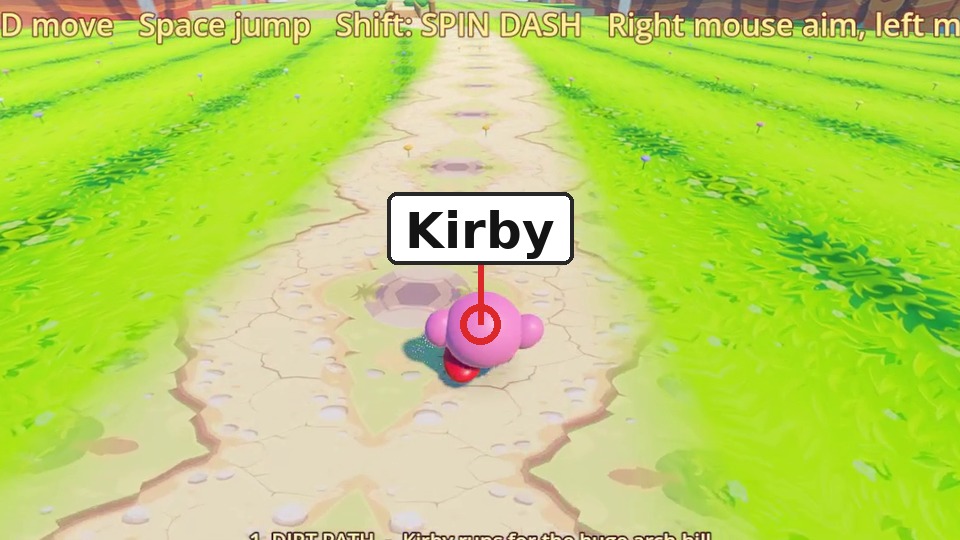}{Kirby on the path} & \pf{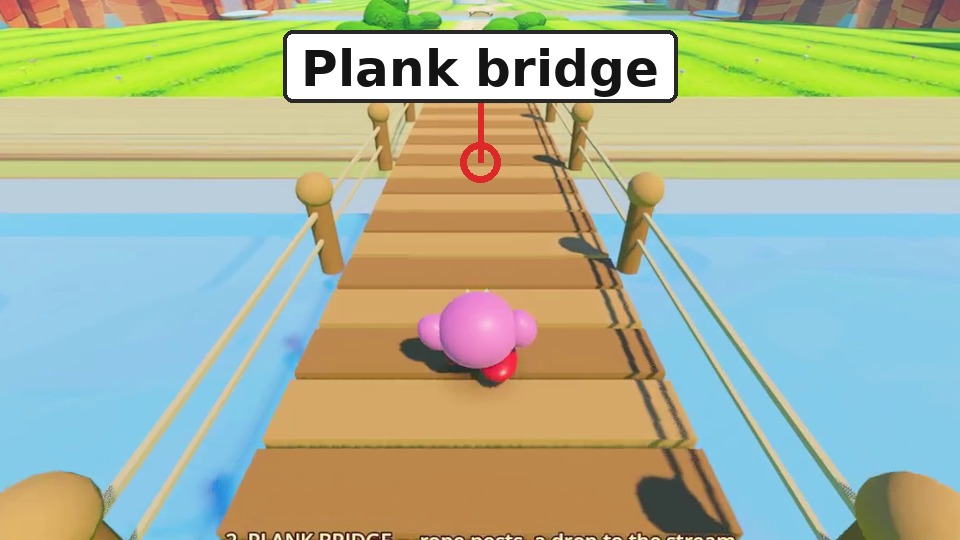}{Plank bridge} & \pf{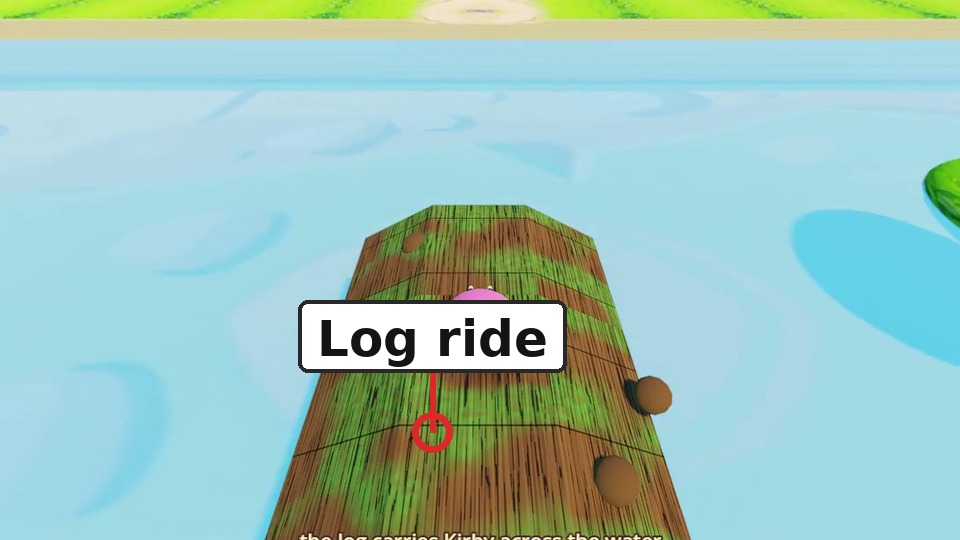}{Log ride} & \pf{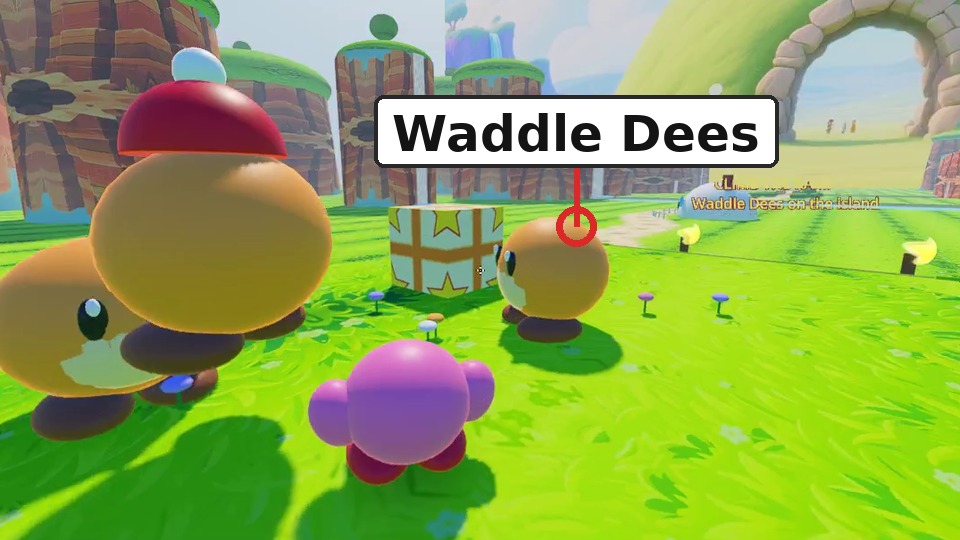}{Meadow and Waddle Dees}} \\[2pt]
\extrow{\multicolumn{5}{@{}l}{\scriptsize\textbf{City}: a golden-hour patrol from the ice rink to the skyline} \\*
\rlab{Reference}{New York city footage} & \rf{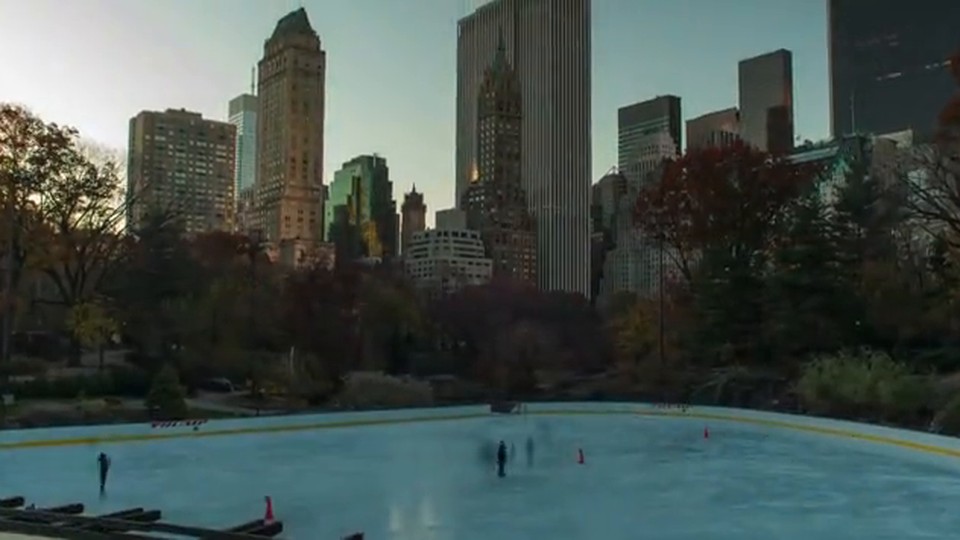} & \rf{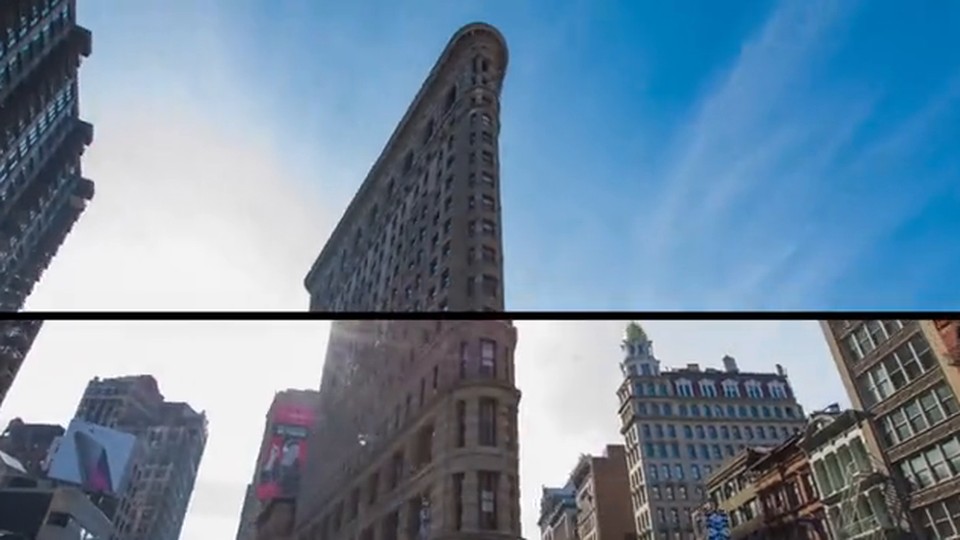} & \rf{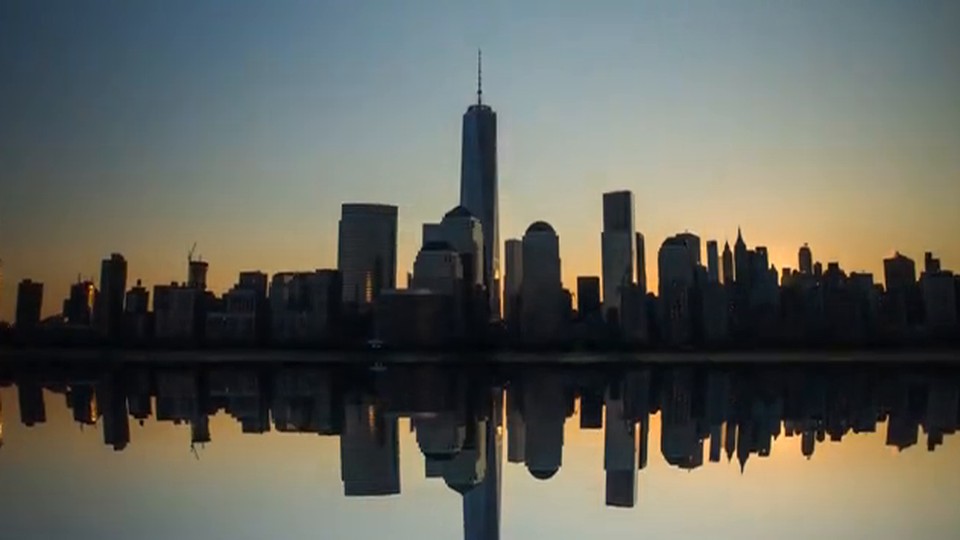} & \rf{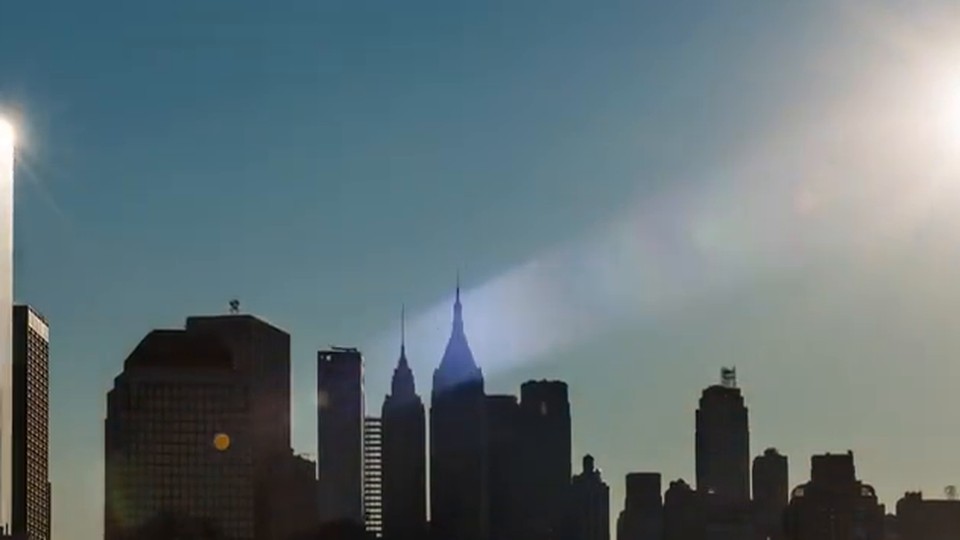} \\*
\rlab{In game}{} & \pf{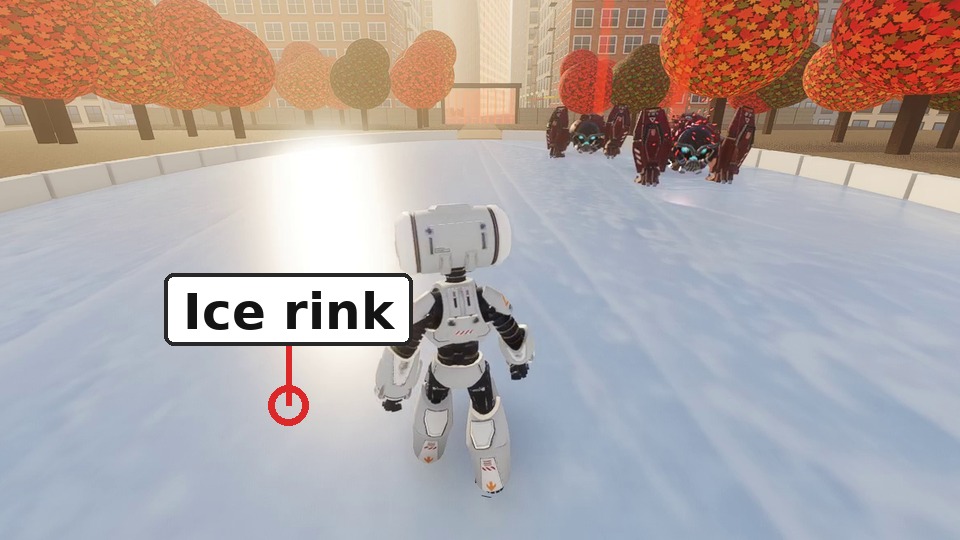}{Park ice rink} & \pf{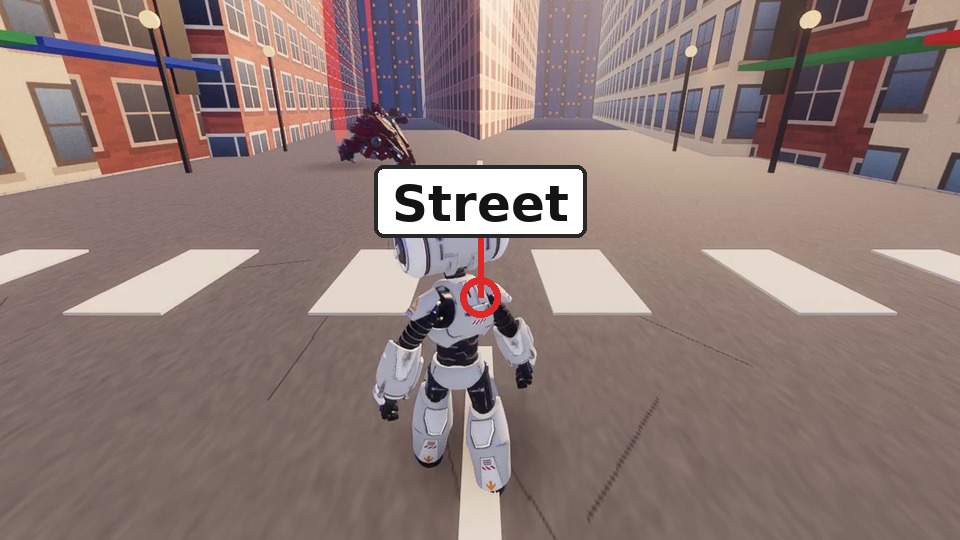}{Street canyon} & \pf{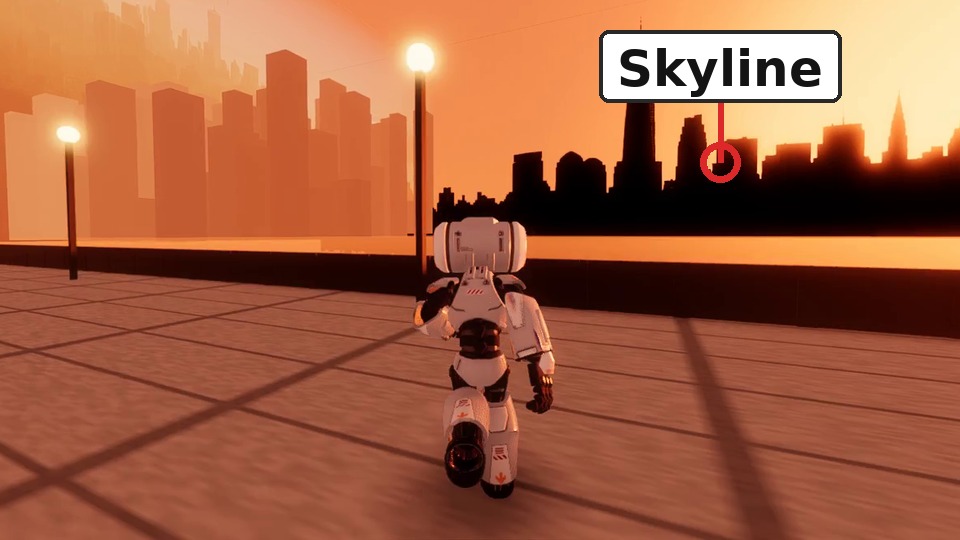}{Skyline at sunset} & \pf{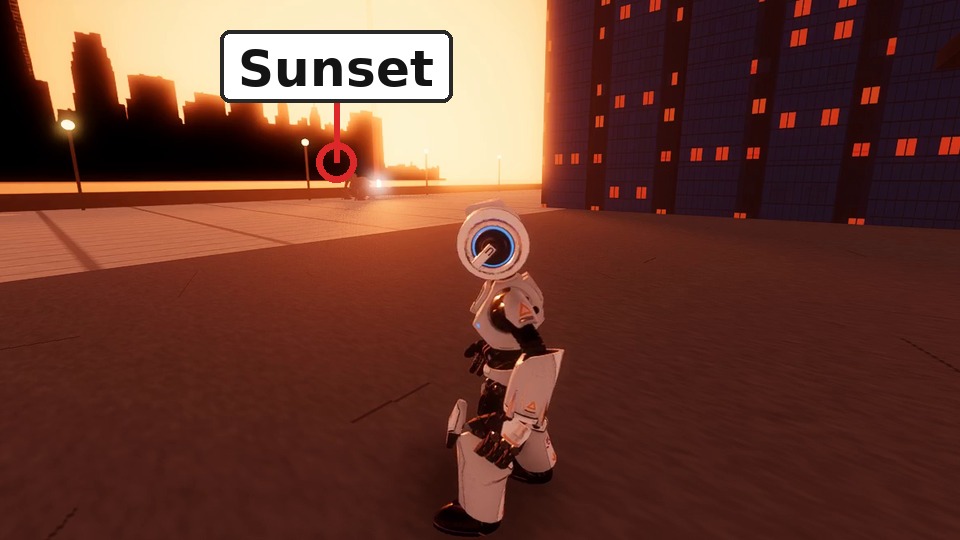}{Sun over the towers}} \\[4pt]
\midrule
\gthumb{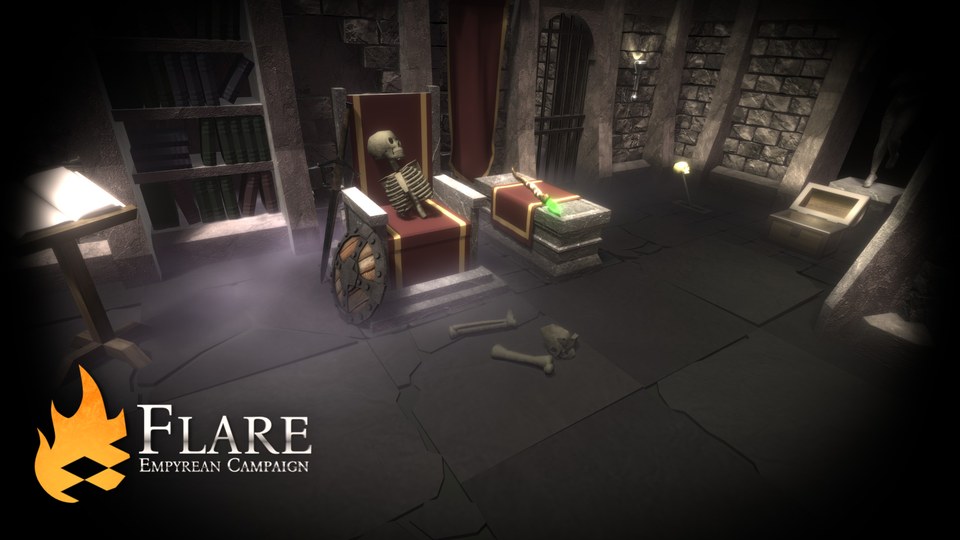} & \gname{Flare}{1.15 \textperiodcentered{} Flare engine \textperiodcentered{} GPL-3.0} & \gtag{Action RPG} & \vtag{2} & Explores isometric maps, fights monsters in real time, collects loot and follows quests. & INI-style content mods, sprites, maps & 5 \\*
\multicolumn{7}{@{}l@{}}{\rule{0pt}{12pt}\includegraphics[width=0.995\textwidth]{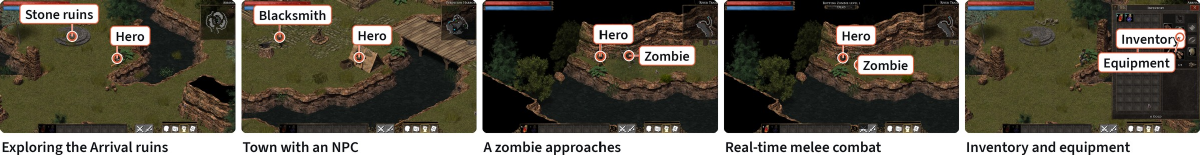}} \\[2pt]
\extlabel
\extrow{\multicolumn{5}{@{}l}{\scriptsize\textbf{Film}: an ice bridge forms and a snowflake spreads} \\*
\rlab{Reference}{Frozen, ``Let It Go''} & \rf{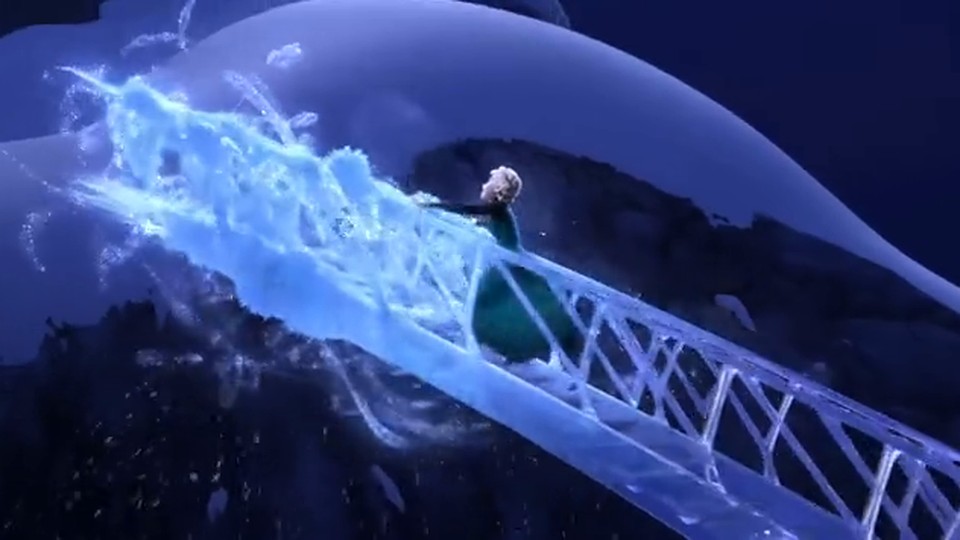} & \rf{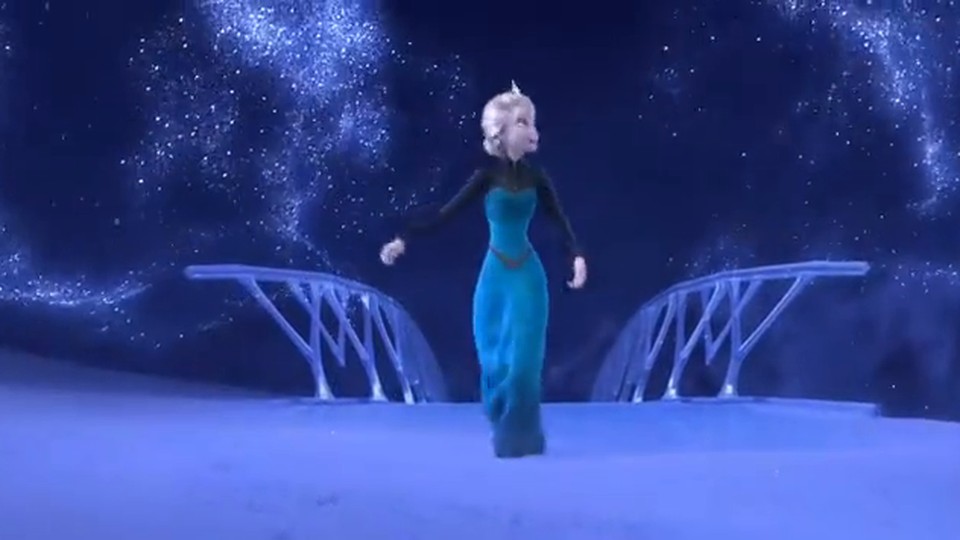} & \rf{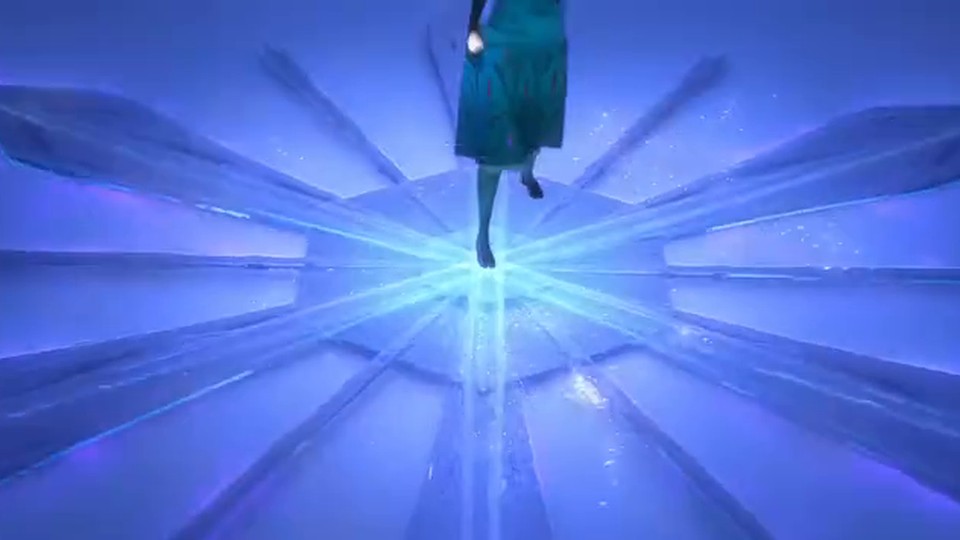} & \rf{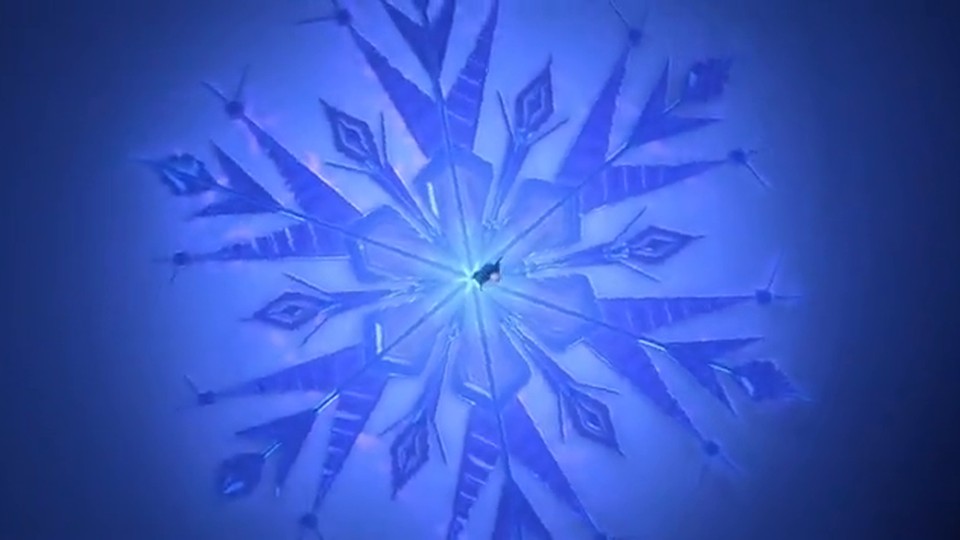} \\*
\rlab{In game}{} & \pf{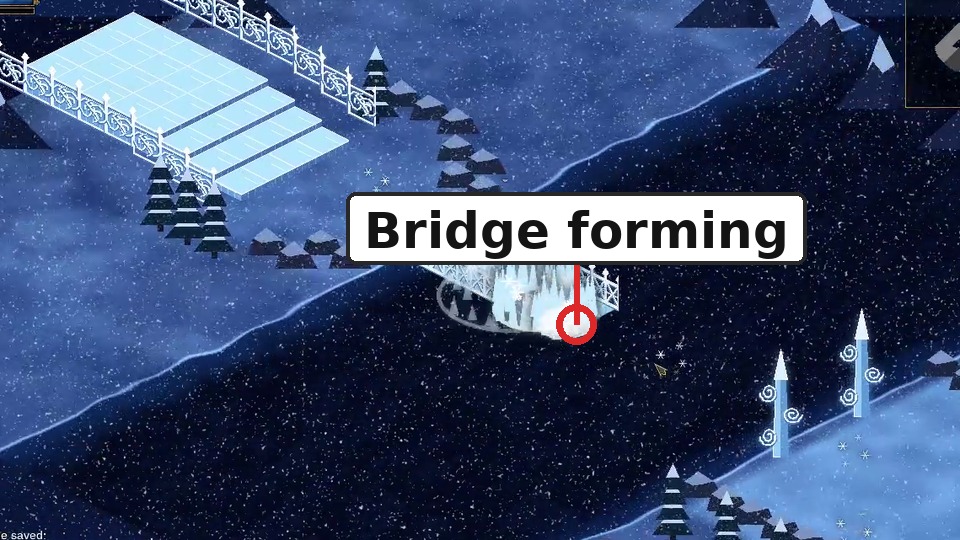}{Ice stairs grow} & \pf{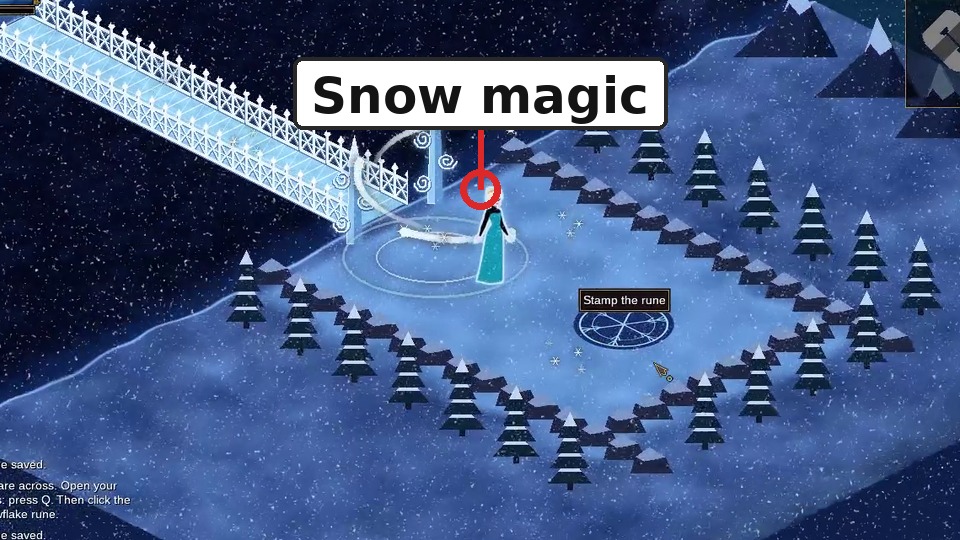}{Snow magic} & \pf{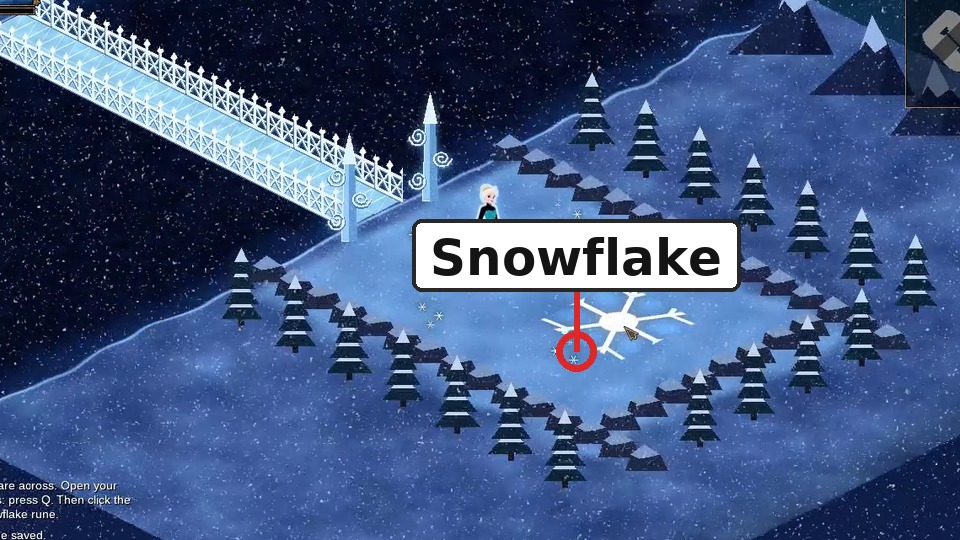}{Snowflake spreads} & \pf{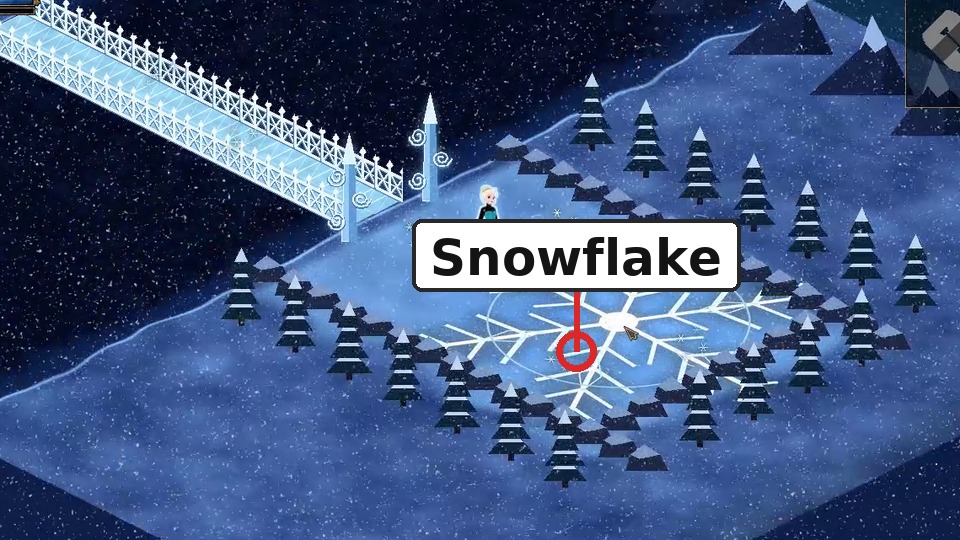}{Full snowflake}} \\[2pt]
\extrow{\multicolumn{5}{@{}l}{\scriptsize\textbf{Game trailer}: Kirby duels King Dedede in the meadow arena} \\*
\rlab{Reference}{Kirby trailer} & \rf{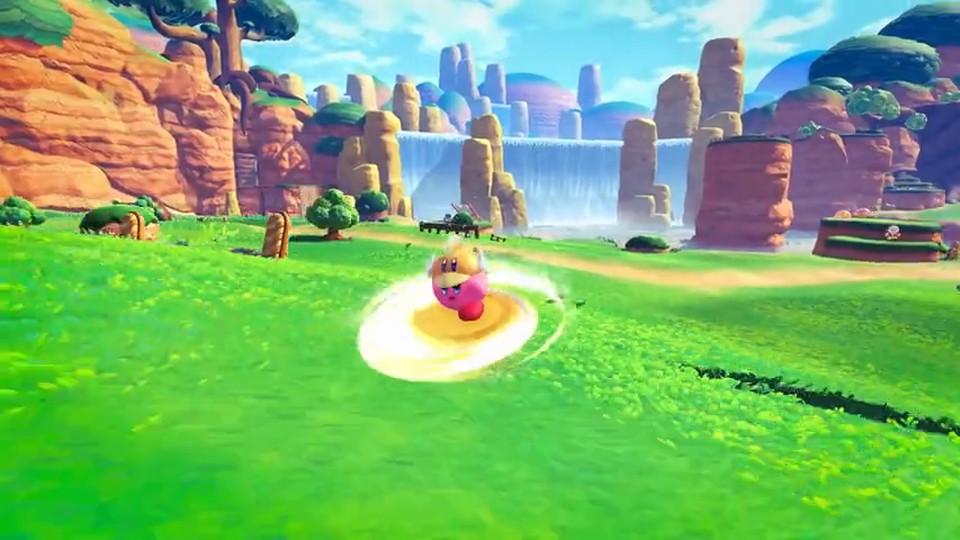} & \rf{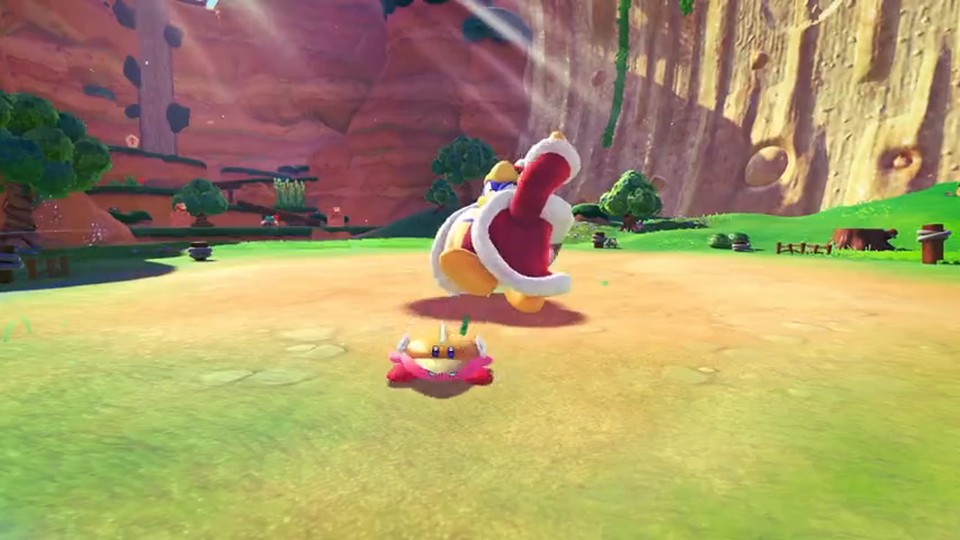} & \rf{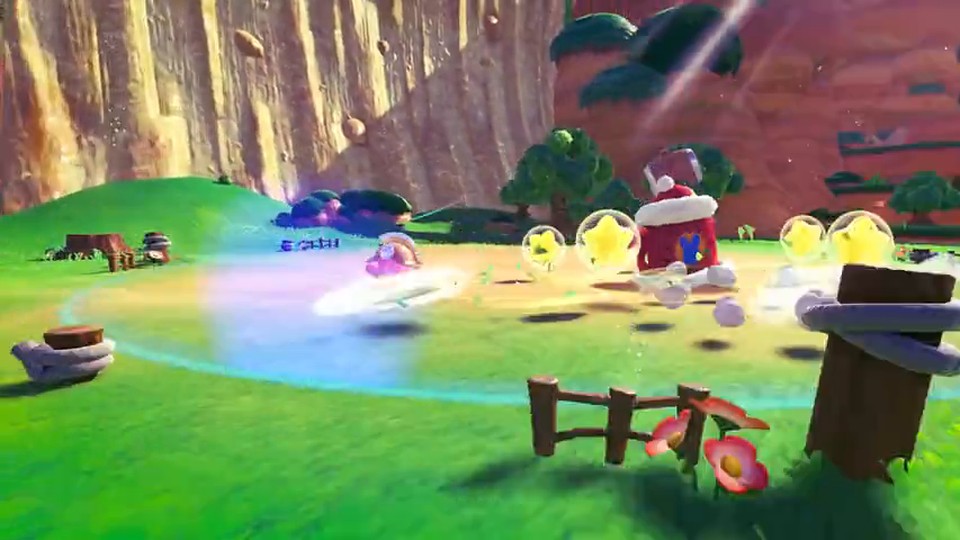} & \rf{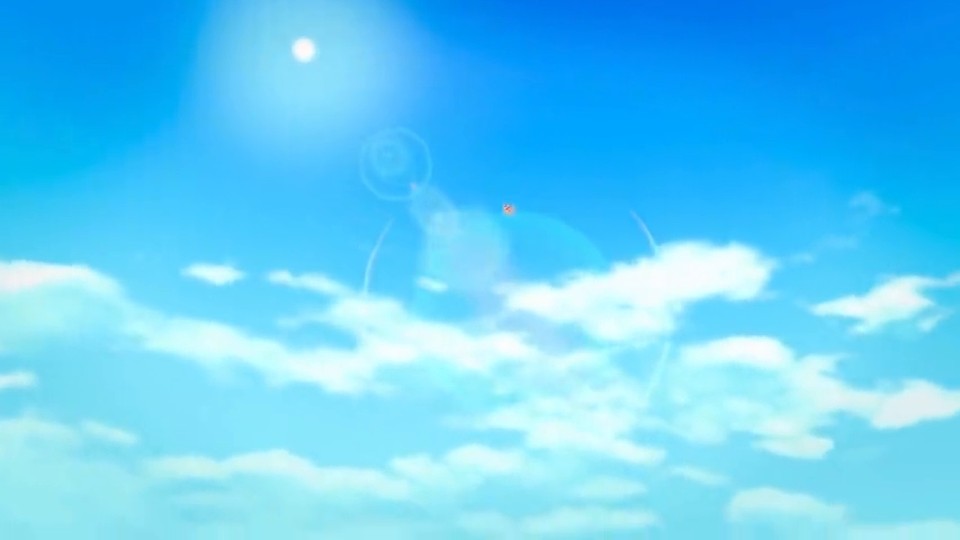} \\*
\rlab{In game}{} & \pf{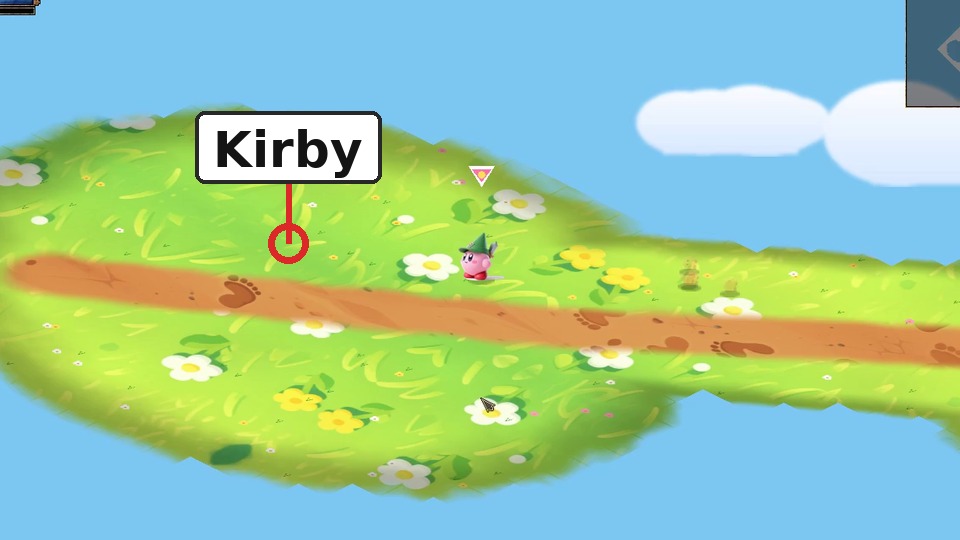}{Kirby on the meadow} & \pf{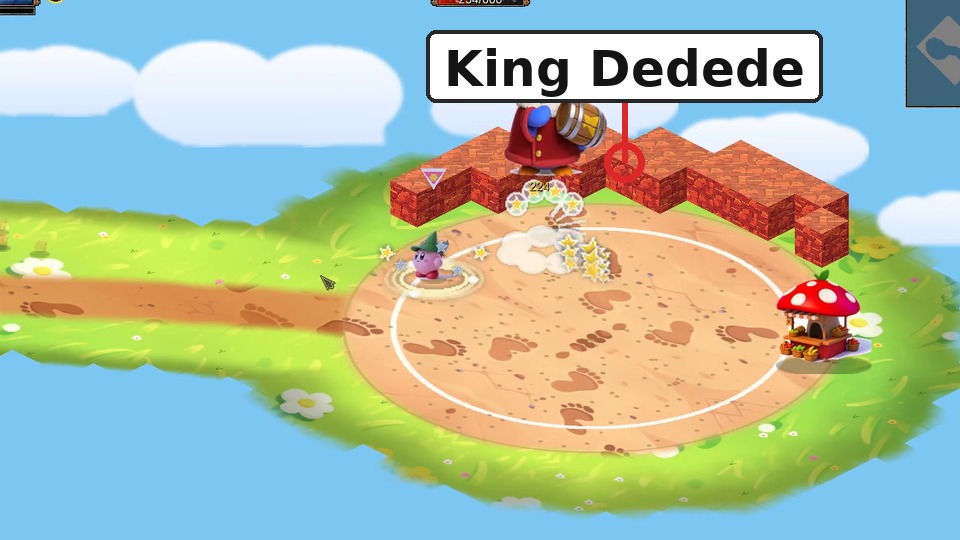}{King Dedede appears} & \pf{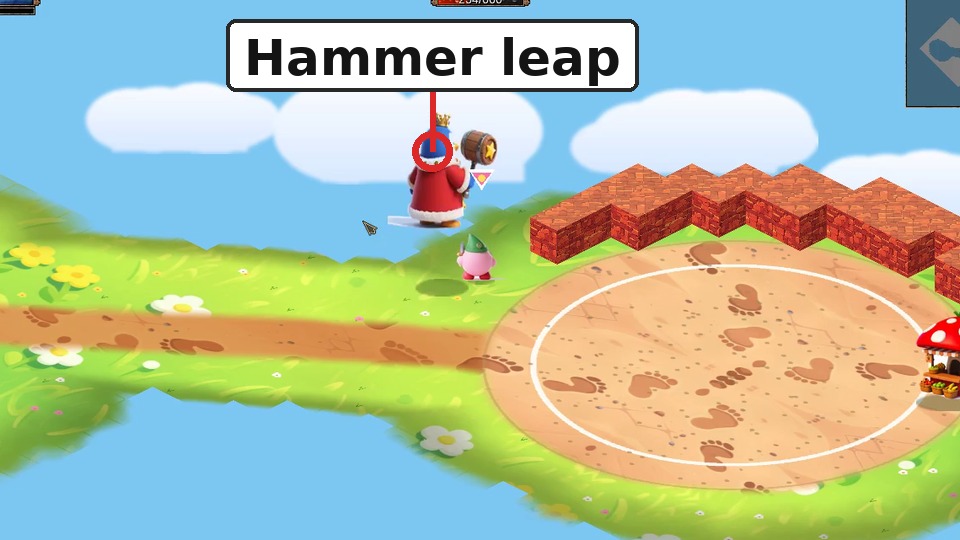}{Hammer leap} & \pf{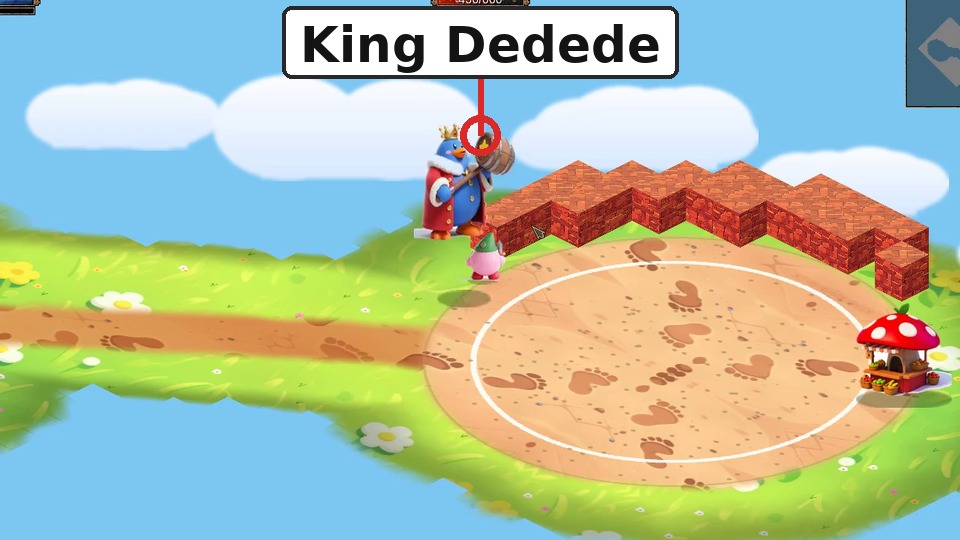}{Dedede sent flying}} \\[4pt]
\midrule
\gthumb{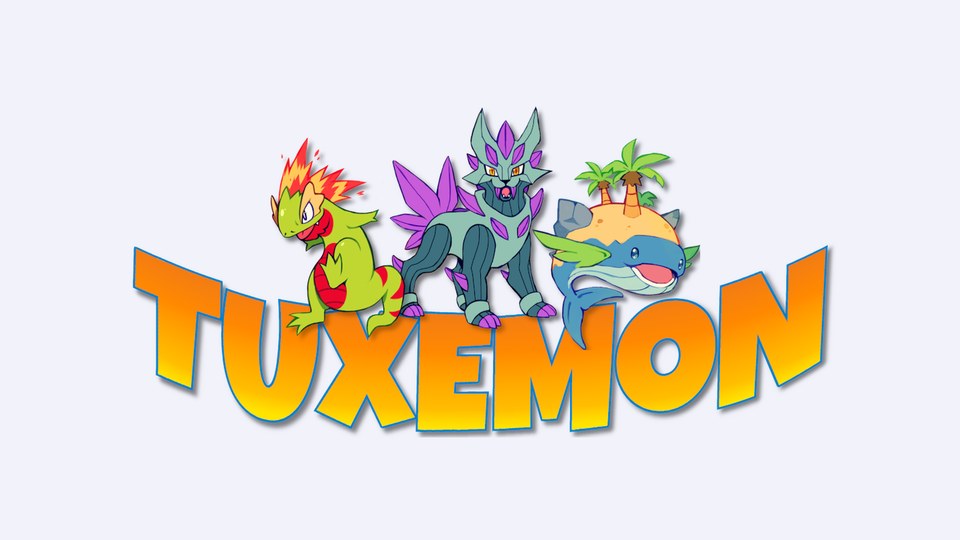} & \gname{Tuxemon}{0.4.35 \textperiodcentered{} Python / pygame \textperiodcentered{} GPL-3.0} & \gtag{Creature RPG} & \vtag{2} & Walks towns and routes, talks to trainers, catches monsters and fights turn-based battles. & Mod folder: sprites, \mf{.tmx} maps, event scripts & 13 \\*
\multicolumn{7}{@{}l@{}}{\rule{0pt}{12pt}\includegraphics[width=0.995\textwidth]{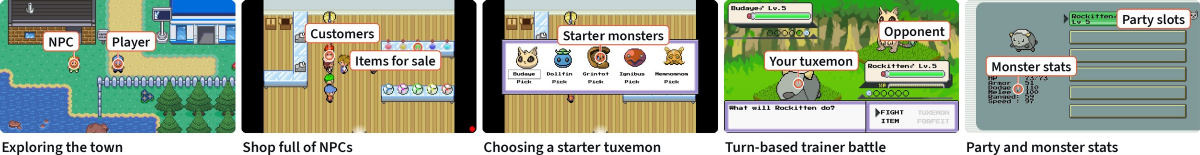}} \\[2pt]
\extlabel
\extrow{\multicolumn{5}{@{}l}{\scriptsize\textbf{City}: a Manhattan walking tour of landmarks} \\*
\rlab{Reference}{New York city footage} & \rf{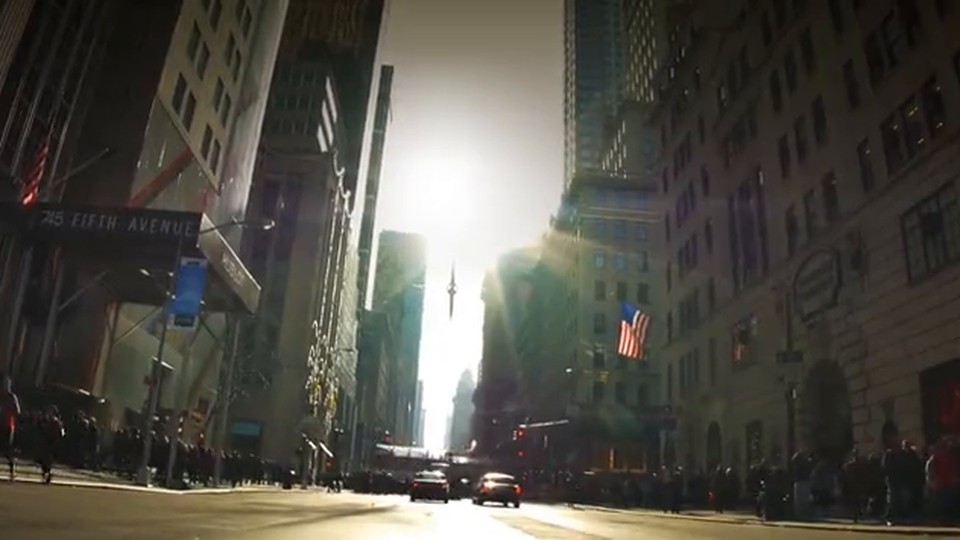} & \rf{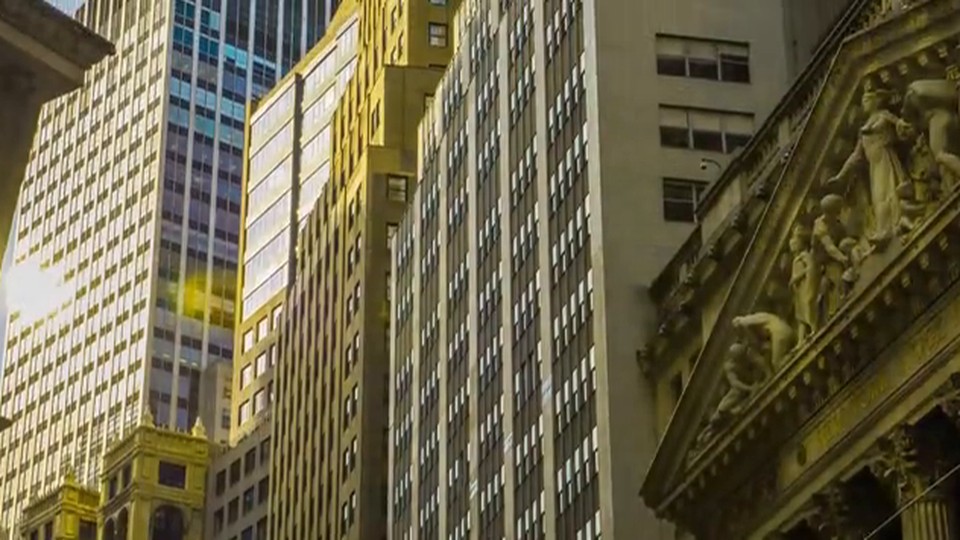} & \rf{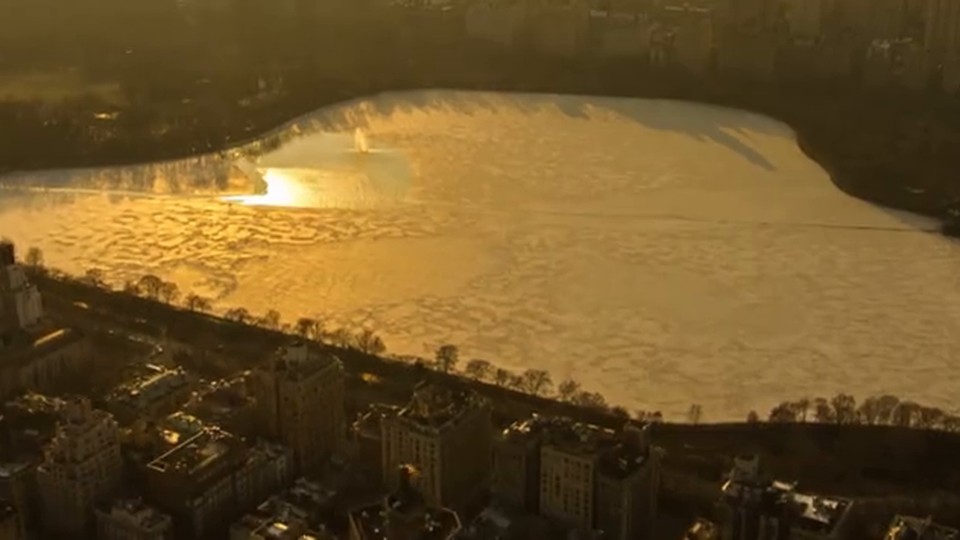} & \rf{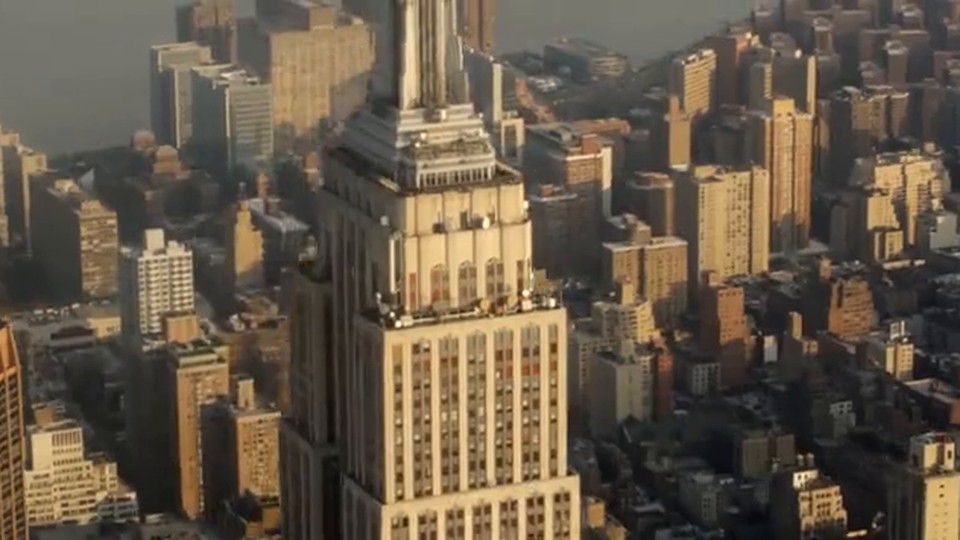} \\*
\rlab{In game}{} & \pf{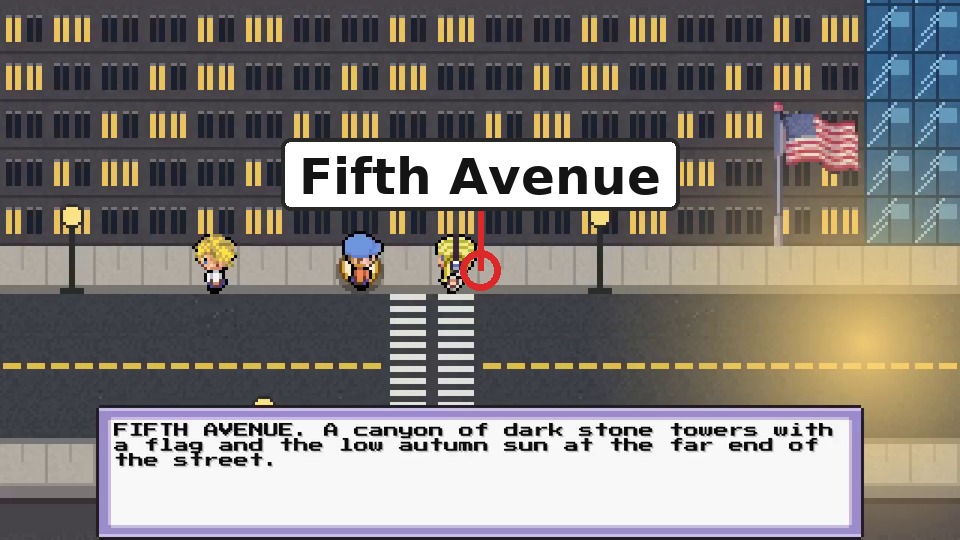}{Fifth Avenue} & \pf{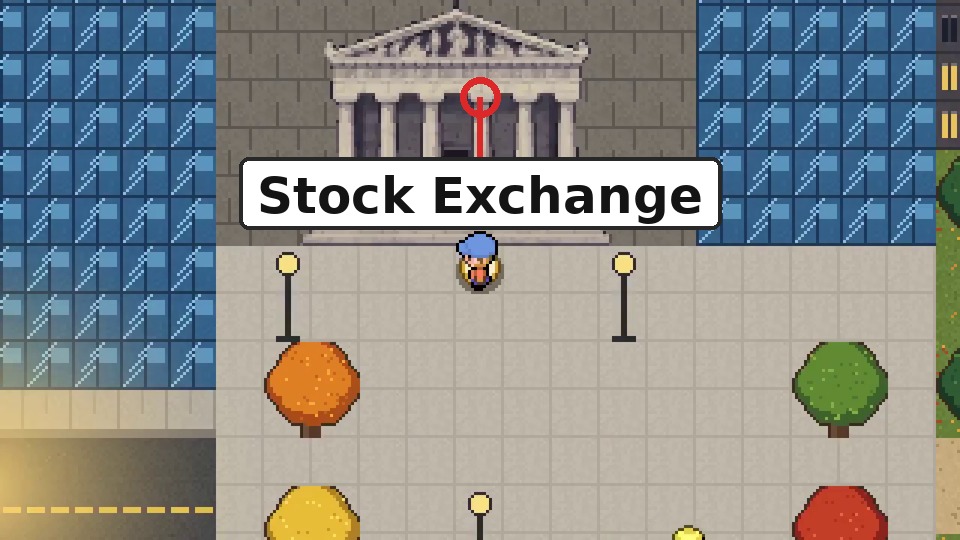}{Stock Exchange} & \pf{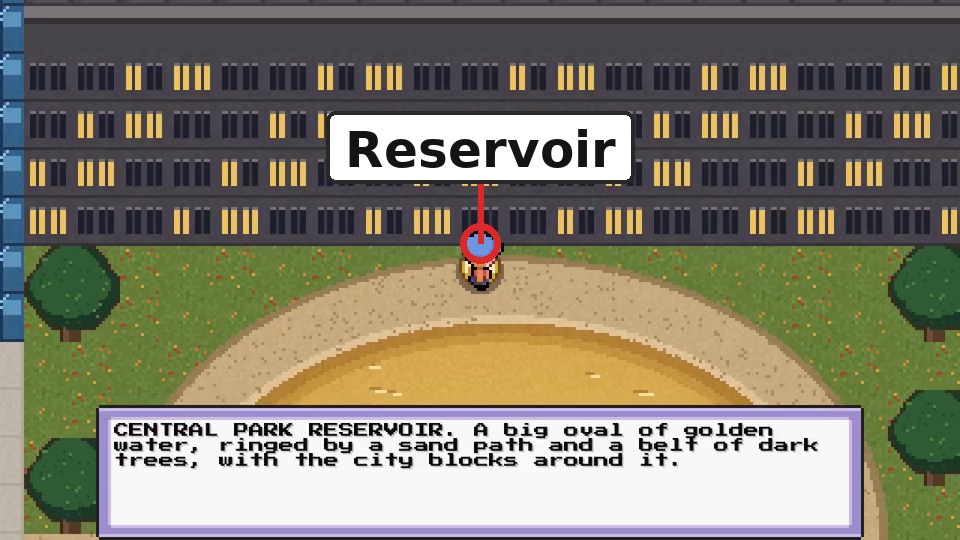}{Central Park reservoir} & \pf{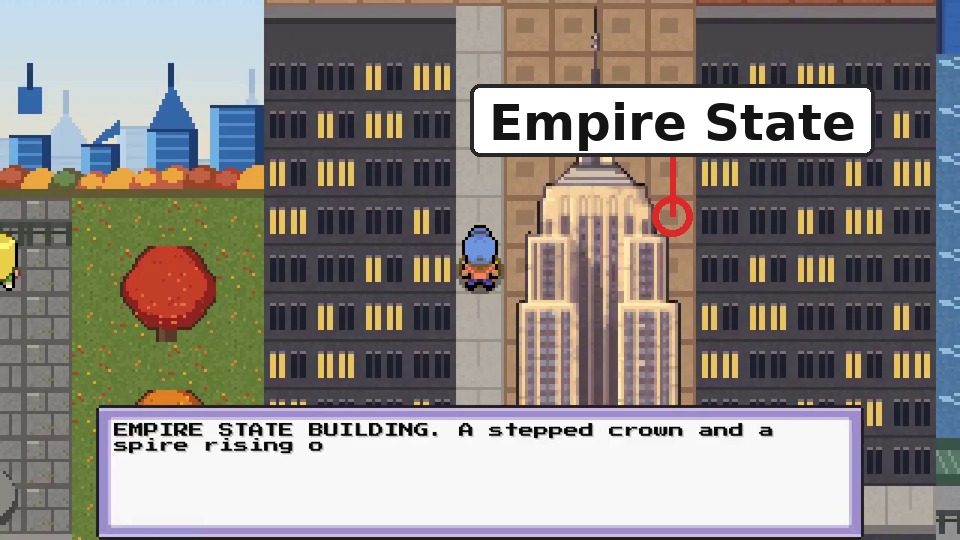}{Empire State Building}} \\[2pt]
\extrow{\multicolumn{5}{@{}l}{\scriptsize\textbf{Game trailer}: jellyfishing, the coaster, and the stage} \\*
\rlab{Reference}{SpongeBob x Minecraft trailer} & \rf{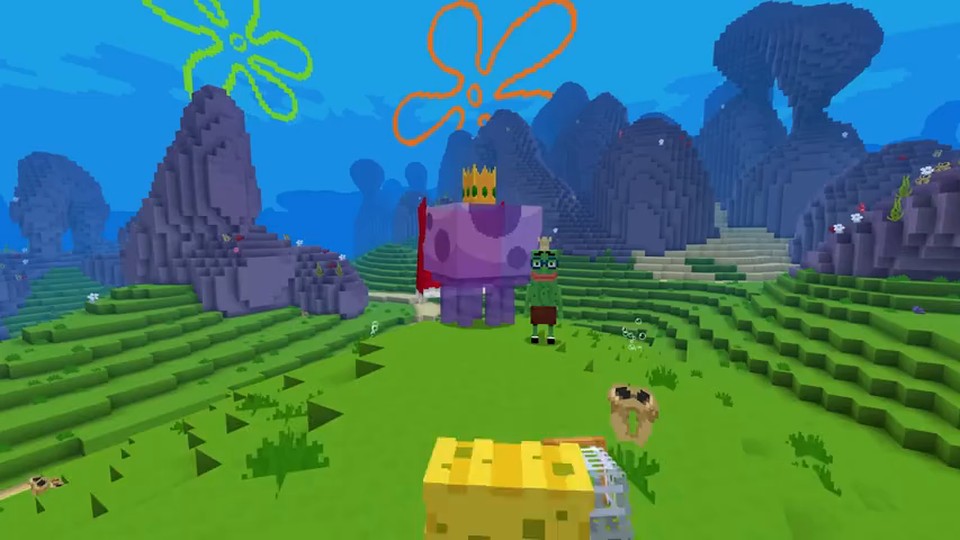} & \rf{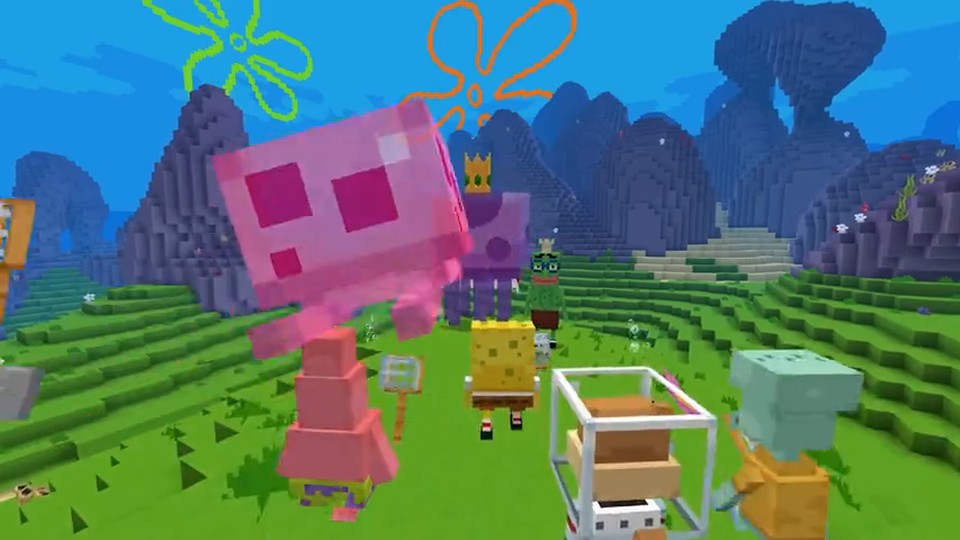} & \rf{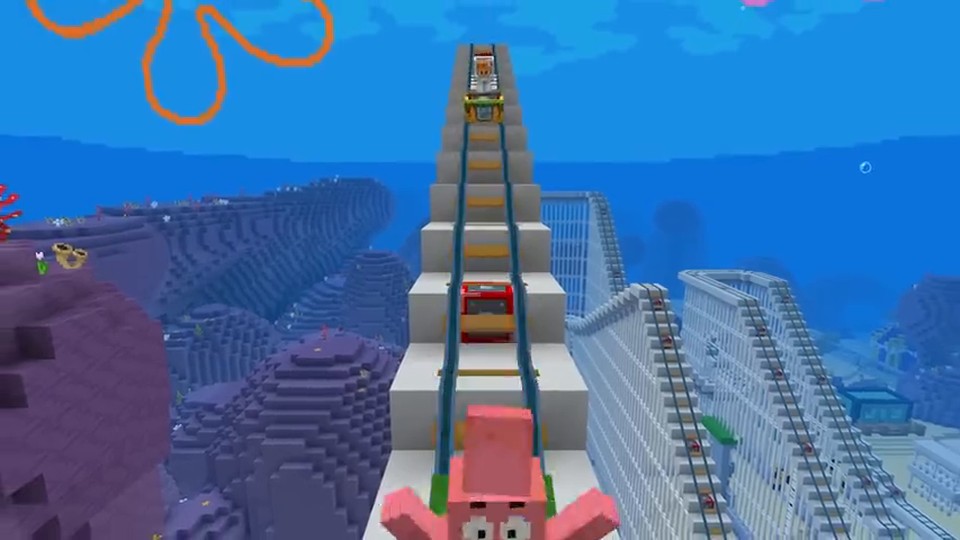} & \rf{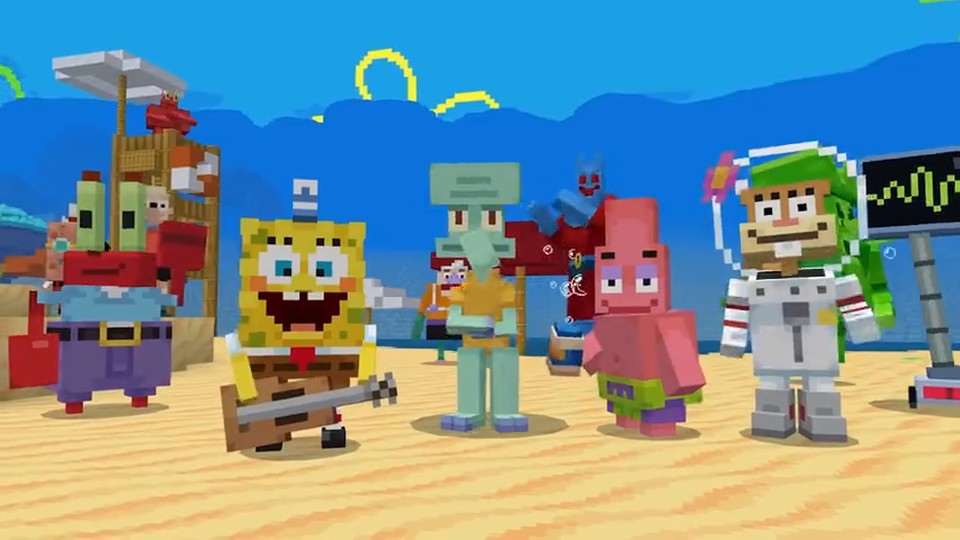} \\*
\rlab{In game}{} & \pf{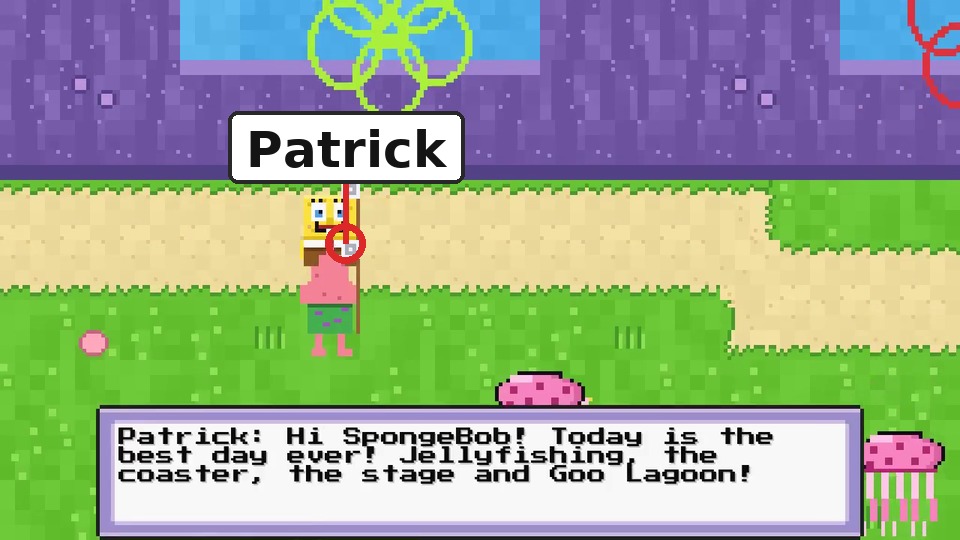}{Jellyfish fields} & \pf{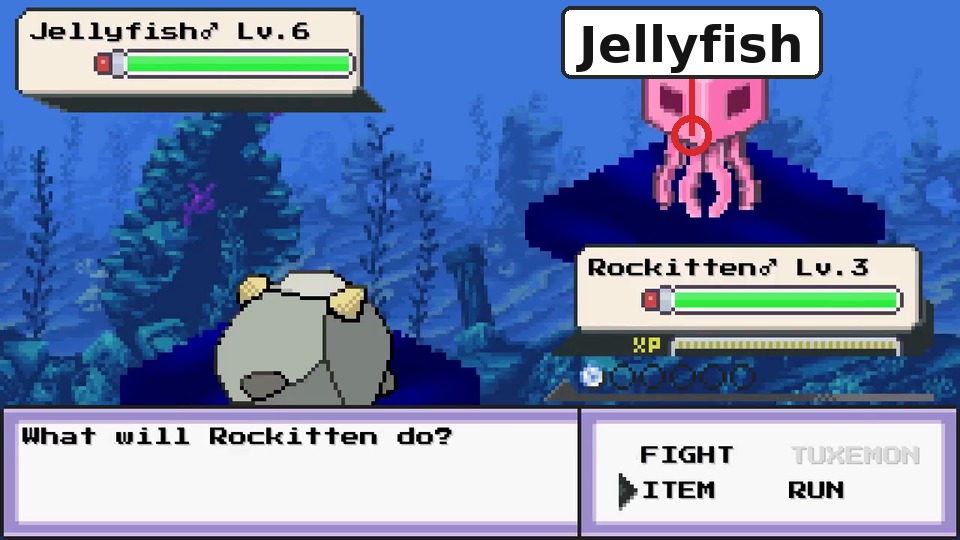}{Catching a jellyfish} & \pf{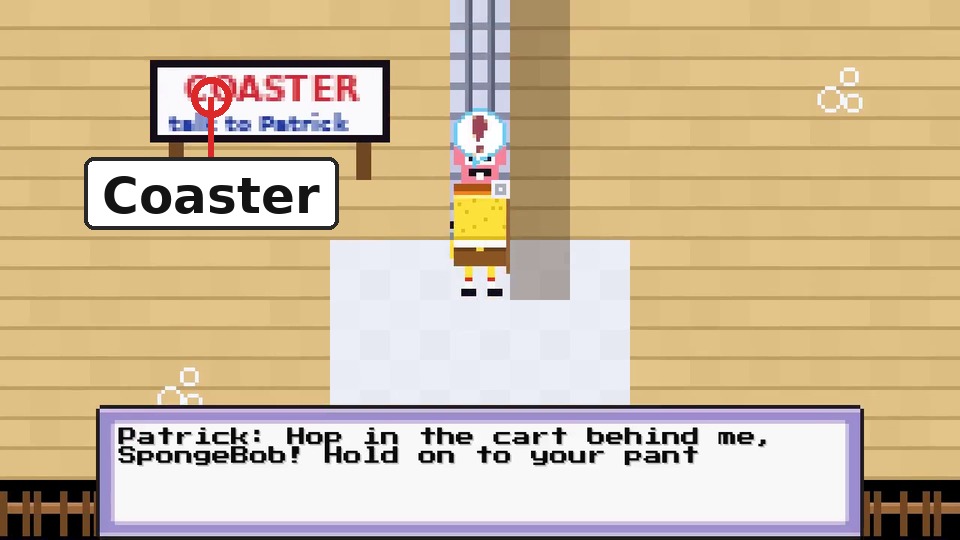}{Coaster} & \pf{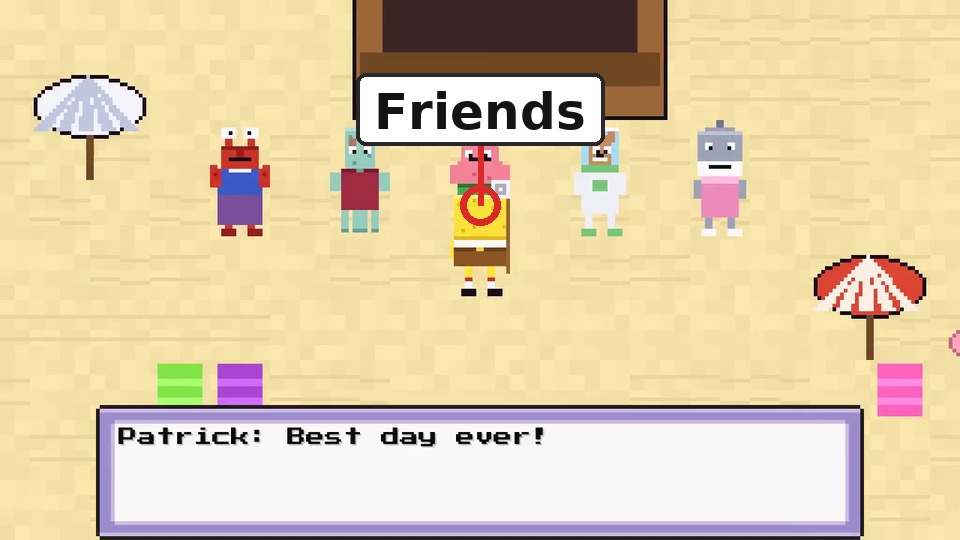}{Friends at Goo Lagoon}} \\[4pt]
\midrule
\gthumb{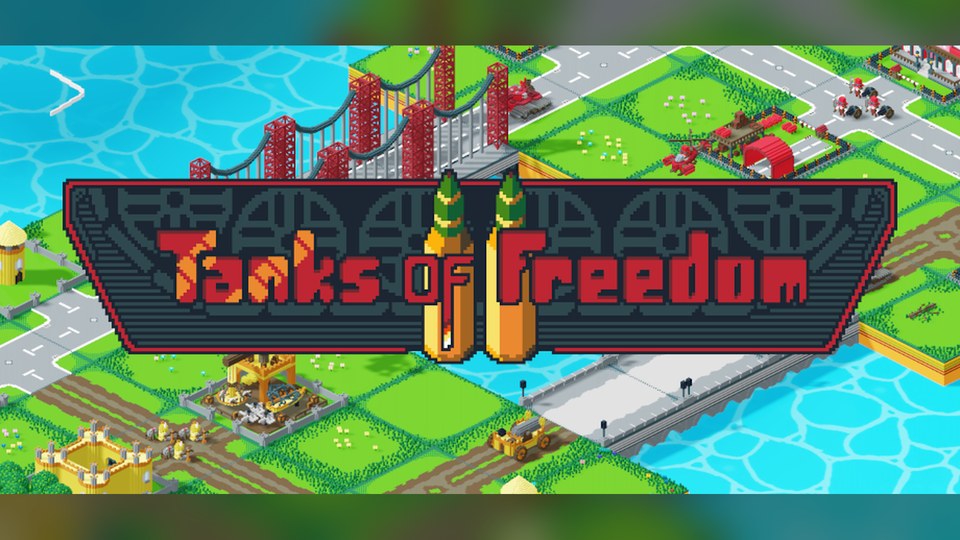} & \gname{Tanks of Freedom II}{1.0.9 \textperiodcentered{} Godot 4.4 \textperiodcentered{} MIT} & \gtag{Turn-based strategy} & \vtag{3} & Commands tanks, infantry and buildings on a voxel map turn by turn to defeat rival armies. & \mf{.json} maps, campaigns, story scripts & 4 \\*
\multicolumn{7}{@{}l@{}}{\rule{0pt}{12pt}\includegraphics[width=0.995\textwidth]{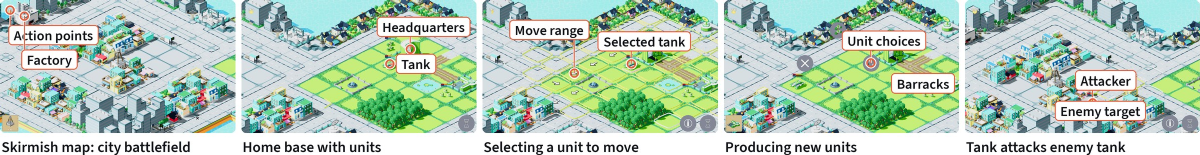}} \\[2pt]
\extlabel
\extrow{\multicolumn{5}{@{}l}{\scriptsize\textbf{City}: stamp the London landmarks in order} \\*
\rlab{Reference}{London city footage} & \rf{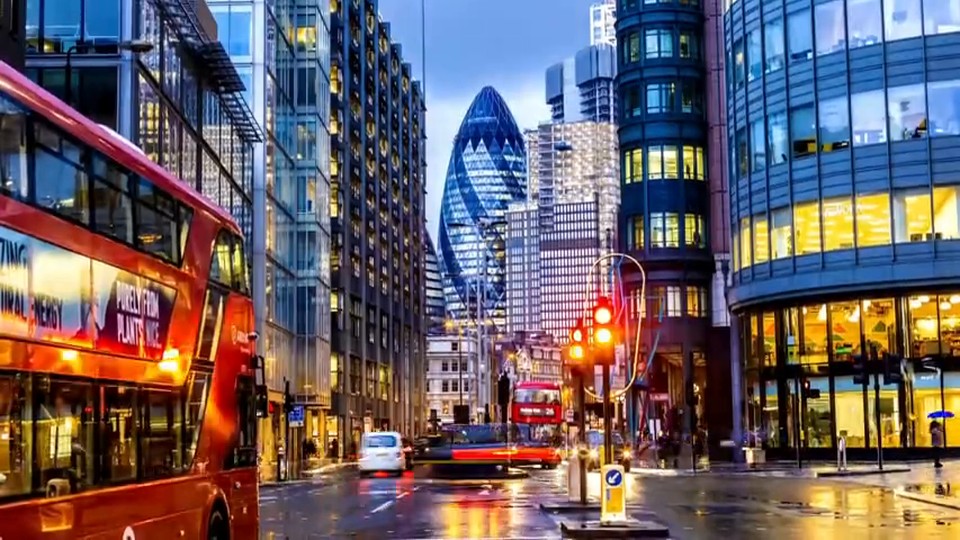} & \rf{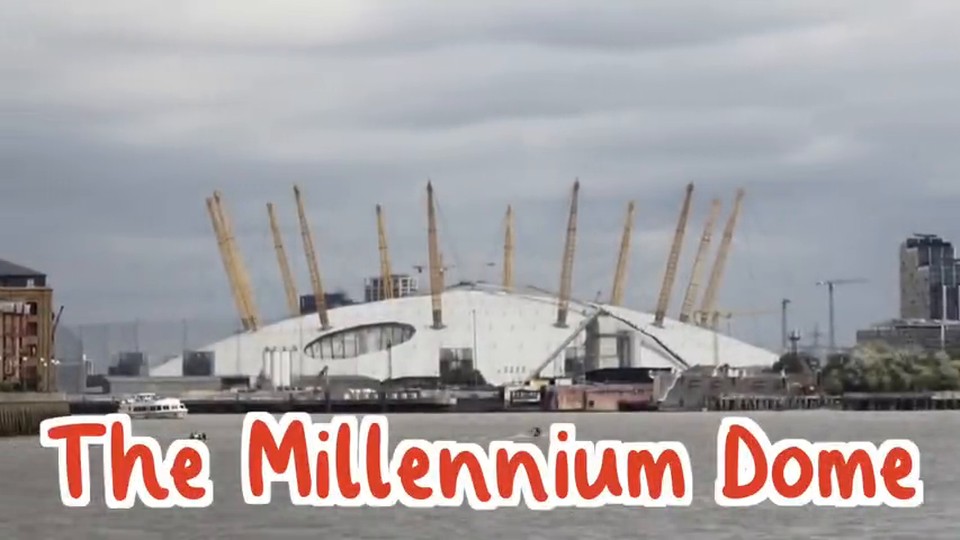} & \rf{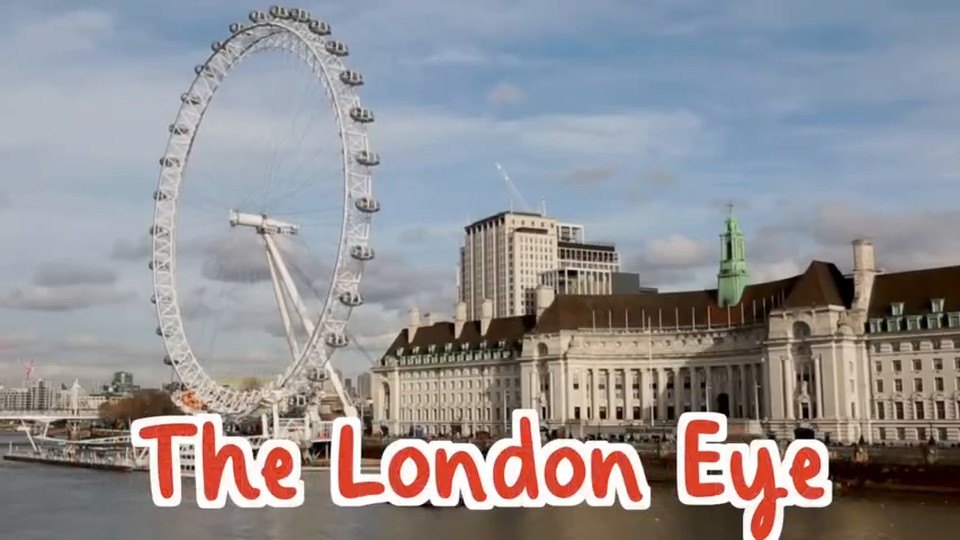} & \rf{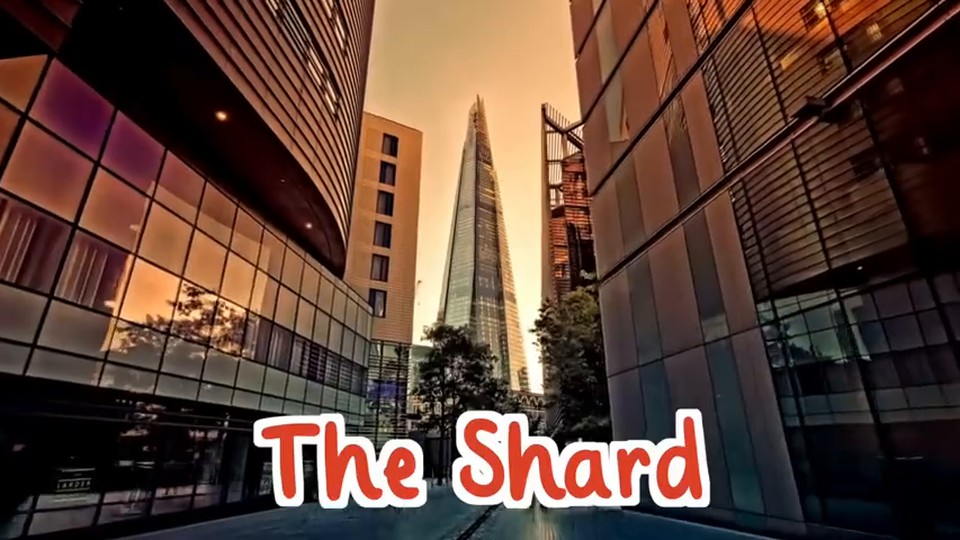} \\*
\rlab{In game}{} & \pf{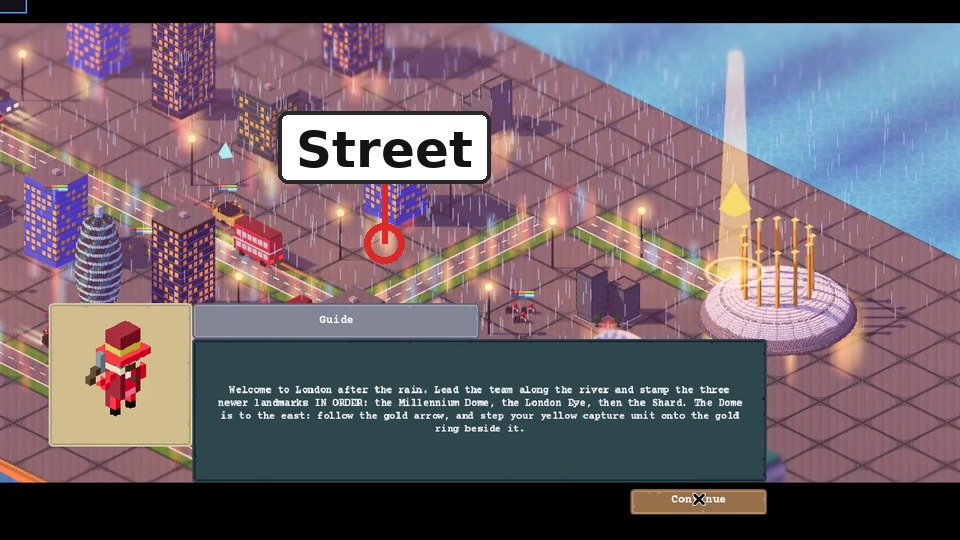}{Rainy street} & \pf{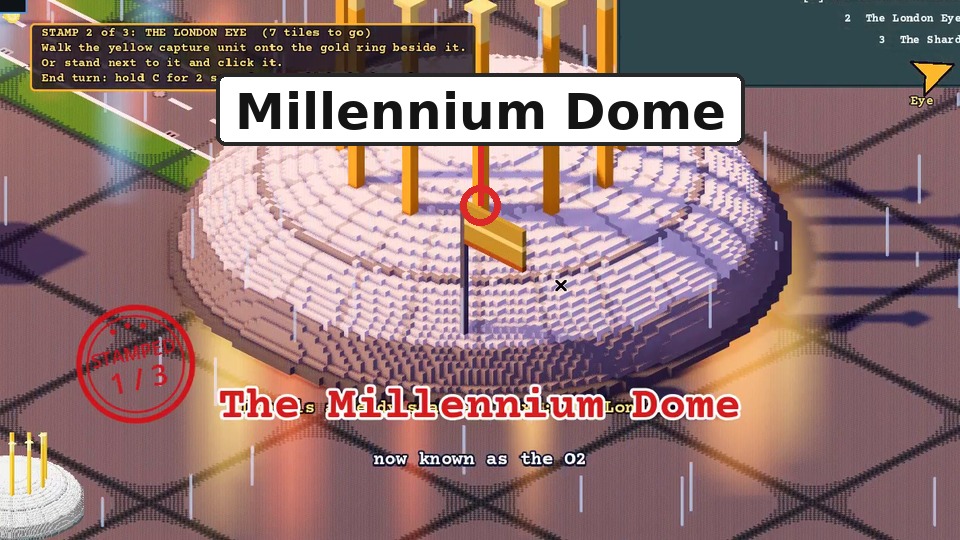}{Millennium Dome} & \pf{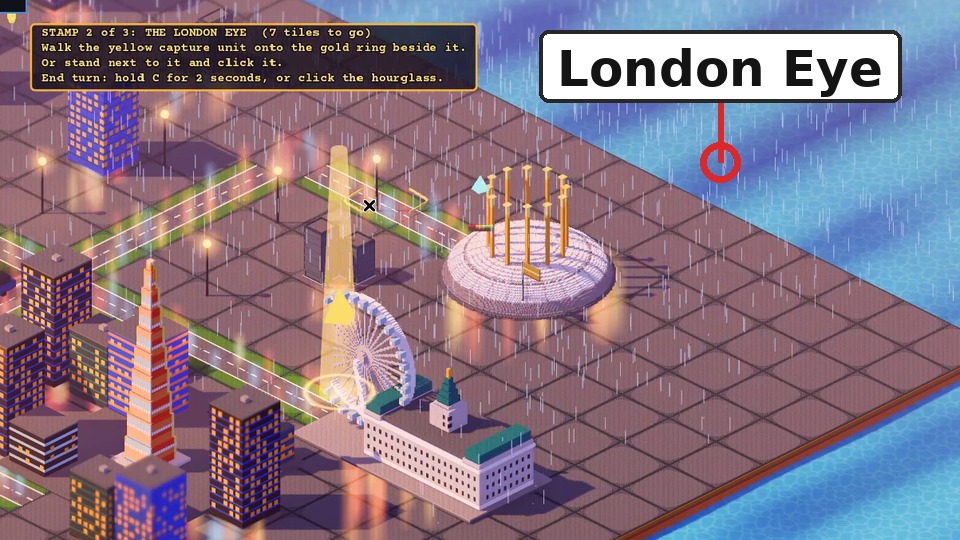}{London Eye} & \pf{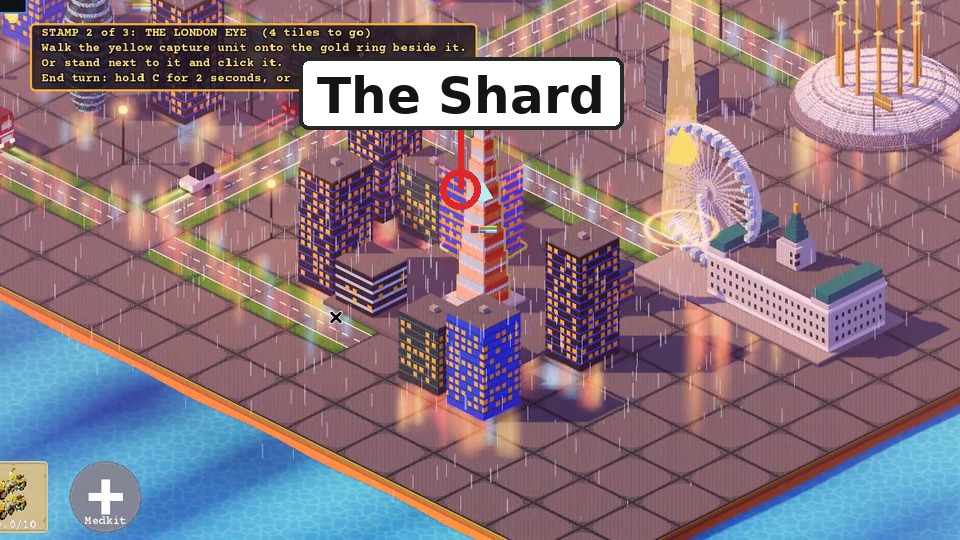}{The Shard}} \\[2pt]
\extrow{\multicolumn{5}{@{}l}{\scriptsize\textbf{Game trailer}: ranger, raptors, and the escape past the predator} \\*
\rlab{Reference}{Jurassic World x Minecraft trailer} & \rf{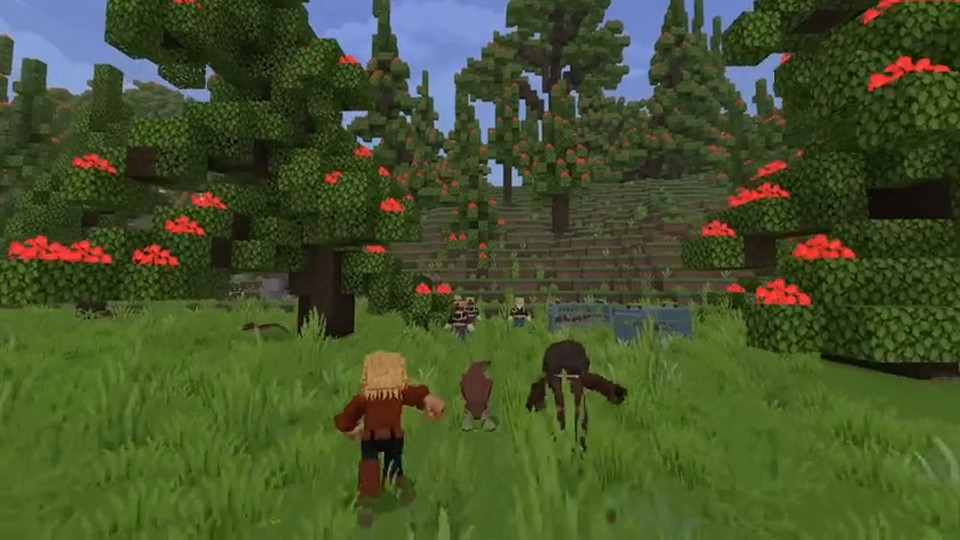} & \rf{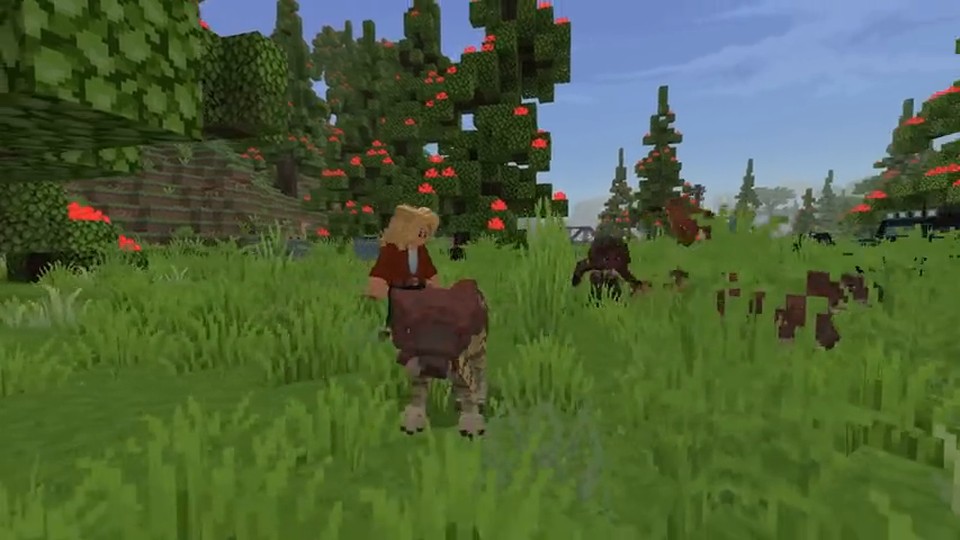} & \rf{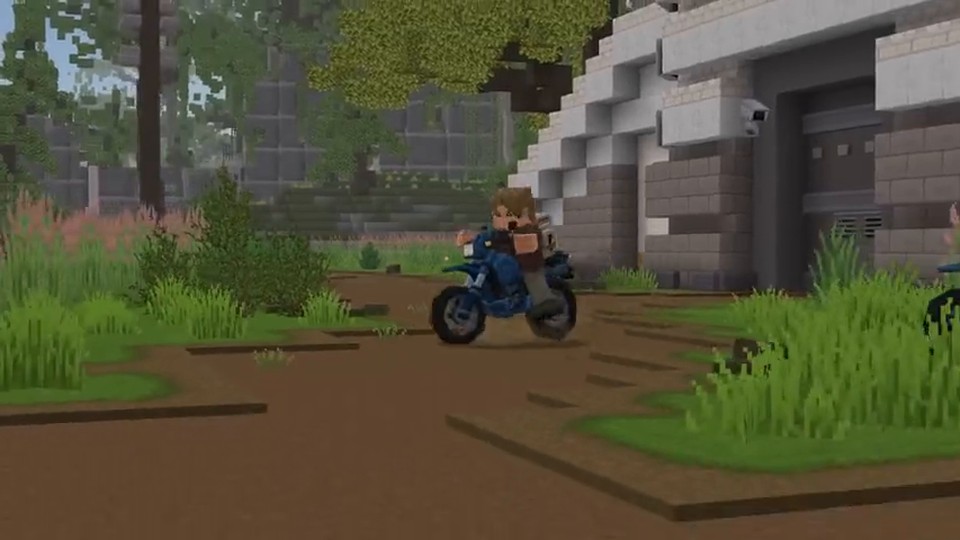} & \rf{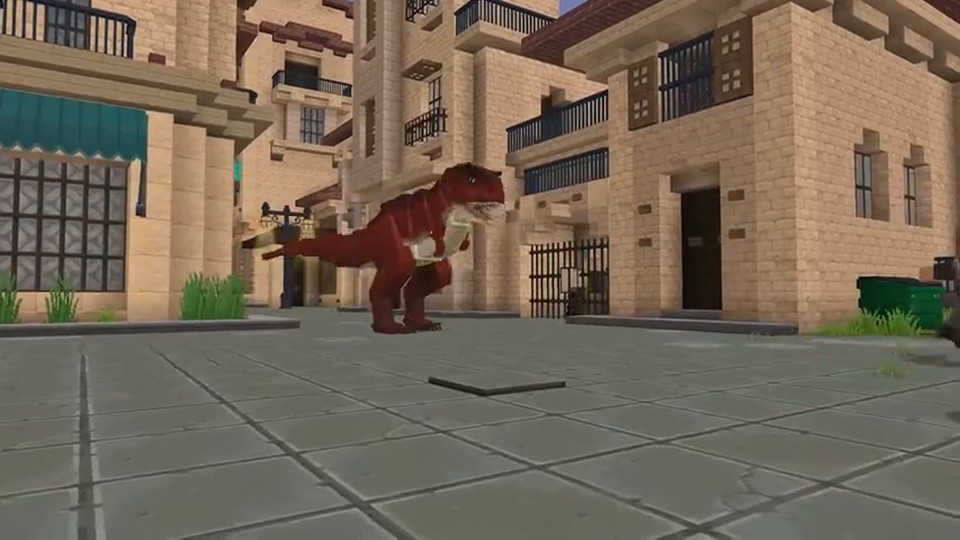} \\*
\rlab{In game}{} & \pf{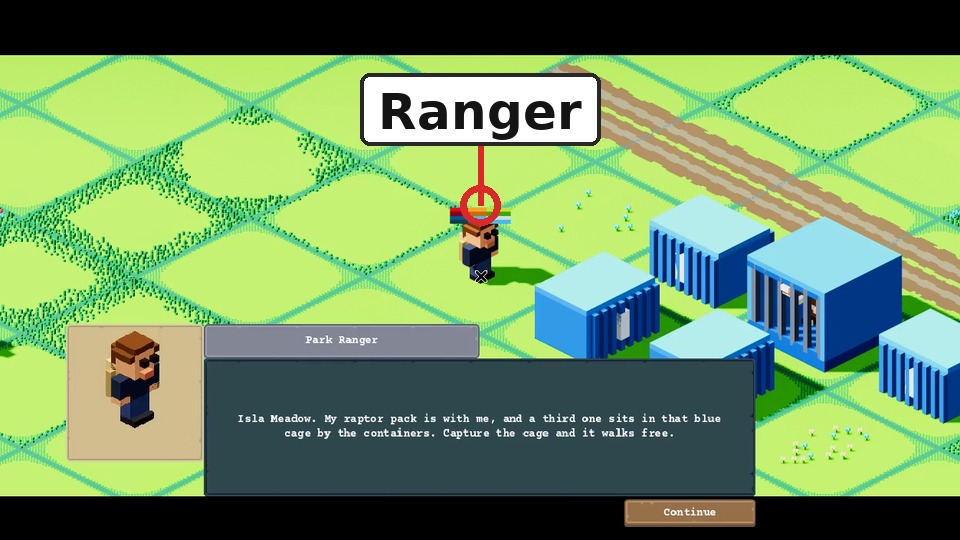}{Ranger in the park} & \pf{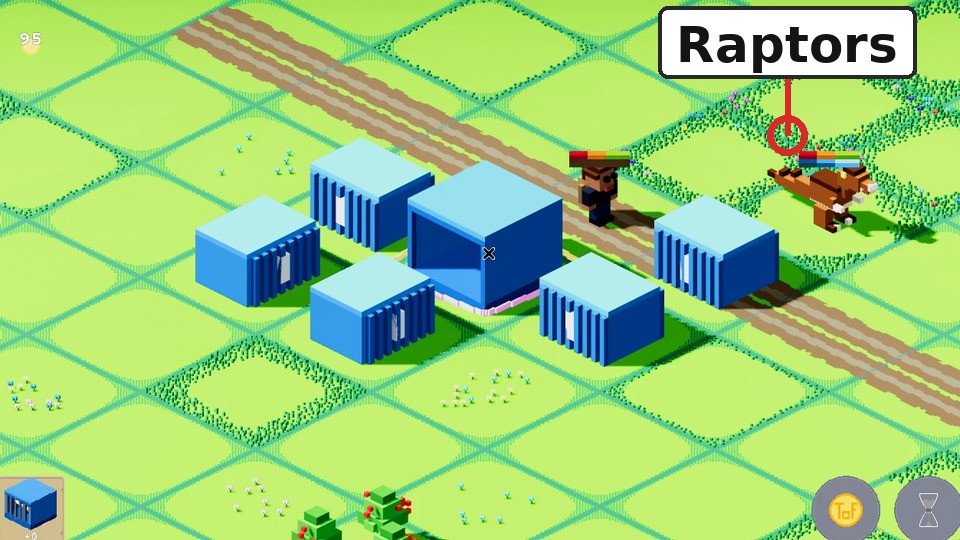}{Raptor pack} & \pf{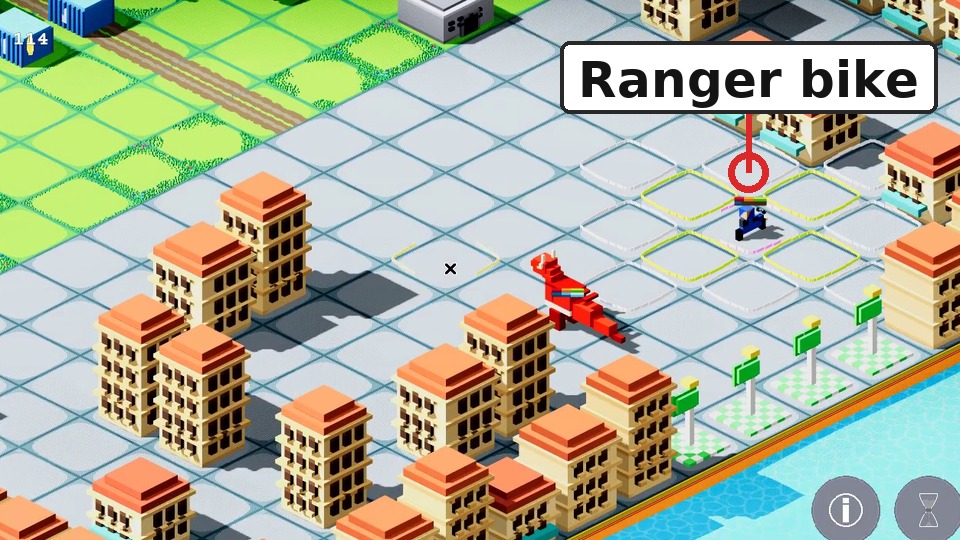}{Bike past the predator} & \pf{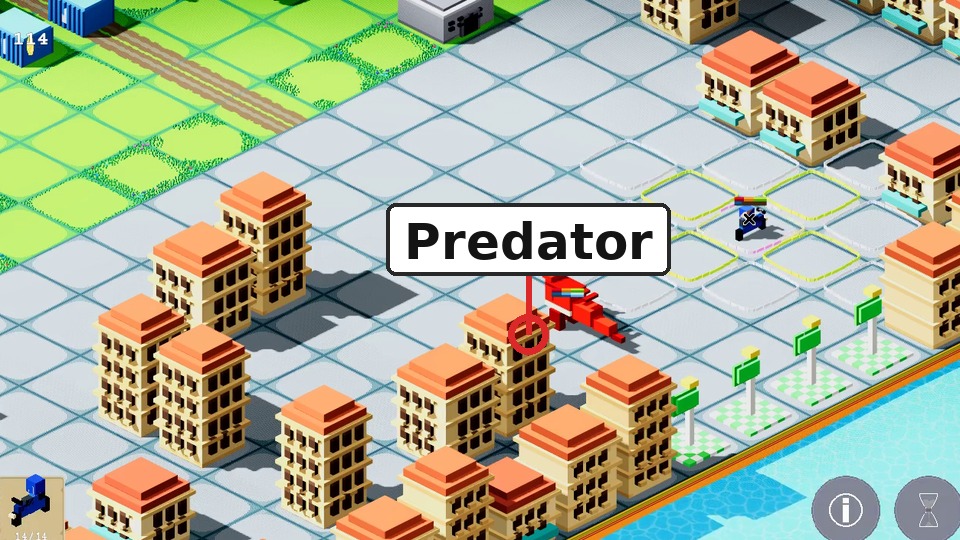}{Predator in town}} \\[4pt]
\midrule
\gthumb{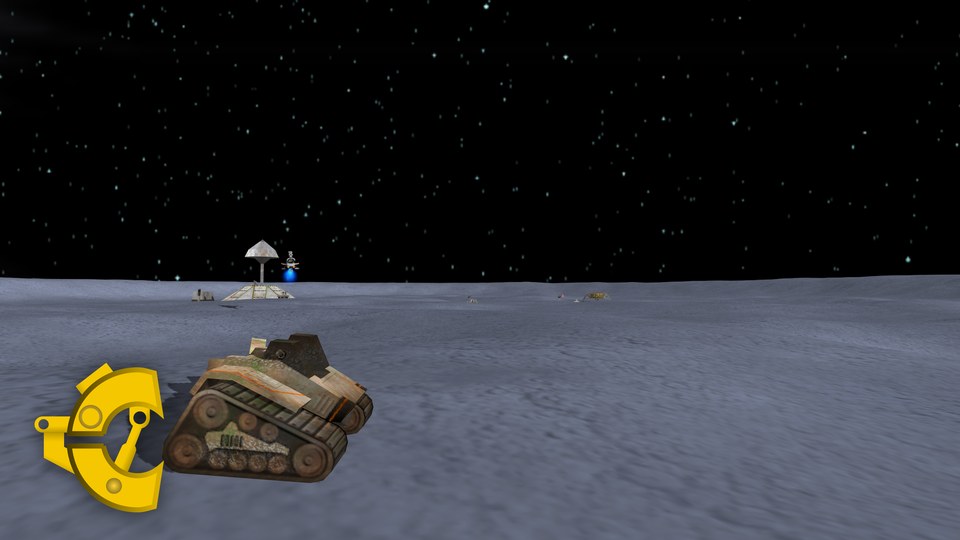} & \gname{Colobot: Gold}{0.2.1 \textperiodcentered{} own engine \textperiodcentered{} GPL-3.0} & \gtag{Real-time strategy} & \vtag{3} & Builds a base on an alien planet and programs robots to mine, carry and defend. & Level files, CBOT robot scripts & 3 \\*
\multicolumn{7}{@{}l@{}}{\rule{0pt}{12pt}\includegraphics[width=0.995\textwidth]{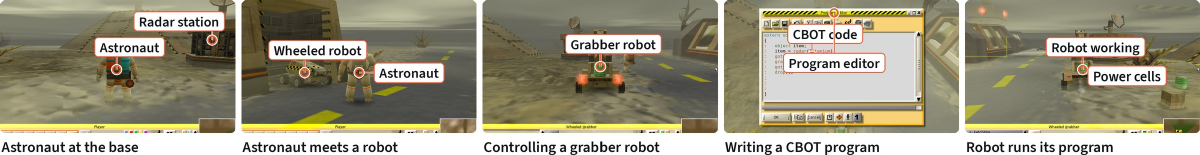}} \\[2pt]
\extlabel
\extrow{\multicolumn{5}{@{}l}{\scriptsize\textbf{Game trailer}: drive the boatmobile to the rocket launch} \\*
\rlab{Reference}{SpongeBob x Minecraft trailer} & \rf{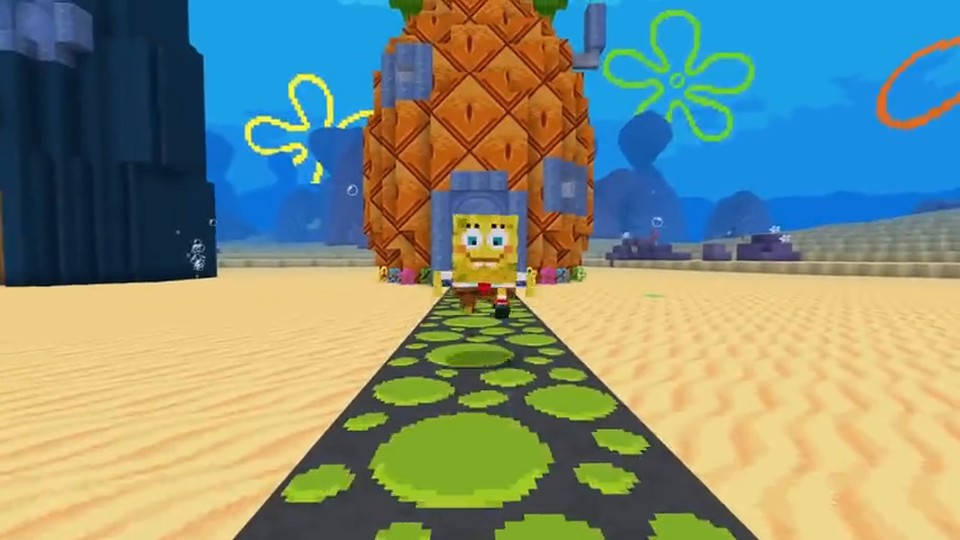} & \rf{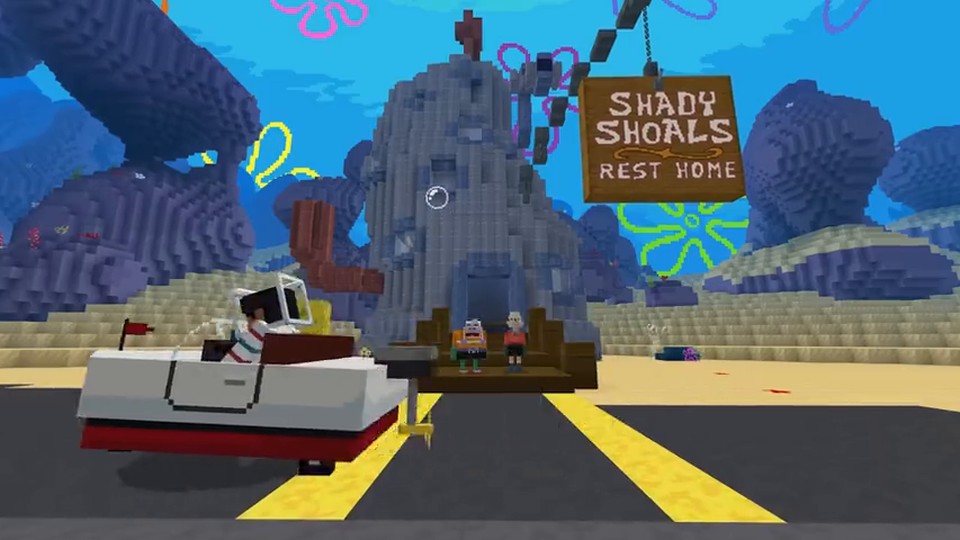} & \rf{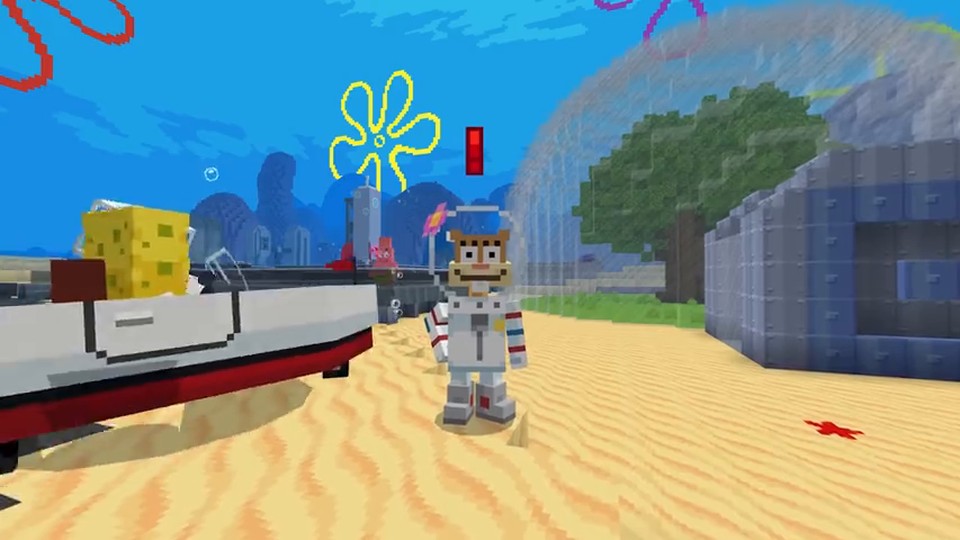} & \rf{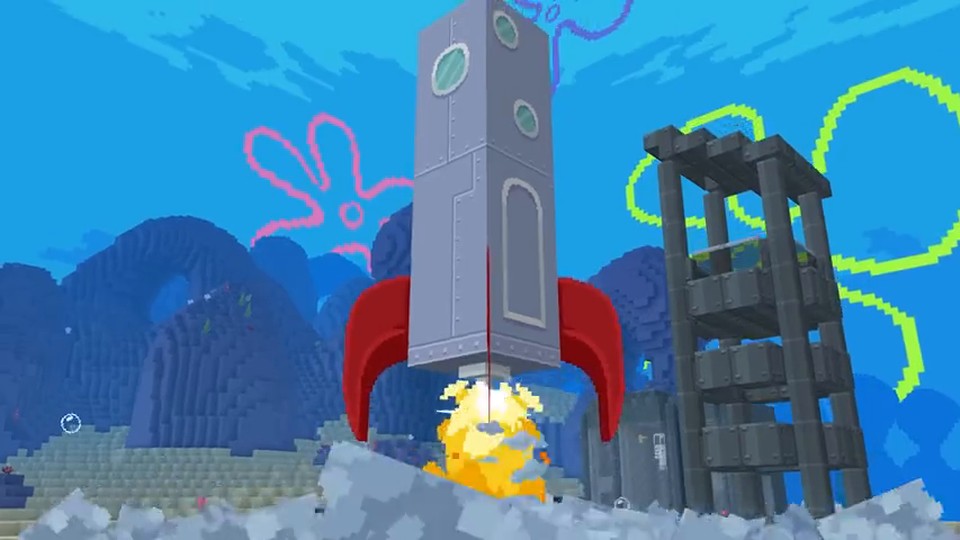} \\*
\rlab{In game}{} & \pf{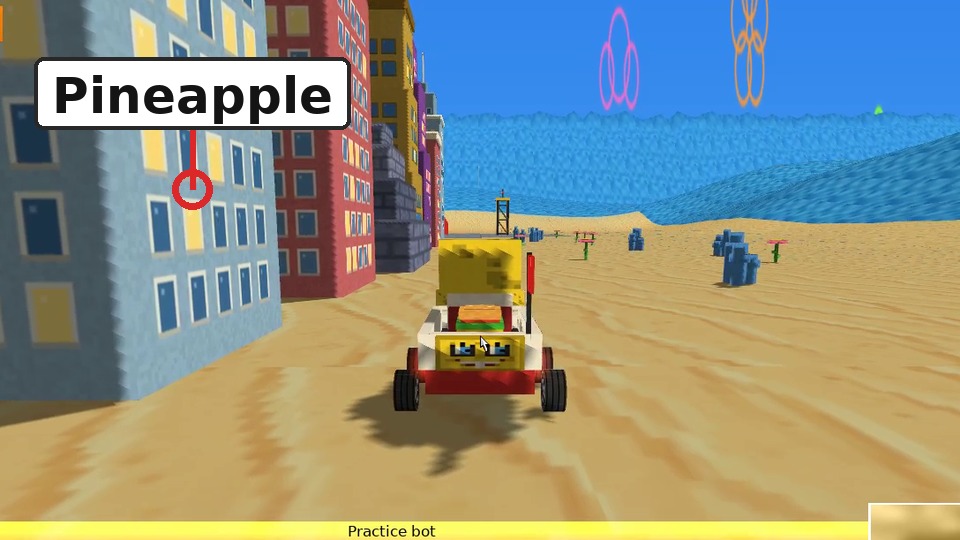}{SpongeBob's pineapple} & \pf{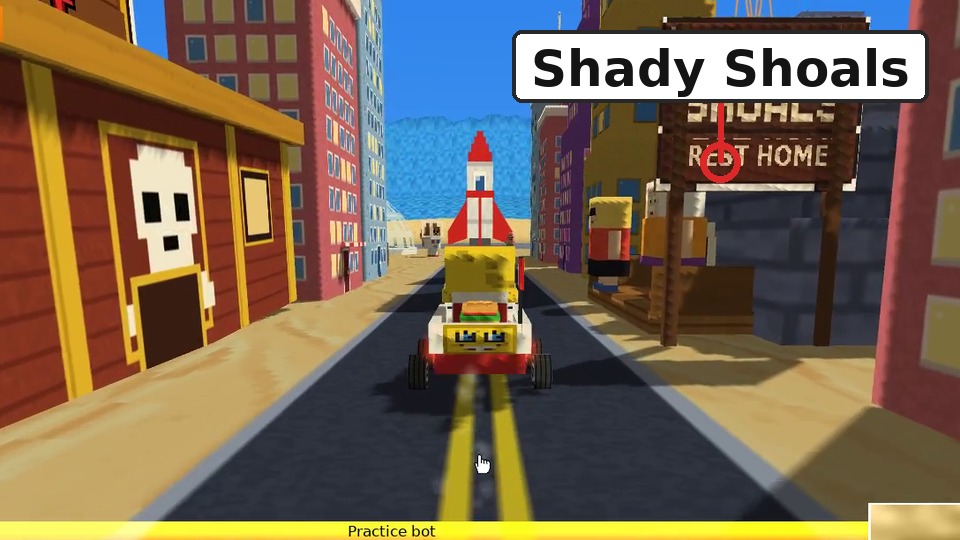}{Shady Shoals} & \pf{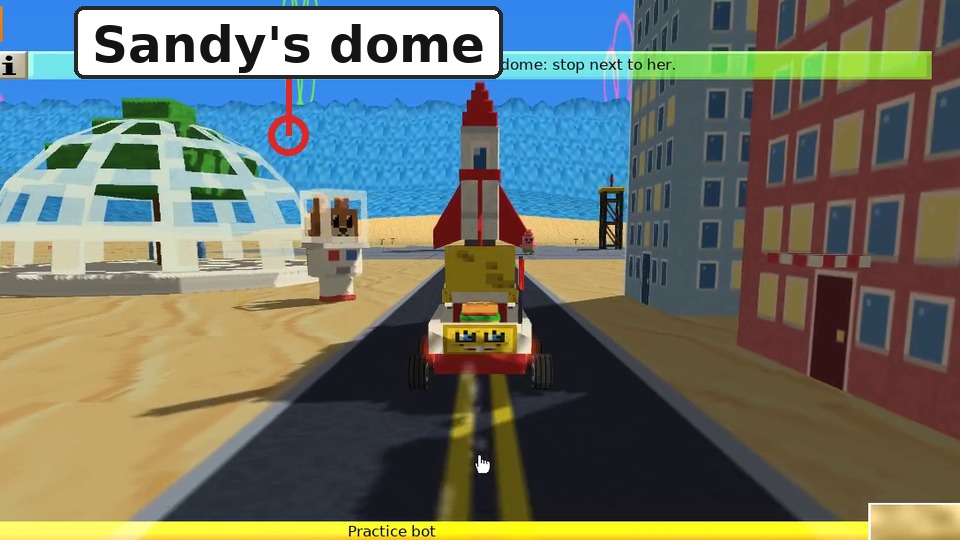}{Sandy's dome} & \pf{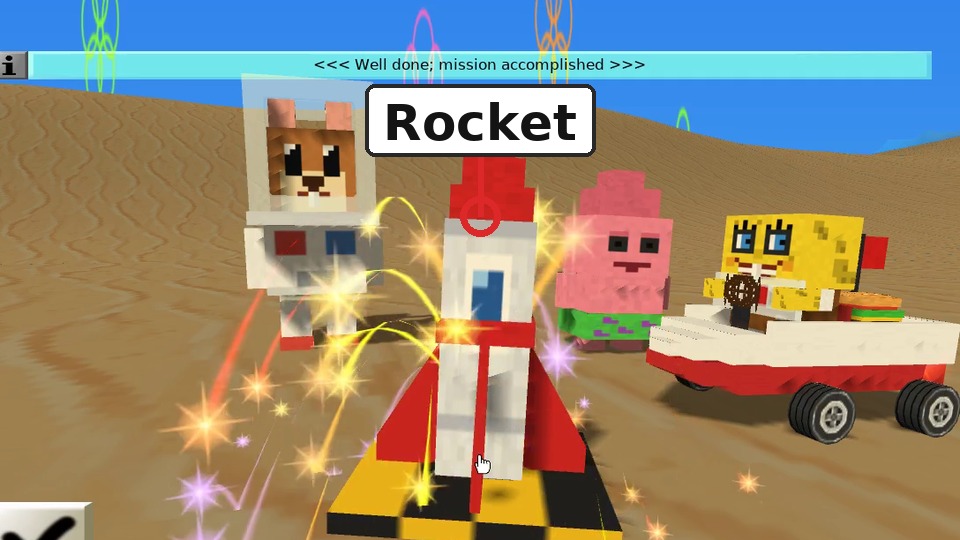}{Rocket launch}} \\[2pt]
\extrow{\multicolumn{5}{@{}l}{\scriptsize\textbf{Sports}: program the fielder to beat the runner} \\*
\rlab{Reference}{Baseball footage} & \rf{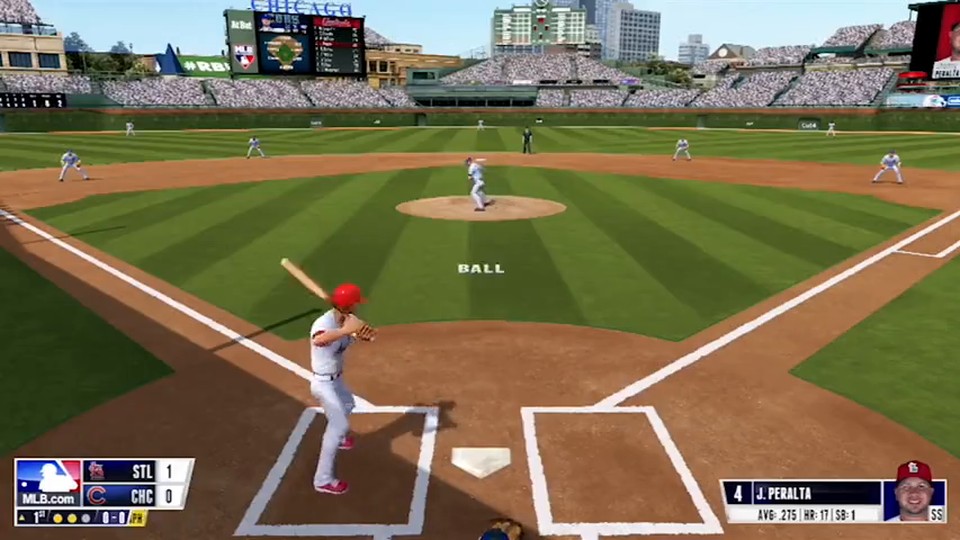} & \rf{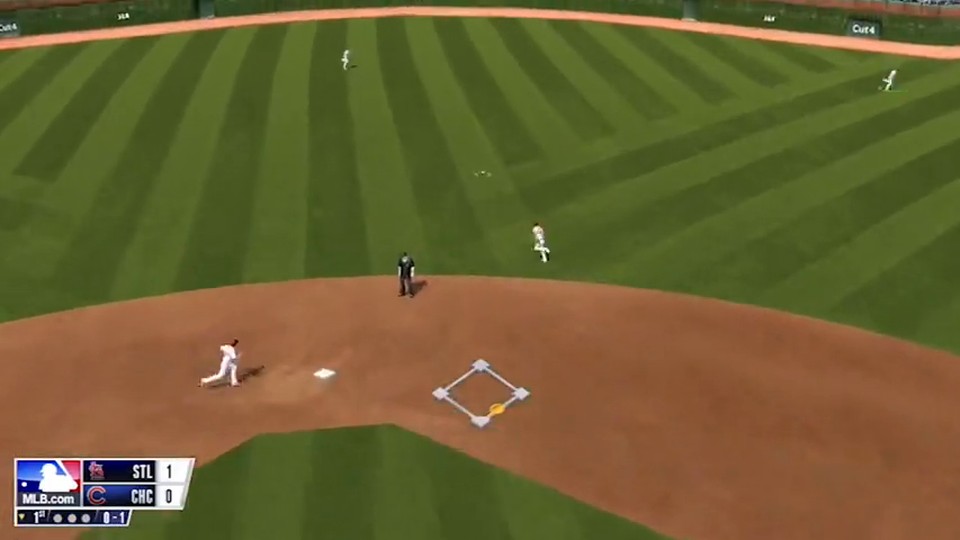} & \rf{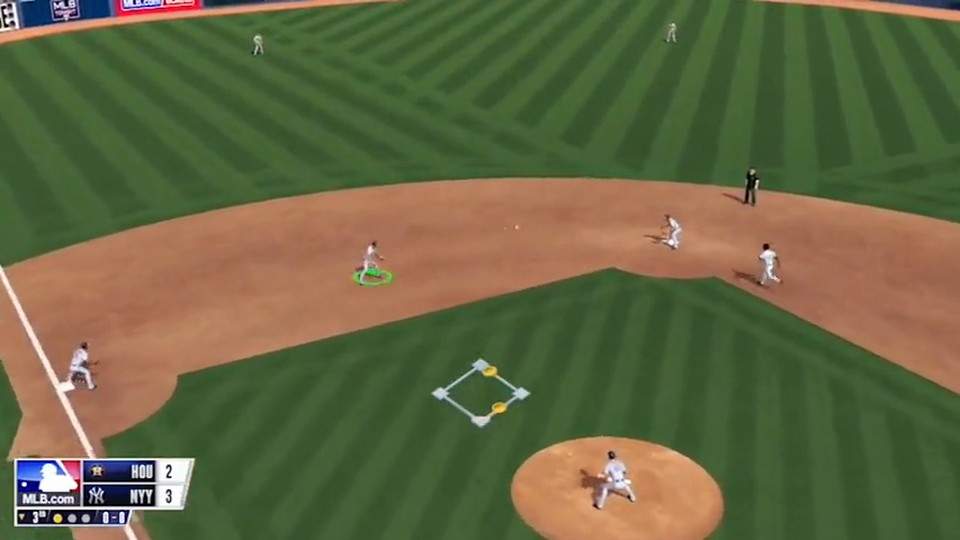} & \rf{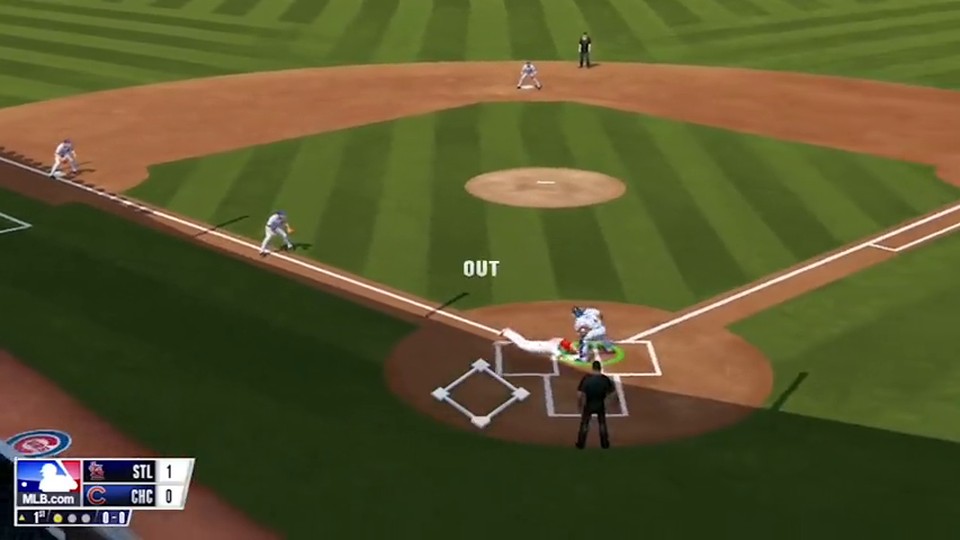} \\*
\rlab{In game}{} & \pf{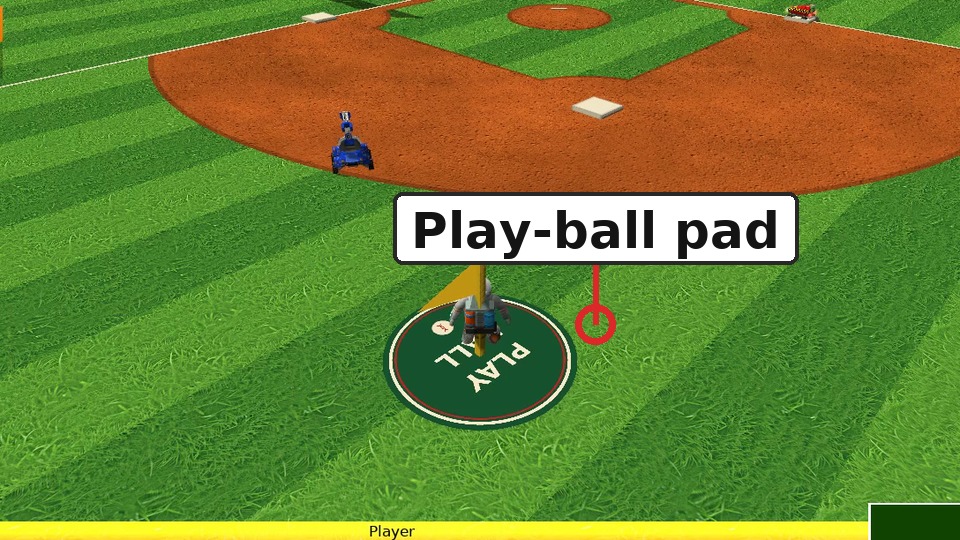}{Pitch at the plate} & \pf{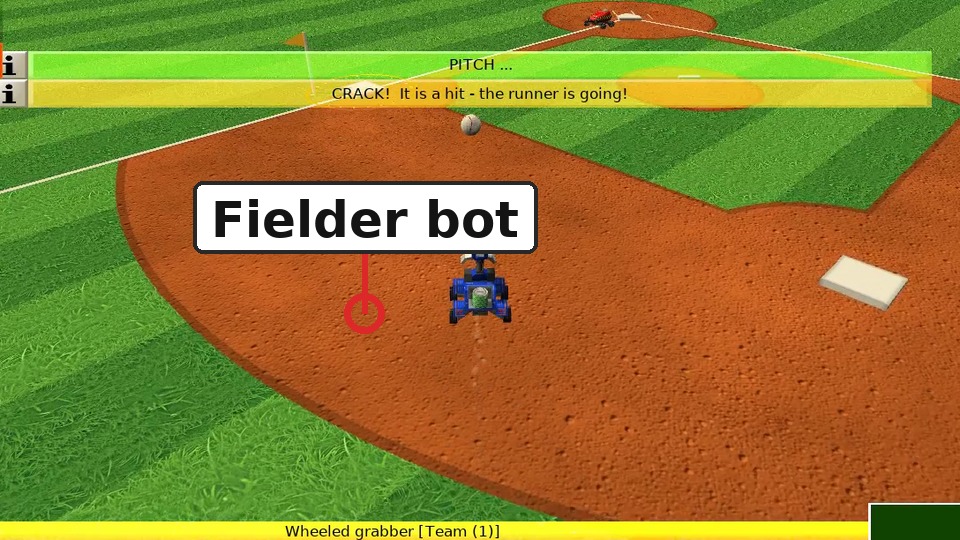}{Ball in play} & \pf{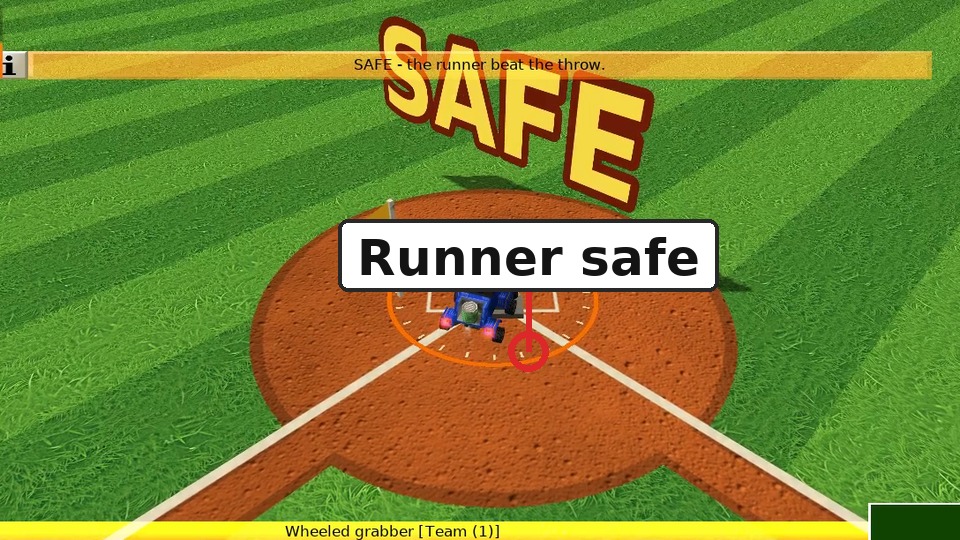}{Race to the base} & \pf{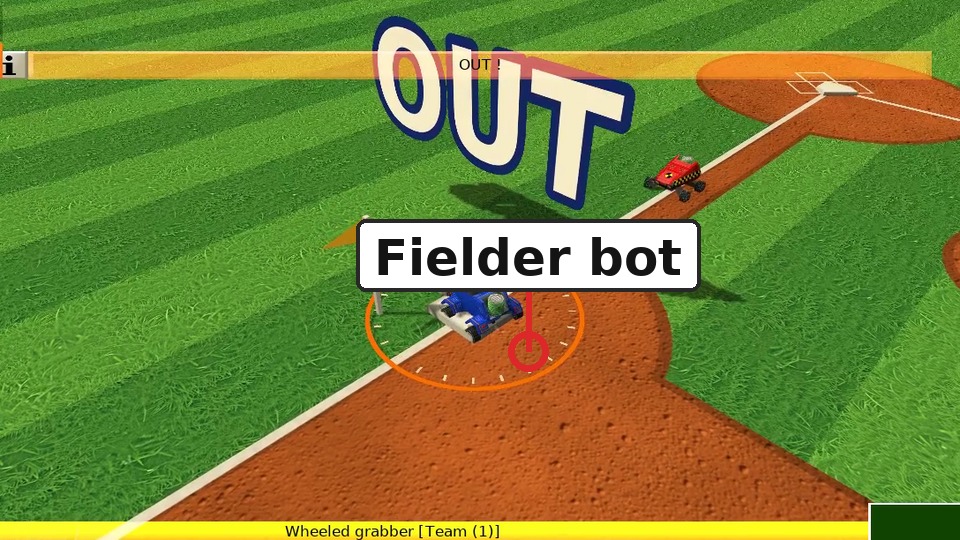}{Out at the base}} \\[4pt]
\midrule
\gthumb{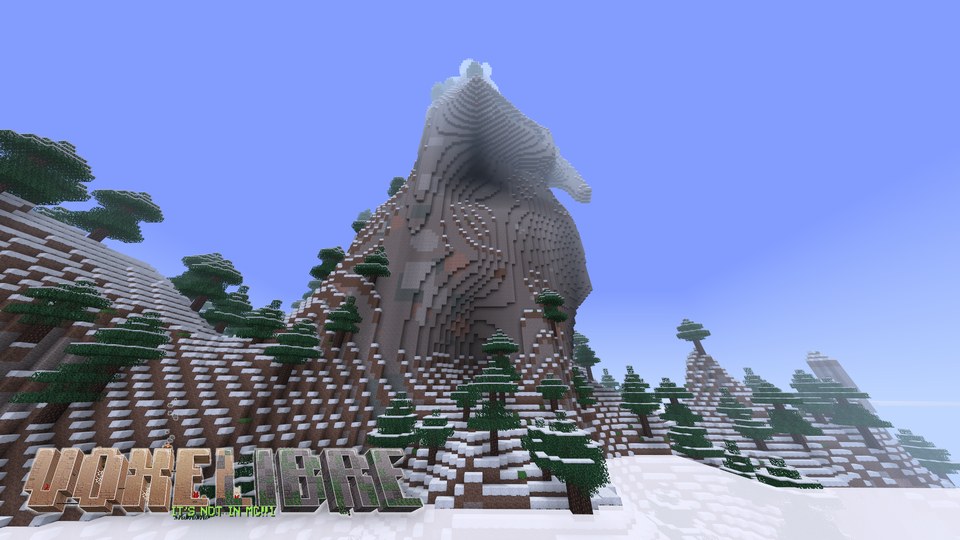} & \gname{VoxeLibre}{0.92.4 \textperiodcentered{} Luanti 5.17 \textperiodcentered{} GPL-3.0} & \gtag{Survival / sandbox} & \vtag{3} & Mines and places blocks, crafts tools and survives mobs and nights in a voxel world. & Lua mods: block, item, entity APIs & 18 \\*
\multicolumn{7}{@{}l@{}}{\rule{0pt}{12pt}\includegraphics[width=0.995\textwidth]{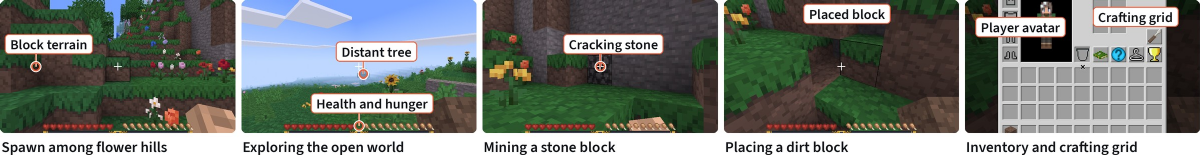}} \\[2pt]
\extlabel
\extrow{\multicolumn{5}{@{}l}{\scriptsize\textbf{City}: light the tower from sunset to blue hour} \\*
\rlab{Reference}{Paris city footage} & \rf{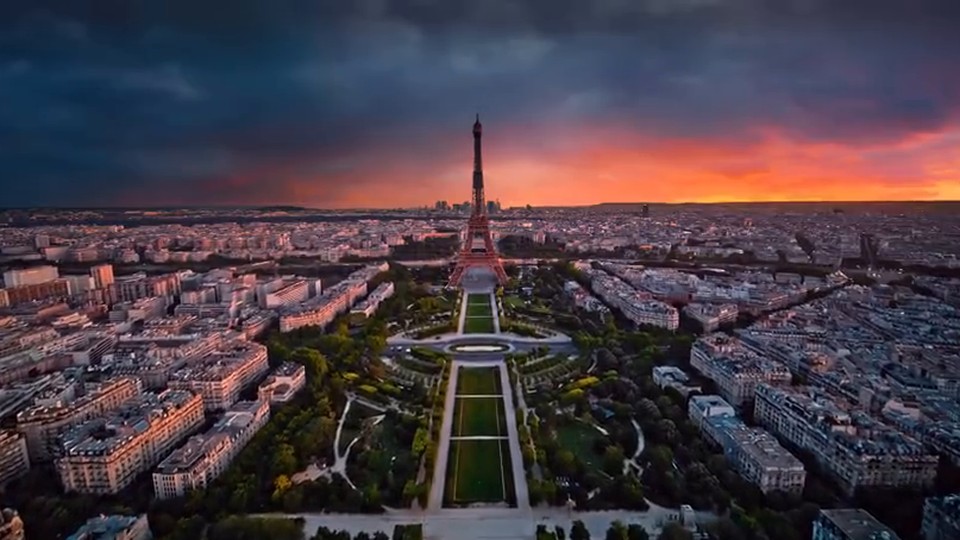} & \rf{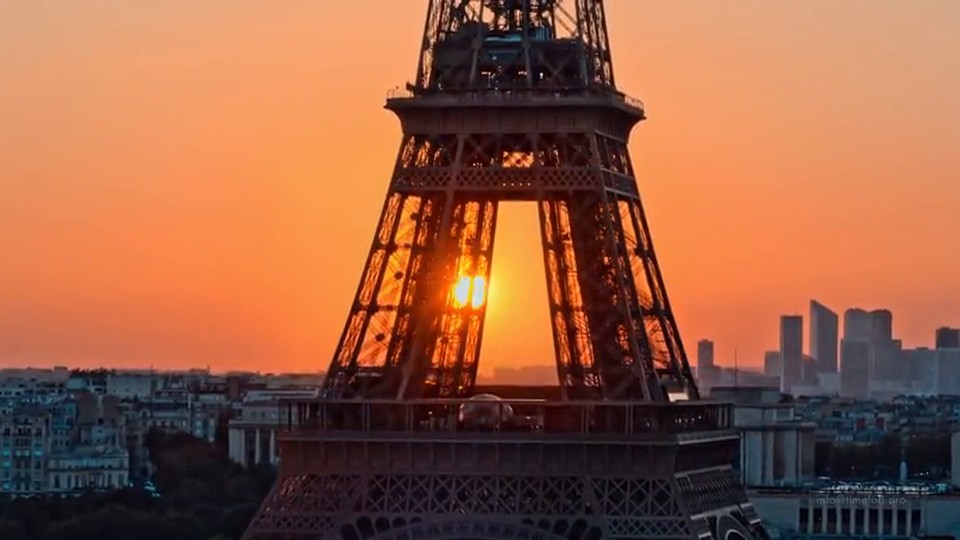} & \rf{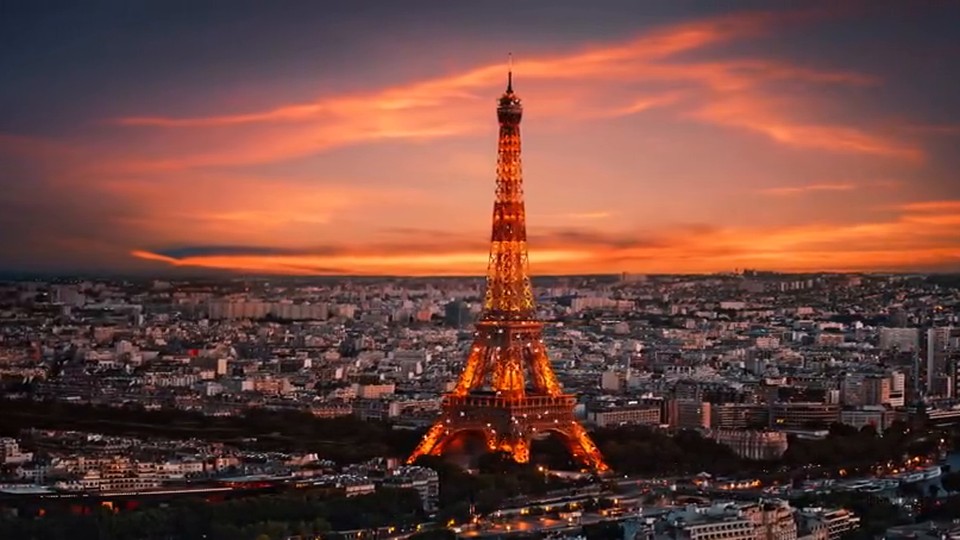} & \rf{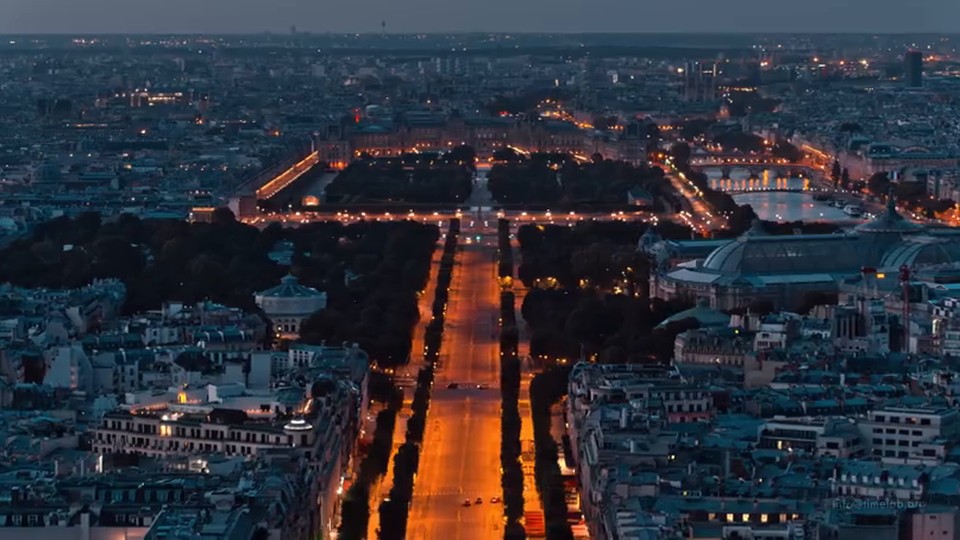} \\*
\rlab{In game}{} & \pf{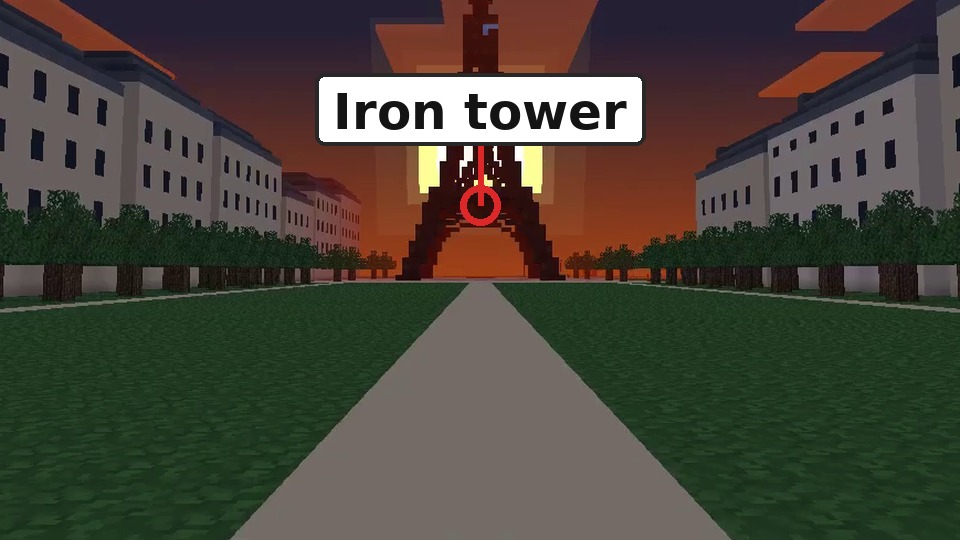}{Park toward the tower} & \pf{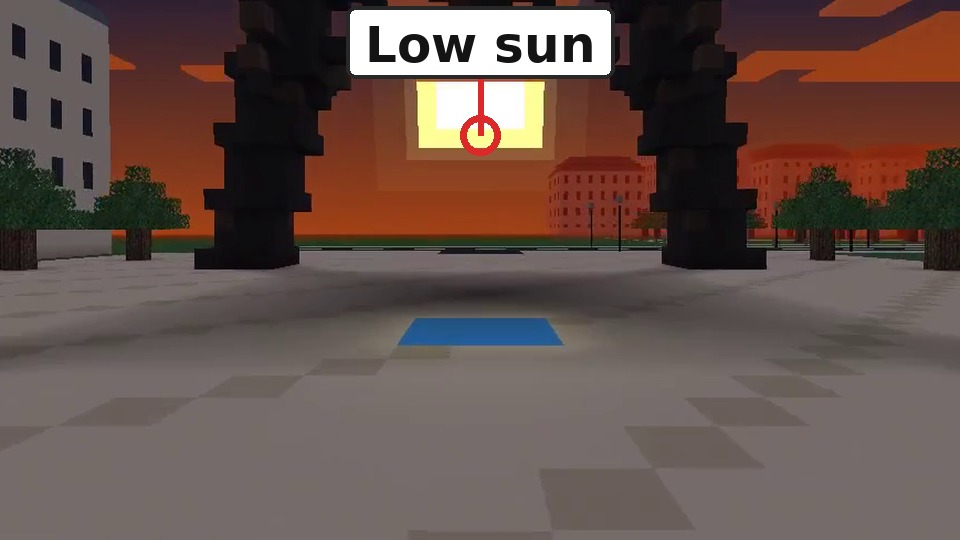}{Sun between the legs} & \pf{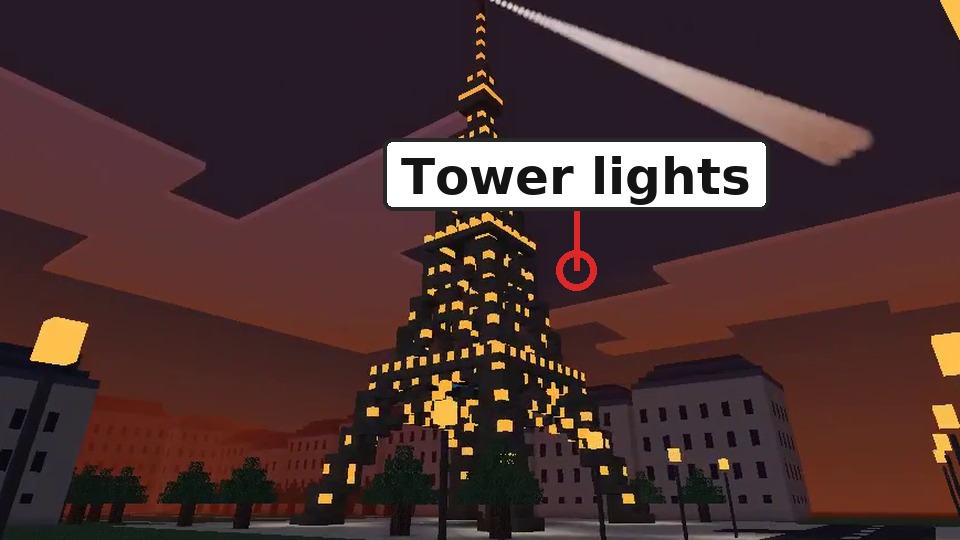}{Tower lights up} & \pf{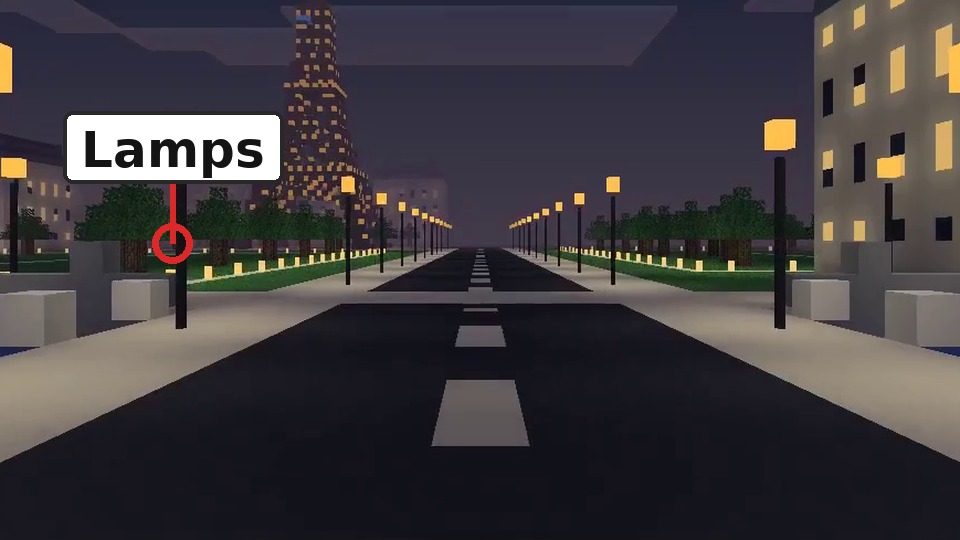}{Lit avenue}} \\[2pt]
\extrow{\multicolumn{5}{@{}l}{\scriptsize\textbf{Game trailer}: tree, star ball, and parasol glide in the toy valley} \\*
\rlab{Reference}{Kirby trailer} & \rf{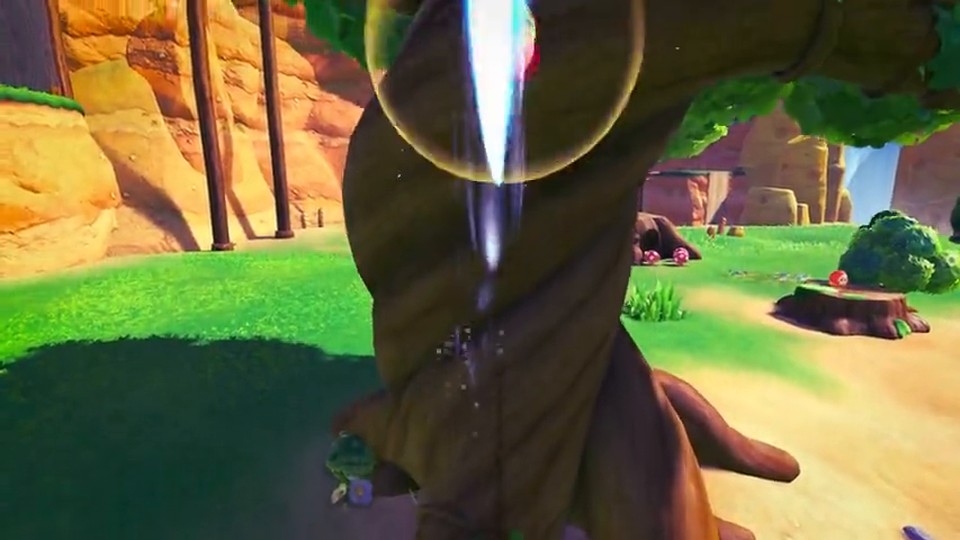} & \rf{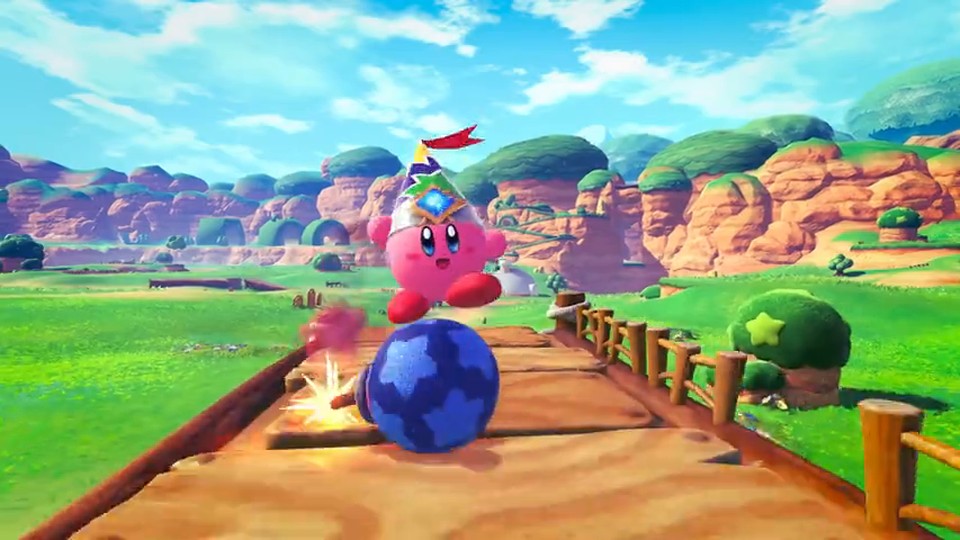} & \rf{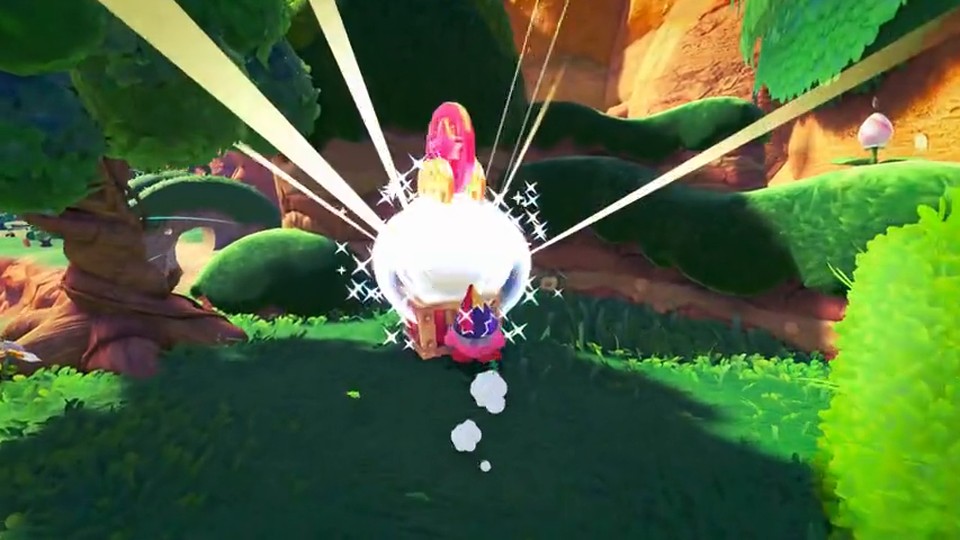} & \rf{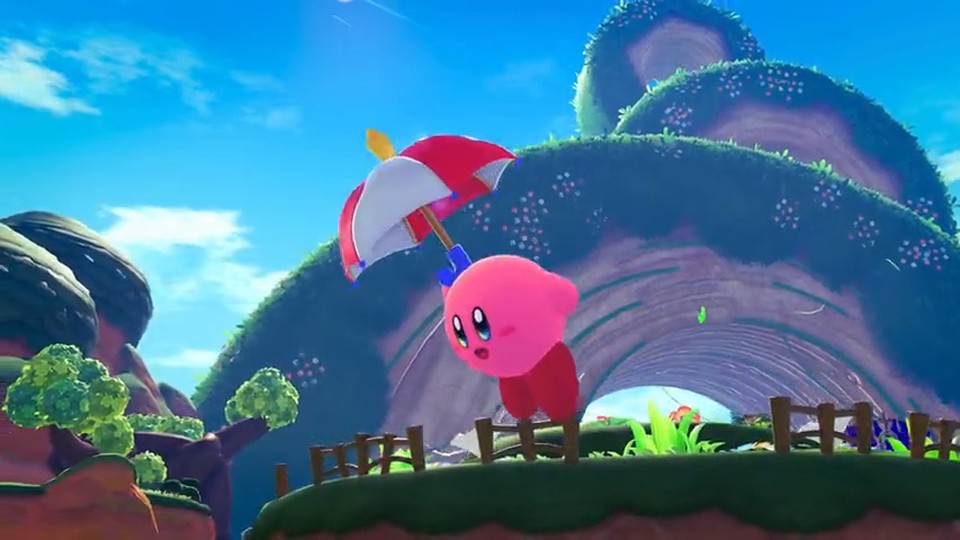} \\*
\rlab{In game}{} & \pf{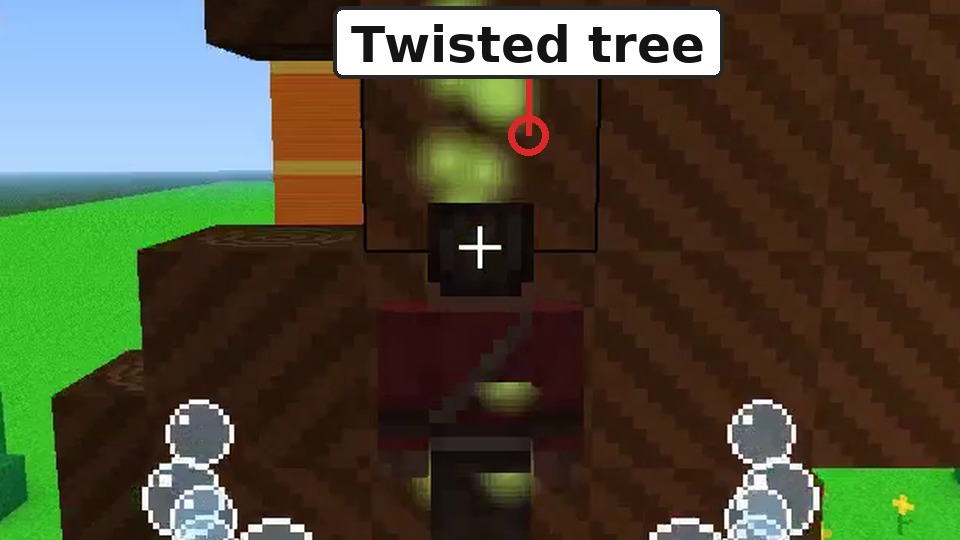}{Climbing the tree} & \pf{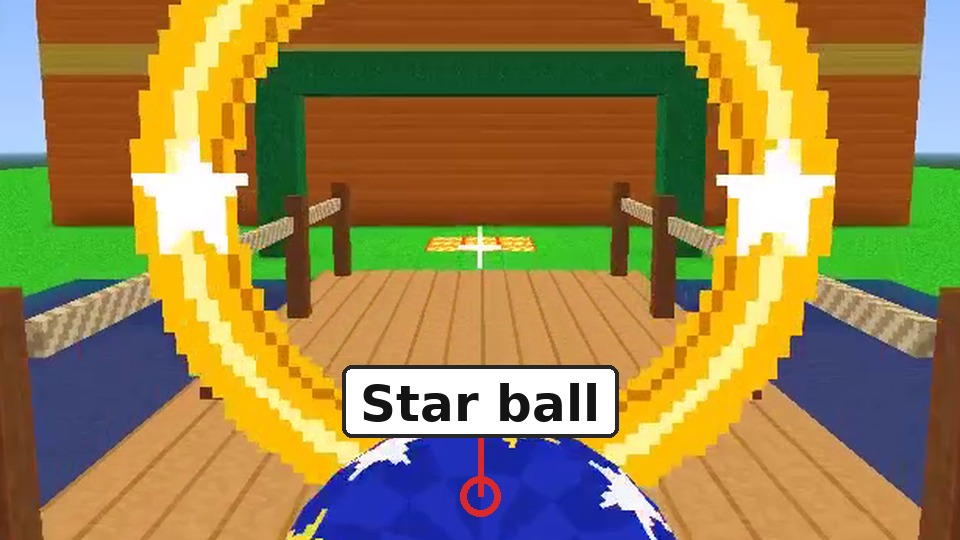}{Star ball} & \pf{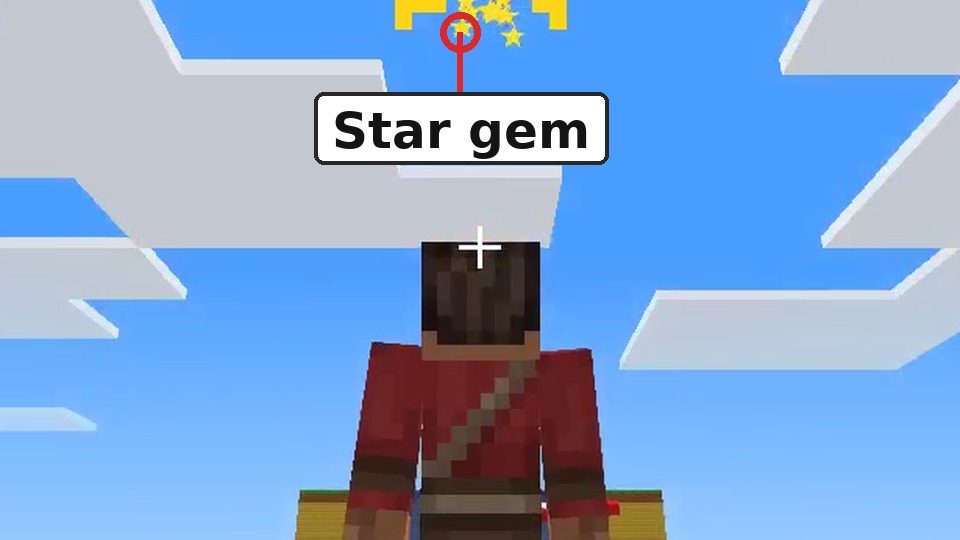}{Star gem rises} & \pf{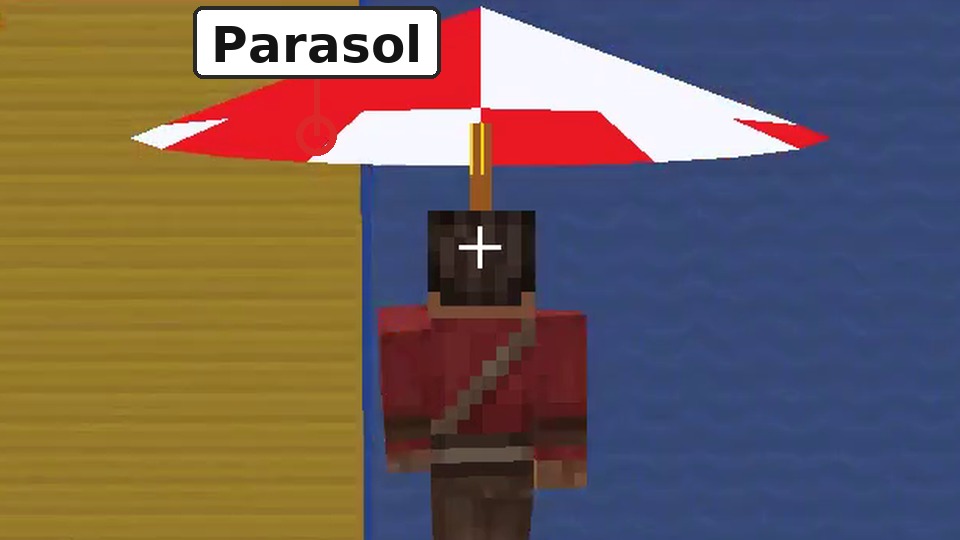}{Parasol glide}} \\[4pt]
\midrule
\gthumb{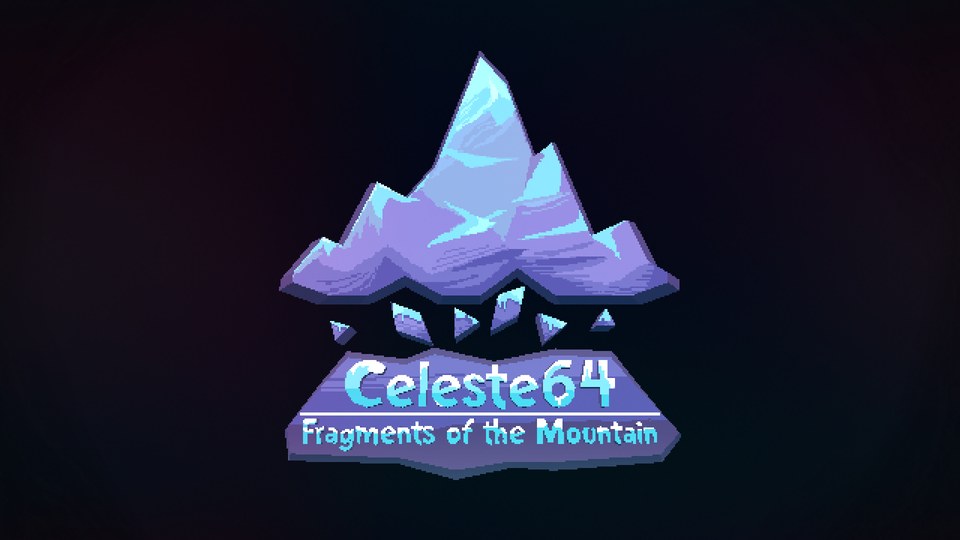} & \gname{Celeste 64}{1.1.1 \textperiodcentered{} Foster / C\# \textperiodcentered{} MIT code} & \gtag{Platformer} & \vtag{3} & Runs, jumps, dashes and climbs through a 3D mountain to collect strawberries. & Map files, level \mf{.json} & 6 \\*
\multicolumn{7}{@{}l@{}}{\rule{0pt}{12pt}\includegraphics[width=0.995\textwidth]{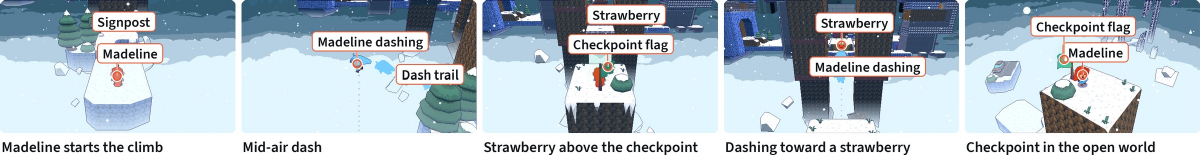}} \\[2pt]
\extlabel
\extrow{\multicolumn{5}{@{}l}{\scriptsize\textbf{City}: climb the iron tower to the summit} \\*
\rlab{Reference}{Paris city footage} & \rf{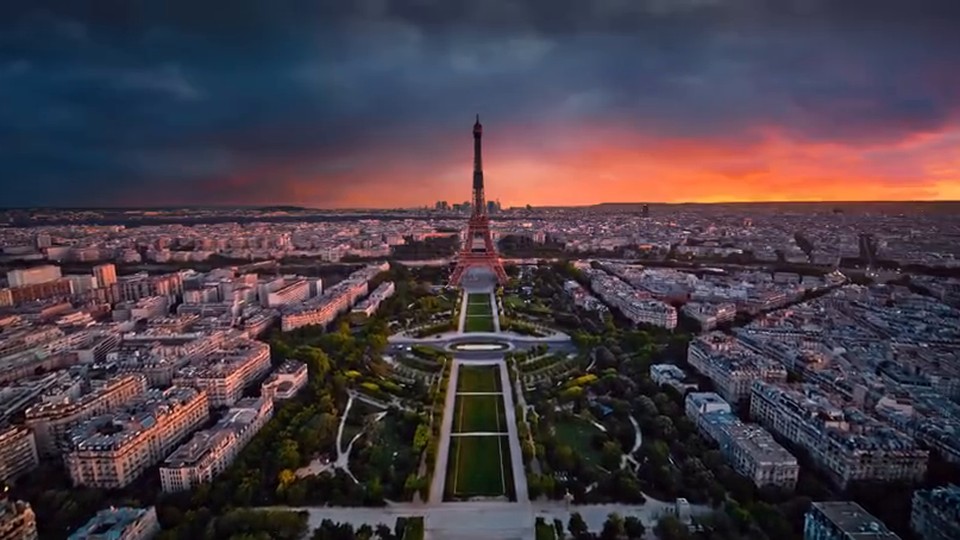} & \rf{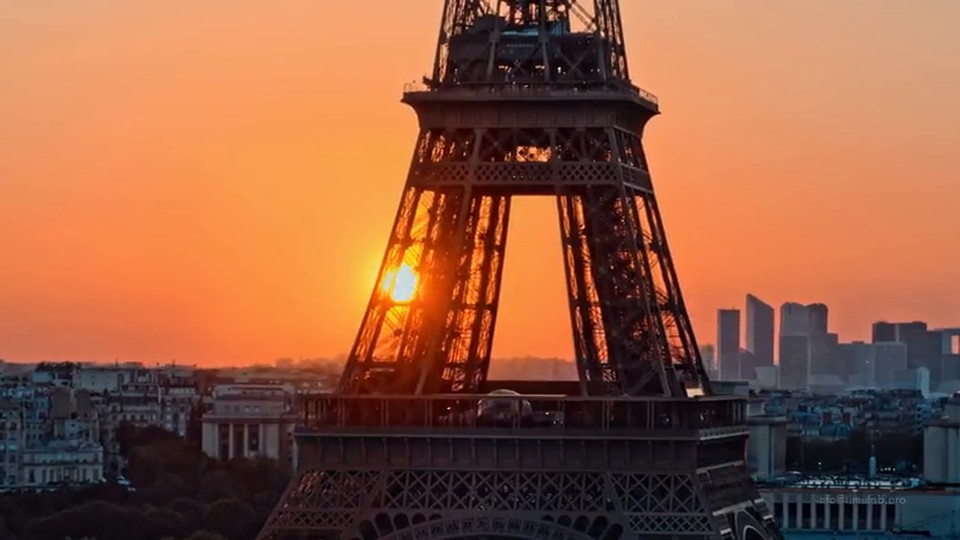} & \rf{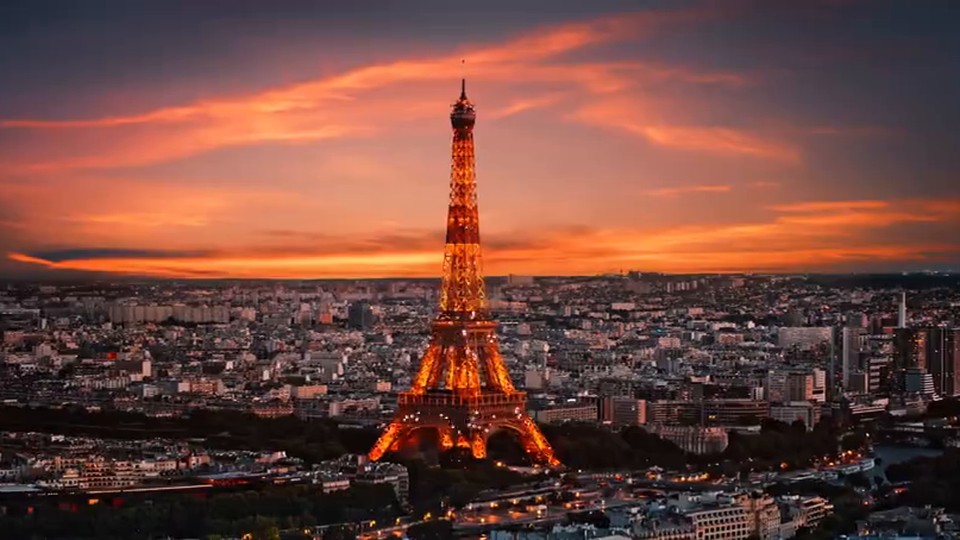} & \rf{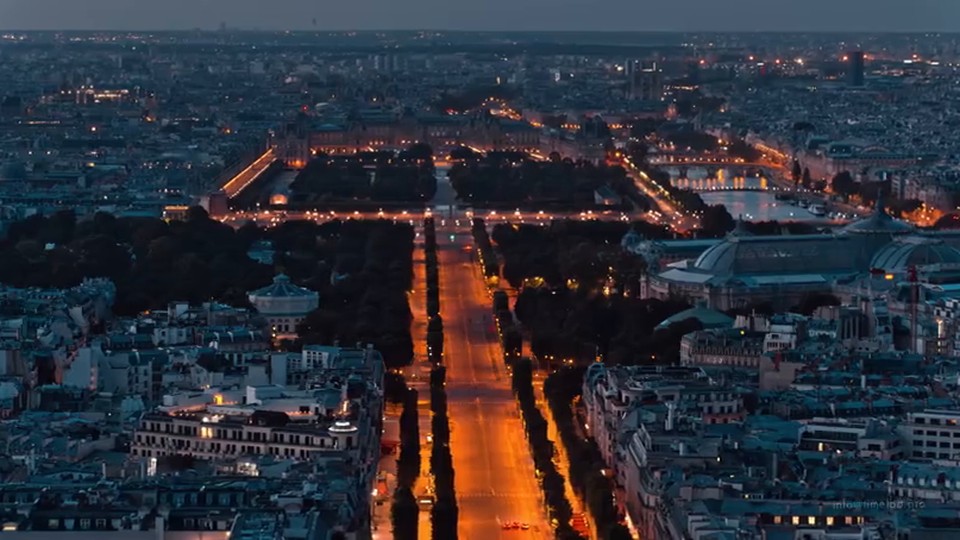} \\*
\rlab{In game}{} & \pf{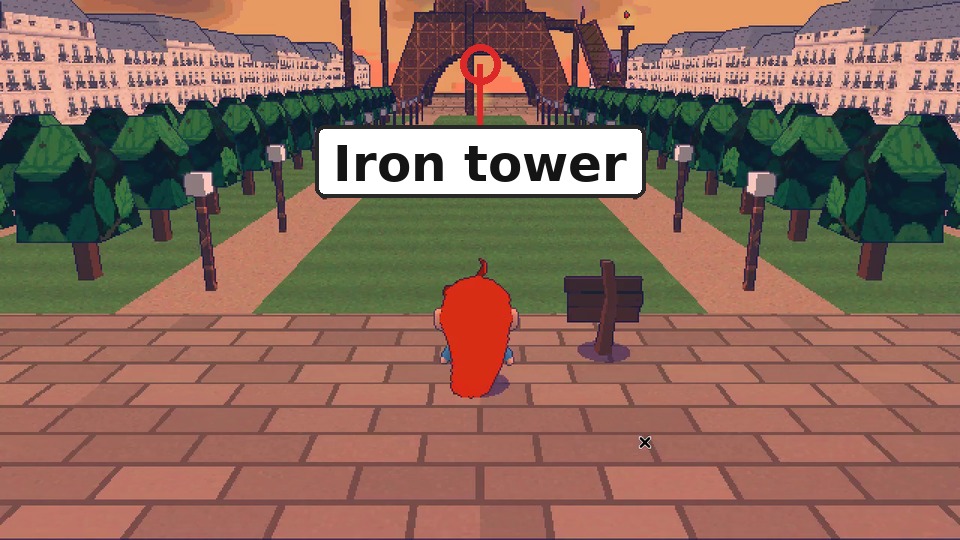}{Park toward the tower} & \pf{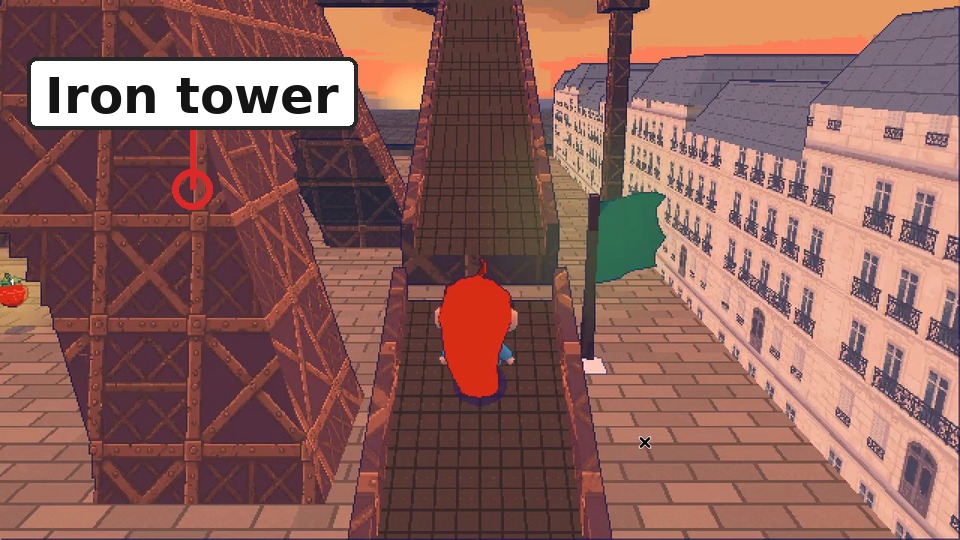}{Tower base at sunset} & \pf{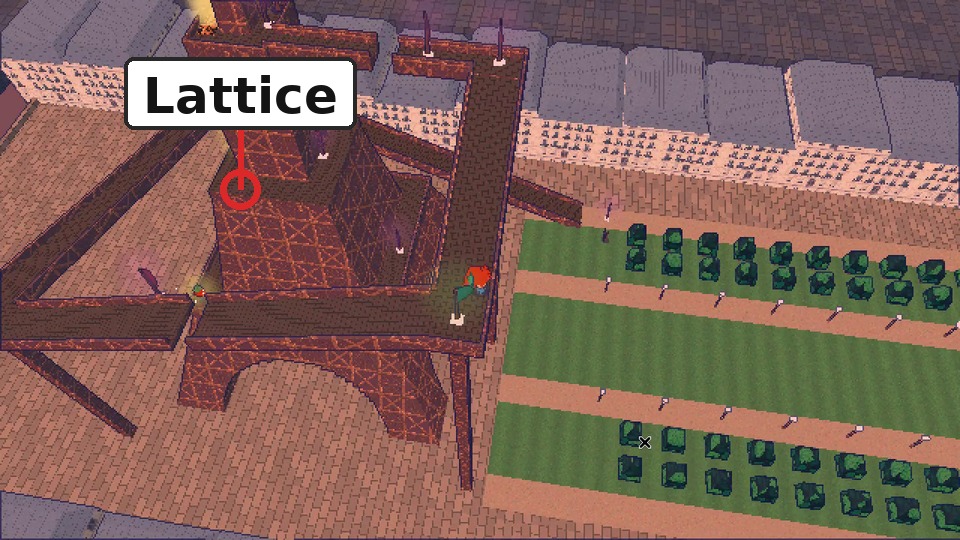}{High on the tower} & \pf{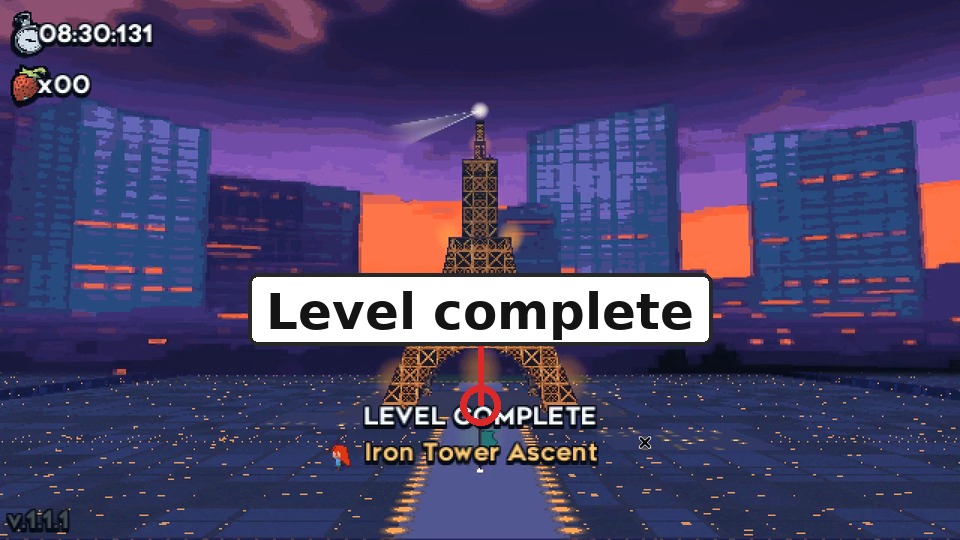}{Summit at night}} \\[2pt]
\extrow{\multicolumn{5}{@{}l}{\scriptsize\textbf{Film}: through Bowser Jr.'s lava traps to the goal flag} \\*
\rlab{Reference}{The Super Mario Bros. Movie} & \rf{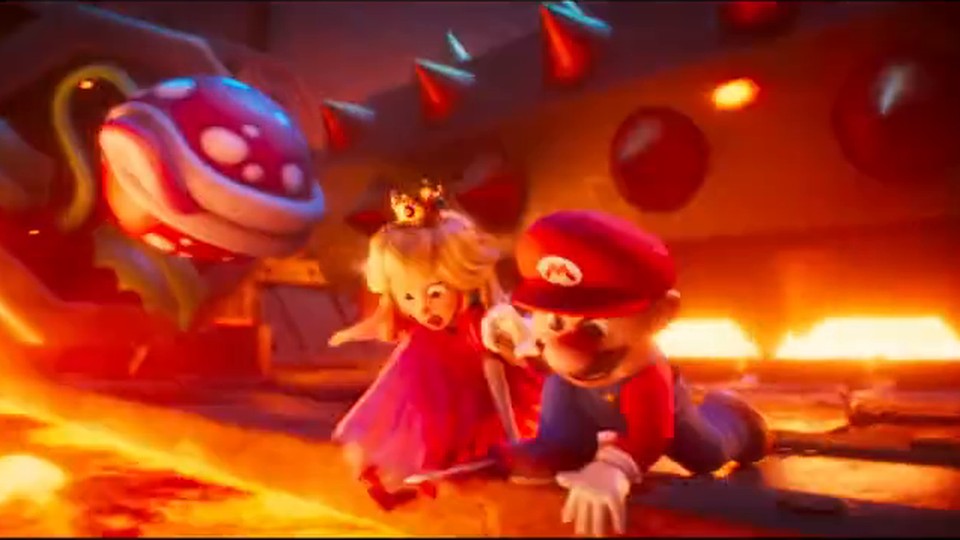} & \rf{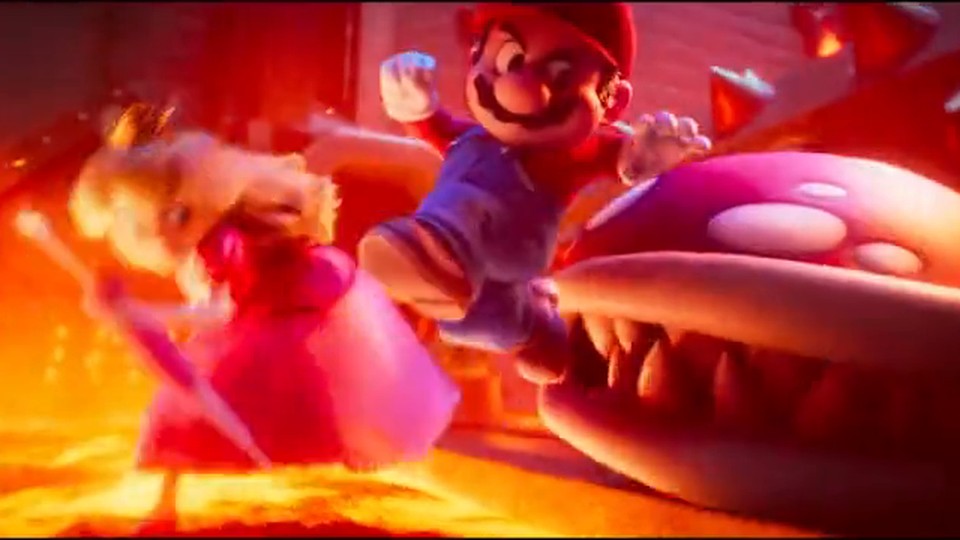} & \rf{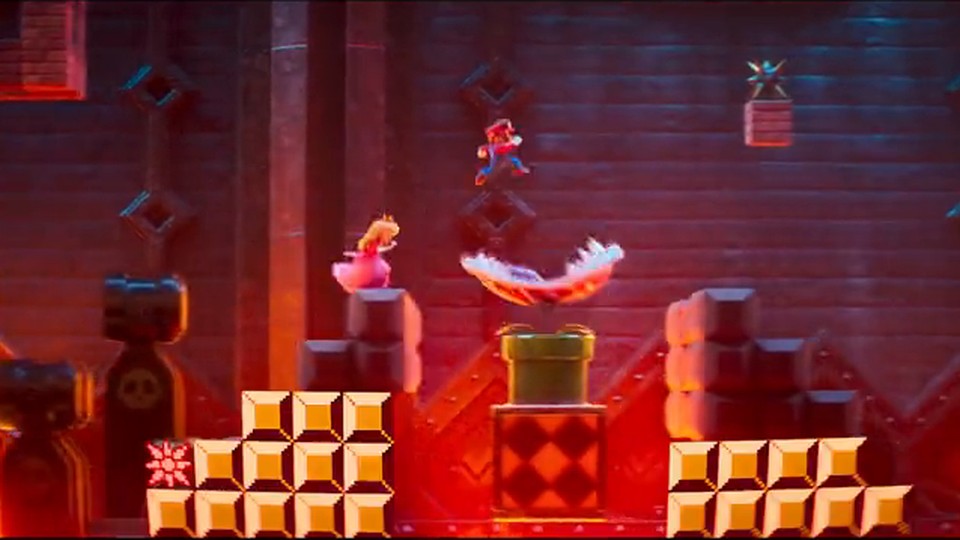} & \rf{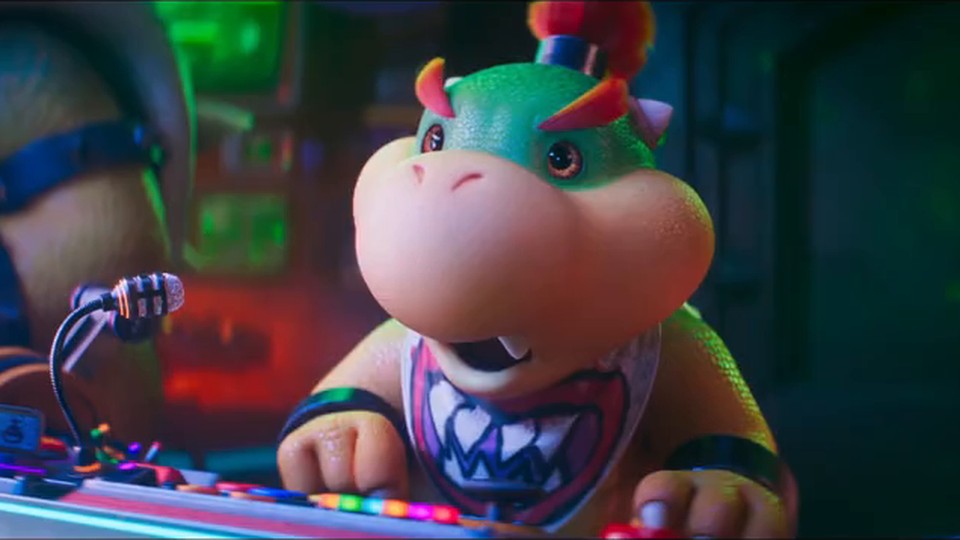} \\*
\rlab{In game}{} & \pf{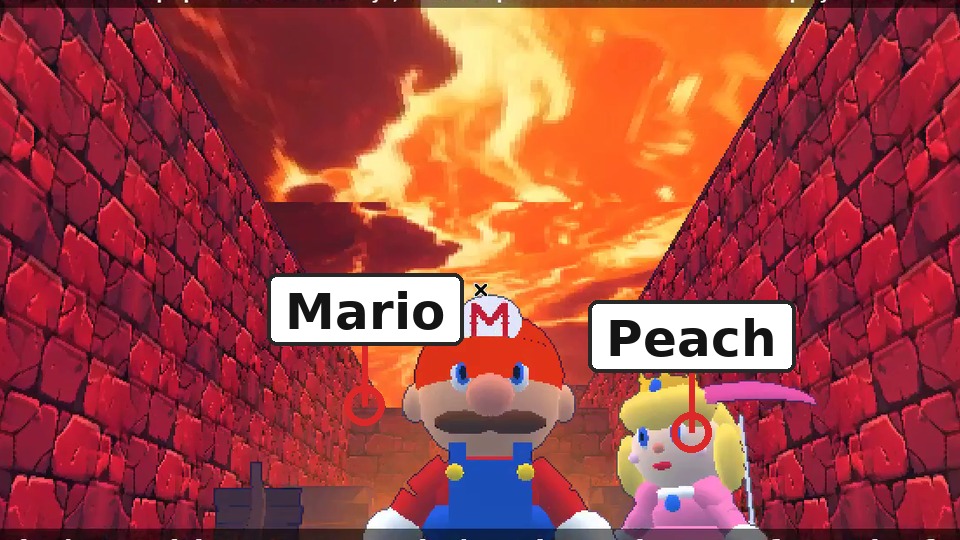}{Mario and Peach} & \pf{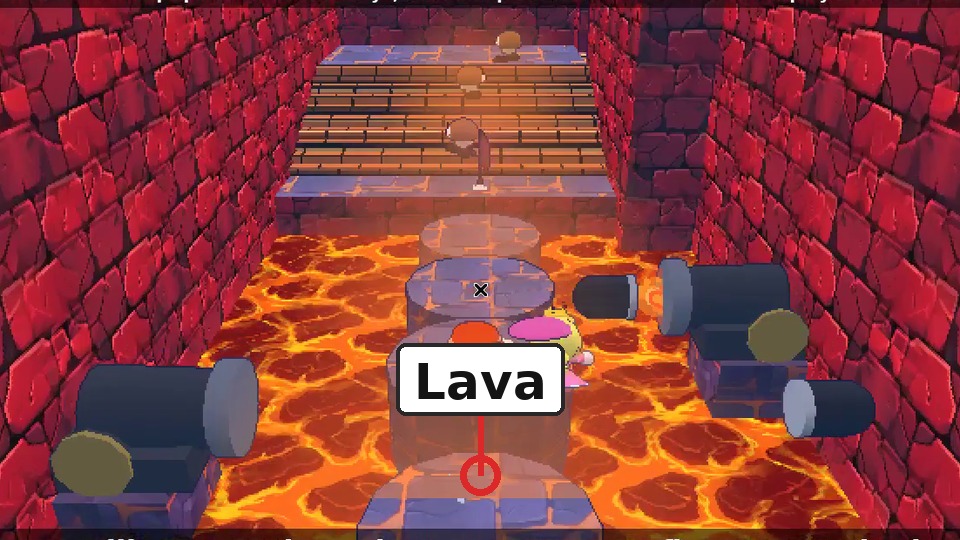}{Lava traps} & \pf{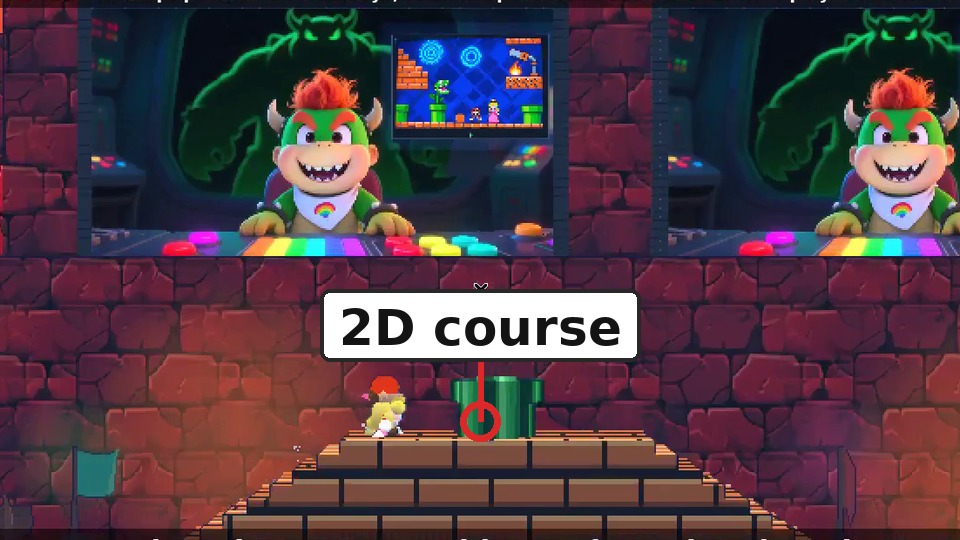}{2D course} & \pf{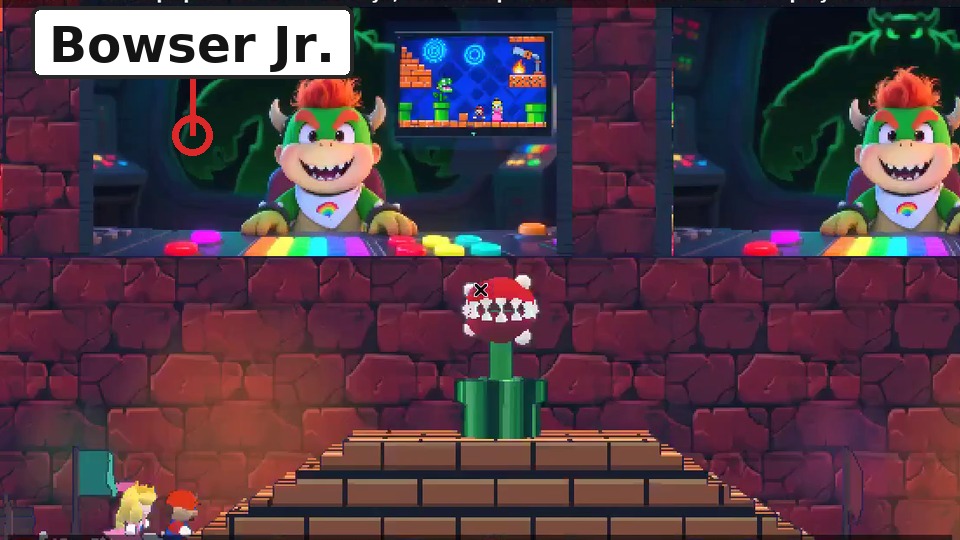}{Bowser Jr. watching}} \\[4pt]
\midrule
\gthumb{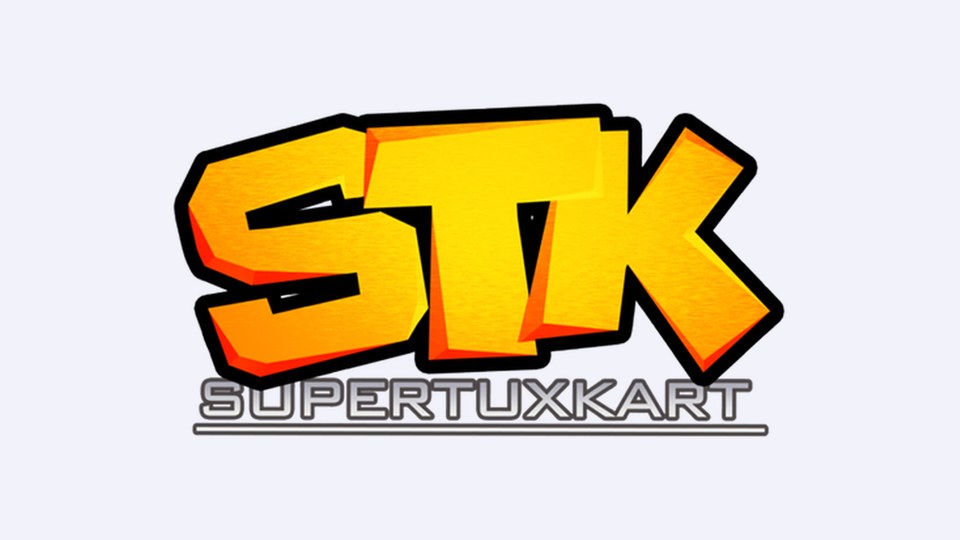} & \gname{SuperTuxKart}{1.5 \textperiodcentered{} own engine \textperiodcentered{} GPL-3.0} & \gtag{Racing} & \vtag{3} & Races karts on themed tracks with items and nitro; battle arenas and a story mode. & Add-on karts and tracks, track scripting & 10 \\*
\multicolumn{7}{@{}l@{}}{\rule{0pt}{12pt}\includegraphics[width=0.995\textwidth]{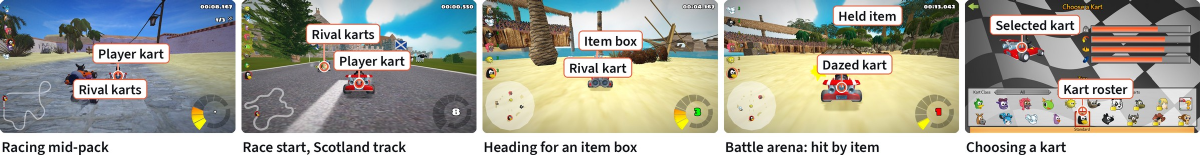}} \\[2pt]
\extlabel
\extrow{\multicolumn{5}{@{}l}{\scriptsize\textbf{Sports}: ride the wave ahead of the whitewater} \\*
\rlab{Reference}{Surfing footage} & \rf{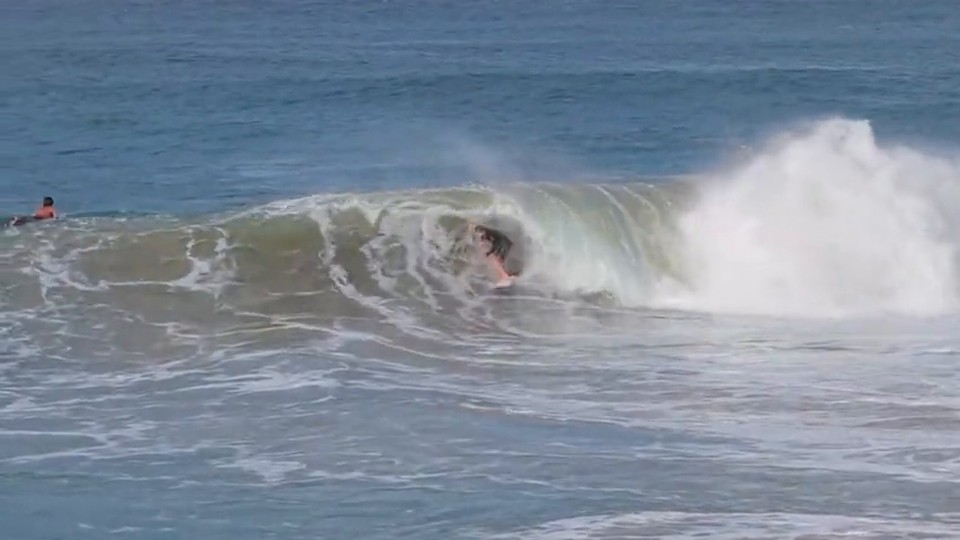} & \rf{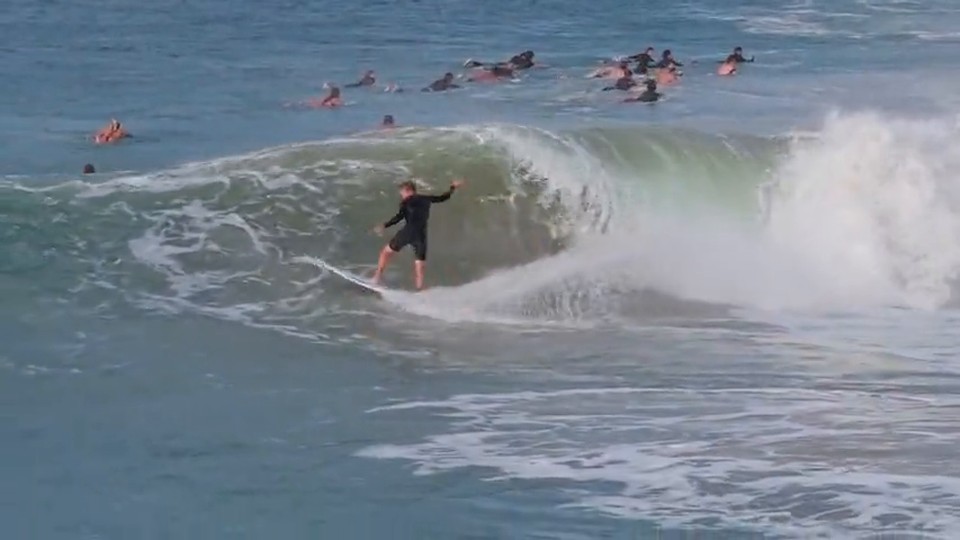} & \rf{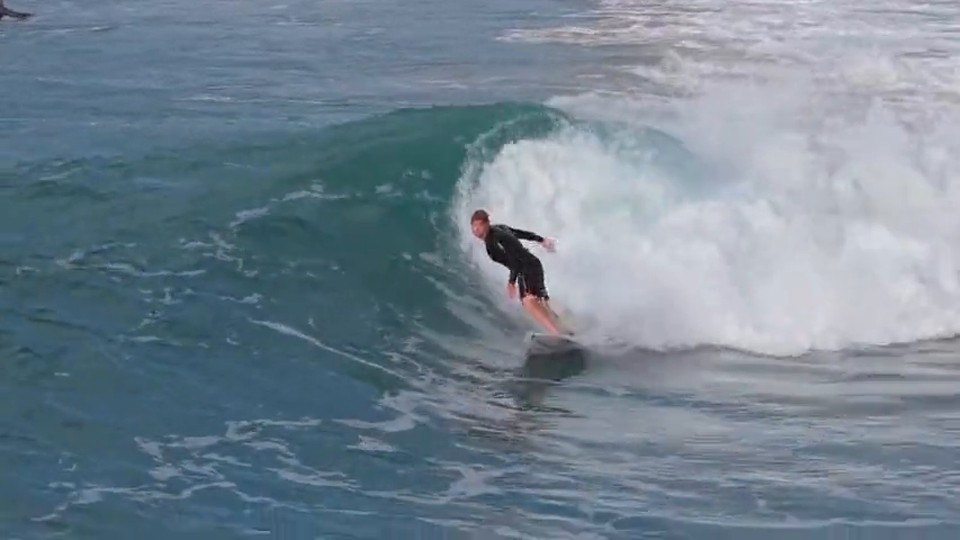} & \rf{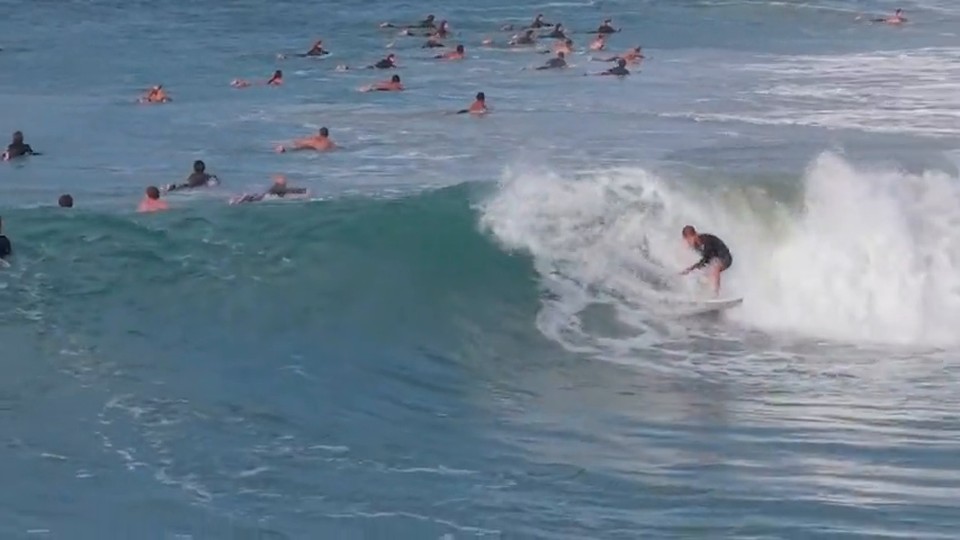} \\*
\rlab{In game}{} & \pf{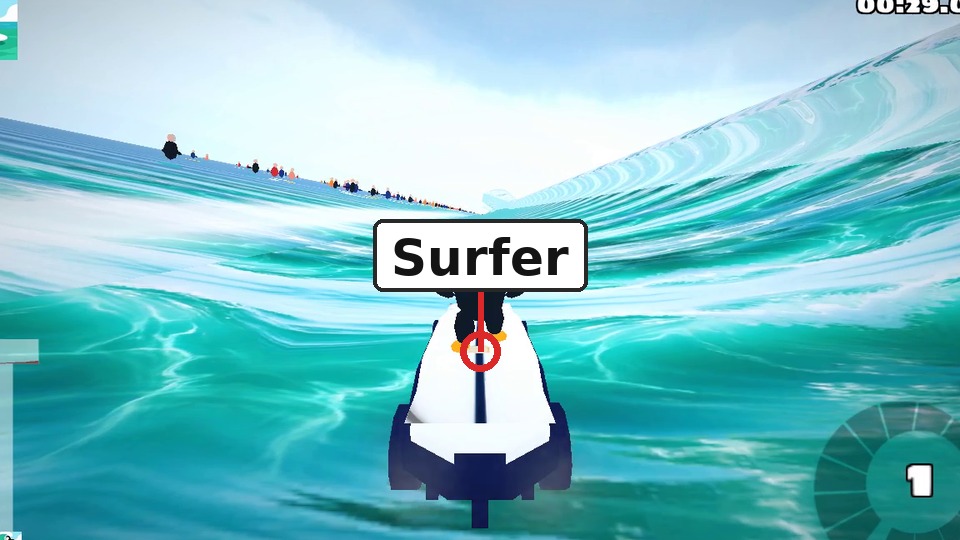}{Dropping in} & \pf{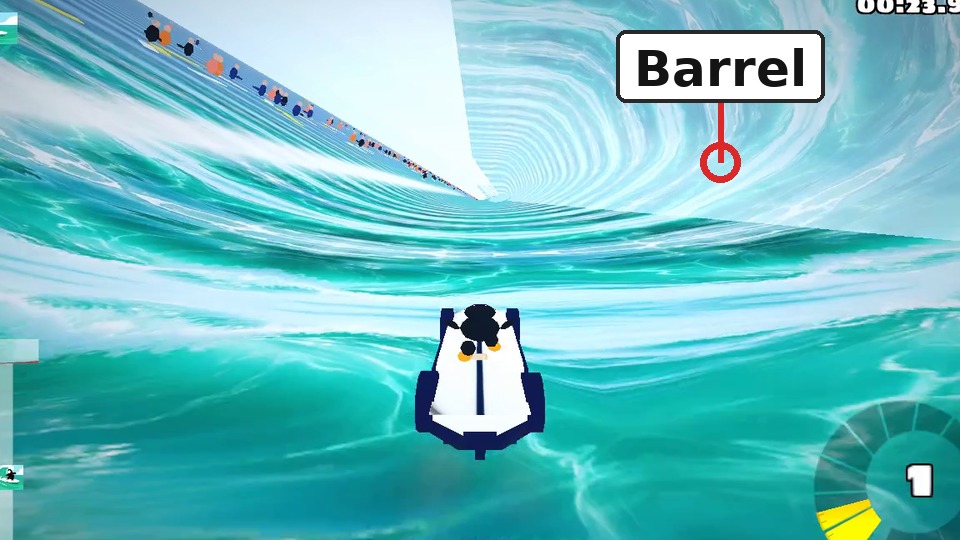}{Under the lip} & \pf{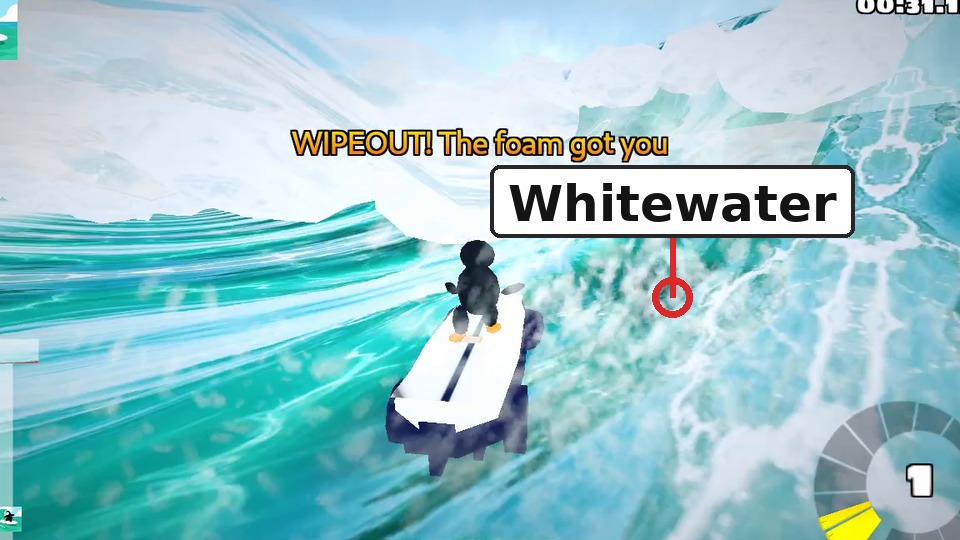}{Carving the face} & \pf{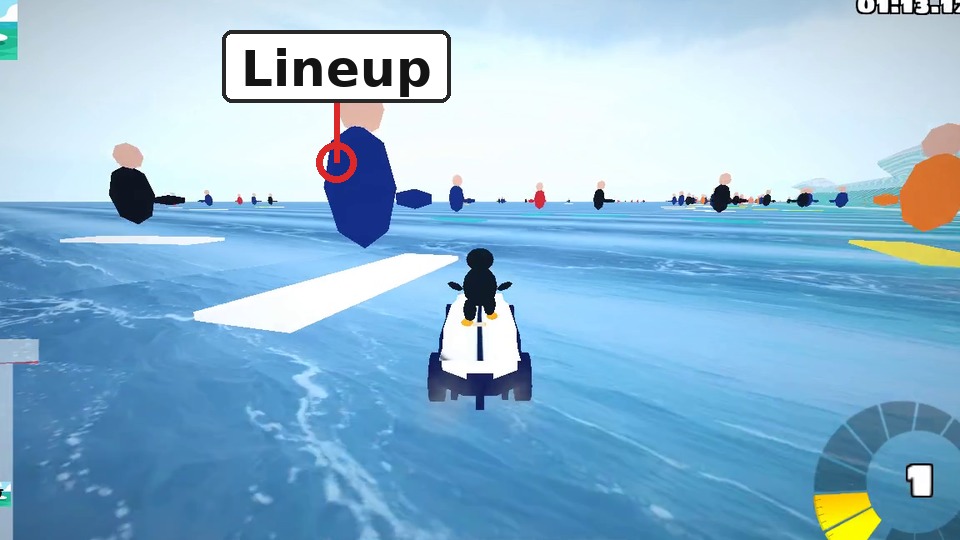}{Lineup behind}} \\[2pt]
\extrow{\multicolumn{5}{@{}l}{\scriptsize\textbf{Game trailer}: a meadow lap collecting stars} \\*
\rlab{Reference}{Kirby trailer} & \rf{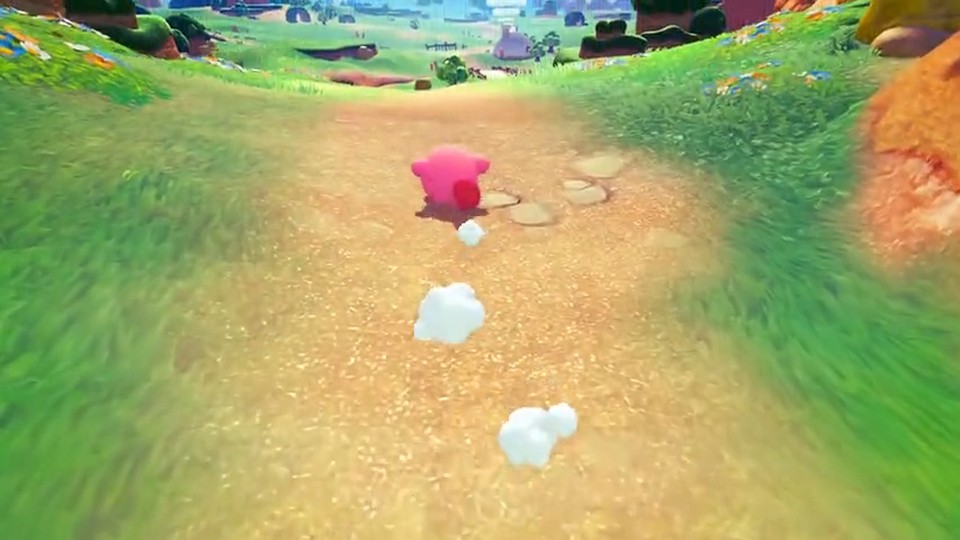} & \rf{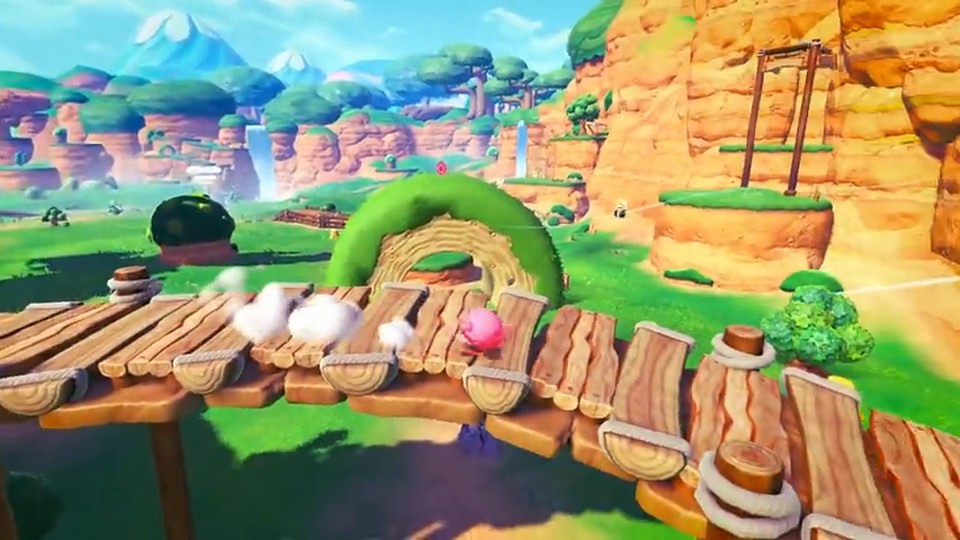} & \rf{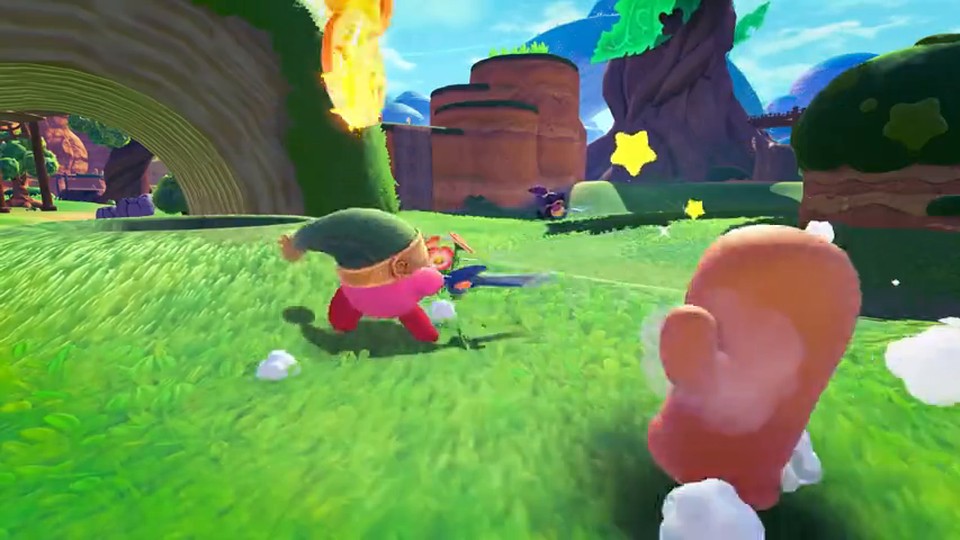} & \rf{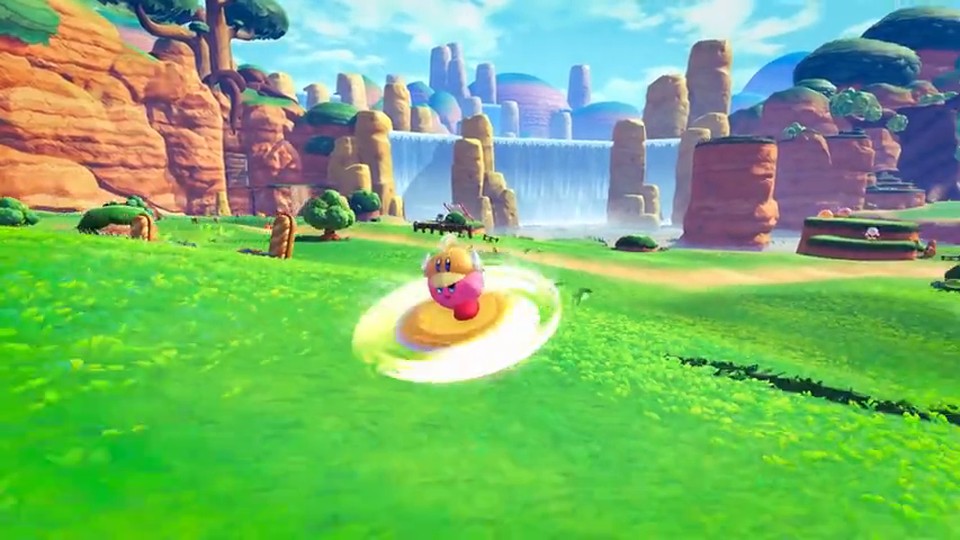} \\*
\rlab{In game}{} & \pf{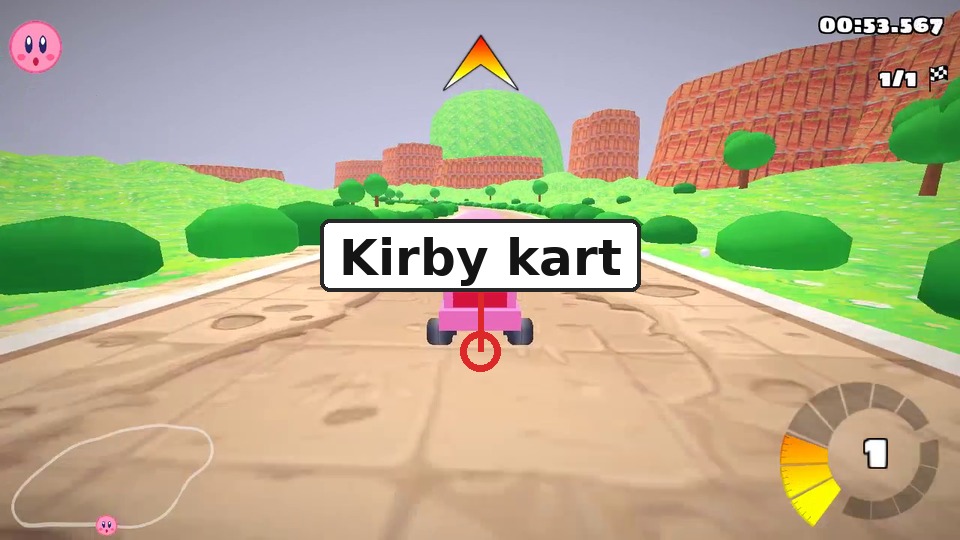}{Down the path} & \pf{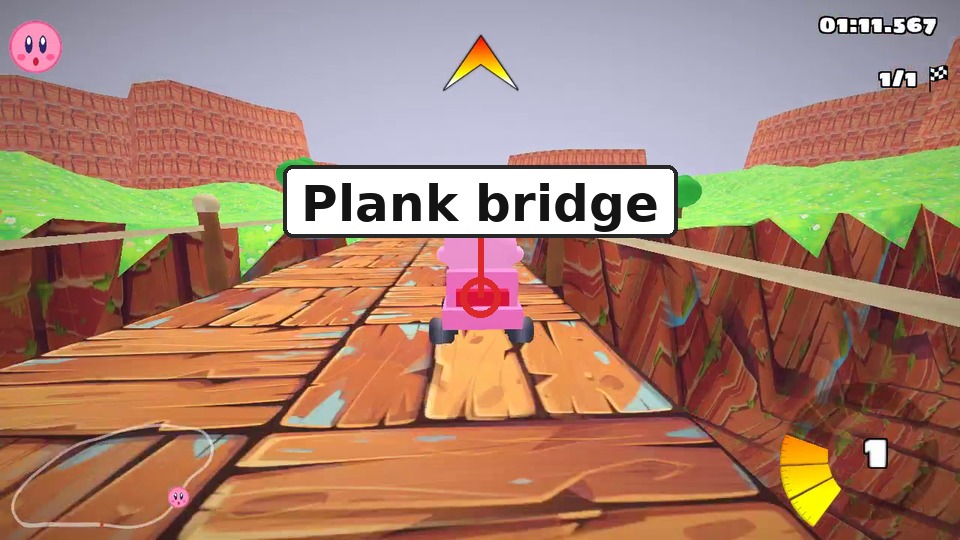}{Plank bridge} & \pf{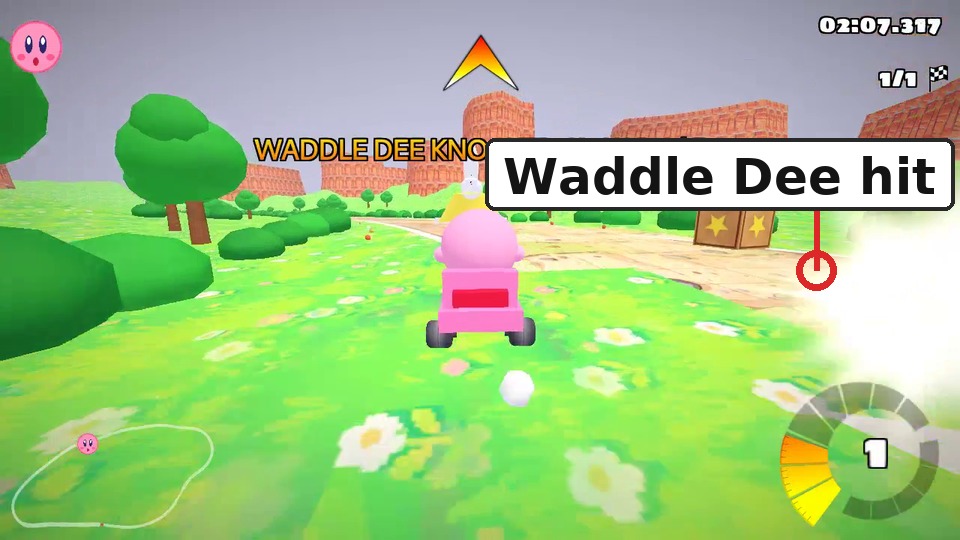}{Waddle Dees} & \pf{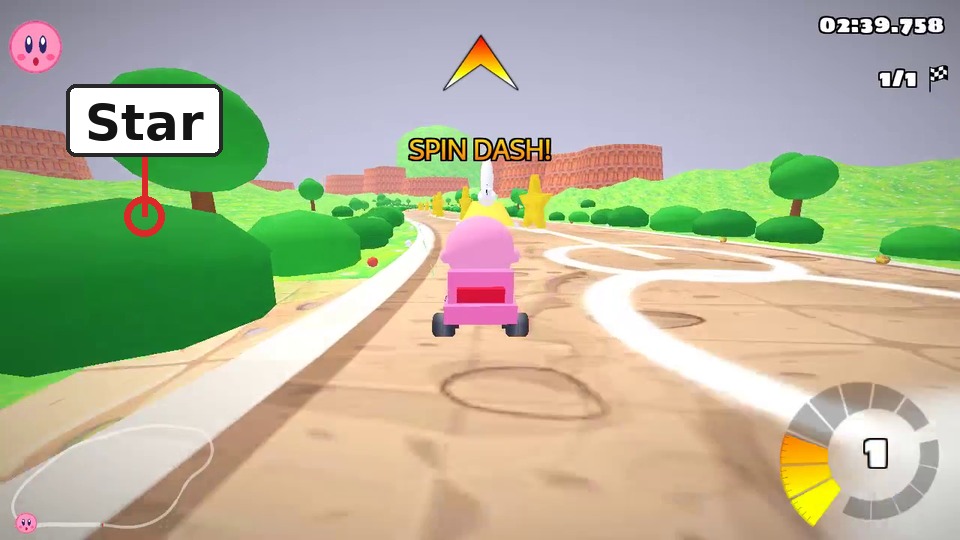}{Collecting stars}} \\[4pt]
\end{longtable}
\endgroup

\subsubsection{Commercial Games}
\label{app:commercial_game_selection}

\paragraph{Candidate Pool}
The commercial track starts from the Steam reference set described in Appendix~\ref{app:opensource_game_selection}, which also seeds the open-source track. This set draws on Steam review counts, Nexus Mods listings, and Steam Workshop entries.
We added Minecraft, which is sold outside Steam but has one of the largest modding communities.
From this set we shortlisted ten moddable commercial games that cover different genres: Minecraft, Stardew Valley, Terraria, Valheim, Cities: Skylines, The Sims~4, Slay the Spire, Civilization~VI, Kerbal Space Program, and Overcooked!~2.

\paragraph{Selection Criteria}
We selected games on three criteria.
\textit{Popularity}: the game has a large player base and an active modding community, so its modding interfaces are mature, documented, and widely used.
\textit{Genre coverage}: the selected games span distinct genres and views, so the same reference can be expressed in different ways across games.
\textit{Modding adaptability}: the game's modding interface can add the scenes, characters, events, and mechanics that our references call for, rather than only tuning numbers or swapping assets.
We also considered runtime cost, since every task requires repeated launches, playtests, and recordings in our GPU sandbox.
Games that are very heavy to install and run were deferred for now: The Sims~4 and Grand Theft Auto~V both carry large installations, require launcher or online components, and are slow to start in our sandbox.
Slay the Spire, Civilization~VI, Kerbal Space Program, and Overcooked!~2 run well, but their emphasis on card battles, turn-based strategy, spaceflight simulation, and timed cooperative cooking leaves little room for the scenes and stories in our references. We therefore did not select them.

\paragraph{Final Game List}
The five selected games cover 3D voxel building in Minecraft, 2D farming and life simulation in Stardew Valley, 2D side-view action in Terraria, 3D open-world survival in Valheim, and city building in Cities: Skylines.

Table~\ref{tab:commercial_games} lists the five commercial games. For each game it gives the key art and facts, screenshots of the unmodified game, and example extensions.

\begingroup
\small
\setlength{\tabcolsep}{4pt}
\setlength{\LTleft}{0pt}\setlength{\LTright}{0pt}
\renewcommand{\arraystretch}{1.3}
\begin{longtable}{@{}p{2.3cm}>{\raggedright\arraybackslash}p{2.5cm}>{\raggedright\arraybackslash}p{1.95cm}>{\centering\arraybackslash}p{0.75cm}>{\raggedright\arraybackslash}p{3.55cm}>{\raggedright\arraybackslash}p{2.75cm}>{\raggedright\arraybackslash}p{0.85cm}@{}}
\caption{\textbf{Commercial target games.} Game information, original-game screenshots, and two example extensions per game from development trials. Reference frames are shown above the corresponding in-game results. Task counts refer to Construction tasks.}
\label{tab:commercial_games}\\
\toprule
\textbf{Key art} & \textbf{Game} & \textbf{Genre} & \textbf{View} & \textbf{What the player does} & \textbf{Modding surface} & \textbf{Tasks} \\
\midrule
\endfirsthead
\multicolumn{7}{@{}l}{\small\textit{Table~\ref{tab:commercial_games} continued}} \\
\toprule
\textbf{Key art} & \textbf{Game} & \textbf{Genre} & \textbf{View} & \textbf{What the player does} & \textbf{Modding surface} & \textbf{Tasks} \\
\midrule
\endhead
\bottomrule
\endlastfoot
\gthumb{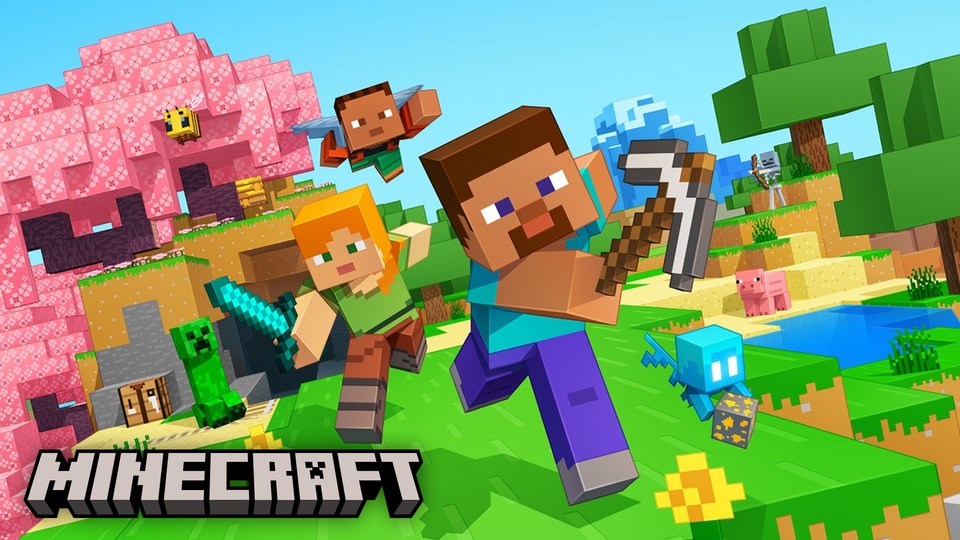} & \gname{Minecraft}{Mojang \textperiodcentered{} Java 26.3 \textperiodcentered{} Fabric 0.19.5} & \gtag{Survival / sandbox} & \vtag{3} & Mines and places blocks, crafts tools, builds and survives mobs and nights in a procedurally generated voxel world. & Fabric Java mods for blocks, items, entities, world generation; data packs & 13 \\*
\multicolumn{7}{@{}l@{}}{\rule{0pt}{12pt}\includegraphics[width=0.995\textwidth]{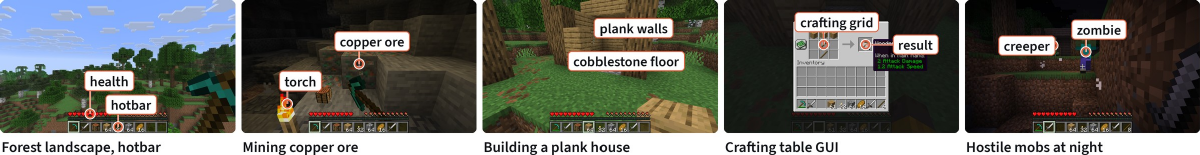}} \\[2pt]
\extlabel
\extrow{\multicolumn{5}{@{}l}{\scriptsize\textbf{Novel}: the Grand View Garden from \textit{Dream of the Red Chamber}} \\*
\rlab{Reference}{Preview pictures drawn from the novel} & \rf{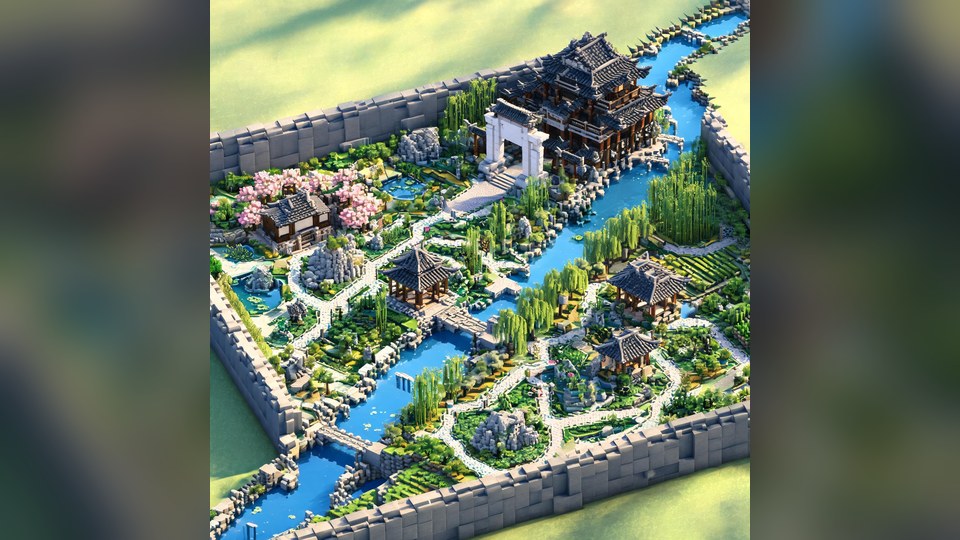} & \rf{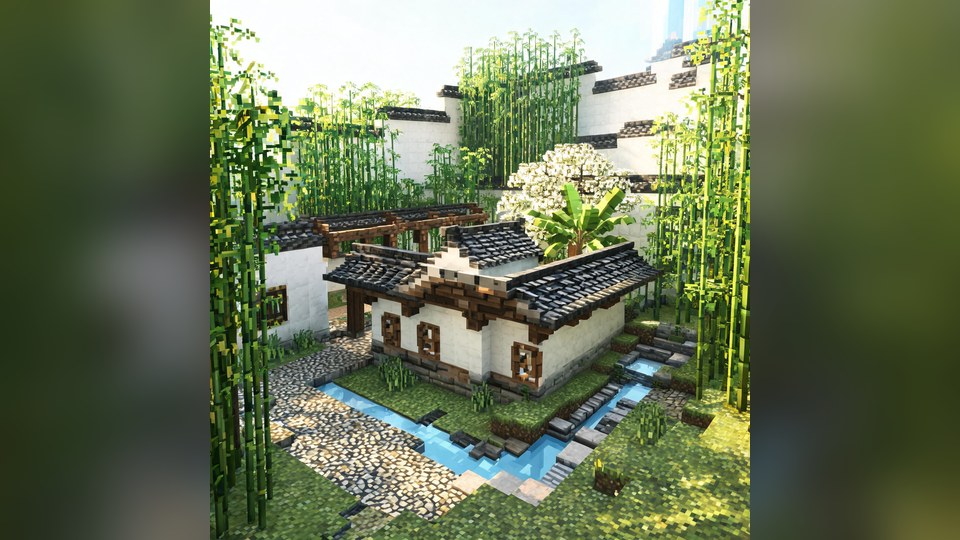} & \rf{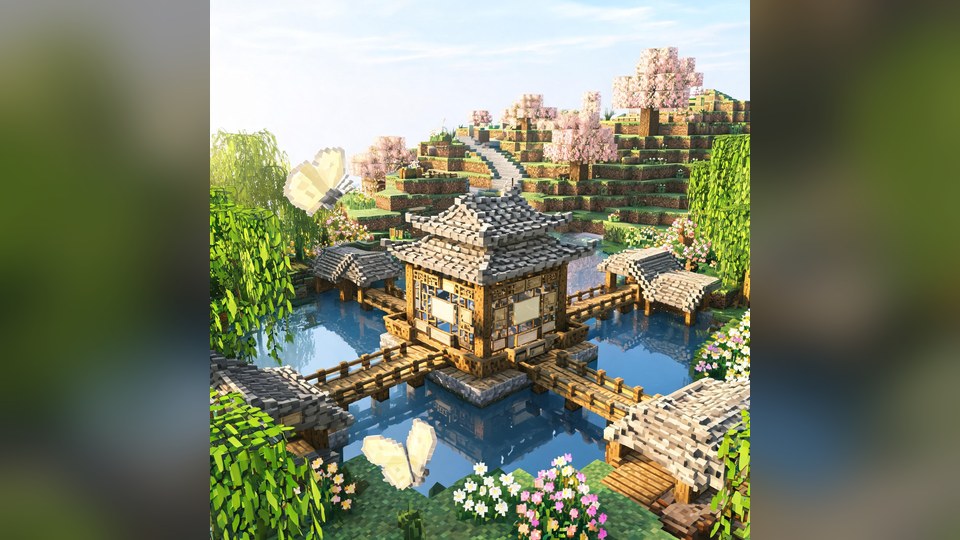} & \rf{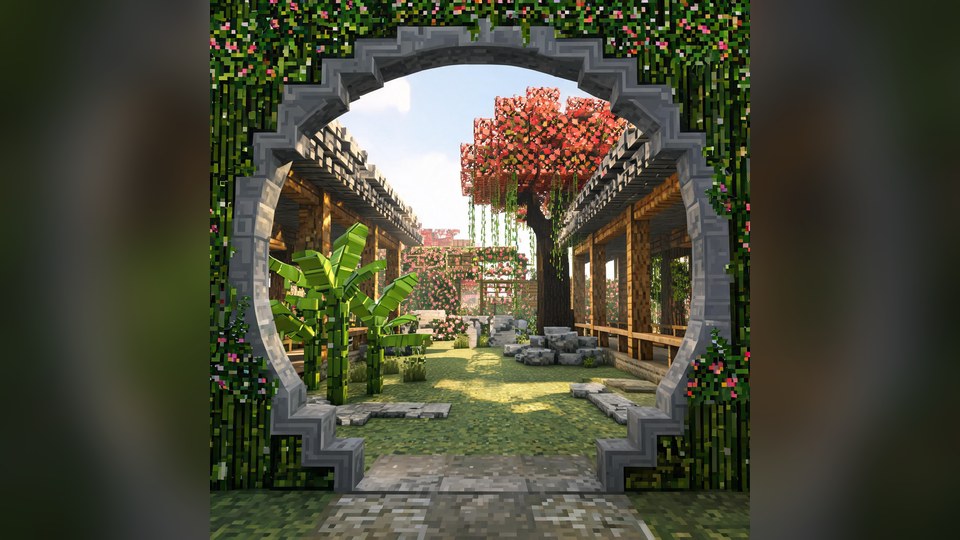} \\*
\rlab{In game}{} & \pf{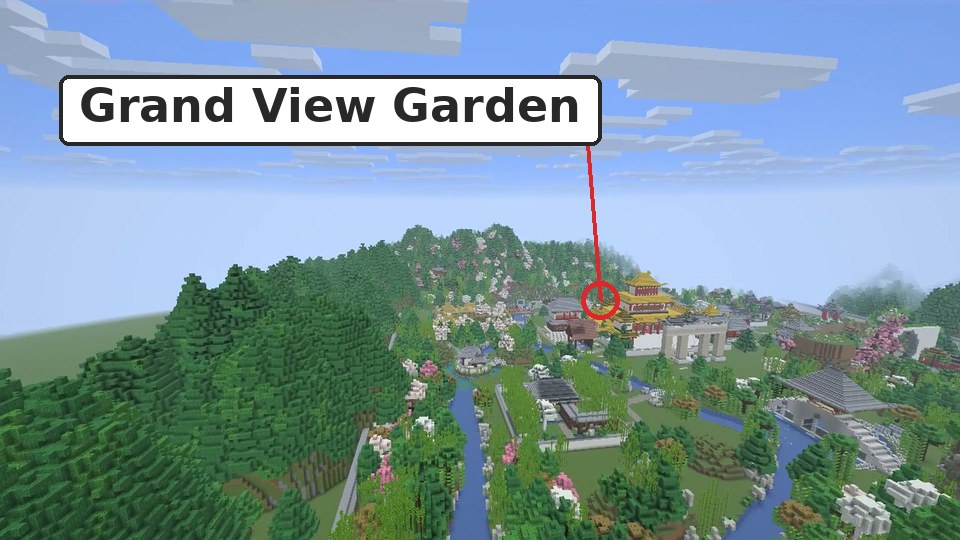}{The whole garden} & \pf{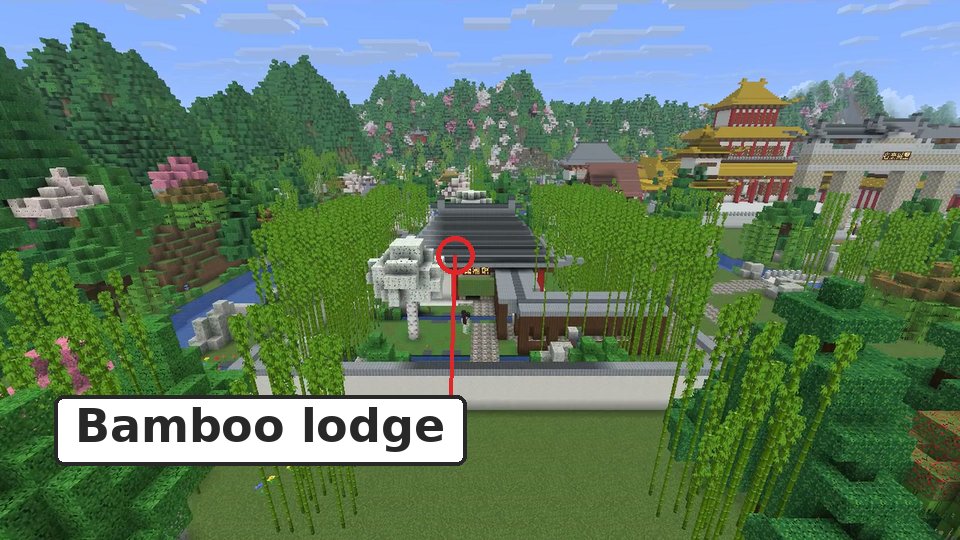}{Xiaoxiang bamboo lodge} & \pf{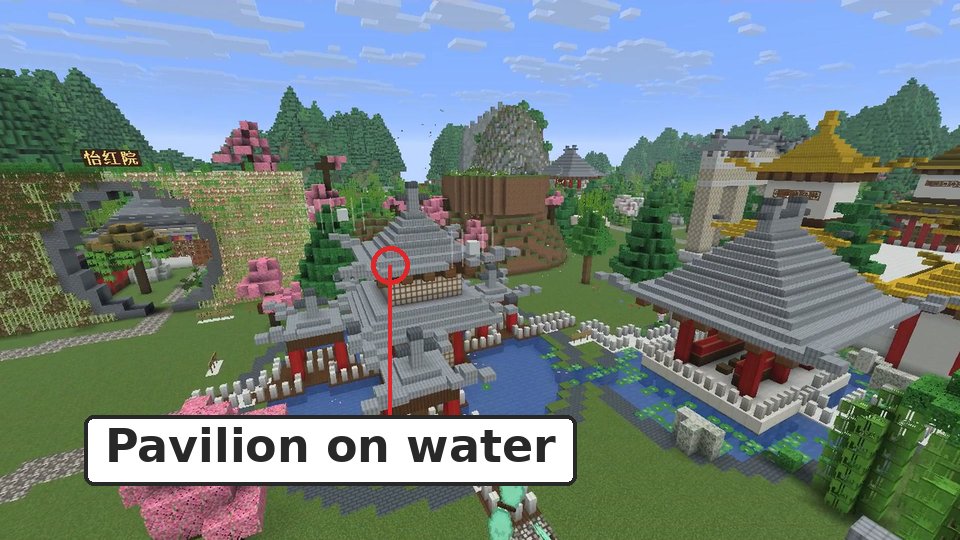}{Dicui pavilion on water} & \pf{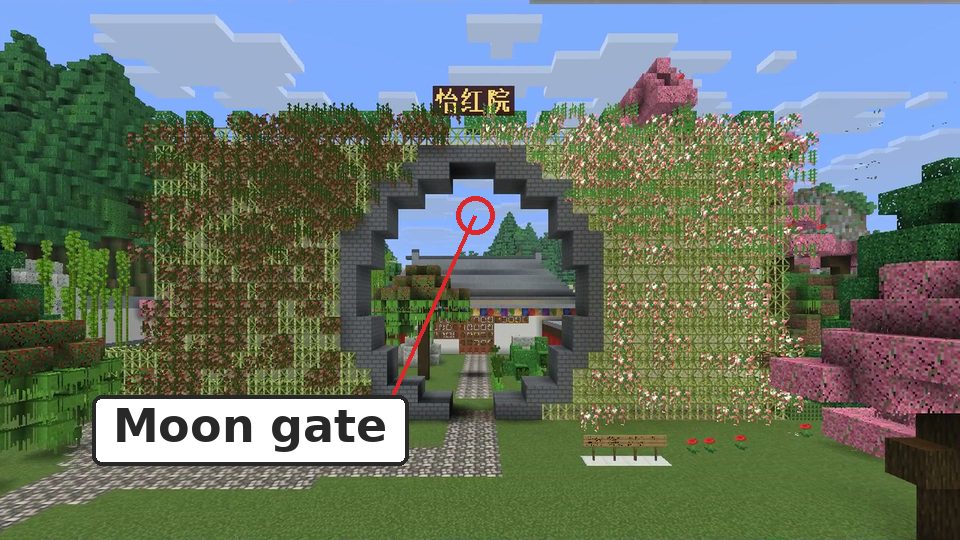}{Yihong moon gate}} \\[2pt]
\extrow{\multicolumn{5}{@{}l}{\scriptsize\textbf{Film}: Hogwarts from the lake to the common room} \\*
\rlab{Reference}{Harry Potter and the Philosopher's Stone} & \rf{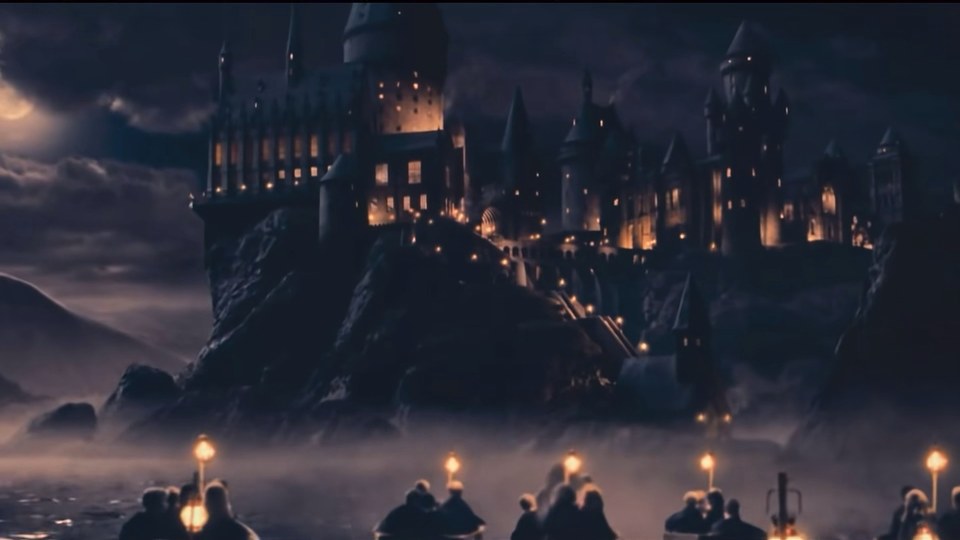} & \rf{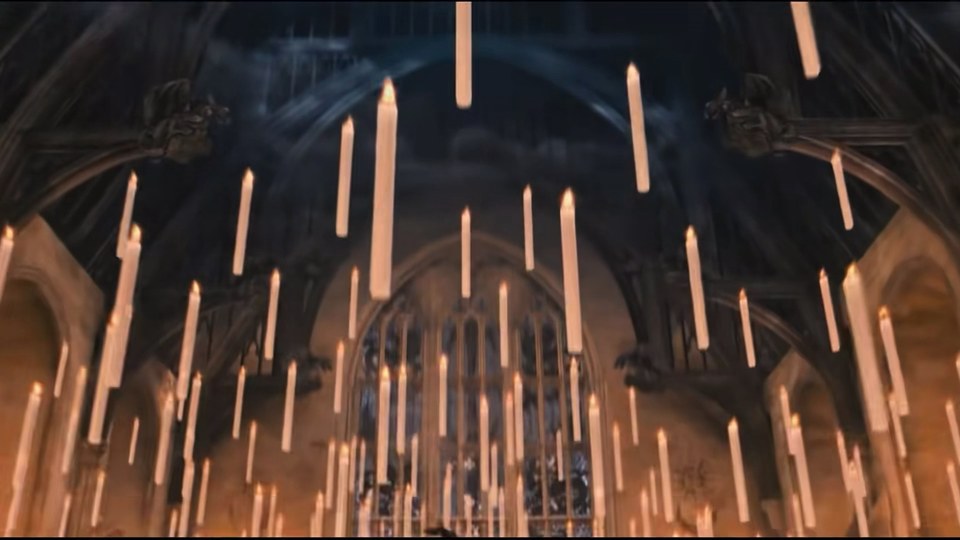} & \rf{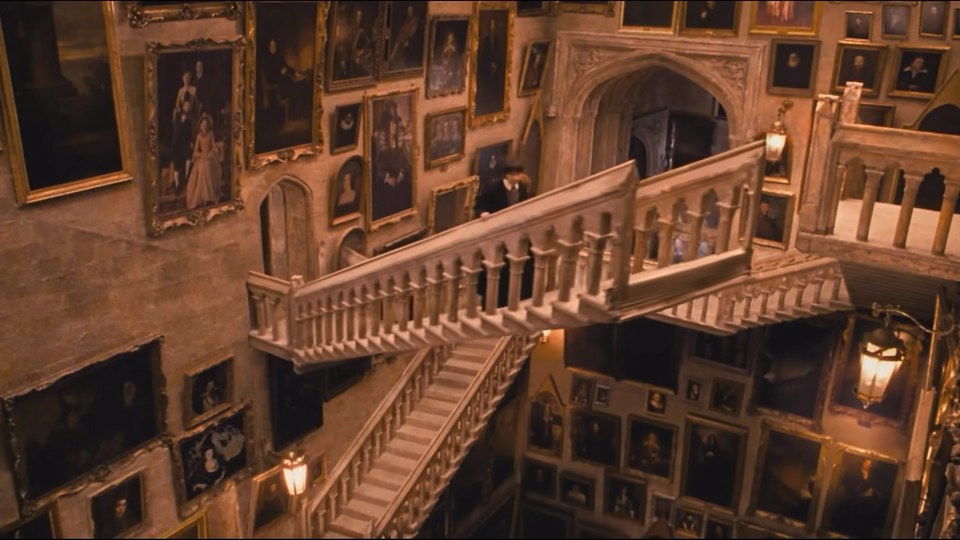} & \rf{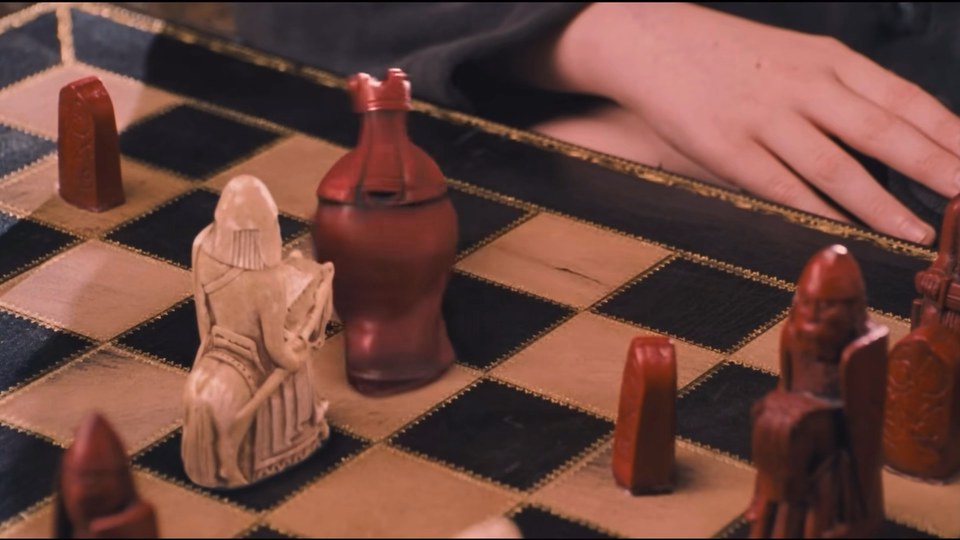} \\*
\rlab{In game}{} & \pf{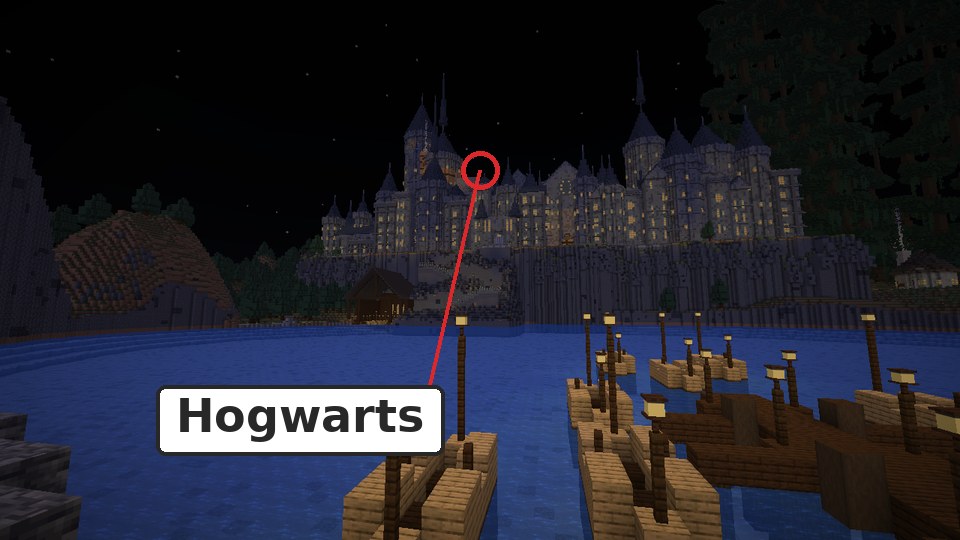}{Castle over the lake} & \pf{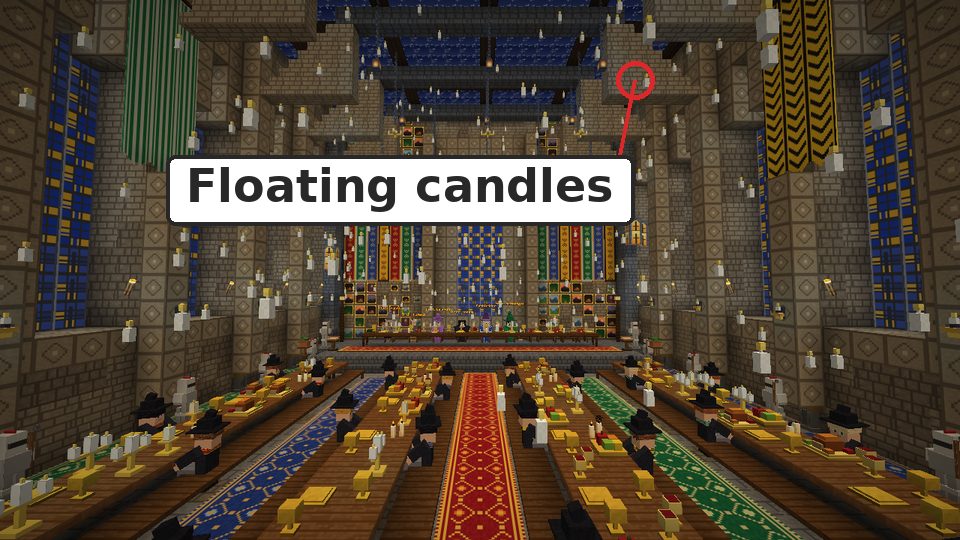}{Great Hall candles} & \pf{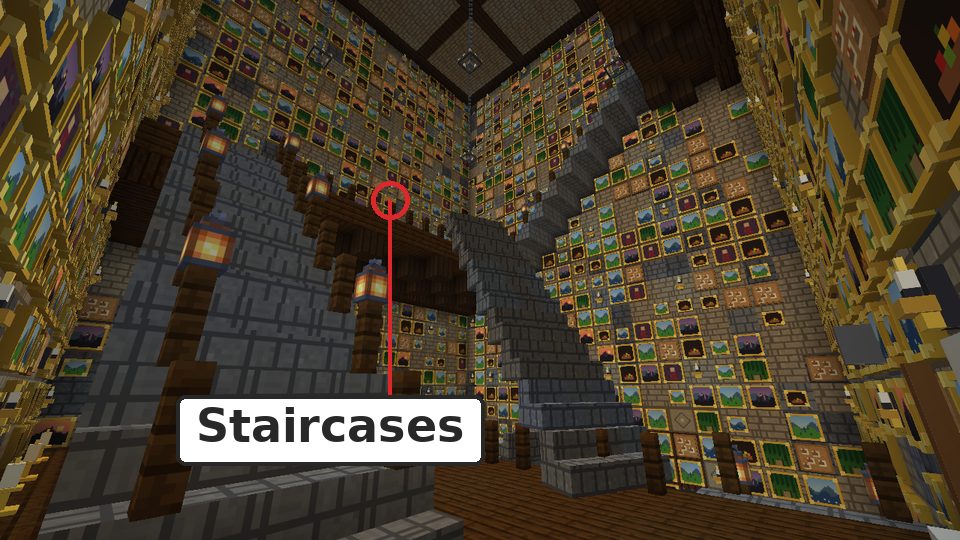}{Moving staircases} & \pf{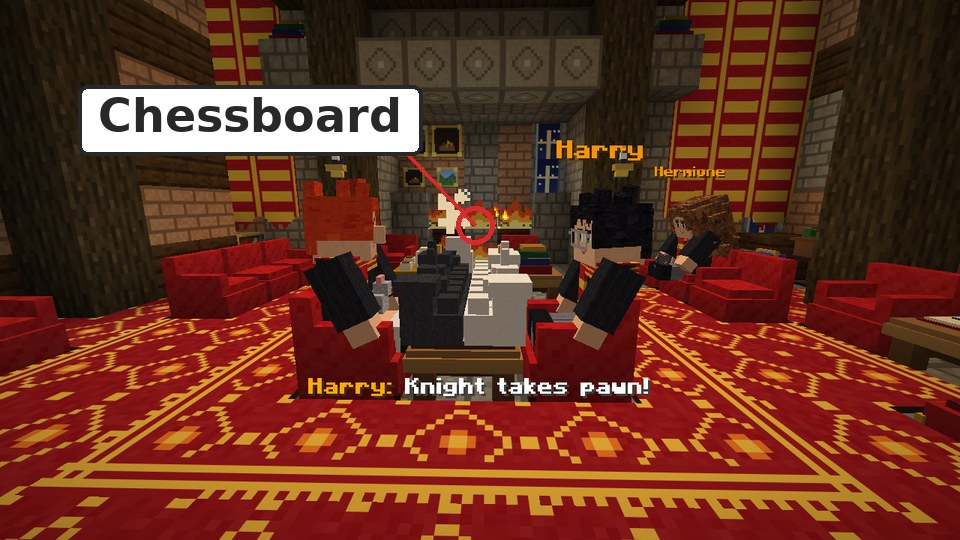}{Wizard's chess}} \\[4pt]
\midrule
\gthumb{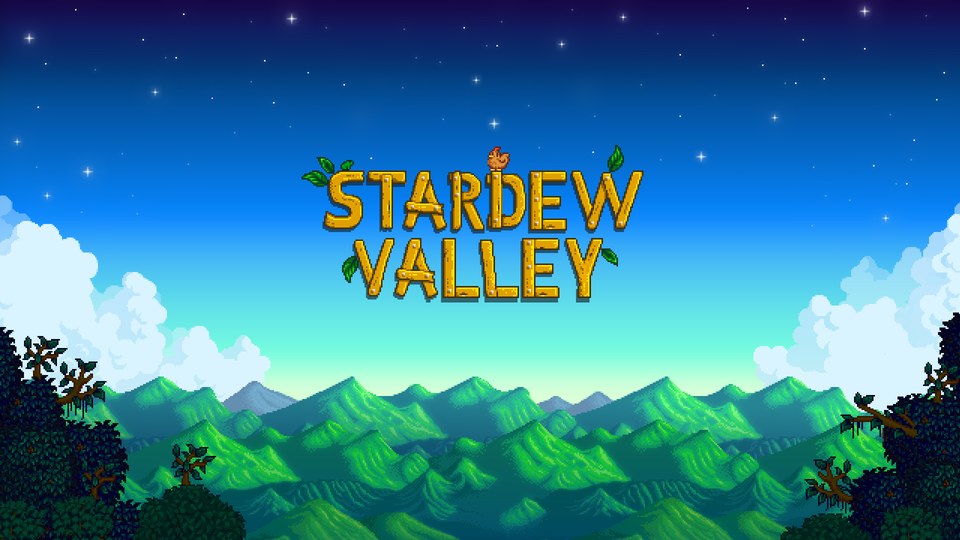} & \gname{Stardew Valley}{ConcernedApe \textperiodcentered{} 1.6.15 \textperiodcentered{} SMAPI 4.5.2} & \gtag{Farming / life sim} & \vtag{2} & Grows crops and raises animals on a farm, befriends villagers in town and explores the mines. & SMAPI C\# mods, Content Patcher \mf{.json} packs & 13 \\*
\multicolumn{7}{@{}l@{}}{\rule{0pt}{12pt}\includegraphics[width=0.995\textwidth]{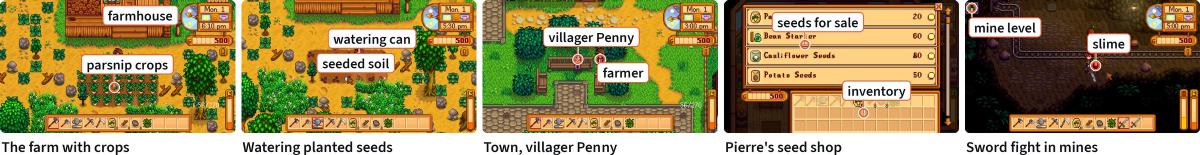}} \\[2pt]
\extlabel
\extrow{\multicolumn{5}{@{}l}{\scriptsize\textbf{Game}: an \textit{Overcooked}-style kitchen rush behind the Saloon} \\*
\rlab{Reference}{Official Overcooked! 2 screenshots} & \rf{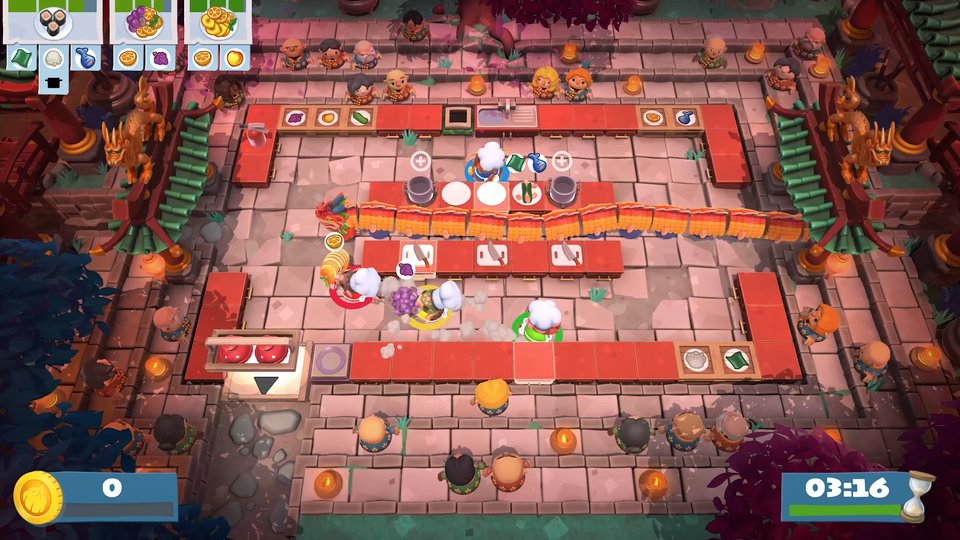} & \rf{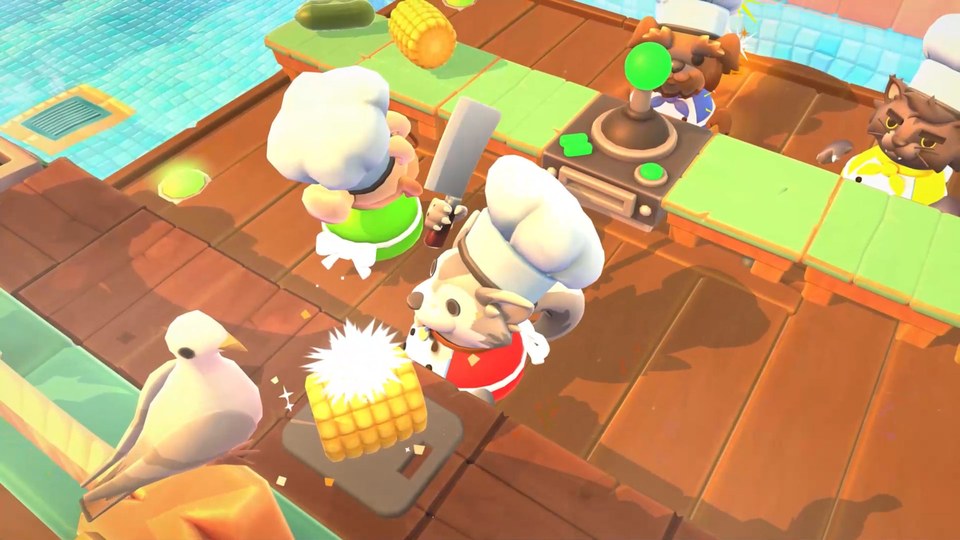} & \rf{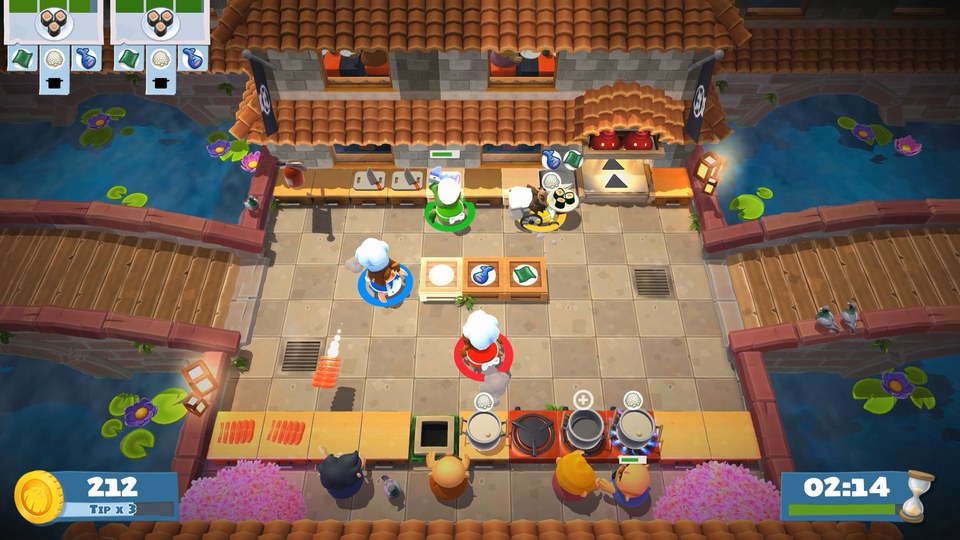} & \rf{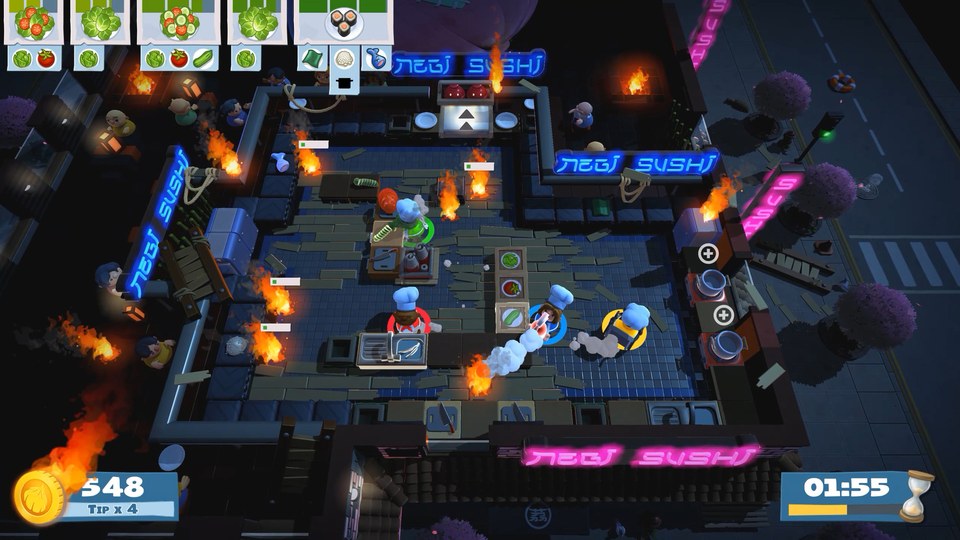} \\*
\rlab{In game}{} & \pf{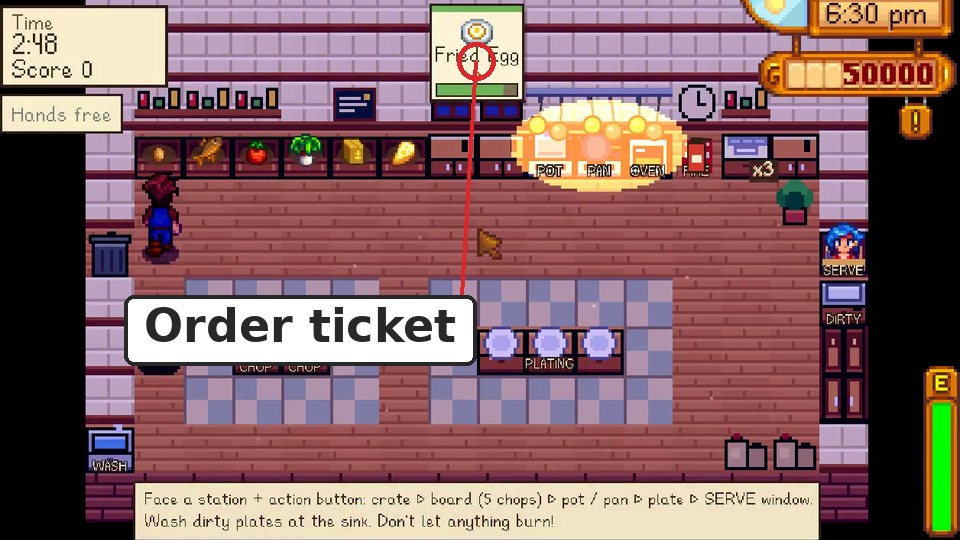}{Order tickets} & \pf{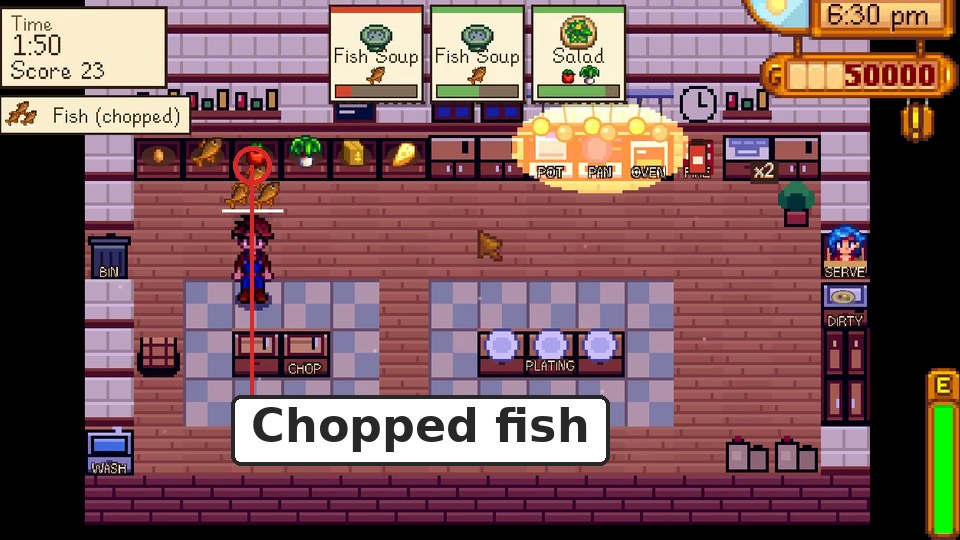}{Chopping the fish} & \pf{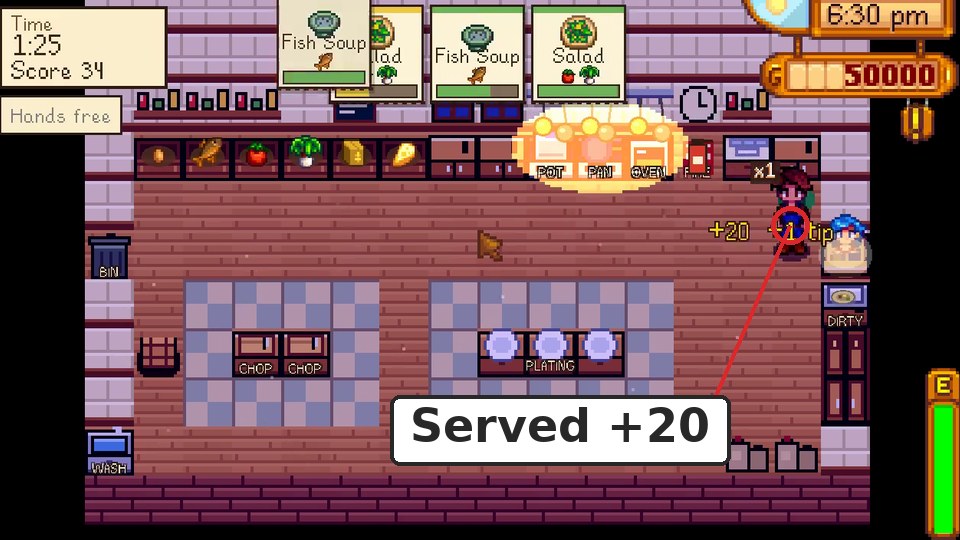}{Serving at the window} & \pf{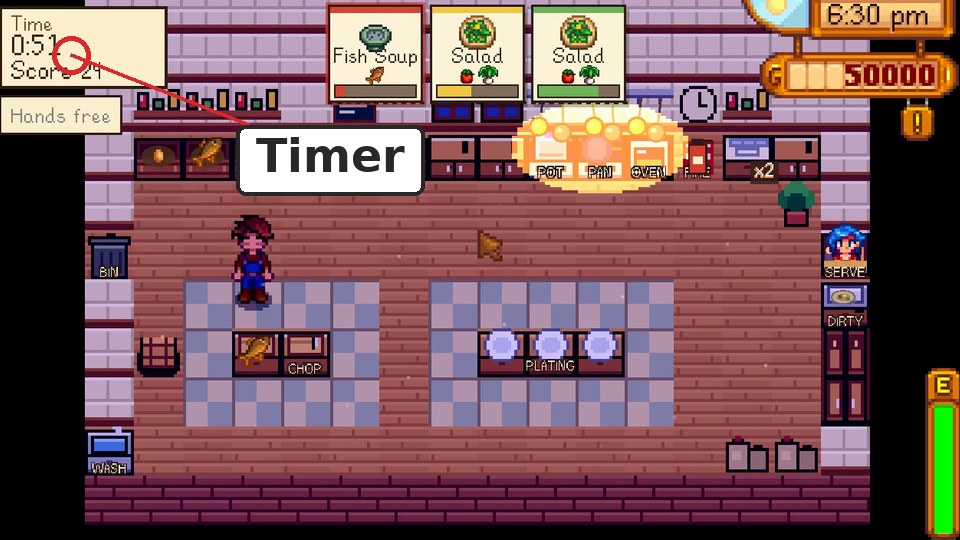}{Round timer and score}} \\[4pt]
\midrule
\gthumb{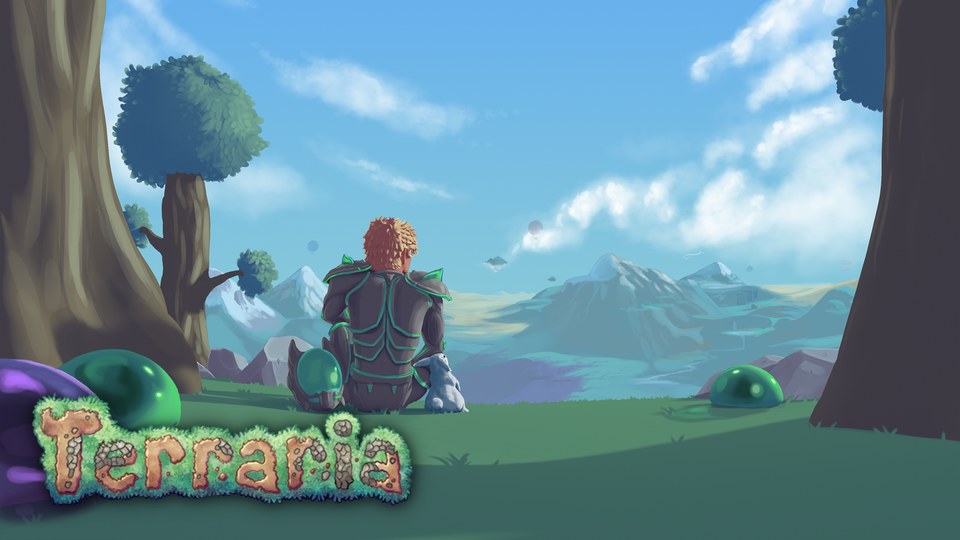} & \gname{Terraria}{Re-Logic \textperiodcentered{} tModLoader} & \gtag{Action / sandbox} & \vtag{2} & Digs through a 2D world, builds houses, crafts gear and fights enemies and bosses. & tModLoader C\# mods: items, NPCs, tiles, world generation & 7 \\*
\multicolumn{7}{@{}l@{}}{\rule{0pt}{12pt}\includegraphics[width=0.995\textwidth]{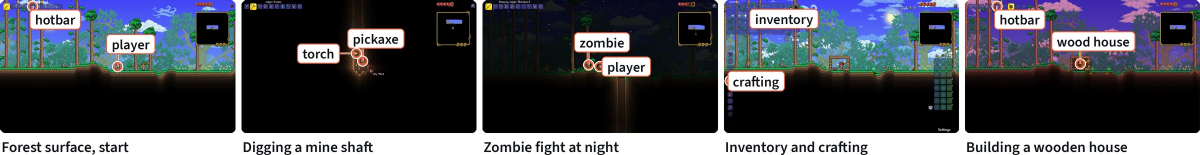}} \\[2pt]
\extlabel
\extrow{\multicolumn{5}{@{}l}{\scriptsize\textbf{Image}: the Derinkuyu underground city, from the entrance to the air shaft} \\*
\rlab{Reference}{Derinkuyu underground city photos} & \rf{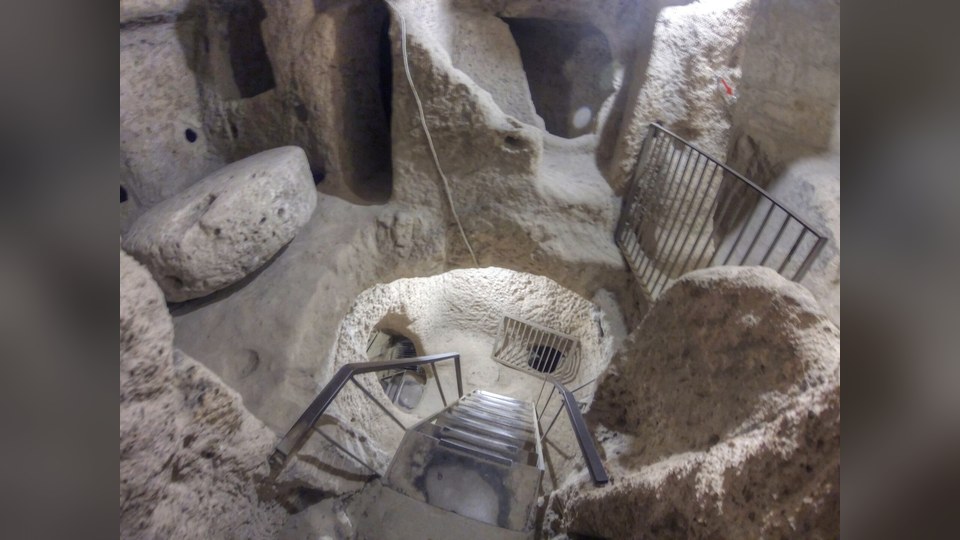} & \rf{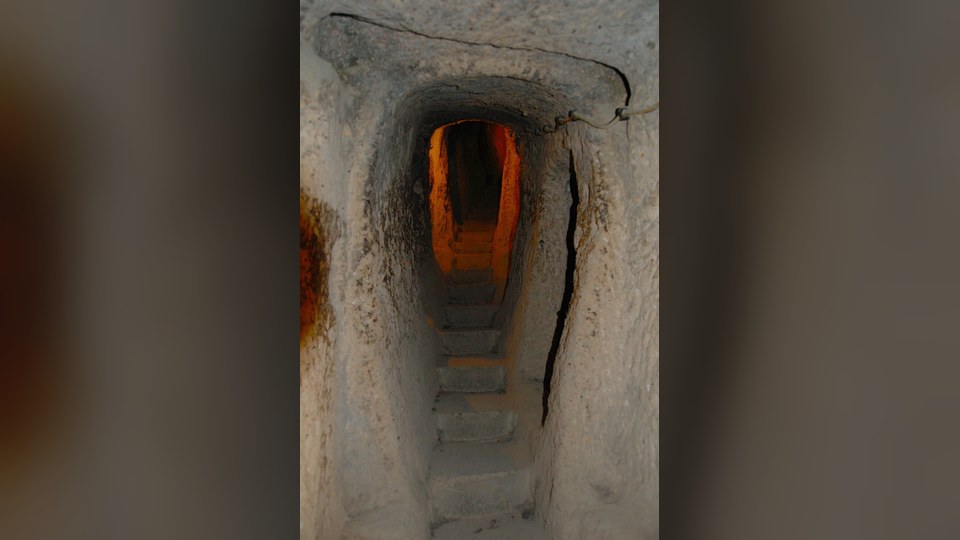} & \rf{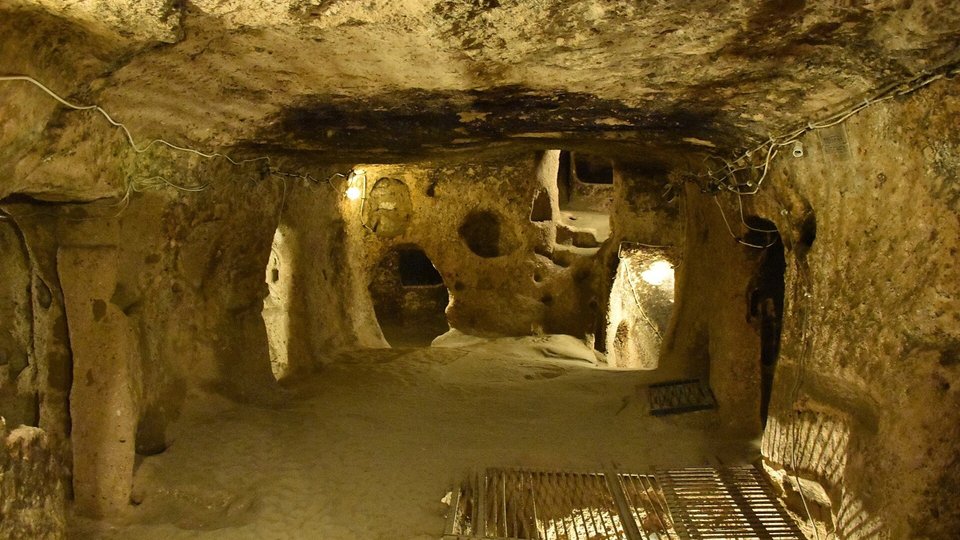} & \rf{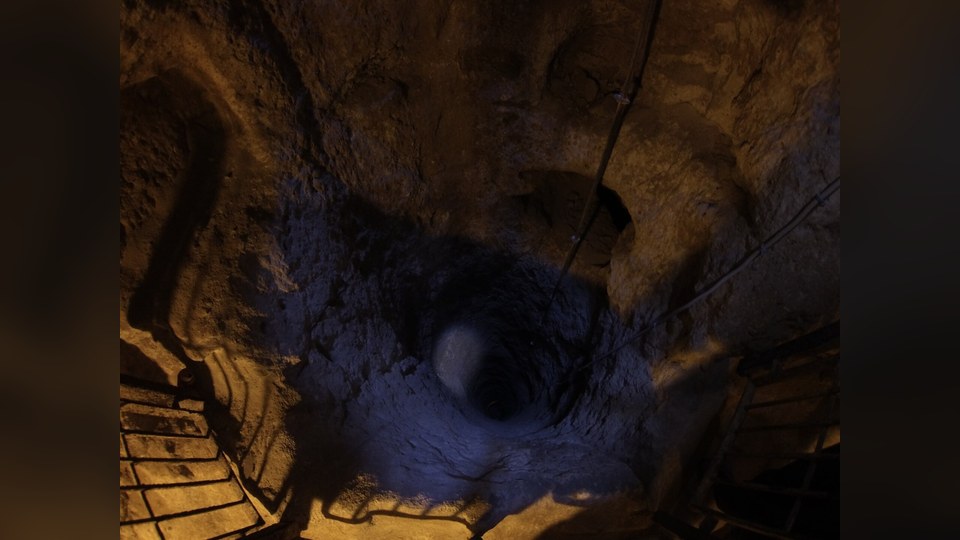} \\*
\rlab{In game}{} & \pf{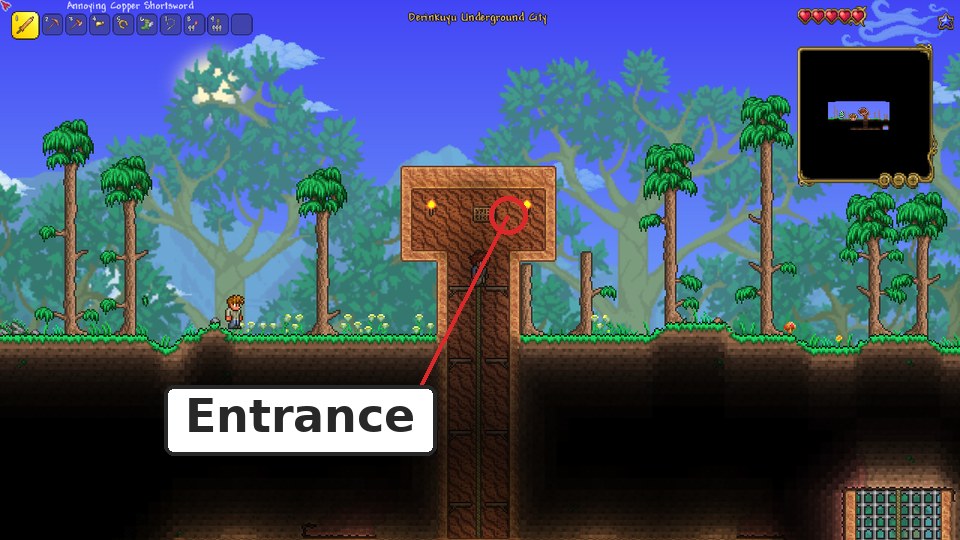}{Entrance on the surface} & \pf{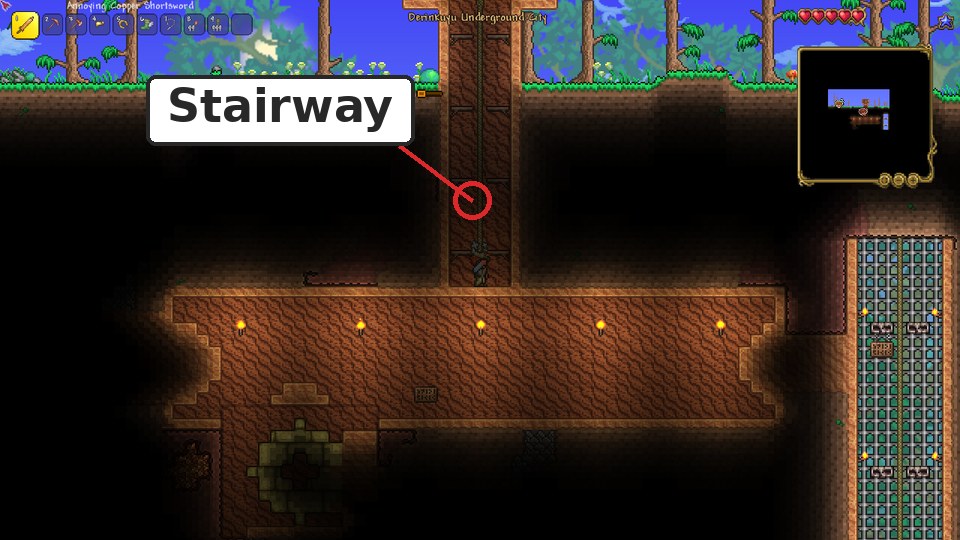}{Stairway down} & \pf{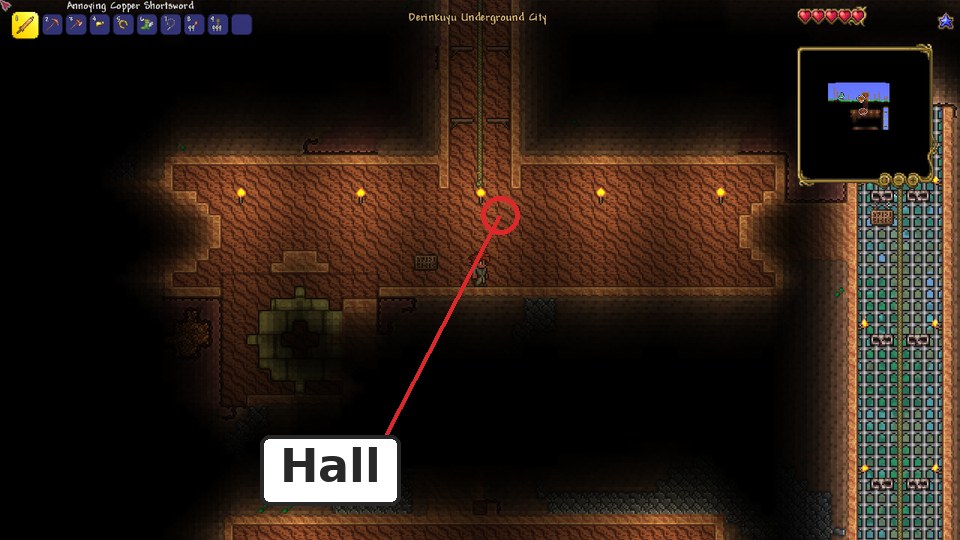}{Carved hall} & \pf{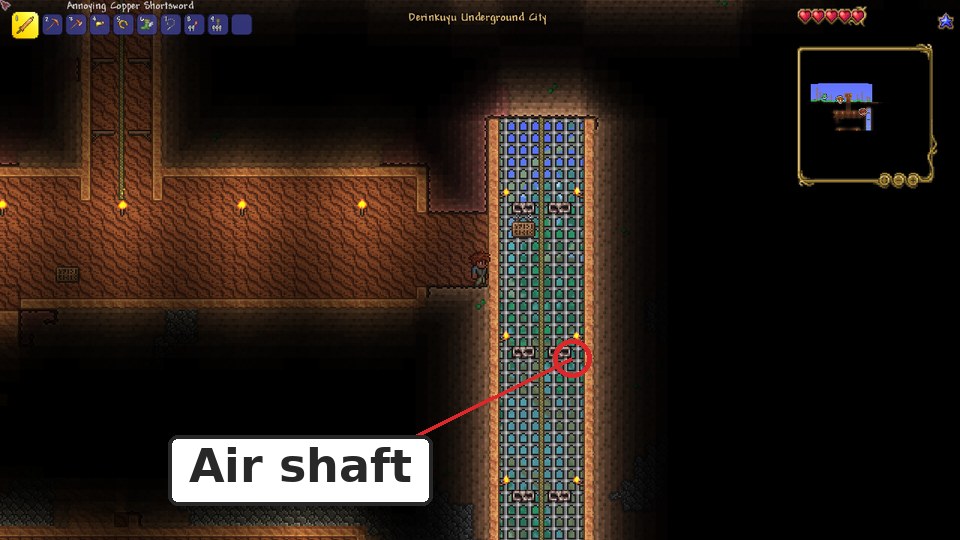}{Ventilation shaft}} \\[4pt]
\midrule
\gthumb{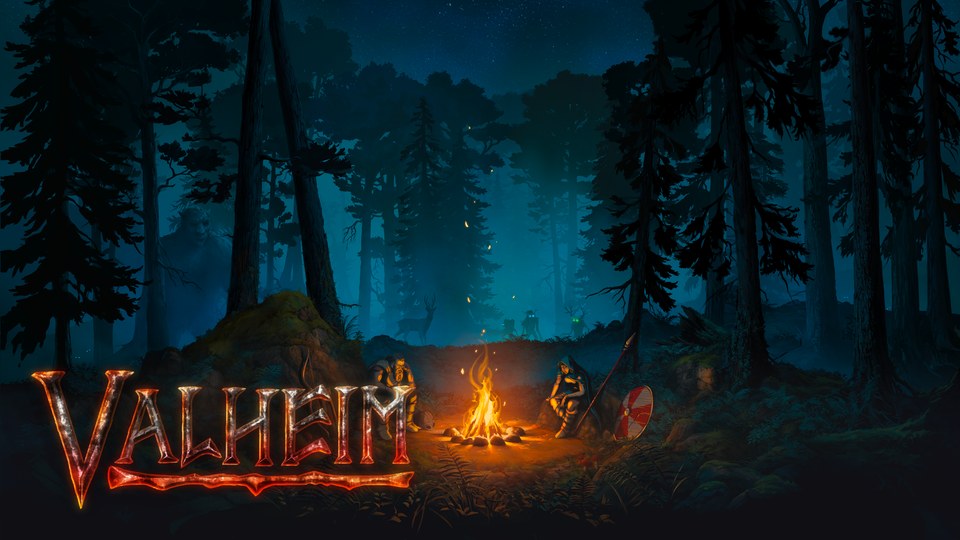} & \gname{Valheim}{Iron Gate \textperiodcentered{} BepInEx 5.4} & \gtag{Survival / exploration} & \vtag{3} & Explores a Norse world, gathers resources, builds bases with a hammer and fights creatures. & BepInEx C\# plugins with Harmony patches and assets & 8 \\*
\multicolumn{7}{@{}l@{}}{\rule{0pt}{12pt}\includegraphics[width=0.995\textwidth]{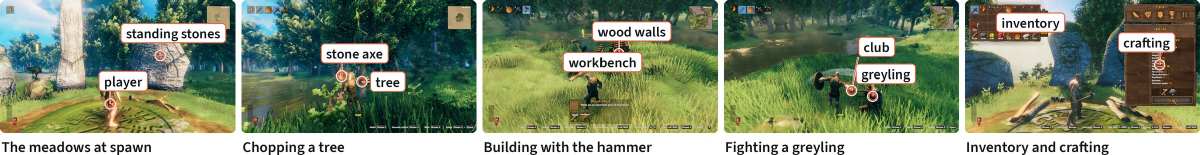}} \\[2pt]
\extlabel
\extrow{\multicolumn{5}{@{}l}{\scriptsize\textbf{Sports}: a timed downhill run with splits and standings} \\*
\rlab{Reference}{Olympic women's downhill} & \rf{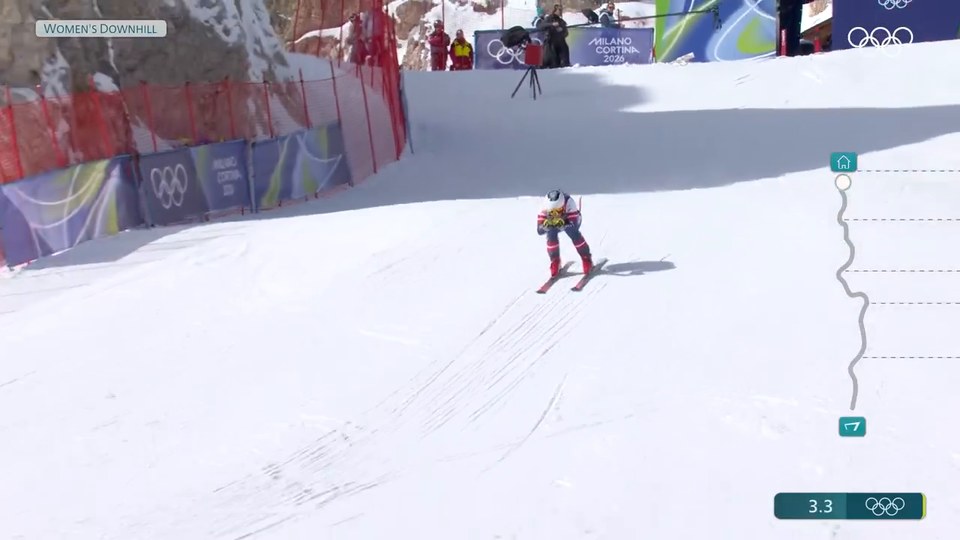} & \rf{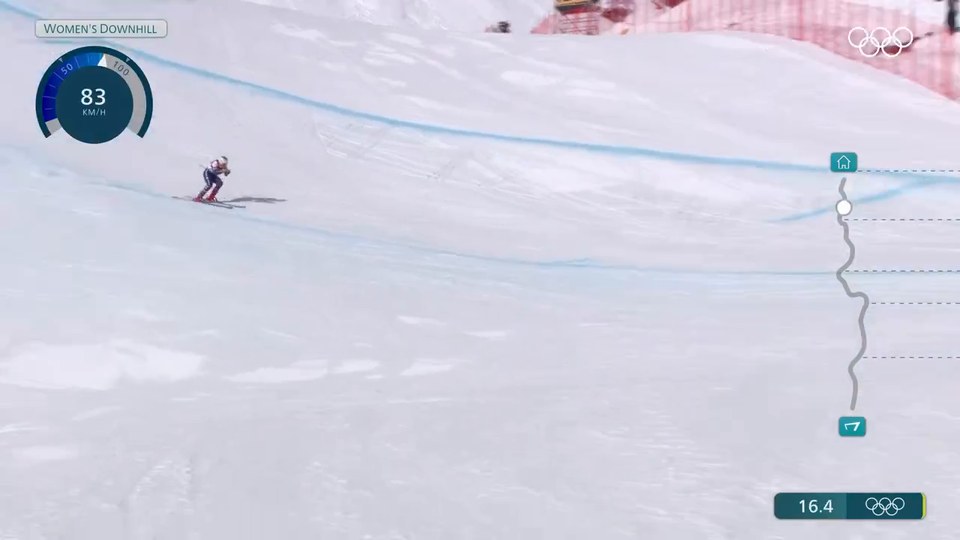} & \rf{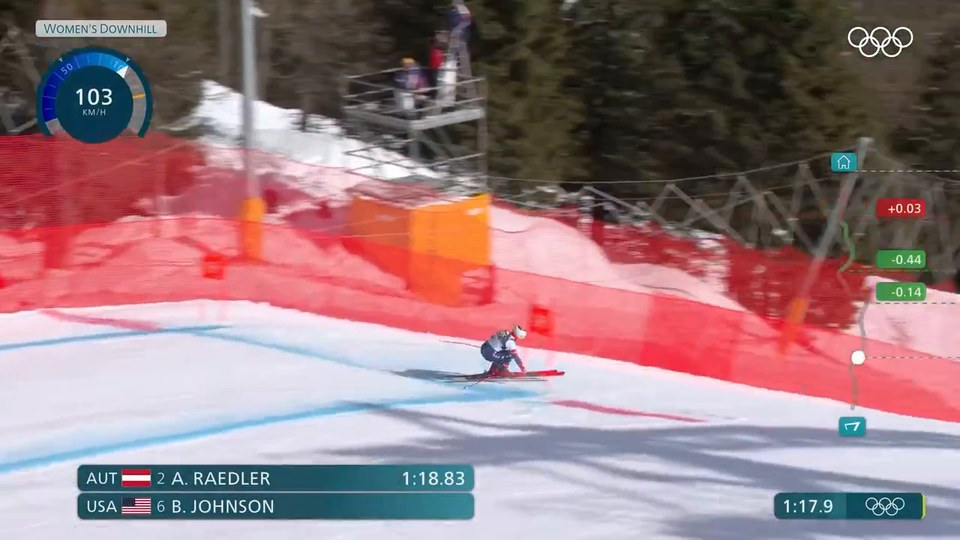} & \rf{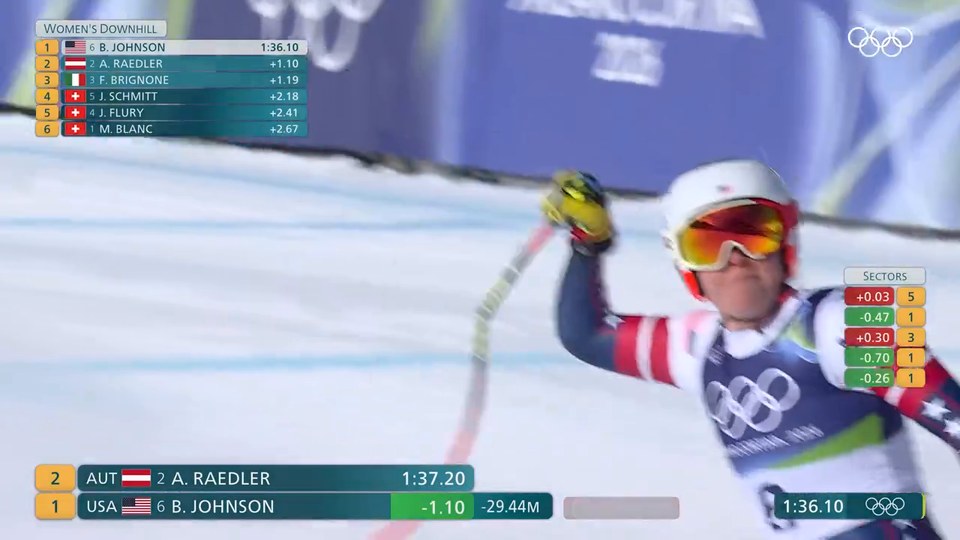} \\*
\rlab{In game}{} & \pf{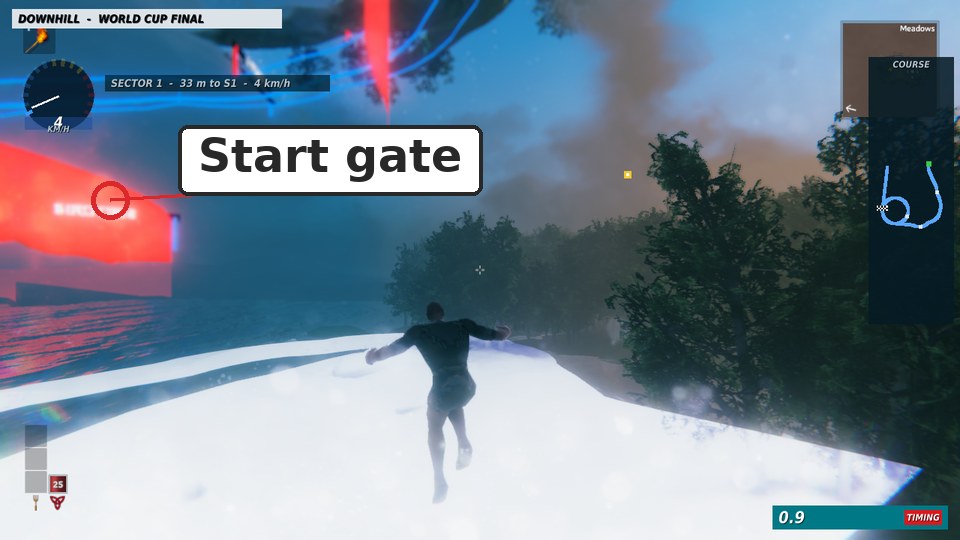}{Out of the start} & \pf{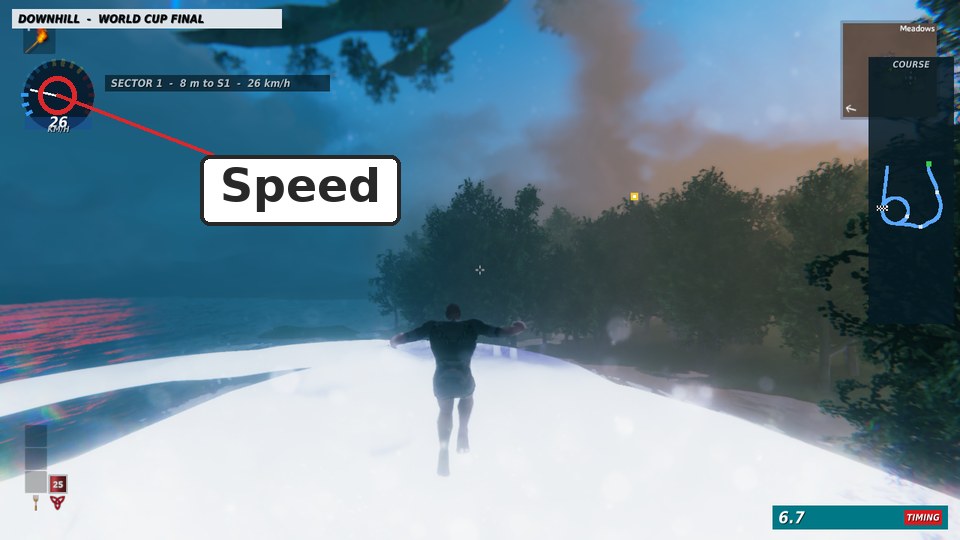}{Live speed dial} & \pf{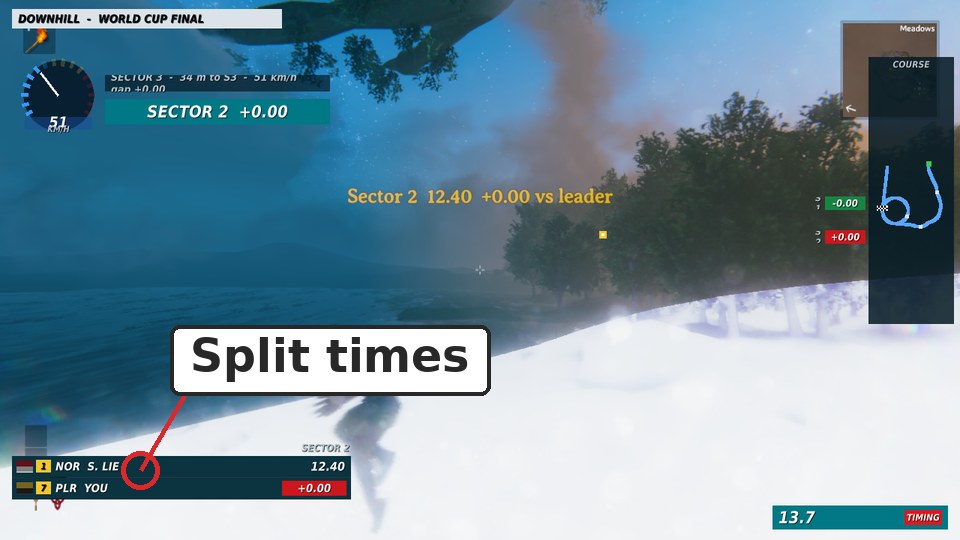}{Sector split times} & \pf{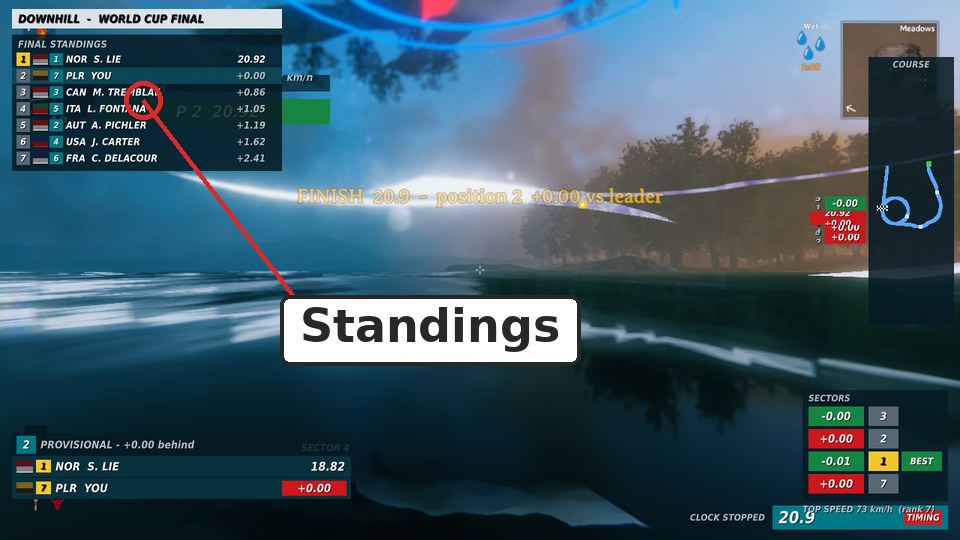}{Final standings}} \\[4pt]
\midrule
\gthumb{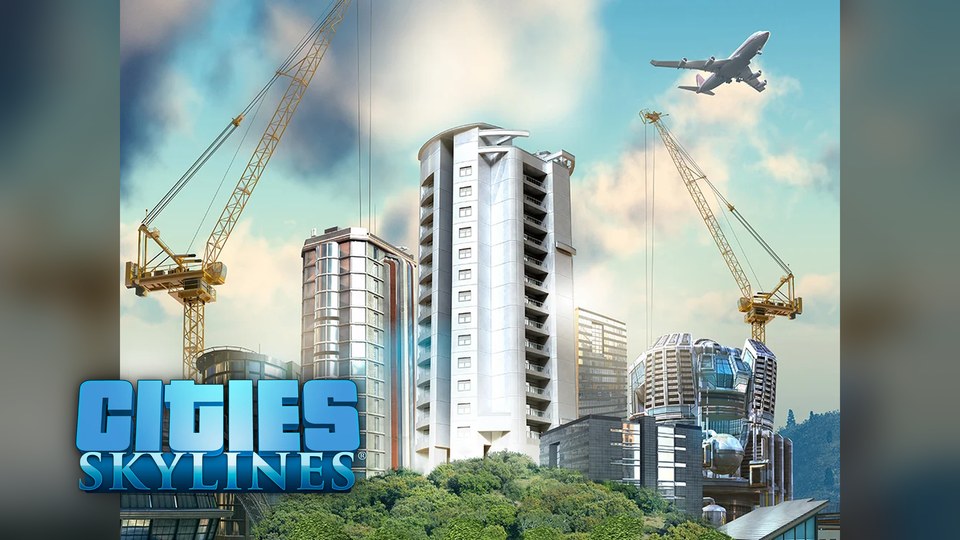} & \gname{Cities: Skylines}{Colossal Order / Paradox \textperiodcentered{} C\# mods} & \gtag{City builder} & \vtag{3} & Lays out roads and zones, provides services and manages traffic and budget as a city grows. & C\# mods using the ICities API and assets & 7 \\*
\multicolumn{7}{@{}l@{}}{\rule{0pt}{12pt}\includegraphics[width=0.995\textwidth]{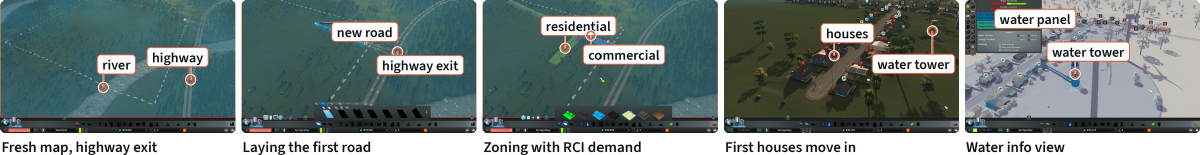}} \\[2pt]
\extlabel
\extrow{\multicolumn{5}{@{}l}{\scriptsize\textbf{Photos}: the San Diego coast from the harbor to La Jolla Cove} \\*
\rlab{Reference}{Landmark photos from Wikimedia Commons} & \rf{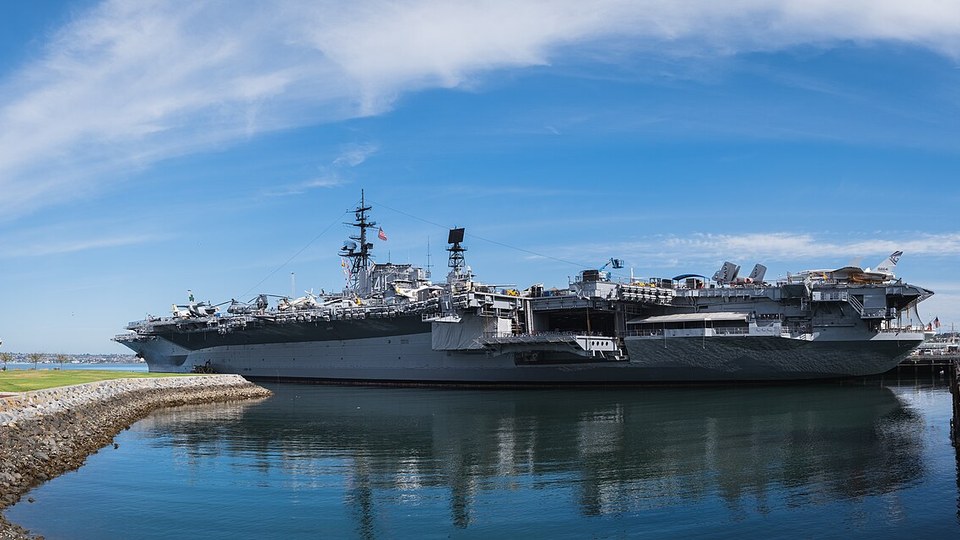} & \rf{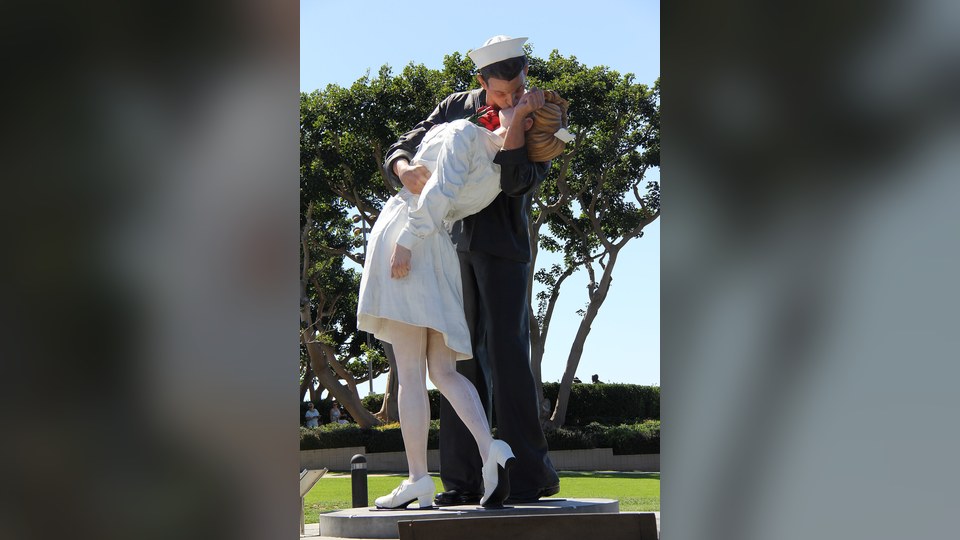} & \rf{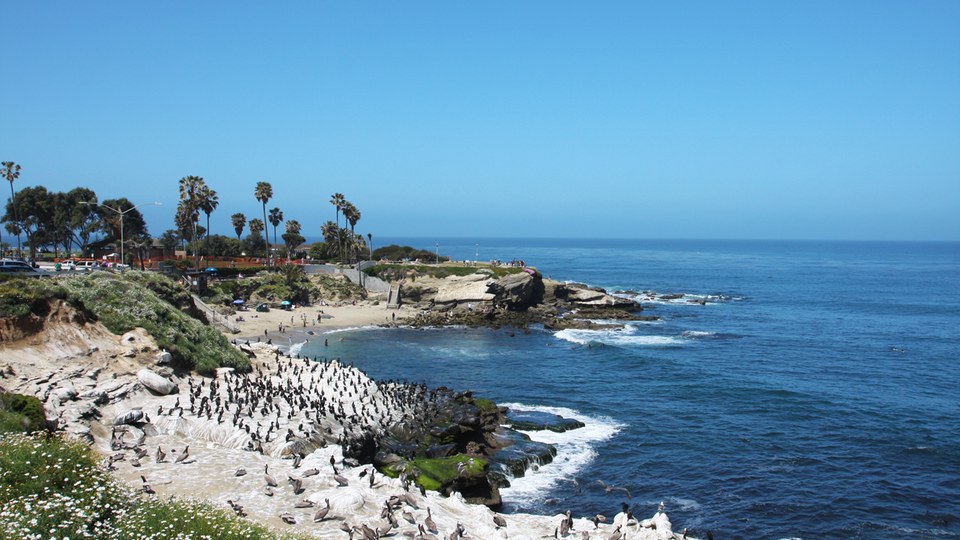} & \rf{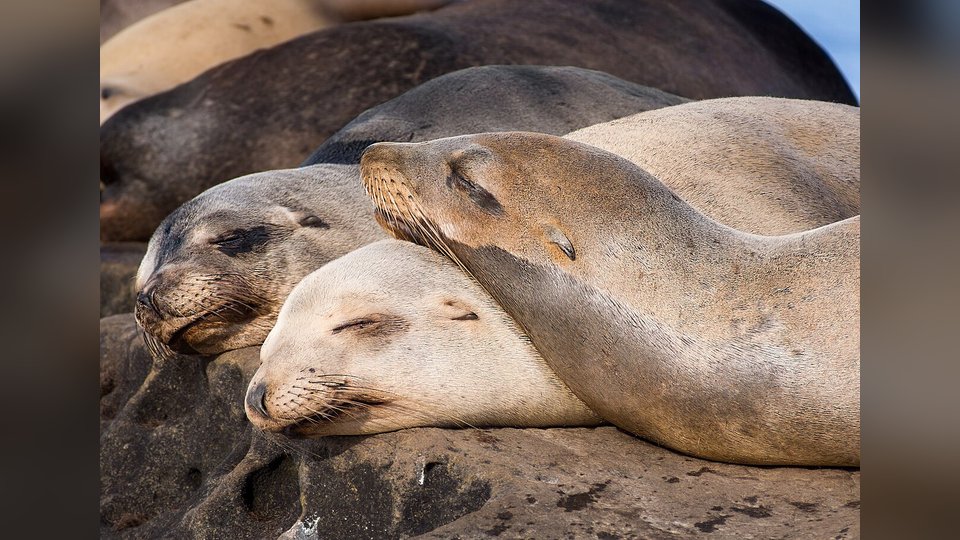} \\*
\rlab{In game}{} & \pf{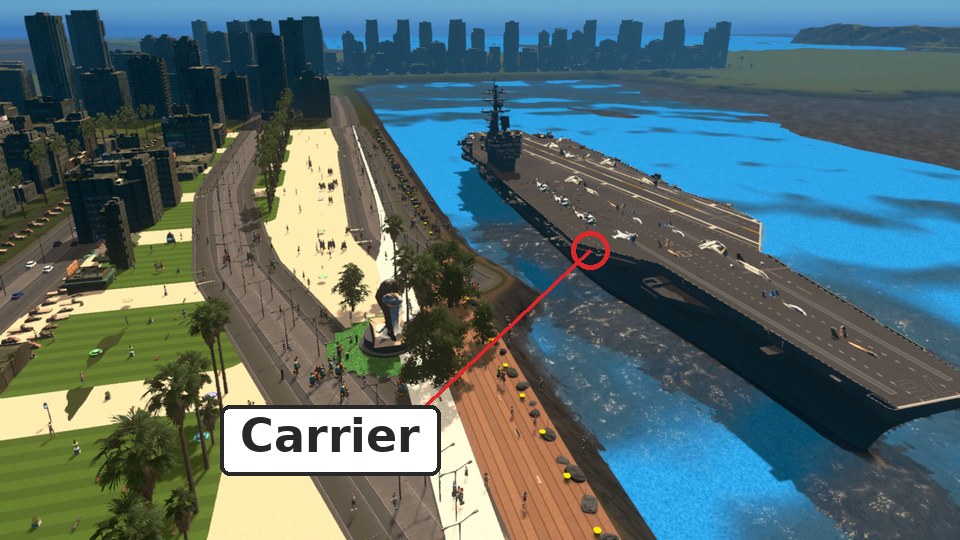}{USS Midway carrier} & \pf{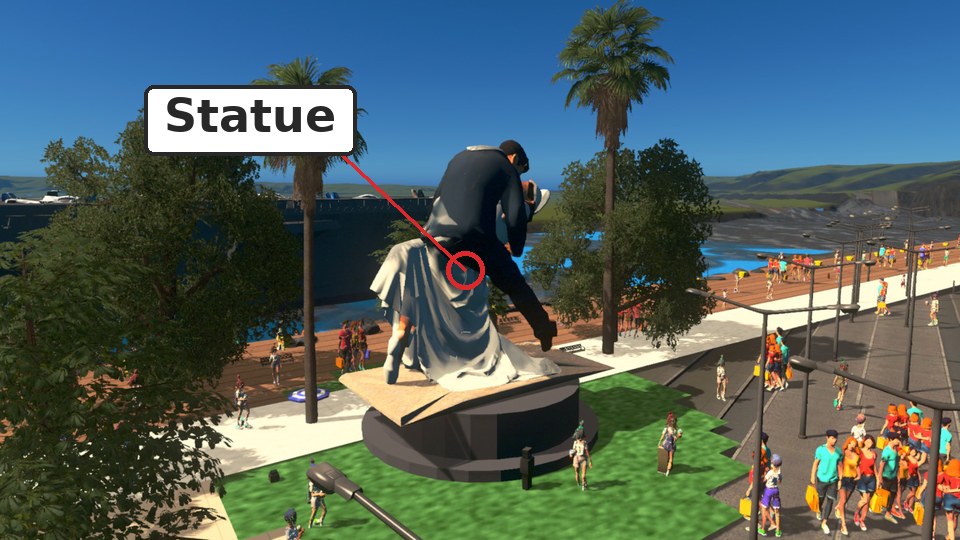}{Kissing sailor statue} & \pf{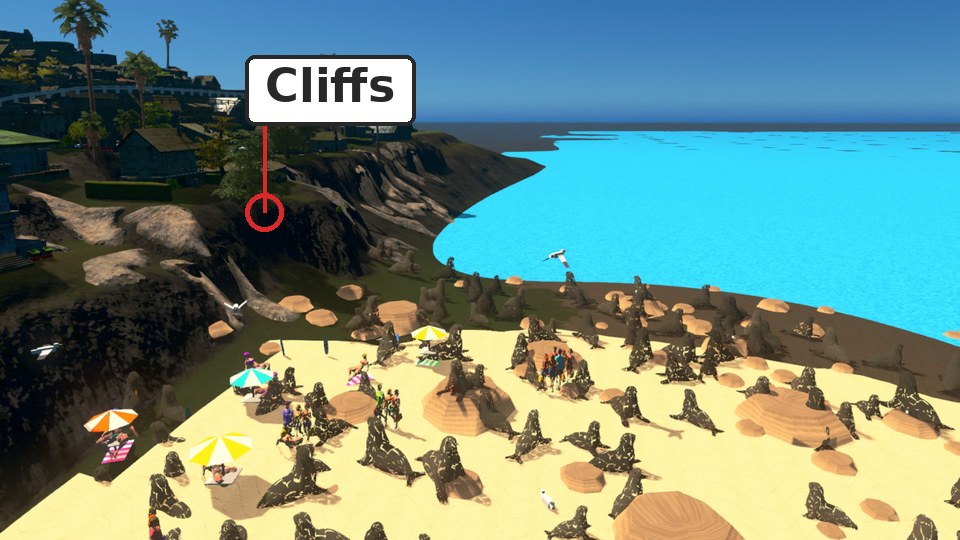}{La Jolla Cove cliffs} & \pf{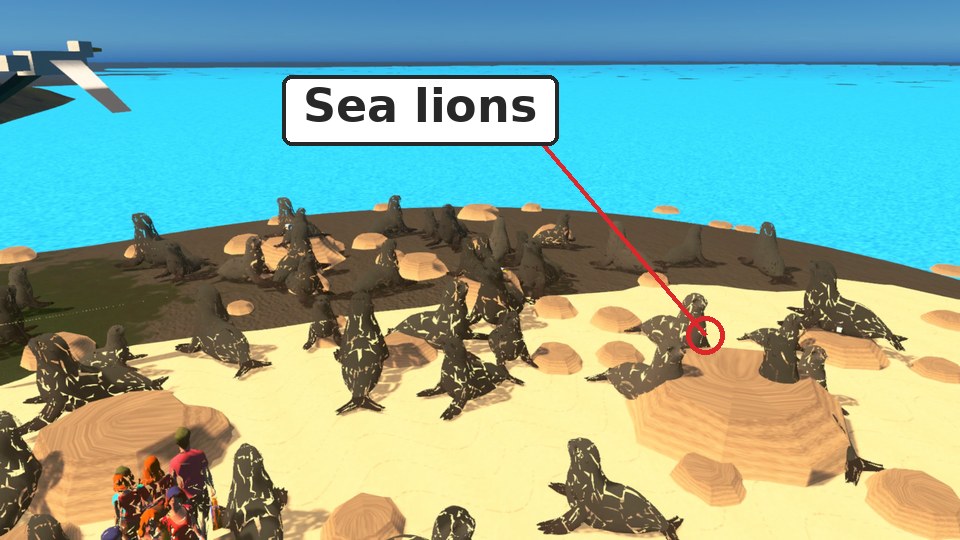}{Sea lions sunbathing}} \\[2pt]
\extrow{\multicolumn{5}{@{}l}{\scriptsize\textbf{Photos}: Balboa Park and the UCSD campus} \\*
\rlab{Reference}{Landmark photos from Wikimedia Commons} & \rf{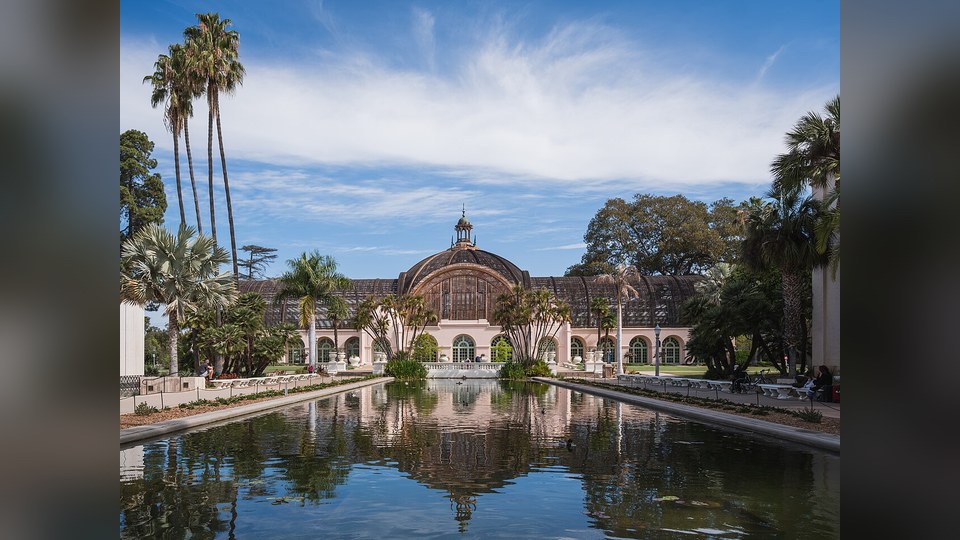} & \rf{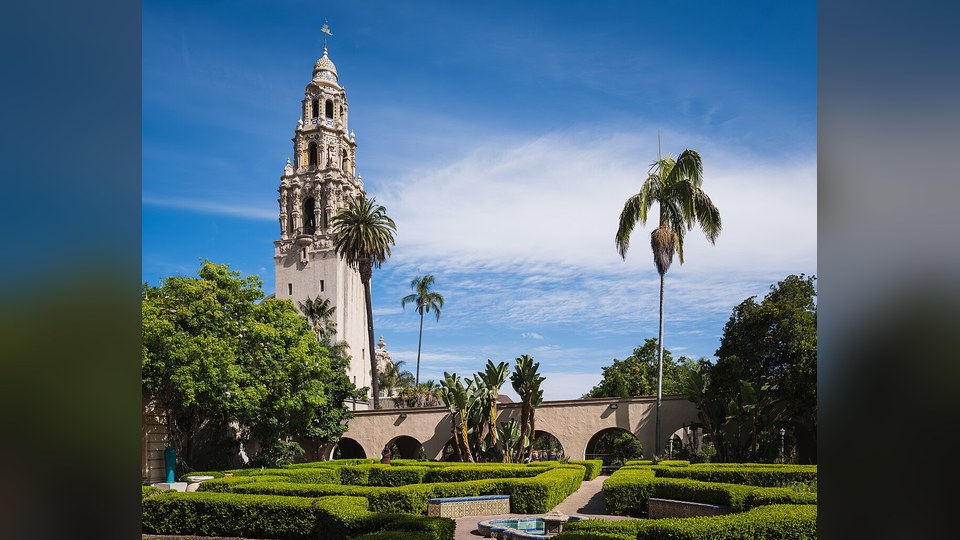} & \rf{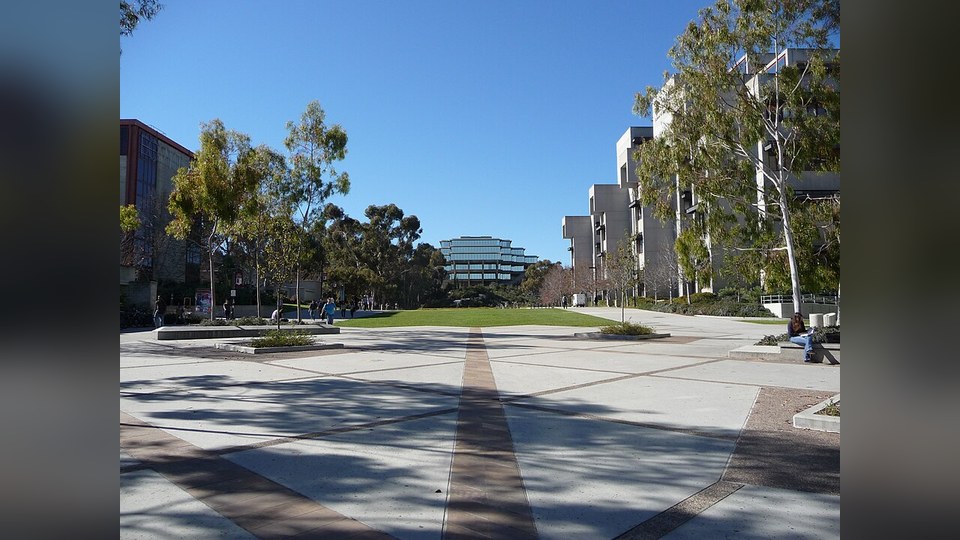} & \rf{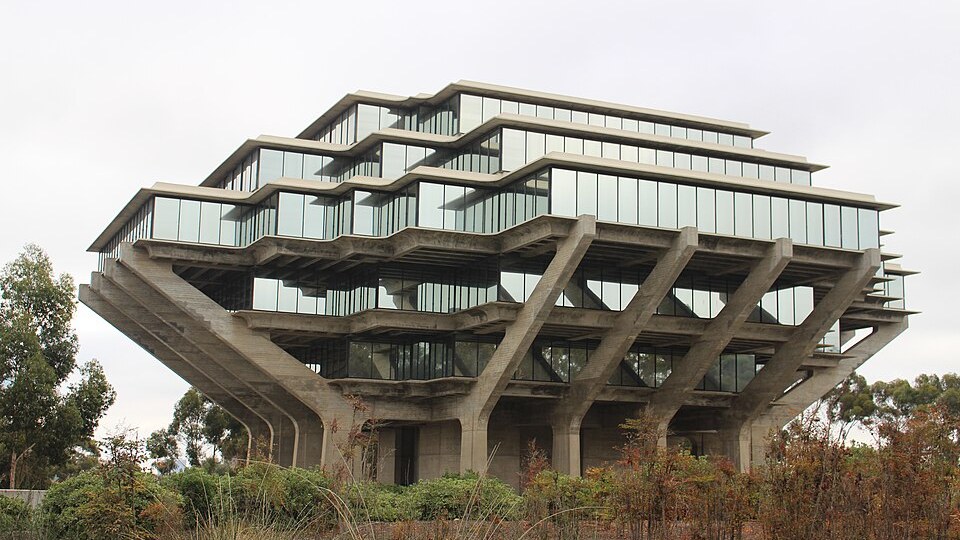} \\*
\rlab{In game}{} & \pf{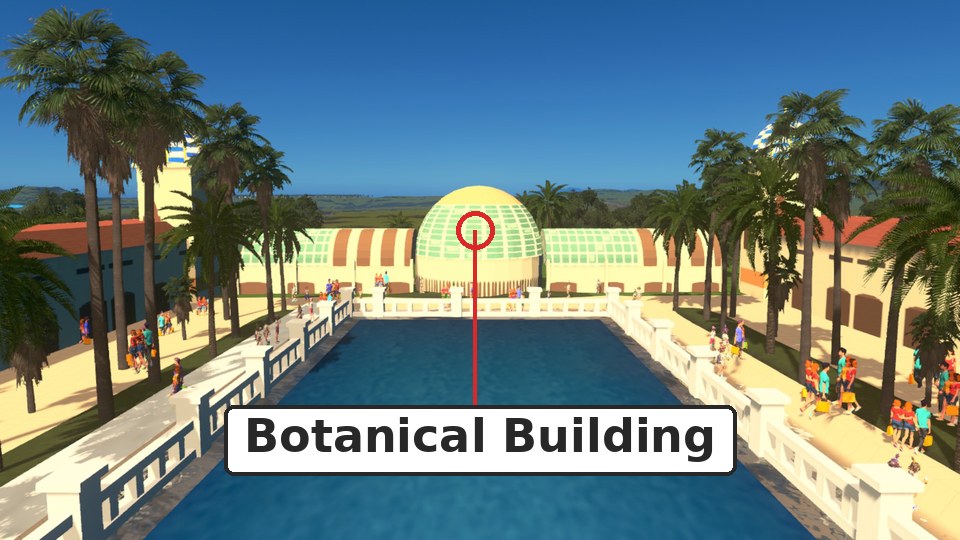}{Botanical Building pond} & \pf{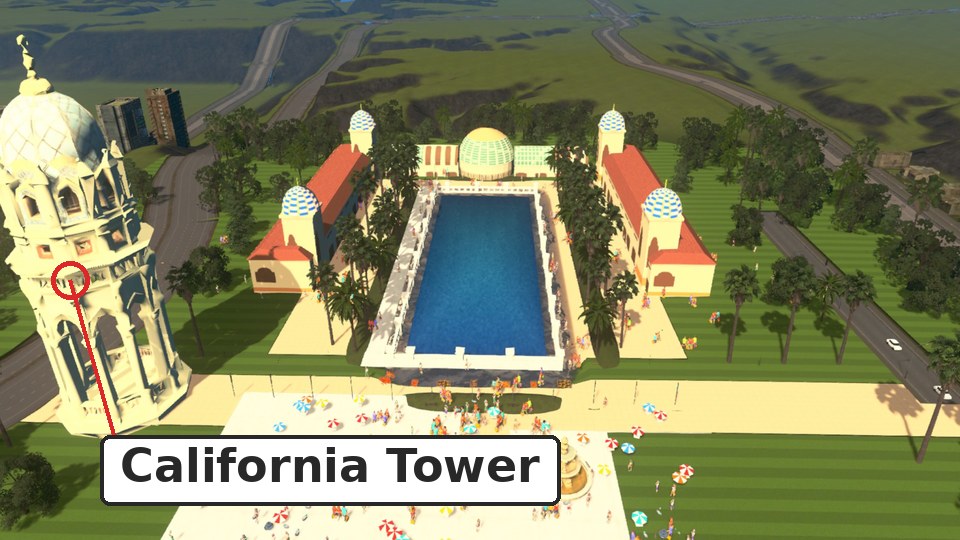}{California Tower} & \pf{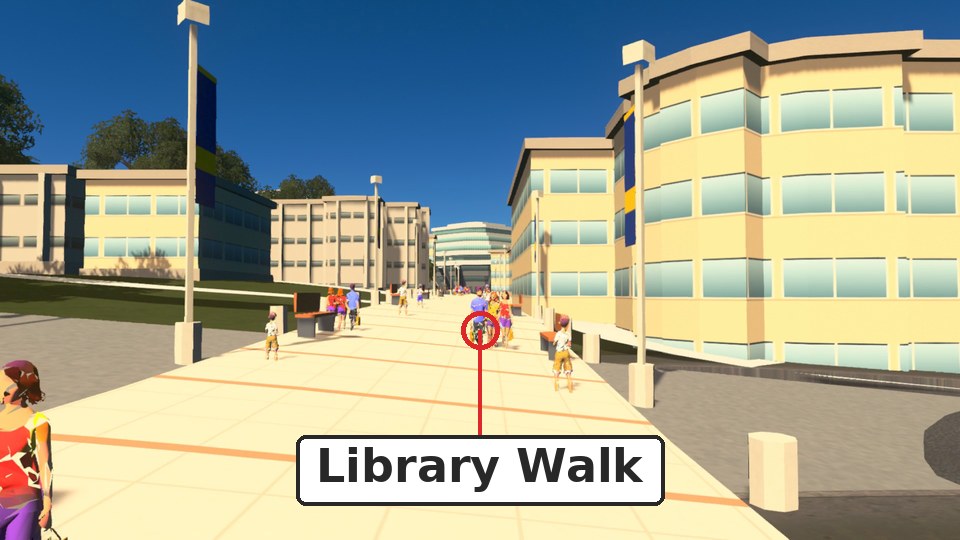}{Library Walk} & \pf{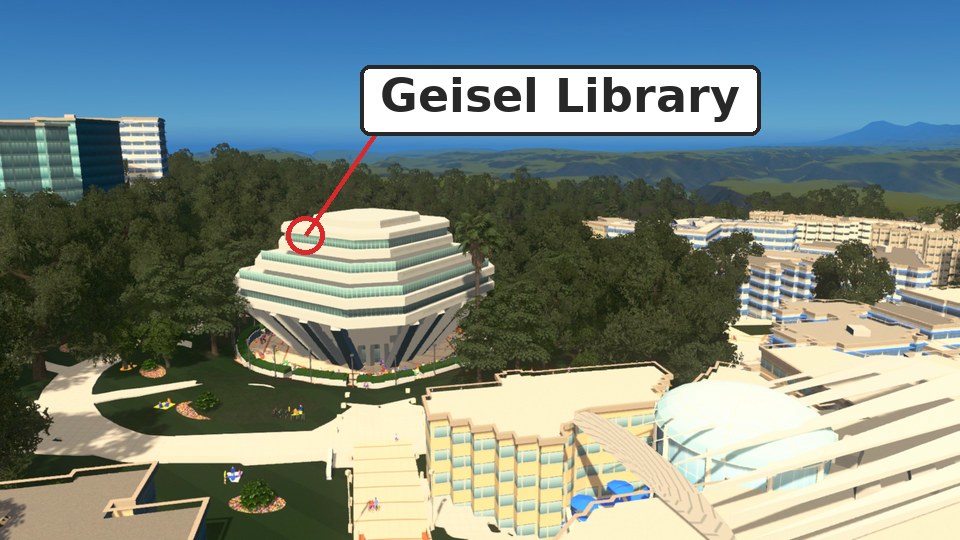}{Geisel Library}} \\[4pt]
\end{longtable}
\endgroup

\subsection{Reference Collection and Annotation}
\label{app:reference_collection}

\paragraph{Reference Sources.}
We collect multimodal references from diverse sources to cover a broad range of creative content.
Our \textit{video collection} includes film clips, gameplay recordings, sports footage, and travel videos.
\textit{Text references} are drawn from novels, short stories, fairy tales, game rules, and descriptions of places and activities.
\textit{Image references} include photographs, paintings, comics, illustrations, and diagrams.
The initial collection contains 180 video files, 79 text works, and 77 image groups.
We record the source and licensing information of each reference to support traceability and dataset release.

\paragraph{Reference Processing and Annotation.}
We process each modality according to its characteristics while adopting a unified annotation scheme, to identify meaningful candidate references.

\textbf{For videos}, we remove duplicates, identify relevant clips, and extract timestamped keyframes and transcripts. Two independent annotation passes identify important mechanisms, entities and their properties, and spatial, temporal, and causal relations.
\textbf{For texts}, we parse the collected works into passages with stable paragraph references.
A model examines the source text to identify candidate passages and annotate relevant content with supporting quotations.
\textbf{For images}, we group related pictures, detect duplicates, and annotate important visual elements and relationships. These image groups also underwent two independent annotation passes.

Across all modalities, annotations are linked to the original evidence through video timestamps, text passages, or image locations.
These annotations support task construction but are not provided to the development agents.

\paragraph{Candidate Ranking and Human Selection.}
We score candidate references based on \textit{semantic clarity, transfer potential, and evaluation feasibility}.
For videos, we combine independent model annotations with rule based filtering to obtain a candidate shortlist.
For texts, Claude identifies and scores candidate passages, while Codex performs comparative ranking to better distinguish candidates with similar scores.
For images, model annotations support the identification of important content and redundant references.
Human reviewers then inspect the original materials and select the final references, considering both candidate quality and coverage across task categories.
The selected references are subsequently used to construct benchmark tasks with suitable target games.
Table~\ref{tab:reference_statistics} summarizes the collection and selection statistics.

The 125 final Construction tasks use 51 distinct videos (74 clips) across the 40 video tasks and 19 hybrid tasks.
They use 39 text passages, 30 from 25 public-domain works for the text tasks and nine in hybrid tasks, and 50 image bundles, 36 for the image tasks and 14 in hybrid tasks.
These are the final counts in Table~\ref{tab:reference_statistics}, and the distribution of the tasks is reported in Table~\ref{tab:app_task_distribution}.

\begin{table*}[!htbp]
\centering
\caption{
Statistics of multimodal reference collection and annotation.
Counts are reported using the collection unit of each modality.
Final List: distinct references used by the 125 final Construction tasks, including those of hybrid tasks.
}
\label{tab:reference_statistics}

\small
\setlength{\tabcolsep}{7pt}
\renewcommand{\arraystretch}{1.15}

\begin{tabular}{@{}lllll@{}}
\toprule
\textbf{Modality}
& \textbf{Collected}
& \textbf{Selected}
& \textbf{Final List}
& \textbf{Evidence Grounding} \\
\midrule

Video
& 180 files
& 57 videos
& 51 videos
& Timestamps and clips \\

Text
& 79 works
& 46 passages
& 39 passages
& Paragraph references and quotations \\

Image
& 77 groups
& 71 image bundles
& 50 image bundles
& Image indices and regions \\

\bottomrule
\end{tabular}

\end{table*}

\subsection{Task and Rubric Construction}
\label{app:task_construction}

\paragraph{Pilot Calibration and Iterative Rubric Construction.}
We first calibrate the task construction process on a small set of representative tasks through multiple rounds of model generation and human feedback.
Based on the feedback, we refine the authoring guidelines, model instructions, and examples before scaling up task generation.
For each task, Claude drafts the evaluation rubrics, which are independently checked by Codex for source grounding, scoring consistency, and relevance to the adaptation objective.
Claude then reconciles the feedback and revises the rubrics.
Every task undergoes human review, where reviewers examine all rubric items and provide feedback until the task is approved.
Feedback on individual items leads to targeted revisions, while recurring issues are incorporated into the shared authoring guidelines and applied to other affected tasks.

\paragraph{Rubric Authoring Principles.}
We follow three principles when constructing evaluation rubrics.
\begin{itemize}
    \item Each rubric item must be supported by reference evidence or explicit user requirements, with clear scoring anchors for scores of 0, 0.5, and 1.
    \item The evaluation criteria focus on the primary transfer objective.
Mechanics tasks assess the underlying rules and behaviors, scene tasks emphasize locations and spatial structures, and narrative tasks focus on characters, events, and causal relationships.
    \item We avoid requiring incidental details that do not define the intended experience.
We apply a removal test to determine whether an element is essential: if removing or replacing it preserves the defining rules, setting, or narrative structure, its exact reproduction is not required.
This allows different valid implementations without sacrificing the intended reference content.
\end{itemize}

Host Fidelity uses the game-specific checklists described in Appendix~\ref{app:eval_details}, while the executability gate uses the validated capability probes described in Appendix~\ref{app:game_capability}.

\paragraph{Instruction Construction.}
Once the rubrics are established, we derive task instructions for Construction and Editing.
Construction instructions specify a creative goal, requiring agents to realize the reference content in the target game with 2-3 explicit user requests to preserve, specify, and modify content.
Editing instructions specify targeted changes to existing entities, properties, mechanisms, or relations, potentially guided by additional references.
The corresponding rubrics define both the required modifications and the content that should remain unchanged.
All instructions undergo human review and iterative refinement to ensure clarity, feasibility, and alignment with the evaluation criteria.

\paragraph{Editing Checkpoint Preparation.}
\label{app:editing_checkpoints}
We select 31 validated Construction checkpoints as the initial states for 80 Editing tasks, primarily using outputs from Claude Fable 5.2. Each checkpoint must pass executability checks, contain sufficiently complete and playable content, and provide access to the entities or mechanics targeted by the edits. We further verify that the requested changes are not already implemented and that preservation rubrics refer to existing content. All instructions and rubrics are reviewed and revised when necessary. All agents receive the same initial checkpoint for each task, which also serves as the reference for evaluating edit success and preservation.

\paragraph{Benchmark Task Distribution.}
Table~\ref{tab:app_task_distribution} reports the distribution of Construction tasks by target game and reference modality.
The open-source core contains 77 Construction tasks across nine games, and the commercial track contains 48 tasks across five games, for 125 tasks in total.
Table~\ref{tab:app_edit_distribution} reports the 80 Editing tasks, which start from the checkpoints of 31 Construction tasks with two or three edits each: 64 tasks on six open-source games and 16 on two commercial games.
Edits fall into three classes. \textit{Appearance} modifies the look of existing content in 26 tasks. \textit{Content \& structure} changes objects or spatial layouts in 26 tasks, while \textit{behavior \& flow} modifies rules, parameters, or event sequences in 28 tasks.
Most edits target a single object, accounting for 61 tasks, while the remaining edits apply to groups of similar objects.

\begin{table}[!htbp]
\centering
\caption{Construction tasks per target game and reference modality (V: video, T: text, I: image, H: hybrid).}
\label{tab:app_task_distribution}
\small
\begin{tabular}{@{}llrrrrr@{}}
\toprule
\textbf{Game} & \textbf{Type} & \textbf{V} & \textbf{T} & \textbf{I} & \textbf{H} & \textbf{Total} \\
\midrule
Celeste 64 & Open-source & 3 & 1 & 2 & 0 & 6 \\
Colobot & Open-source & 1 & 1 & 1 & 0 & 3 \\
FLARE & Open-source & 5 & 0 & 0 & 0 & 5 \\
Godot TPS Demo & Open-source & 4 & 1 & 5 & 1 & 11 \\
SuperTuxKart & Open-source & 5 & 2 & 3 & 0 & 10 \\
Tanks of Freedom II & Open-source & 1 & 0 & 2 & 1 & 4 \\
Tuxemon & Open-source & 2 & 4 & 4 & 3 & 13 \\
VoxeLibre & Open-source & 6 & 6 & 4 & 2 & 18 \\
Xonotic & Open-source & 2 & 2 & 2 & 1 & 7 \\
\midrule
Cities: Skylines & Commercial & 2 & 2 & 2 & 1 & 7 \\
Minecraft & Commercial & 4 & 3 & 2 & 4 & 13 \\
Stardew Valley & Commercial & 1 & 4 & 4 & 4 & 13 \\
Terraria & Commercial & 3 & 2 & 2 & 0 & 7 \\
Valheim & Commercial & 1 & 2 & 3 & 2 & 8 \\
\midrule
\textbf{Total} & & \textbf{40} & \textbf{30} & \textbf{36} & \textbf{19} & \textbf{125} \\
\bottomrule
\end{tabular}

\vspace{12pt}
\centering
\caption{Editing tasks per target game and edit class (AP: appearance, CS: content \& structure, BF: behavior \& flow). Parents: Construction tasks whose checkpoints the edits start from.}
\label{tab:app_edit_distribution}
\small
\begin{tabular}{@{}llrrrrr@{}}
\toprule
\textbf{Game} & \textbf{Type} & \textbf{Parents} & \textbf{AP} & \textbf{CS} & \textbf{BF} & \textbf{Total} \\
\midrule
Celeste 64 & Open-source & 3 & 3 & 2 & 3 & 8 \\
Godot TPS Demo & Open-source & 5 & 4 & 4 & 5 & 13 \\
SuperTuxKart & Open-source & 3 & 2 & 3 & 3 & 8 \\
Tanks of Freedom II & Open-source & 2 & 2 & 1 & 1 & 4 \\
VoxeLibre & Open-source & 8 & 8 & 7 & 7 & 22 \\
Xonotic & Open-source & 3 & 3 & 3 & 3 & 9 \\
\midrule
Minecraft & Commercial & 4 & 3 & 3 & 3 & 9 \\
Stardew Valley & Commercial & 3 & 1 & 3 & 3 & 7 \\
\midrule
\textbf{Total} & & \textbf{31} & \textbf{26} & \textbf{26} & \textbf{28} & \textbf{80} \\
\bottomrule
\end{tabular}

\end{table}

\subsection{Task Construction Case Studies}
\label{app:case_studies}
Tables~\ref{tab:case_text_wild}--\ref{tab:case_hybrid_zootopia} trace task construction from the original references through finalized instructions and rubrics. The examples cover text, image, video, and hybrid inputs. The hybrid task combines the clip used in the video-only case with separately collected images.

\begin{table*}[!htbp]
\centering
\footnotesize
\renewcommand{\arraystretch}{1.25}
\caption{Construction of a text-only narrative task from a public-domain novel. Rubric items are abridged; evidence anchors such as \texttt{c06.p020} identify the corresponding chapter and paragraph.}
\label{tab:case_text_wild}
\begin{tabular}{@{}>{\raggedright\arraybackslash}p{2.3cm}p{\dimexpr\textwidth-2.3cm-2\tabcolsep\relax}@{}}
\toprule
\textbf{Raw reference} &
\textit{The Call of the Wild} by Jack London, published in 1903. The public-domain Standard Ebooks edition is based on Project Gutenberg \#215 and contains 31{,}701 words across seven chapters.\newline
{\setlength{\tabcolsep}{2pt}\scriptsize
\begin{tabular}{@{}*{7}{>{\raggedright\arraybackslash}p{\dimexpr(\linewidth-14\tabcolsep)/7\relax}}@{}}
\cellcolor{black!6}c01 a pampered dog is stolen and shipped north &
\cellcolor{black!6}c02 learns the law of club and fang &
\cellcolor{black!6}c03 trail to Dawson; rivalry with the lead dog &
\cellcolor{black!6}c04 takes the lead; record run &
\cellcolor{blue!8}c05 overloaded sled, starving team, thin ice, candidate \textbf{e2} &
\cellcolor{orange!25}c06 nursed by a new master; rapids rescue; the sled wager, candidate \textbf{e1} &
\cellcolor{black!6}c07 the lost mine; the wolves' call
\end{tabular}}\\
\midrule
\textbf{Segmentation \& annotation} &
The book is split into chapters and anchored paragraphs. Two independent annotation passes write a chapter map and propose candidate excerpts with a primary category, a summary, the renaming needed, knowledge leakage and content safety:
\textbf{e1} covers \texttt{c06.p001--p062} with 4{,}742 words for a narrative task. \textbf{e2} covers \texttt{c05.p001--p066} with 5{,}335 words for a mechanics task.
Candidate e1 has elevated recognition risk because the novel is well known. Three sentences contain graphic injury details, while the remaining fight scenes present less severe content. \\
\midrule
\textbf{Selection} &
Across the pool, 67 narrative candidates are ranked; the curator keeps \textbf{e1}, ranked 21st. The delivered text is then prepared:
a 106-word recap of earlier events that is excluded from scoring, replacement of 15 names across 232 occurrences to reduce recognition, and rewriting eight sensitive sentences. The name changes include Buck $\rightarrow$ Ranger and John Thornton $\rightarrow$ Daniel Harlow. The rewrite rate is 1.7 sentences per 1{,}000 words. For example,\
\textit{``drove in again for the throat \ldots his throat was torn open''} $\rightarrow$ \textit{``Ranger drove in again, and the man was badly hurt.''}
Delivered: 4{,}788 words, no title or author. Target game: \textit{Stardew Valley}. \\
\midrule
\textbf{Rubric} &
\begin{minipage}[t]{\linewidth}
\textbf{Content Realization, 6 items.} CR1 Ranger is devoted to Harlow alone ~\texttt{c06.p006--p013}; CR2 the bond starts with Harlow nursing the rescued, weakened Ranger ~\texttt{p001--p004};
CR3 the rapids: Harlow is swept toward deadly rapids and Ranger saves him ~\texttt{p020--p028}; CR4 Harlow's own boast pulls him into a bet on Ranger ~\texttt{p029--p034};
CR5 the haul test: Ranger alone in harness against an extraordinarily heavy load over a marked distance ~\texttt{p029--p043}; CR6 a call from the forest pulls Ranger away, and his love for Harlow brings him back ~\texttt{p008--p012}.\\[2pt]
\textbf{Functions \& Playability, 4 items.} FP1 the river rescue is a challenge: reach Harlow against the current and bring him to the bank; FP2 the haul is an effortful challenge whose success wins the bet;
FP3 the story plays in causal order: bond, rapids, wager and haul; FP4 \textit{Modified requirement} training days between bet and haul decide whether the haul can succeed.\\[2pt]
\textbf{Visual \& Presentation, 3 items.} VP1 a harsh northern frontier of wilderness and rough settlements; VP2 Harlow held midstream in wild water, out of reach of the bank; VP3 one dog straining in harness against a sled that looks far too heavy.\\[2pt]
\textit{Kept out as peripheral:} what exactly is staked in the bet, and the offer for Ranger that Harlow refuses after the haul.\\[2pt]
\textit{Anchors, e.g.} CR4: 1 = Harlow boasts about Ranger, and that boast is what leads to the bet; 0.5 = a bet exists but does not come from his boast; 0 = no bet.
\vspace{2pt}\end{minipage} \\
\midrule
\textbf{Final task} &
\textit{``Adapt the story told in the reference text into a playable narrative experience in Stardew Valley. Build an opening that establishes the main characters and the relationship between them before the story's later events. Turn the story's key situations into playable challenges with clear goals, effort, feedback, and a clear outcome or a way to retry. \ldots Specifically, I want: Harlow's own boasting about Ranger should be what drags him into the bet. Also, unlike the text, put a few in-game days between the bet and the haul where I train Ranger, and make how well I trained him decide whether he can move the sled.''}\newline
Explicit requirements specify CR4 and modify FP4. The agent receives only the delivered text and this instruction. \\
\bottomrule
\end{tabular}
\end{table*}

\begin{table*}[!htbp]
\centering
\footnotesize
\renewcommand{\arraystretch}{1.25}
\caption{Construction of an image-only task about minecart mechanics. Rubrics are abridged and image numbers match those shown to the agent. Wikimedia Commons credits are img01 by Flocci Nivis under CC BY 4.0, img02 by Ronald Werner under CC BY-SA 4.0, img03 by the Mining Museum of the Basque Country under CC BY-SA 4.0, img04 by Fernost in the public domain, and img05 by Timothy Titus under CC BY-SA 4.0.}
\label{tab:case_image_minecart}
\begin{tabular}{@{}>{\raggedright\arraybackslash}p{2.3cm}p{\dimexpr\textwidth-2.3cm-2\tabcolsep\relax}@{}}
\toprule
\textbf{Raw reference} &
\begin{minipage}[t]{\linewidth}\vspace{2pt}
\begin{tabular}{@{}c@{\hspace{3pt}}c@{\hspace{3pt}}c@{\hspace{3pt}}c@{\hspace{3pt}}c@{}}
\includegraphics[width=0.192\linewidth]{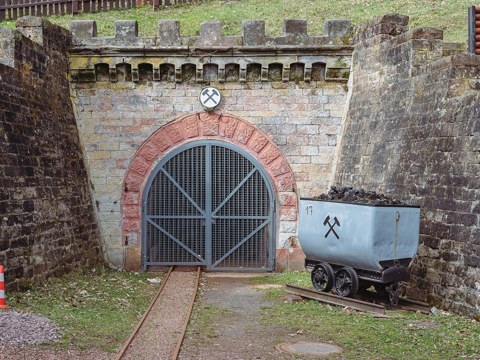} &
\includegraphics[width=0.192\linewidth]{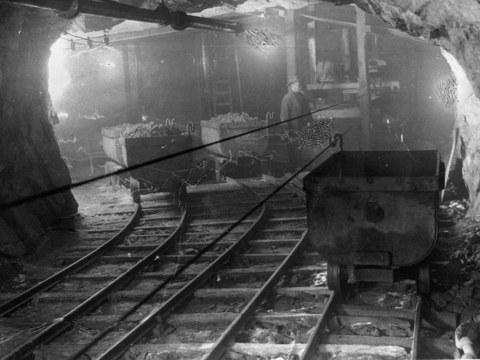} &
\includegraphics[width=0.192\linewidth]{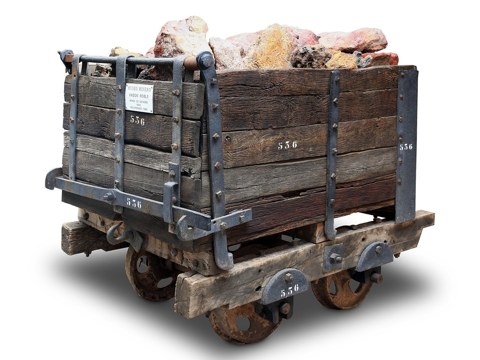} &
\includegraphics[width=0.192\linewidth]{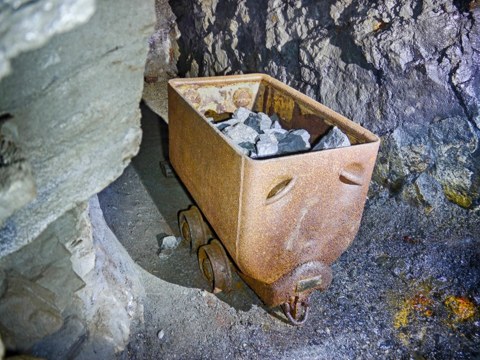} &
\includegraphics[width=0.192\linewidth]{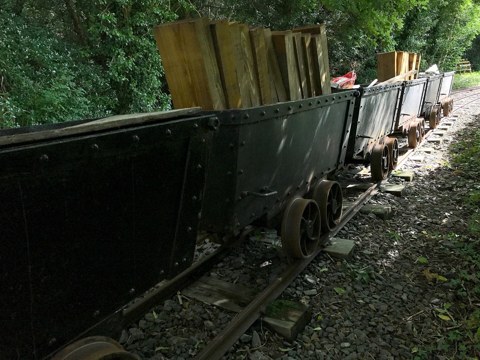} \\[-1pt]
\scriptsize img01 mine portal & \scriptsize img02 tunnel tracks & \scriptsize img03 museum cart & \scriptsize img04 loaded cart & \scriptsize img05 cart train
\end{tabular}\vspace{2pt}

A curator collects six openly licensed photographs of mine carts being loaded and hauled along rails. A short description is withheld from the agent. img06, a historical photograph of workers pushing a cart, is part of the task but not reproduced here.
\vspace{2pt}\end{minipage} \\
\midrule
\textbf{Screening \& annotation} &
The inventory records image dimensions, license types, duplicate checks, and numbered views. Licenses include CC BY, CC BY-SA, and public-domain material, with no near duplicates.
Two independent annotation passes record each image's role and necessity, the bundle's coverage, possible recognition cues, and content safety. Recognition risk is low to medium, and no severe content is identified.
The annotators disagree on whether the task concerns mechanics or scenes because motion is implied rather than directly observed. The curator chooses mechanics, and the rubrics evaluate only supported properties. \\
\midrule
\textbf{Selection} &
The curator keeps the group for the image-only set after viewing the images alone. The selected target is \textit{Terraria}, extended through tModLoader. Its side-view mines and rail systems suit the task. \\
\midrule
\textbf{Rubric} &
\begin{minipage}[t]{\linewidth}
\textbf{Content Realization, 3 items.} CR1 mine carts that hold mined material, with both loaded and empty carts, see img01--img04;
CR2 the track is a network: several lines join or cross, see img02, img06; CR3 an underground gallery with track along its floor and an entrance through which the track runs out to the surface, see img01, img02, img04.\\[2pt]
\textbf{Functions \& Playability, 4 items.} FP1 carts are guided by the track and cannot leave it, see img01, img02, img05; FP2 the player sets a cart moving by their own action, see img06;
FP3 a loaded cart keeps its load while it moves, see img02--img04; FP4 a loaded cart can be hauled from the workings out through the entrance, see img01, img02.\\[2pt]
\textbf{Visual \& Presentation, 3 items.} VP1 the underground space reads as a rock-cut mine tunnel, see img02, img04; VP2 loaded and empty carts are told apart at a glance;
VP3 \textit{Modified requirement} the ground at the entrance is snowy and icy instead of grassy.\\[2pt]
\textit{Not scored.} Speeds, timing, capacities, and incidental photographic details cannot be reliably inferred from the images.\\[2pt]
\textit{Anchors, e.g.} CR2: 1 = several track lines that join or cross one another; 0.5 = several separate tracks that do not connect; 0 = a single line only, or no track.
\vspace{2pt}\end{minipage} \\
\midrule
\textbf{Final task} &
\textit{``Adapt the mine-cart system shown in the reference images into a playable tModLoader mod for Terraria. Implement how the carts move and how the player sets them in motion, what the carts carry and how that state is shown to the player, and a route through the mine along which the player works the carts. Build the mine spaces in which this system is played. \ldots Specifically, I want the following. The track should be a network like in the pictures, with several lines that join or cross, not just one single line. Set the mine entrance in a snowy, icy landscape instead of on grass.''}\newline
Explicit requirements specify CR2 and modify VP3. The agent receives the six images without titles, descriptions, or captions, together with the instruction. \\
\bottomrule
\end{tabular}
\end{table*}

\begin{table*}[!htbp]
\centering
\footnotesize
\renewcommand{\arraystretch}{1.25}
\caption{Construction of a video-only scene and world task from a film trailer. Rubrics are abridged, and timestamps refer to the source video.}
\label{tab:case_video_zootopia}
\begin{tabular}{@{}>{\raggedright\arraybackslash}p{2.3cm}p{\dimexpr\textwidth-2.3cm-2\tabcolsep\relax}@{}}
\toprule
\textbf{Raw reference} &
\includegraphics[width=\linewidth]{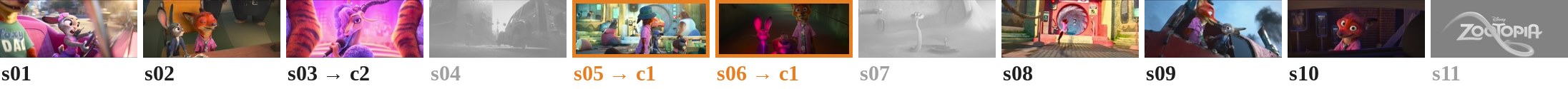}\newline
The animated film's international trailer runs for 134.5 seconds. Two independent annotation passes divide it into eleven segments. Three segments show transitions, marketing content, or title cards, and are excluded from scoring; they appear in grey. The orange frames mark the selected clip c1. \\
\midrule
\textbf{Segmentation \& annotation} &
The annotated segments cover a purple-lit gala in s03, the canal-side Marsh Market with docks, slides, and a kiosk hatch in s05, a neon reptile club in s06, and a water chase in s08. Each observation is grounded in a source timestamp.
The annotators propose three clips. Candidate \textbf{c1} covers 0:43.1 to 1:07.4, spanning scene segments s04 to s06. Candidate \textbf{c2} covers narrative segment s03 from 0:22.5 to 0:43.1. The second pass adds mechanics candidate \textbf{c3}, showing a water chase from 1:18.2 to 1:29.3.
The source film is recognizable but is not named in the task. Rapid cuts and text overlays provide weak evidence of layout, and the clip does not reveal the full route between the market and club. \\
\midrule
\textbf{Selection} &
The curator selects \textbf{c1} as a scene and world task with a consensus score of 5.0 out of 6. The same clip anchors the hybrid task in Table~\ref{tab:case_hybrid_zootopia}. Candidate c2 scores 3.0, and c3 falls below the selection threshold. The target game is the voxel survival sandbox \textit{VoxeLibre}. \\
\midrule
\textbf{Rubric} &
\begin{minipage}[t]{\linewidth}\vspace{2pt}
\begin{tabular}{@{}c@{\hspace{3pt}}c@{\hspace{3pt}}c@{\hspace{3pt}}c@{\hspace{3pt}}c@{}}
\includegraphics[width=0.192\linewidth]{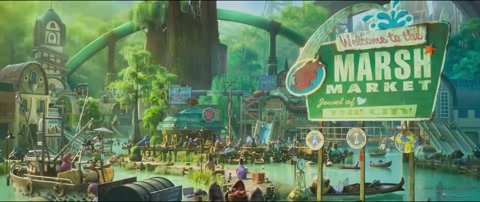} &
\includegraphics[width=0.192\linewidth]{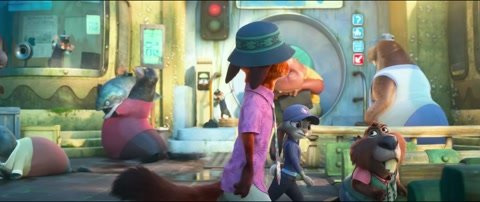} &
\includegraphics[width=0.192\linewidth]{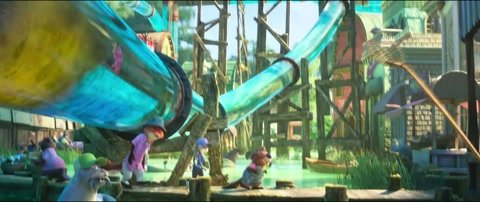} &
\includegraphics[width=0.192\linewidth]{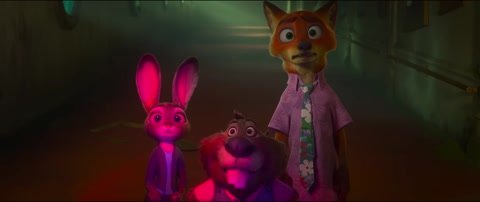} &
\includegraphics[width=0.192\linewidth]{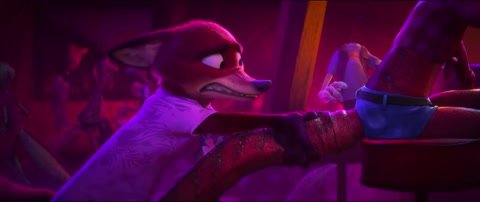} \\[-1pt]
\scriptsize 0:46 CR1, VP2 & \scriptsize 0:47 CR3, CR6, FP3 & \scriptsize 0:51 CR2 & \scriptsize 0:56 CR4, VP1, VP3 & \scriptsize 1:04 CR5, VP2
\end{tabular}\vspace{3pt}

\textbf{Content Realization, 6 items.} CR1 a market built on water: a canal with boats, docks with stalls and stilted buildings under a large hanging market sign;
CR2 its landmark, teal see-through tube water slides looping between wooden scaffold towers; CR3 a transit kiosk, a counter with a round hatch that opens onto a tube;
CR4 the market and the club are two distinct places joined by an enclosed corridor into the club; CR5 the club is a reptile hangout with a bar and a piano;
CR6 animal characters of different kinds and clearly different sizes on the docks and in the water.\\[2pt]
\textbf{Functions \& Playability, 3 items.} FP1 the water market can be explored on foot along walkways and docks, and the water behaves consistently; FP2 the player walks through the corridor into the club without a break;
FP3 entering the round hatch carries the player down and away through a tube.\\[2pt]
\textbf{Visual \& Presentation, 3 items.} VP1 the corridor is dim, lit by warm wall lamps, with a glowing porthole and a red-lit round doorway at its end;
VP2 palette contrast: the market in bright daylight with green water and teal accents, the club dark in magenta and purple neon; VP3 entering the club is a reveal, from the narrow corridor into the wide neon room.\\[2pt]
\textit{Each item has 1 / 0.5 / 0 anchors, e.g.} CR4: 1 = an open market area and a separate enclosed club, with a corridor that leads into the club; 0.5 = both places exist but the corridor is missing or does not lead into the club, or the club is just a corner of the market; 0 = one of the two places is missing.
\vspace{2pt}\end{minipage} \\
\midrule
\textbf{Final task} &
\textit{``Recreate the place shown in the reference video as an explorable location in VoxeLibre. Make it freely explorable on foot, with its layout and landmarks reachable and walkable connections between its areas, so that the player can move from one area into the next without a break. Make the place's interactive features usable by the player as the video shows them. \ldots Specifically, I want: I really want those teal see-through tube water slides looping between wooden scaffold towers in the market. The club should be a reptile hangout, dark and lit in magenta and purple neon, with a bar and a piano and lizards as the regulars.''}\newline
Explicit requirements preserve CR2 and specify CR5 and VP2. The agent receives only the clip 0:43.1--1:07.4 and this instruction. \\
\bottomrule
\end{tabular}
\end{table*}

\begin{table*}[!htbp]
\centering
\footnotesize
\renewcommand{\arraystretch}{1.25}
\caption{Hybrid video and image task construction for a scene and world objective. The video provides the ground-level experience, while separately collected images provide missing layout information. The top row shows clip frames and the lower row shows images. Rubric items are abridged, with a superscript x indicating cross-reference criteria. Studio concept art and theme-park photographs are used as private references.}
\label{tab:case_hybrid_zootopia}
\begin{tabular}{@{}>{\raggedright\arraybackslash}p{2.3cm}p{\dimexpr\textwidth-2.3cm-2\tabcolsep\relax}@{}}
\toprule
\textbf{Raw references} &
\begin{minipage}[t]{\linewidth}\vspace{2pt}
\begin{tabular}{@{}c@{\hspace{3pt}}c@{\hspace{3pt}}c@{\hspace{3pt}}c@{\hspace{3pt}}c@{}}
\includegraphics[width=0.192\linewidth]{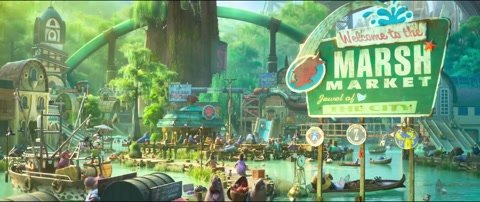} &
\includegraphics[width=0.192\linewidth]{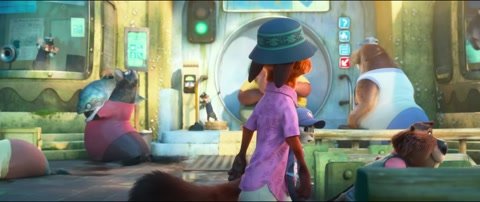} &
\includegraphics[width=0.192\linewidth]{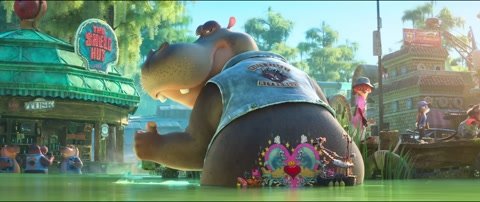} &
\includegraphics[width=0.192\linewidth]{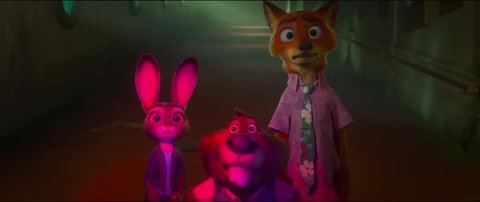} &
\includegraphics[width=0.192\linewidth]{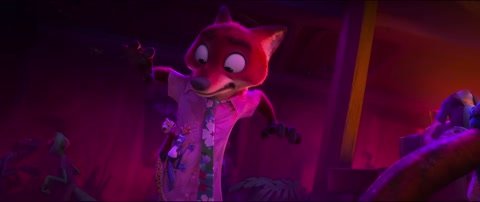} \\[-1pt]
\scriptsize clip 0:46 welcome sign & \scriptsize 0:47 the market kiosk & \scriptsize 0:53 on the water & \scriptsize 0:57 the corridor & \scriptsize 1:00 the club \\[0pt]
\includegraphics[width=0.192\linewidth]{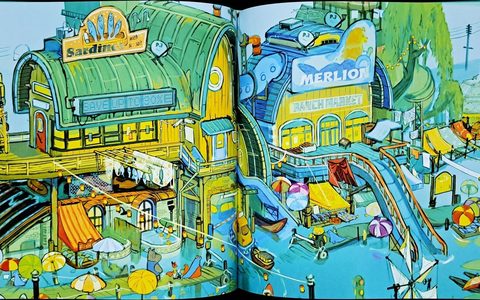} &
\includegraphics[width=0.192\linewidth]{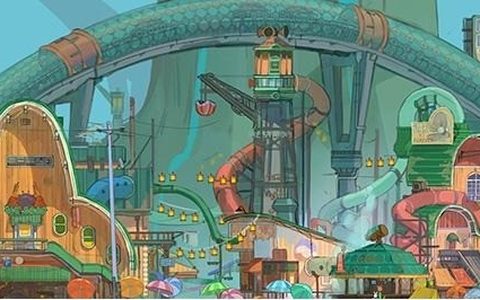} &
\includegraphics[width=0.192\linewidth]{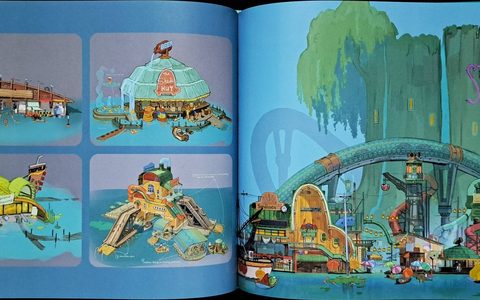} &
\includegraphics[width=0.192\linewidth]{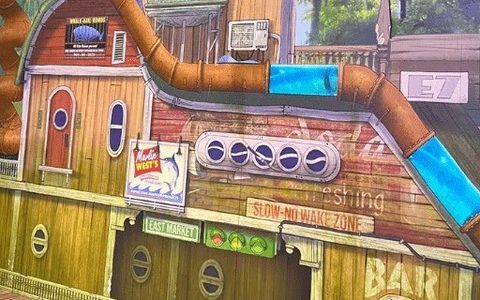} &
\includegraphics[width=0.192\linewidth]{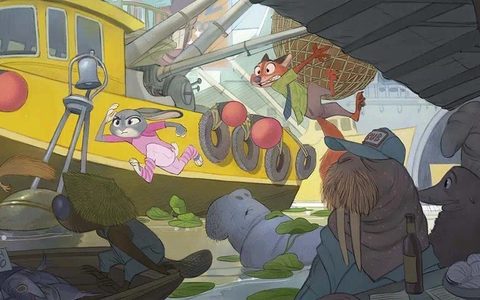} \\[-1pt]
\scriptsize img01 market & \scriptsize img02 skyline & \scriptsize img03 buildings & \scriptsize img04 facade & \scriptsize img05 waterfront
\end{tabular}\vspace{0pt}

\textbf{Video:} the same clip c1 from 0:43.1 to 1:07.4 as in Table~\ref{tab:case_video_zootopia}. It shows the market at eye level and the walk into the club, but not the layout of the district. \textbf{Images} collected separately for this task, shown as crops, supply the missing layout.
\vspace{2pt}\end{minipage} \\
\midrule
\textbf{Planning \& collection} &
At video selection, the curator marks c1 as a paired hybrid: same clip, plus independent images with a \emph{co-referential} relation showing the same film location through another medium. The plan states what the images must add --- a wider view of the district, its stilted buildings and slide scaffolds --- and that the built district must follow the images' arrangement while matching the clip's corridor and palette transition.
The collector finds official concept art and art-book pages of the district, used only as private references. Content-safety pass: no flagged element. Target game: \textit{VoxeLibre}. \\
\midrule
\textbf{Rubric} &
\begin{minipage}[t]{\linewidth}
\textit{Roles.} Video: the eye-level look, the marsh water, the sign, the slide tubes, and the corridor into the club. Images: the layout and buildings.\\[2pt]
\textbf{Content Realization, 5 items.} CR1 the market is built on the water: buildings on their own stilts, piers or platforms, joined by boardwalks and stairs, with the water as the street, see clip 0:45--0:55, img01, img03, img05;
\textsuperscript{\textbf{x}} CR2 tubes thread the district: an enormous tube arches over the market, and slide tubes on scaffolds run down into the water, see clip 0:45--0:51, img01--img03;
CR3 the club is an enclosed nightclub whose patrons are reptiles, see clip 0:57--1:02; CR4 the club is reached from the market through a dim corridor lined with round doors, with a pink glow at its end, see clip 0:55--0:58;
CR5 a large welcome sign naming the market stands at the water's edge, with the market behind it, see clip 0:45--0:46, img02.\\[2pt]
\textbf{Functions \& Playability, 2 items.} FP1 the player moves through the water market along boardwalks and docks, with the water as part of the route;
\textsuperscript{\textbf{x}} FP2 one continuous place: the player walks from the market laid out as in the images through the clip's corridor into the club and back, without a break.\\[2pt]
\textbf{Visual \& Presentation, 3 items.} VP1 marsh character: huge tree trunks and hanging greenery over green-tinted water and weathered wooden buildings, see keyframes 0:46, 0:52; VP2 the giant arching tube dominates the skyline, and the slides are translucent teal tubes on tall scaffolds, see 0:45, 0:50--0:51; VP3 two lighting zones, the market in daylight and the club in magenta-purple neon, with a dim corridor between them, see 0:52, 0:56, 0:58.
\vspace{2pt}\end{minipage} \\
\midrule
\textbf{Final task} &
\textit{``Recreate the place shown in the reference video and the reference images as an explorable scene in VoxeLibre using Luanti, combining the references into one coherent place. Let the player move freely through the place along all the routes it offers and reach its landmarks. \ldots Specifically, I want: the club should be a dark, enclosed place that you reach from the sunny market by walking down a dim corridor with round doors and a pink glow at its end, just like in the clip.''}\newline
Explicit requirements specify CR4 and VP3. The agent receives the clip, the five images without captions, and this instruction. Compared with the paired video-only task, the pier-built layout and the giant arching tube come from the images, while the kiosk hatch of the clip is no longer scored. \\
\bottomrule
\end{tabular}
\end{table*}

Tables~\ref{tab:case_edit_starwars} and~\ref{tab:case_edit_littleprince} follow two Construction checkpoints into Editing tasks: from each validated checkpoint, one appearance, one content \& structure, and one behavior \& flow edit, each with its own edit-success and preservation items.

\begin{table*}[!htbp]
\centering
\footnotesize
\renewcommand{\arraystretch}{1.25}
\caption{Editing task case study based on a video and text Construction task in Godot TPS. Three edits, one per class, share a validated checkpoint. The top row compares two reference frames with three checkpoint frames. Rubrics are abridged, and a superscript t marks trap items.}
\label{tab:case_edit_starwars}
\begin{tabular}{@{}>{\raggedright\arraybackslash}p{2.3cm}p{\dimexpr\textwidth-2.3cm-2\tabcolsep\relax}@{}}
\toprule
\textbf{Parent task \& checkpoint} &
\begin{minipage}[t]{\linewidth}\vspace{2pt}
\begin{tabular}{@{}c@{\hspace{3pt}}c@{\hspace{3pt}}c@{\hspace{3pt}}c@{\hspace{3pt}}c@{}}
\includegraphics[width=0.192\linewidth]{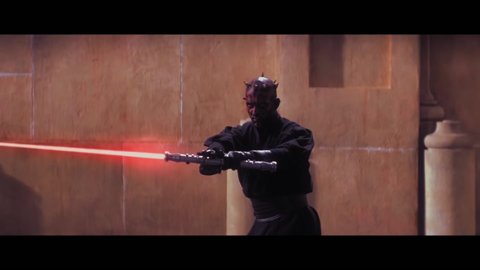} &
\includegraphics[width=0.192\linewidth]{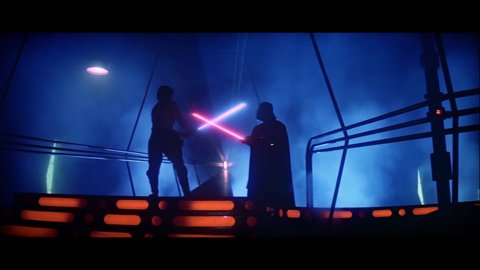} &
\includegraphics[width=0.192\linewidth]{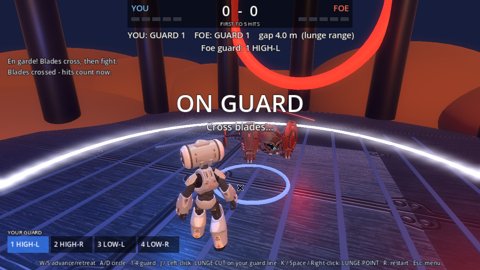} &
\includegraphics[width=0.192\linewidth]{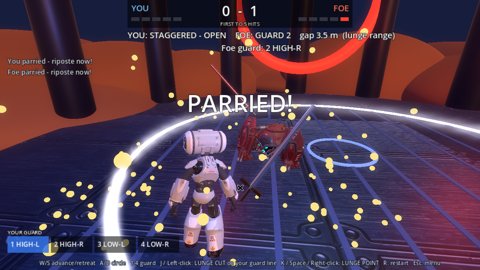} &
\includegraphics[width=0.192\linewidth]{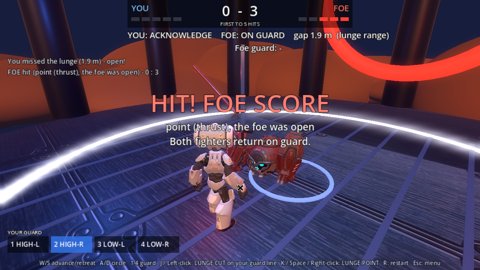} \\[-1pt]
\scriptsize 3:01 double blade & \scriptsize 3:04 the duel & \scriptsize game: on guard & \scriptsize game: point parried & \scriptsize game: hit and count
\end{tabular}\vspace{2pt}

\textbf{Parent.} Two clips of a montage video of the Star Wars films, ending in lightsaber duels, and a text on the rules of a single-stick bout. The Construction task asks for a playable sword duel in Godot TPS: lunging cuts and point attacks on guarded lines, glowing blades and a flare where they meet, first to five hits.\\[2pt]
\textbf{Checkpoint.} The refinement critic rates Claude Sonnet 5.5 highest at 3.25, compared with 3.0 for the next agent. The checkpoint turns the level into a duel against a small red drone: four guards, cut and point lunges, a score counter, and a way to fight again.
\vspace{2pt}\end{minipage} \\
\midrule
\textbf{Edit derivation} &
A generator proposes 17 candidate edits anchored in the parent's rubric items and the reference, a cross-check by Claude Opus 5.5 and GPT-5.6 Sol reviews them, and one edit per class is selected; each then receives a rubric, an instruction ending in ``Keep everything else as it is'', and a final edit check. A reviewer agent verifies that each target is present in the checkpoint, and a human reviewer approves all three edits. Labels record the edit subtype, scope, and reference dependence. R0 indicates no reference, while R1 indicates a reference-based request. \\
\midrule
\textbf{AP: appearance} &
\textit{``Give the opponent a double-ended energy blade like the dark fighter's in the video, with a glowing beam coming out of each end of the hilt.''} \quad local style, single, R1.\newline
\textbf{Edit success, 3 items.} ES1 one hilt with a beam out of each end; ES2 the second beam matches the first in colour and glow; \textsuperscript{\textbf{t}} ES3 the blade stays whole and in the opponent's hands through every state of the fight.
\textbf{Preservation, 6 items.} The player's blade is unchanged; the opponent keeps its own colour; the second beam adds no new way to score and no extra guard; clashes still flare; the opponent still fights on its own.\newline
\textit{Anchors, e.g.} ES1: 1 = on guard, one hilt with a full-length beam leaving each end; 0.5 = the second beam is a stub, sits at an angle, or looks like a second weapon; 0 = a single-beam blade. \\
\midrule
\textbf{CS: content \& structure} &
\textit{``Add two tall light columns at the sides of the duelling floor, one in each fighter's blade colour. Each column has five segments and lights one more every time that fighter scores a hit.''} \quad add, group, R1.\newline
\textbf{Edit success, 4 items.} ES1 two columns of five segments beside the floor; ES2 one column per fighter, in its blade colour; \textsuperscript{\textbf{t}} ES3 lit segments follow that fighter's hits; \textsuperscript{\textbf{t}} ES4 both columns go dark when a new bout starts.
\textbf{Preservation, 5 items.} The floor keeps its bounds and nothing blocks the footwork; the fighters stay in view; the score display stays; the bout still ends at five hits; simultaneous hits still add nothing. \\
\midrule
\textbf{BF: behavior \& flow} &
\textit{``Make a landed point thrust count as two hits toward the bout instead of one, for either fighter.''} \quad parameter, single, R1.\newline
\textbf{Edit success, 3 items.} ES1 the player's landed thrust adds two; \textsuperscript{\textbf{t}} ES2 so does the opponent's; ES3 a fighter on three hits who lands a thrust wins at once.
\textbf{Preservation, 5 items.} Cuts still count one; a thrust at four hits still ends the bout; a guarded thrust scores nothing; the counter rule for simultaneous hits holds; both fighters return on guard after a thrust. \\
\midrule
\textbf{Evaluation} &
Each Editing task starts from this checkpoint for every agent. Edit-success items check the requested change; preservation items check that the rest of the checkpoint is unchanged, with the initial checkpoint as the reference state; trap items check the parts of a change that are easy to miss. \\
\bottomrule
\end{tabular}
\end{table*}

\begin{table*}[!htbp]
\centering
\footnotesize
\renewcommand{\arraystretch}{1.25}
\caption{Editing task case study based on a video, text, and image Construction task in Stardew Valley. Three edits, one per class, share a validated checkpoint. The top row pairs trailer frames with matching checkpoint scenes. Rubrics are abridged, and a superscript t marks trap items.}
\label{tab:case_edit_littleprince}
\begin{tabular}{@{}>{\raggedright\arraybackslash}p{2.3cm}p{\dimexpr\textwidth-2.3cm-2\tabcolsep\relax}@{}}
\toprule
\textbf{Parent task \& checkpoint} &
\begin{minipage}[t]{\linewidth}\vspace{2pt}
\begin{tabular}{@{}c@{\hspace{2pt}}c@{\hspace{2pt}}c@{\hspace{2pt}}c@{\hspace{2pt}}c@{\hspace{2pt}}c@{}}
\includegraphics[width=0.162\linewidth]{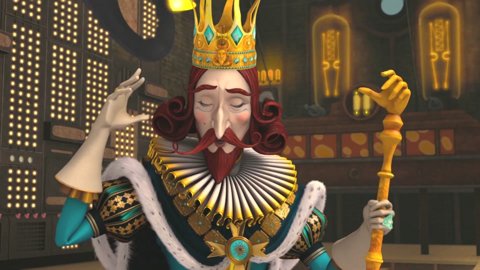} &
\includegraphics[width=0.162\linewidth]{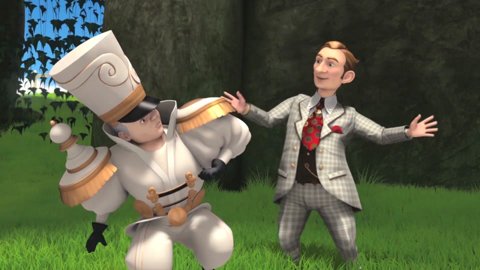} &
\includegraphics[width=0.162\linewidth]{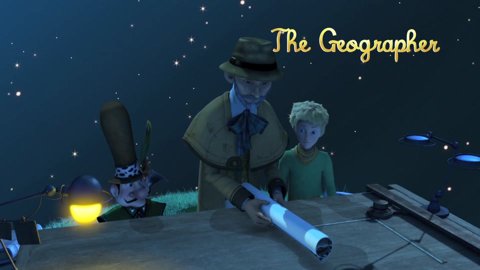} &
\includegraphics[width=0.162\linewidth]{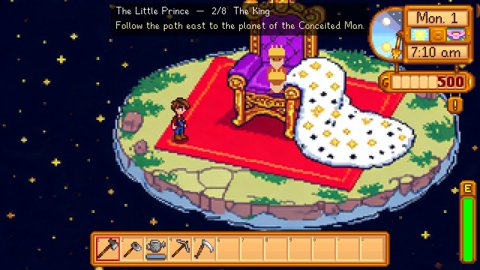} &
\includegraphics[width=0.162\linewidth]{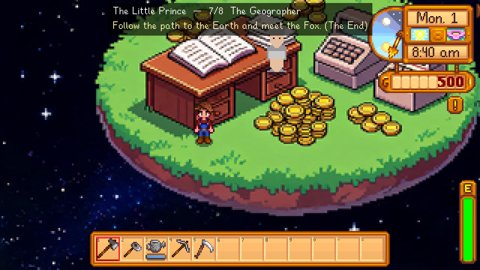} &
\includegraphics[width=0.162\linewidth]{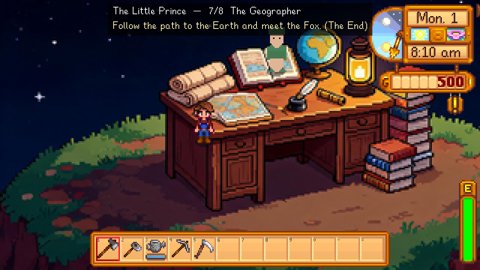} \\[-1pt]
\scriptsize 1:45 king & \scriptsize 1:31 businessman & \scriptsize 1:40 geographer & \scriptsize game: king & \scriptsize game: businessman & \scriptsize game: geographer
\end{tabular}\vspace{2pt}

\textbf{Parent.} A clip of an animated \textit{The Little Prince} trailer, a short text of the story, and six storybook images. The Construction task asks for the prince's journey as a playable story in Stardew Valley: the inhabitants of the planets with their single traits, in the order of the text, ending with the geographer sending the prince to Earth.\\[2pt]
\textbf{Checkpoint.} The refinement critic rates DeepSeek V4.1 Flash highest at 3.25, compared with 3.0 for the next agent. The checkpoint, a Content Patcher pack and a small SMAPI mod, has eight stages from asteroid B-612 through six planets to Earth, each with its inhabitant, dialogue and a chapter banner.
\vspace{2pt}\end{minipage} \\
\midrule
\textbf{Edit derivation} &
A generator proposes 16 candidate edits anchored in the parent's rubric items and the reference, a cross-check by Claude Opus 5.5 and GPT-5.6 Sol reviews them, and one edit per class is selected; each then receives a rubric, an instruction ending in ``Keep everything else as it is'', and a final edit check. A reviewer agent verifies that each target is present in the checkpoint, and a human reviewer approves all three edits. Labels record the edit subtype, scope, and reference dependence. R0 indicates no reference, while R1 indicates a reference-based request. \\
\midrule
\textbf{AP: appearance} &
\textit{``Make the whole globe of the geographer's world look made of paper, with its ground one vast hand-inked map and rolled charts and book towers rising from it.''} \quad local style, single, R1.\newline
\textbf{Edit success, 4 items.} ES1 the ground reads as one huge inked map; \textsuperscript{\textbf{t}} ES2 it holds in every view of that world; ES3 the rim reads as paper against starry space; ES4 rolled charts stand on the map.
\textbf{Preservation, 6 items.} The other worlds keep their own ground; the walk to the geographer, his map table and book towers remain; the world is still a small globe in space; he still sends the prince to Earth. \\
\midrule
\textbf{CS: content \& structure} &
\textit{``Remove the businessman and his bare rocky planet from the journey, so it goes straight from the world before his to the world after it.''} \quad remove, single, R0.\newline
\textbf{Edit success, 4 items.} ES1 the businessman no longer appears; ES2 his planet is no longer a stop; \textsuperscript{\textbf{t}} ES3 the world before his leads straight to the one after; \textsuperscript{\textbf{t}} ES4 what the game tells the player about the journey no longer includes him.
\textbf{Preservation, 6 items.} The journey can still be finished; the remaining worlds keep their order, inhabitants, exchanges, role names, looks and settings. \\
\midrule
\textbf{BF: behavior \& flow} &
\textit{``Let me leave the king's world only after I get him to issue an order that can be obeyed. \ldots trying to leave without an order should turn me back.''} \quad new rule, single, R1.\newline
\textbf{Edit success, 4 items.} ES1 leaving without his order turns the player back; ES2 the player is told the order is needed; ES3 the king refuses an order that cannot be obeyed; \textsuperscript{\textbf{t}} ES4 a reasonable order lets the player move on.
\textbf{Preservation, 6 items.} Every other world is left as before; the journey still reaches Earth; the king's world stays first, with his crown, robe, sceptre and seat of rule; after a refusal, leaving stays blocked.\newline
\textit{Anchors, e.g.} ES1: 1 = every attempt to leave without the order fails; 0.5 = the first attempt is stopped, but a repeated attempt or another way out lets the player leave; 0 = the player travels on without the order. \\
\bottomrule
\end{tabular}
\end{table*}

\section{Evaluation Protocol Details}
\label{app:eval_protocol}

\subsection{Game Capability Probes and Executability Gate}
\label{app:game_capability}

\paragraph{Game Capability Probes.}
Modifications to existing games may break or remove the original game.
For each target game, we build executable probes that drive the game through its own inputs and read its state, covering core mechanics, interactions, and system behaviors.
We run the same probes on the original and modified games under matched conditions and compare their outcomes.
The original game serves as the reference, avoiding the need to manually specify expected states for every capability.

\paragraph{Probe Validation.}
We validate the probes through two complementary tests.
First, we perform A/A testing on independent instances of the original game to identify unstable measurements and establish appropriate tolerances.
Second, we introduce controlled modifications that deliberately break selected game capabilities and check that only the corresponding probes fail.
We exclude probes that cannot be evaluated reliably under the benchmark environment.

\paragraph{Executability Gate.}
The executability gate uses a small set of validated probes for each game, as listed in Table~\ref{tab:capability_pools}. These include a common stability probe and two checks that the original playable content remains accessible.
These probes do not penalize intended changes such as new content or modified values; a submission fails only if it crashes, produces new errors, or removes or replaces the original game wholesale.
The gate uses 27 probes across the nine open-source games and 15 probes across the five commercial games.
Host Fidelity separately assesses whether the new experience uses the target game's original environment and systems, as detailed in Appendix~\ref{app:eval_details}.

\providecommand{\pname}[1]{\textit{#1}}
\newcommand{\pthumb}[1]{\includegraphics[width=1.9cm]{figures/eval/probes/#1}}
\begingroup
\footnotesize
\setlength{\tabcolsep}{5pt}
\setlength{\LTleft}{0pt}\setlength{\LTright}{0pt}
\renewcommand{\arraystretch}{0.92}
\begin{longtable}{@{}>{\raggedright\arraybackslash}m{2.95cm} >{\raggedright\arraybackslash}m{1.85cm} >{\centering\arraybackslash}m{1.9cm} >{\raggedright\arraybackslash}m{\dimexpr\textwidth-6.7cm-6\tabcolsep\relax}@{}}
\caption{Executability gate probes. Each probe runs the same inputs on the modified and an unmodified game under matched conditions and compares the outcomes; the thumbnail shows the function it checks.}
\label{tab:capability_pools}\\
\toprule
\textbf{Game} & \textbf{Probe} & & \textbf{What it checks} \\
\midrule
\endfirsthead
\multicolumn{4}{@{}l}{\small\textit{Table~\ref{tab:capability_pools} continued}} \\
\toprule
\textbf{Game} & \textbf{Probe} & & \textbf{What it checks} \\
\midrule
\endhead
\bottomrule
\endlastfoot
All games & \pname{Stable} & \pthumb{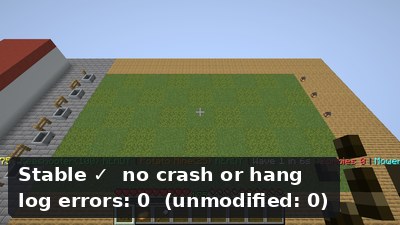} & No crash or hang during the evaluation; no more error lines in the game log than the unmodified game \\
\midrule
\multicolumn{4}{@{}l}{\textit{Open-source games}} \\[2pt]
Celeste 64 & \pname{Levels} & \pthumb{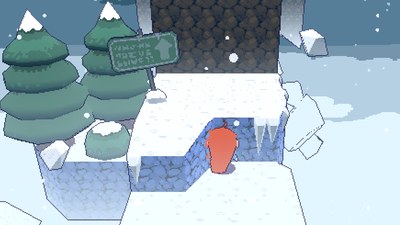} & Every stock map loads and the player spawns alive \\*
 & \pname{Run} & \pthumb{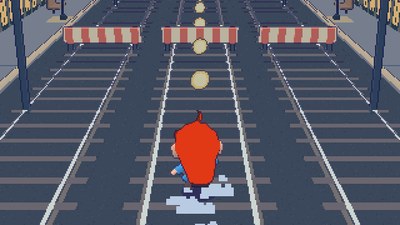} & Running speed and acceleration match \\
Colobot & \pname{Levels} & \pthumb{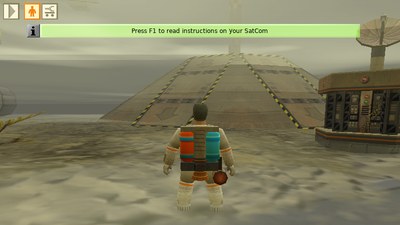} & Stock missions, exercises, challenges, and free game load and run \\*
 & \pname{Bots} & \pthumb{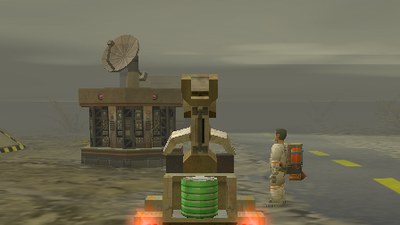} & Stock robots of all four chassis move and turn with matching time, distance, and energy \\
Flare & \pname{New game} & \pthumb{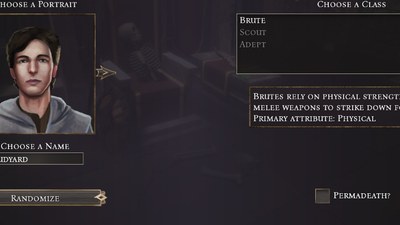} & A new game starts in the first map with the stock class kit \\*
 & \pname{Move} & \pthumb{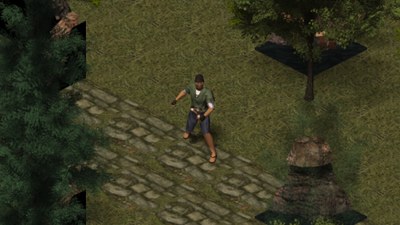} & Keyboard movement speed and in-map teleport match \\
Godot TPS & \pname{Level} & \pthumb{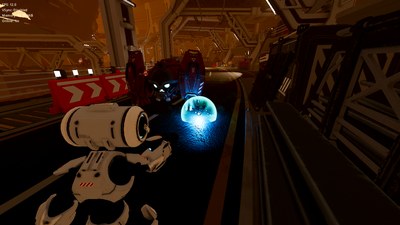} & The level loads with its geometry, collision, spawn points, and robots \\*
 & \pname{Move} & \pthumb{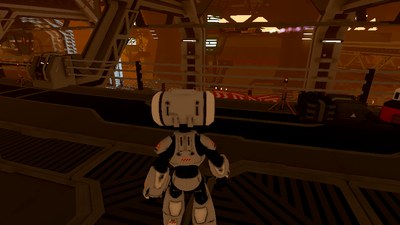} & Third-person walking distance and direction match \\
SuperTuxKart & \pname{Tracks} & \pthumb{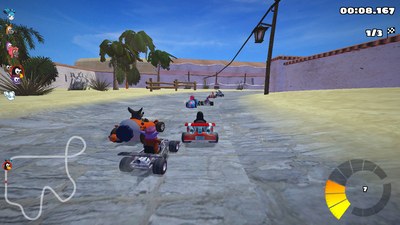} & All 21 stock tracks load and AI karts drive on them \\*
 & \pname{Acceleration} & \pthumb{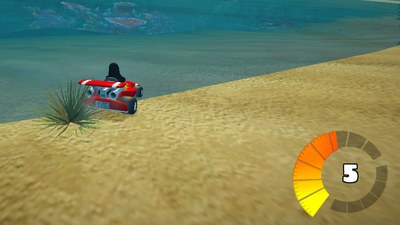} & Acceleration and top speed match \\
Tanks of Freedom II & \pname{Menu} & \pthumb{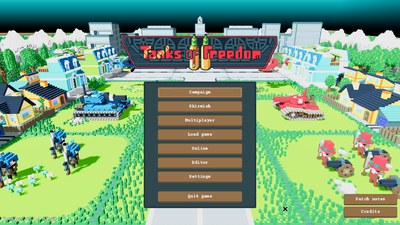} & Map picker, skirmish setup, campaign, and settings work \\*
 & \pname{Maps} & \pthumb{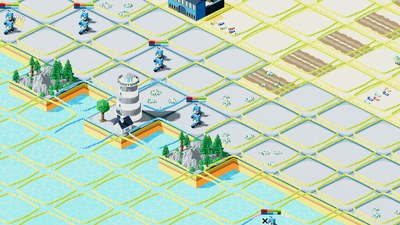} & All 12 stock maps start with the stock sides, terrain, buildings, and units \\
Tuxemon & \pname{Move} & \pthumb{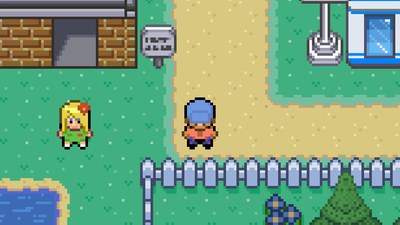} & Grid walking and wall collisions match on a stock map \\*
 & \pname{Warp} & \pthumb{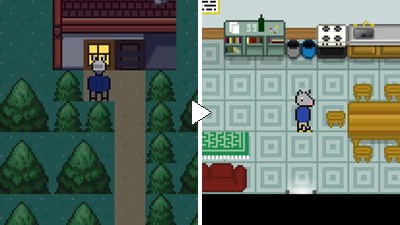} & Stairs and doors move the player between stock maps \\
VoxeLibre & \pname{Join} & \pthumb{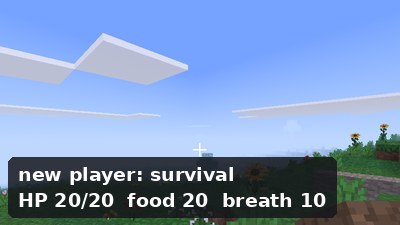} & A new player starts in survival with full health, food, and breath \\*
 & \pname{Dig and place} & \pthumb{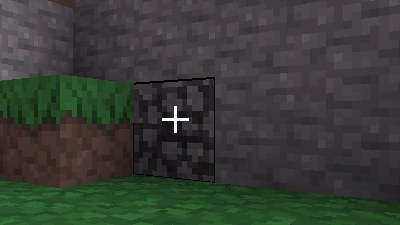} & Blocks can be placed and dug, and dug blocks drop \\
Xonotic & \pname{Maps} & \pthumb{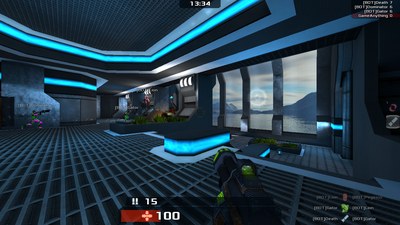} & Stock maps load and a player spawns \\*
 & \pname{Loadout} & \pthumb{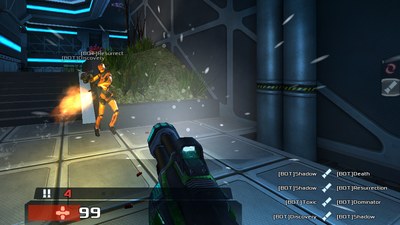} & A new player spawns with the stock health, armor, and weapons \\
\midrule
\multicolumn{4}{@{}l}{\textit{Commercial games}} \\[2pt]
Minecraft & \pname{Join} & \pthumb{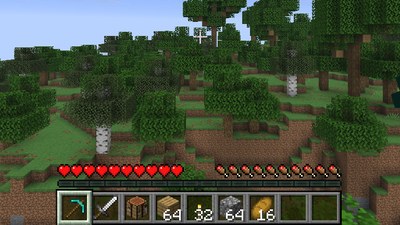} & A new player starts in survival with full health and food \\*
 & \pname{Move} & \pthumb{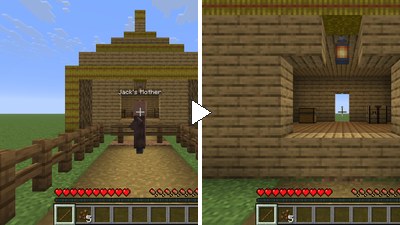} & Walking and jumping work \\
Stardew Valley & \pname{Walk} & \pthumb{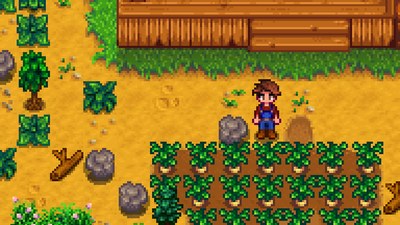} & The farmer walks at the stock speed \\*
 & \pname{Warp} & \pthumb{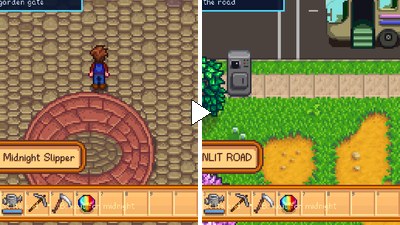} & Walking off a map edge moves between stock locations \\
Terraria & \pname{Move} & \pthumb{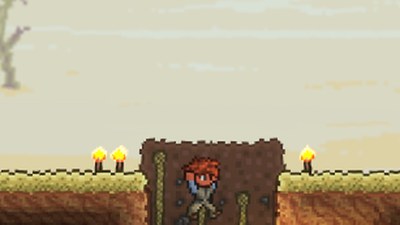} & Running, jumping, falling, and the grappling hook match \\*
 & \pname{Dig and place} & \pthumb{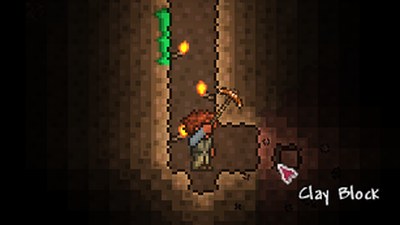} & Mining, block placement, and wall removal work \\
Valheim & \pname{Move} & \pthumb{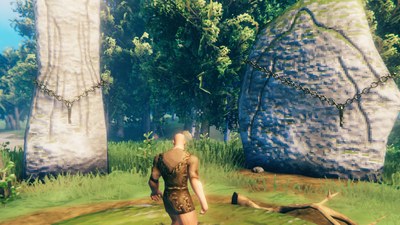} & Walking, running, jumping, swimming, stamina, and fall damage match \\*
 & \pname{Build} & \pthumb{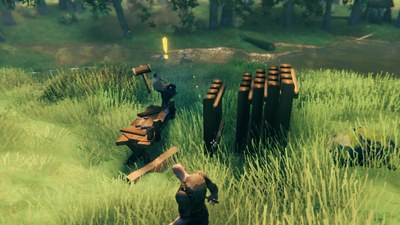} & The hammer places a workbench and structural support holds \\
Cities: Skylines & \pname{Simulation} & \pthumb{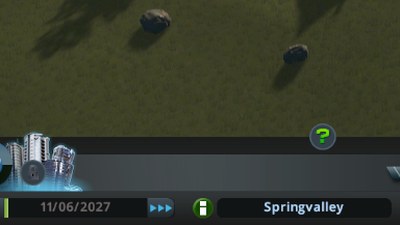} & The simulation runs at the stock speeds and pauses \\*
 & \pname{Roads} & \pthumb{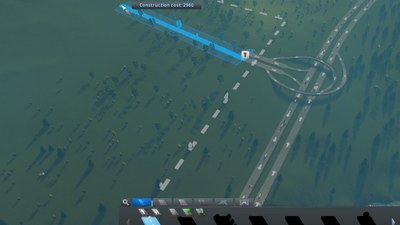} & The road tool builds, upgrades, and removes roads \\
\midrule
\multicolumn{4}{@{}l}{\textbf{Total}: 27 probes on the nine open-source games and 15 on the five commercial games with \pname{Stable} counted once per game} \\
\end{longtable}
\endgroup

\subsection{Agent Execution and Independent Evaluation}
\label{app:agentic_eval}

\paragraph{Agent Execution and Submission.}
Each agent receives the reference materials, the task instruction, the target game, and a per-game development kit.
Execution is a single stage with a development budget of 120 minutes of wall-clock time per task, within which the agent analyzes the references, implements the modification, and tests it in the running game.
Agents interact with the game through a shared command-line interface that builds the modification, starts a fresh game, sends keyboard and mouse input, captures screenshots, reads the structured game state, and replays input traces.
The run ends when the agent marks its submission as complete or the budget is exhausted; an agent that stops early without marking completion is asked to continue.

Each task runs in its own Docker container built from an image pinned by digest, with a fresh game instance and no network access.
Agents run inside these containers using Claude Code, Codex CLI, or OpenCode with their default prompts and tools.
Model requests leave the container only through a host-side proxy that attaches the credentials, so no credential is visible to the agent.
Failures of the environment rather than the agent, such as an unavailable container or a model API error, are detected and classified automatically; the affected task is rerun and is not scored as an agent failure.

The final submission includes the game modification, gameplay traces, screenshots, and instructions for accessing the created content.
We freeze the submitted artifacts and evaluate them in a separate, clean game instance.
This separation prevents temporary development states from affecting the evaluation.

\paragraph{Action Replay and Independent Agent Playtesting.}
We employ two complementary methods to evaluate the implemented functions and gameplay experience.
\textit{Action Replay} executes the player actions submitted by the development agent, while \textit{Independent Agent Playtesting} actively explores the modified game.

Let $G'$ denote the modified game and $A_{\mathrm{sub}}$ the submitted action sequence.
The replay trace is obtained by
\begin{equation}
\tau_{\mathrm{rep}}
=
\operatorname{Execute}(G', A_{\mathrm{sub}}).
\label{eq:replay}
\end{equation}
For independent playtesting, an evaluation agent with policy $\pi_{\mathrm{eval}}$ interacts with the same modified game:
\begin{equation}
\tau_{\mathrm{play}}
\sim
\operatorname{Interact}(G', \pi_{\mathrm{eval}}).
\label{eq:eval_playtest}
\end{equation}

Action Replay verifies the functions demonstrated by the development agent.
The evaluator reexecutes the submitted actions in a clean game instance and records the resulting observations.
Independent Agent Playtesting further examines the game without relying solely on these demonstrations.
It first explores the game without access to the developer's traces or navigation instructions.
In a second stage, it receives the submitted traces and access instructions to verify specific functions and investigate potential failures.
This allows the evaluator to assess both demonstrated functionality and behaviors encountered through independent exploration.
Each playtest is limited to 40 actions and 30 minutes; the judge may stop earlier once it has the evidence it needs. Appendix~\ref{app:eval_stability} shows that agreement with human labels stops improving beyond 30 to 40 actions.

\paragraph{Evidence Collection and Adjudication.}
Both evaluators collect screenshots and runtime game states during interaction.
\textit{Screenshots provide evidence of visual outcomes and spatial arrangements, while state changes help verify functional behaviors.}
For mechanisms and narrative events, we examine whether player actions produce the expected state transitions.
This helps distinguish functional implementations from static visual demonstrations.
Runtime states are read as structured JSON containing the engine fields listed in Table~\ref{tab:state-fields}, content added by the modification, and on-screen text. Every replay and playtest screenshot is paired with a state snapshot, allowing the judge to compare the effects of actions. For Tanks of Freedom II and Cities: Skylines, which lack a directly controlled avatar, state changes in buildings, units, resources, or game time provide much of the functional evidence and must agree with the screenshots.

\begin{table*}[!htbp]
\centering
\caption{Structured runtime state returned by the \texttt{game state} call. Top: per-game \texttt{engine} fields by kind; every game also returns content added by the modification in the \texttt{mod} field and on-screen text. $^\dagger$Read from adapter code only. Bottom: five evaluated submissions, each a replay screenshot with an excerpt of its paired snapshot.}
\label{tab:state-fields}
\footnotesize
\setlength{\tabcolsep}{4pt}
\renewcommand{\arraystretch}{1.0}
\begin{tabular}{@{}l >{\raggedright\arraybackslash}p{0.29\textwidth} >{\raggedright\arraybackslash}p{0.11\textwidth} >{\raggedright\arraybackslash}p{0.16\textwidth} >{\raggedright\arraybackslash}p{0.175\textwidth}@{}}
\toprule
\textbf{Game} & \textbf{Player} & \textbf{World \& place} & \textbf{Entities \& objects} & \textbf{Progress, time \& UI} \\
\midrule
Celeste 64 & position, velocity, dash state & map, checkpoint & actor counts & strawberries, deaths, completion record; menu \\
Colobot & astronaut position, shield, energy, load & level & object counts by type & mission time \\
Flare & position, HP/MP, level, XP; gold, inventory, equipment & map & enemies, NPCs & dialog, campaign flags; menus \\
Godot TPS & position, heading, aiming/shooting, camera & -- & robot enemies with health; bullets & menu; level time \\
SuperTuxKart & kart position and speed & track & -- & race setup fields for mode, laps, karts, and difficulty; screen \\
Tanks of Freedom II & cursor, selection, action points & map & units and buildings per side & mission, turn; story dialog, triggers \\
Tuxemon & tile, facing, money, steps; bag; monster party & map & NPC counts & story-variable counts; state stack; clock \\
VoxeLibre & position, look, HP, food, breath, XP; wielded item, inventory & weather & nearby entities & day time \\
Xonotic & position, health, armor, weapons, ammo, frags & map, gametype & other players and bots & match phase \\
\midrule
Minecraft & position, look, health, food, XP, mode; inventory, effects & dimension, biome, weather & nearby entities & time \\
Stardew Valley & tile, health, stamina, money, skills; inventory & location & nearby NPCs, friendship & quests, flags, mail; date \\
Terraria & position, life, mana, defense, coins, inventory, buffs & world & nearby NPCs, bosses & bosses downed; menus; clock \\
Valheim & position, health, stamina, skills; inventory, equipment & biome & nearby creatures and items & progress keys; raids; time \\
Cities: Skylines$^\dagger$ & camera, tool & money, population, happiness, demand & building, road, vehicle, citizen counts; transport lines & date \\
\bottomrule
\end{tabular}

\vspace{3pt}
\noindent\begin{minipage}[t]{0.19\textwidth}\raggedright
\includegraphics[width=\linewidth]{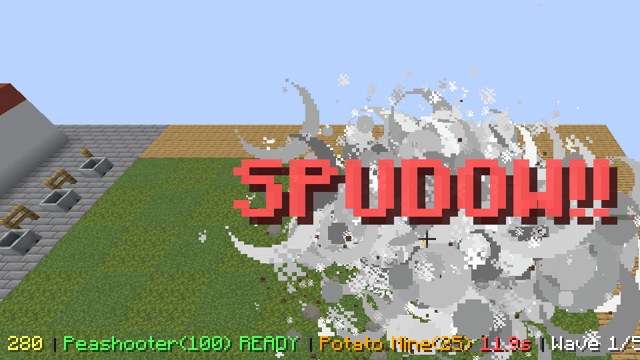}\par\vspace{2pt}
{\scriptsize\textbf{Mod-added entities}}\par\vspace{1pt}
\fontsize{5.6}{6.4}\selectfont
\begin{verbatim}
"mod": {"mods":
  {"spudow": ...},
 "entities": {
  "minecraft:zombie#9":
   {"name": "Zombie 20"},
  "minecraft:minecart#6":
   {"name": "Lawn Mower"},
  ...}}
\end{verbatim}
\end{minipage}\hfill
\begin{minipage}[t]{0.19\textwidth}\raggedright
\includegraphics[width=\linewidth]{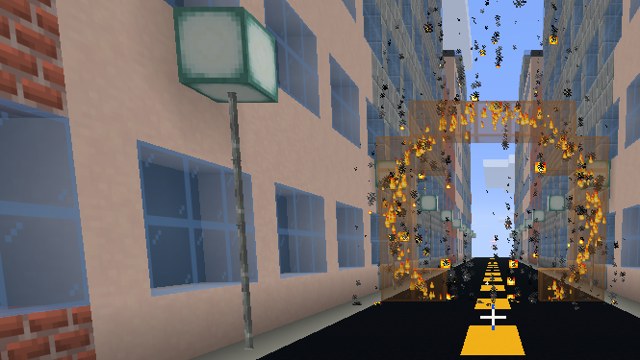}\par\vspace{2pt}
{\scriptsize\textbf{Avatar, on-screen text}}\par\vspace{1pt}
\fontsize{5.6}{6.4}\selectfont
\begin{verbatim}
"pos": [22.5, 151.0, 0.5],
"held":
  "minecraft:blaze_rod",
"screen_text": [
 "actionbar: ring 16m |
  pursuer 18m | city:
  flat", ...]
\end{verbatim}
\end{minipage}\hfill
\begin{minipage}[t]{0.19\textwidth}\raggedright
\includegraphics[width=\linewidth]{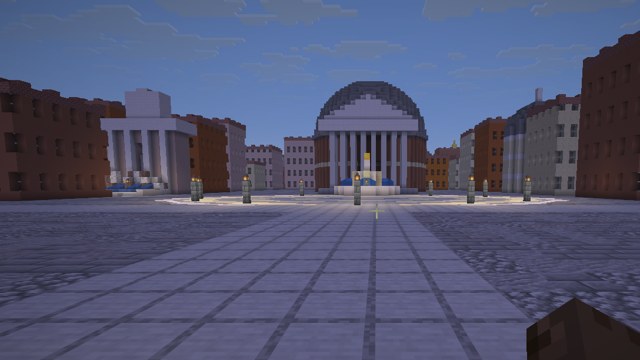}\par\vspace{2pt}
{\scriptsize\textbf{Position in a built city}}\par\vspace{1pt}
\fontsize{5.6}{6.4}\selectfont
\begin{verbatim}
"pos": [-87.9, -60.0,
  27.5],
"mod": {"mods":
  {"rome": ...}},
"screen_text": [
 "chat: ROME - ...
  Pantheon: west
  (-90,-55). ..."]
\end{verbatim}
\end{minipage}\hfill
\begin{minipage}[t]{0.19\textwidth}\raggedright
\includegraphics[width=\linewidth]{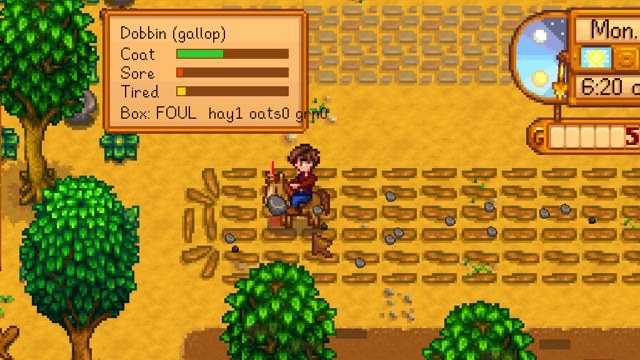}\par\vspace{2pt}
{\scriptsize\textbf{Riding an added horse}}\par\vspace{1pt}
\fontsize{5.6}{6.4}\selectfont
\begin{verbatim}
"location": "Farm",
"riding": "Dobbin",
"inventory": {...,
  "Hoof Lotion": 4,
  "Hoof Pick": 1, ...},
"smapi_mods": [
  "HorseKeeping 1.0.0
  ..."]
\end{verbatim}
\end{minipage}\hfill
\begin{minipage}[t]{0.19\textwidth}\raggedright
\includegraphics[width=\linewidth]{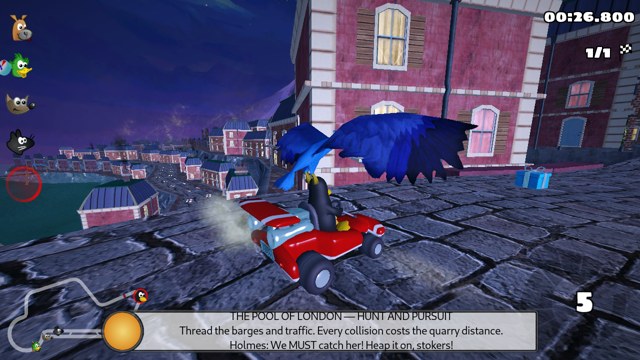}\par\vspace{2pt}
{\scriptsize\textbf{Kart on an added track}}\par\vspace{1pt}
\fontsize{5.6}{6.4}\selectfont
\begin{verbatim}
"track":
 "addon_thames_pursuit",
"karts": [{"kart": 0,
  "speed": 17.0,
  "max_speed": 20.0,
  ...}],
"tracks_added": {
 "addon_thames_pursuit":
  "race"}
\end{verbatim}
\end{minipage}
\end{table*}

For visual evaluation, a vision judge compares screenshots of the generated content against the corresponding reference evidence and task rubrics.
The comparison focuses on semantic and visual characteristics rather than exact reproduction, allowing different visual styles across target games.
Candidate screenshots come from the action replay and the independent playtest, both captured by the evaluator in a clean game instance; for each rubric item, the judge selects the screenshots that best show the item before scoring it.

The replay and playtesting evaluators independently score each rubric item.
When their judgments agree, the score is accepted directly.
Otherwise, an additional judge examines the supporting evidence and scoring criteria to resolve the disagreement.
We record the initial judgments and adjudication results; Appendix~\ref{app:evidence_ablation} reports how often the verifiers agree and how adjudication changes the scores.

\subsection{Scoring Rules and Aggregation}
\label{app:eval_details}

\paragraph{Rubric Scoring and Executability Gate.}
Construction is evaluated on Content, Functions \& Playability, Visual \& Presentation, and Host Fidelity; Editing on Edit Success, Preservation, and Host Fidelity.
Each rubric item includes explicit scoring criteria and supporting evidence.
Rubric items use three scoring levels of 0, 0.5, and 1; Host Fidelity is also scored 0, 0.5, or 1, derived from a per-game host-feature checklist as described below.
We assign equal importance to all applicable rubric items within each dimension and do not use additional critical item penalties.

Before evaluation, we check whether the submitted modification can be built, installed, and loaded in a clean game instance, and the gate probes of Appendix~\ref{app:game_capability} check that the game stays stable and that the original game is still present.
If this check fails because of the submitted modification, all quality scores for the task are set to zero.
Evaluation environment failures are handled separately.
Environment failures, such as a container or game start failure caused by the evaluation infrastructure or a judge call that fails after retries, mark the evaluation as an infrastructure error rather than a submission failure. Transient errors are retried automatically up to two times; remaining infrastructure errors are inspected and rerun, and are never scored as zero.

\paragraph{Construction and Editing Scoring Rules.}
The two task settings share the same evaluation dimensions but differ in their scoring objectives.
For Construction, the Content, Functions \& Playability, and Visual \& Presentation rubrics assess how well the generated game realizes the reference materials and user instructions, while Host Fidelity checks whether the adaptation is built on and played through the target game.
For Editing, Edit Success items check whether the requested change to content, behavior, or appearance is realized; an item that cannot be tested from the evidence counts as zero.
Preservation items compare the edited game against its initial checkpoint to detect unintended changes to content and functions the edit should not touch; items that do not apply to the edit are excluded.
Host Fidelity uses the same per-game checklist as in Construction.
The Editing Overall calculation, including penalties for failures in Preservation and Host Fidelity, is specified below.

\paragraph{Host Fidelity Scoring.}
Host Fidelity assesses whether the created experience builds upon the original game rather than replacing its world and systems with an independent implementation. It does not require preserving the game's original genre activities. For each game, we evaluate five host features, including the world or scenes, avatar and controls, two groups of native game systems, and visual style. A judge assesses each feature using gameplay evidence and assigns a score of 0, 0.5, or 1. A score of 0 indicates that the original world or avatar and controls are replaced, or no feature is fully used. A score of 0.5 indicates that another feature is replaced or more than half of the judgeable features are only partially used. Otherwise, the score is 1. Submissions with fewer than two judgeable features are excluded from Host Fidelity aggregation.

\paragraph{Score Aggregation and Statistical Analysis.}
Let $\mathcal{R}_{i,d}^{t}$ denote the applicable rubric items for task $i$, dimension $d$, and task setting $t \in \{\mathrm{Con}, \mathrm{Edit}\}$.
Let $q_{i,r}^{t}$ denote the score of rubric item $r$, and $z_i^t$ indicate whether the submitted modification passes the executability check.
The task score for each applicable dimension is
\begin{equation}
s_{i,d}^{t}
=
z_i^t
\frac{1}{|\mathcal{R}_{i,d}^{t}|}
\sum_{r \in \mathcal{R}_{i,d}^{t}}
q_{i,r}^{t}.
\label{eq:dim_score}
\end{equation}

We average the scores over all tasks with applicable rubrics for each dimension:
\begin{equation}
S_d^t
=
\frac{1}{|\mathcal{I}_d^t|}
\sum_{i \in \mathcal{I}_d^t}
s_{i,d}^{t},
\label{eq:dim_mean}
\end{equation}
where $\mathcal{I}_d^t$ denotes the corresponding set of tasks.
All quality scores are reported as percentages.

For each task, let $C$, $F$, $V$, and $H$ denote Content, Functions \& Playability, Visual \& Presentation, and Host Fidelity, while $E$ and $P$ denote Edit Success and Preservation. All dimension scores range from zero to one. We compute the Overall scores as
\begin{align}
\mathrm{Overall}_{\mathrm{Con}} &= 0.3C + 0.3F + 0.3V + 0.1H, \label{eq:overall_task} \\
\mathrm{Overall}_{\mathrm{Edit}} &= E[1 - 0.4(1-P) - 0.1(1-H)]. \label{eq:overall_edit}
\end{align}
For Editing, Edit Success determines the base score, while failures in Preservation and Host Fidelity incur penalties. When a dimension is unavailable, its contribution is excluded and the remaining weights are renormalized. Submissions failing the executability gate receive zero. Overall scores are averaged across tasks and reported as percentages.

Model comparisons use identical tasks and initial game states. We compute 95\% confidence intervals using 4,000 paired bootstrap resamples over tasks. Since multiple tasks may share references or target games, we additionally perform cluster bootstrap resampling over reference groups and games. The agent rankings remain unchanged across all three resampling schemes, although game-level resampling yields wider confidence intervals. Most model differences remain statistically significant, while small differences between closely performing agents should not be over-interpreted. Detailed confidence intervals for native agents and refinement gains are reported in Tables~\ref{tab:app_ci_native} and~\ref{tab:app_ci_refine}.

\begin{table}[!htbp]
\centering\small
\caption{Native agent Overall scores with 95\% bootstrap confidence intervals from 4{,}000 resamples. Task resampling is paired across configurations, while game resampling draws complete games from the 14 Construction and eight Editing target games.}
\label{tab:app_ci_native}
\setlength{\tabcolsep}{4pt}
\begin{tabular}{@{}l ccc ccc@{}}
\toprule
& \multicolumn{3}{c}{\textbf{Construction}} & \multicolumn{3}{c}{\textbf{Editing}} \\
\cmidrule(lr){2-4}\cmidrule(lr){5-7}
\textbf{Model} & Overall & Task CI & Game CI & Overall & Task CI & Game CI \\
\midrule
Claude Fable 5.2 & 84.1 & 81.7 to 86.4 & 79.7 to 87.5 & 86.5 & 84.0 to 88.9 & 83.6 to 88.7 \\
Claude Sonnet 5.5 & 59.2 & 54.9 to 63.4 & 51.3 to 65.8 & 72.8 & 68.1 to 77.3 & 69.8 to 76.6 \\
DeepSeek V4.1 Flash & 51.4 & 46.9 to 56.0 & 41.3 to 59.7 & 69.9 & 64.9 to 74.7 & 65.6 to 74.5 \\
GPT-5.6 Sol & 40.9 & 36.7 to 45.0 & 33.6 to 46.9 & 72.5 & 68.5 to 76.7 & 68.1 to 77.5 \\
Kimi K3 & 35.7 & 30.7 to 40.7 & 25.8 to 44.6 & 67.7 & 62.3 to 72.8 & 62.1 to 72.6 \\
Qwen3.8-Flash & 35.2 & 30.7 to 39.9 & 28.5 to 41.4 & 61.7 & 54.8 to 68.4 & 55.8 to 69.5 \\
Qwen3.8-Max & 32.2 & 28.1 to 36.6 & 25.8 to 38.2 & 68.5 & 62.4 to 74.1 & 60.8 to 75.3 \\
GLM-5.3-Flash & 29.0 & 24.9 to 33.1 & 22.5 to 34.6 & 63.4 & 57.0 to 69.7 & 55.8 to 70.9 \\
Gemini 3.8 Flash & 27.6 & 24.1 to 31.1 & 24.0 to 30.9 & 61.3 & 54.8 to 67.8 & 53.1 to 67.7 \\
\bottomrule
\end{tabular}
\end{table}
\begin{table}[!htbp]
\centering\small
\caption{Overall gains from iterative refinement relative to native development. The table reports the changes shown in Table~\ref{tab:refinement_results} with 95\% confidence intervals from 4{,}000 paired task bootstrap resamples. Bold gains have intervals excluding zero.}
\label{tab:app_ci_refine}
\setlength{\tabcolsep}{3.5pt}
\resizebox{\textwidth}{!}{%
\begin{tabular}{@{}l cccc cccc@{}}
\toprule
& \multicolumn{4}{c}{\textbf{Construction}} & \multicolumn{4}{c}{\textbf{Editing}} \\
\cmidrule(lr){2-5}\cmidrule(lr){6-9}
\textbf{Model} & $\Delta$R1 & CI & $\Delta$R3 & CI & $\Delta$R1 & CI & $\Delta$R3 & CI \\
\midrule
Claude Sonnet 5.5 & \textbf{+4.5} & +2.0 to +7.3 & \textbf{+10.5} & +7.0 to +14.1 & +1.4 & $-$4.1 to +6.9 & +3.6 & $-$0.6 to +8.1 \\
Qwen3.8-Flash & +0.8 & $-$2.8 to +4.1 & \textbf{+14.8} & +10.2 to +19.2 & \textbf{+10.2} & +3.5 to +16.8 & \textbf{+16.4} & +10.1 to +23.0 \\
DeepSeek V4.1 Flash & +1.1 & $-$1.2 to +3.3 & \textbf{+7.6} & +3.9 to +11.6 & \textbf{+7.0} & +3.0 to +10.9 & \textbf{+7.3} & +3.1 to +11.7 \\
\midrule
GPT-5.6 Sol & \textbf{+6.0} & +2.2 to +9.9 & -- & -- & +0.8 & $-$3.0 to +4.8 & -- & -- \\
Kimi K3 & \textbf{+3.9} & +0.3 to +7.4 & -- & -- & \textbf{+5.6} & +1.2 to +10.3 & -- & -- \\
Qwen3.8-Max & \textbf{+4.6} & +1.3 to +8.2 & -- & -- & \textbf{+6.6} & +2.0 to +11.7 & -- & -- \\
GLM-5.3-Flash & +1.0 & $-$2.0 to +4.1 & -- & -- & \textbf{+9.3} & +4.2 to +14.9 & -- & -- \\
Gemini 3.8 Flash & \textbf{+10.1} & +7.0 to +13.1 & -- & -- & \textbf{+5.6} & +0.1 to +11.5 & -- & -- \\
\bottomrule
\end{tabular}}
\end{table}

We also report executability rates, development time, and average inference cost per task.
Construction and Editing costs are reported separately.
For iterative configurations, costs include all development rounds.
Evaluation costs are excluded from the reported development costs.
Repeated evaluations of the same submissions on a representative subset are analyzed in Appendix~\ref{app:eval_stability}.

\section{Experimental Setup Details}
\label{app:exp_settings}

\subsection{Agent Configurations and Execution Settings}
\label{app:agent_config}

\paragraph{Agent Configurations.}
We treat each model and its coding harness as a complete agent configuration. Claude Code, Codex CLI, and OpenCode retain their native prompts, tools, and reasoning settings, without manual reasoning effort overrides. With these provider defaults, Gemini 3.8 Flash in OpenCode uses only about 3k reasoning tokens per run, compared with 37k to 103k for the other OpenCode backbones. It also frequently stops early. In 18\% of its Construction runs, it ends after three consecutive empty replies, which contributes to its low scores. We keep the default for all configurations rather than tune one backbone. Table~\ref{tab:app_agent_configs} lists the model identifiers and harness versions used for reproducibility.

\begin{table}[!htbp]
\centering
\caption{Coding agent configurations.}
\label{tab:app_agent_configs}
\small
\setlength{\tabcolsep}{4pt}
\begin{tabular}{@{}llll@{}}
\toprule
\textbf{Harness} & \textbf{Version} & \textbf{Model} & \textbf{Model ID} \\
\midrule
\multirow{2}{*}{Claude Code} & \multirow{2}{*}{2.1.318} & Claude Fable 5.2 & \texttt{claude-fable-5-2} \\
 & & Claude Sonnet 5.5 & \texttt{claude-sonnet-5-5} \\
\midrule
Codex CLI & 0.160.0 & GPT-5.6 Sol & \texttt{gpt-5.6-sol} \\
\midrule
\multirow{6}{*}{OpenCode} & \multirow{6}{*}{1.19.5} & Gemini 3.8 Flash & \texttt{gemini-3.8-flash} \\
 & & DeepSeek V4.1 Flash & \texttt{deepseek-v4.1-flash} \\
 & & GLM-5.3-Flash & \texttt{glm-5.3-flash} \\
 & & Qwen3.8-Max & \texttt{qwen3.8-max} \\
 & & Qwen3.8-Flash & \texttt{qwen3.8-flash} \\
 & & Kimi K3 & \texttt{kimi-k3} \\
\bottomrule
\end{tabular}
\end{table}

\begin{figure}[!htbp]
\centering
\definecolor{flowV}{HTML}{5B3A9E}\definecolor{flowBand}{HTML}{F1EBFD}\definecolor{flowGrey}{HTML}{666C75}\definecolor{flowRef}{HTML}{FFF4E8}\definecolor{flowOr}{HTML}{E08A2B}
\begin{tikzpicture}[font=\footnotesize, node distance=13mm and 5mm,
  box/.style={draw=flowV, fill=flowBand, rounded corners=2pt, align=center, text width=3.25cm, minimum height=1.75cm, inner sep=3pt, execute at begin node={\hyphenpenalty=10000\exhyphenpenalty=10000}},
  rbox/.style={box, draw=flowOr, fill=flowRef},
  arr/.style={-{Stealth[length=1.8mm]}, semithick, flowV},
  rarr/.style={-{Stealth[length=1.8mm]}, semithick, flowOr},
  lab/.style={font=\scriptsize, text=flowGrey, align=center}]
\node[box] (in) {\textbf{Task inputs}\\reference, instruction, target game; Editing: + fixed checkpoint};
\node[box, right=of in] (dev) {\textbf{Development}\\native harness; 120 min initially, 60 min per revision};
\node[box, right=of dev] (sub) {\textbf{Frozen submission}\\modification, access guide, 2--4 traces, 3--6 screenshots};
\node[box, right=of sub, fill=white, draw=black!70] (ev) {\textbf{Independent evaluation}\\gate, replay, playtest, judges; see Appendix~\ref{app:eval_details}};
\draw[arr] (in) -- (dev); \draw[arr] (dev) -- (sub); \draw[arr] (sub) -- (ev);
\node[rbox, below=of sub] (rep) {\textbf{1. Replay}\\the submitted traces on a fresh game instance};
\node[rbox, below=of dev] (cr) {\textbf{2. Critic review}\\separate session; checks coverage and reference likeness};
\node[rbox, below=of in] (mem) {\textbf{3. Memory}\\retrieves up to 6k chars of lessons; adds up to 3 new ones};
\draw[rarr] (sub) -- (rep);
\draw[rarr] (rep) -- (cr);
\draw[rarr] (cr) -- (mem);
\draw[rarr] (cr) -- node[lab, right]{feedback} (dev);
\draw[rarr] (mem.north) -- node[lab, left, pos=0.45]{lessons} (dev.south west);
\node[below=of ev, text width=3.25cm, align=center, font=\scriptsize, text=flowGrey] (note) {The critic never sees the hidden rubrics or evaluation results, and its scores are not used in the evaluation.};
\begin{scope}[on background layer]
\node[draw=flowOr, dashed, rounded corners=3pt, fit=(mem)(cr)(rep), inner sep=5pt, label={[font=\scriptsize\bfseries, text=flowOr, anchor=north west]south west:+Ours refinement: R1 for all, R3 for three, then return to Development}] (loop) {};
\end{scope}
\end{tikzpicture}
\caption{Execution of one task. The top row shows native development followed by independent evaluation. The lower loop shows +Ours refinement, in which an independent critic reviews replay evidence and the same agent revises the project with feedback and retrieved lessons.}
\label{fig:app_exec_flow}
\end{figure}
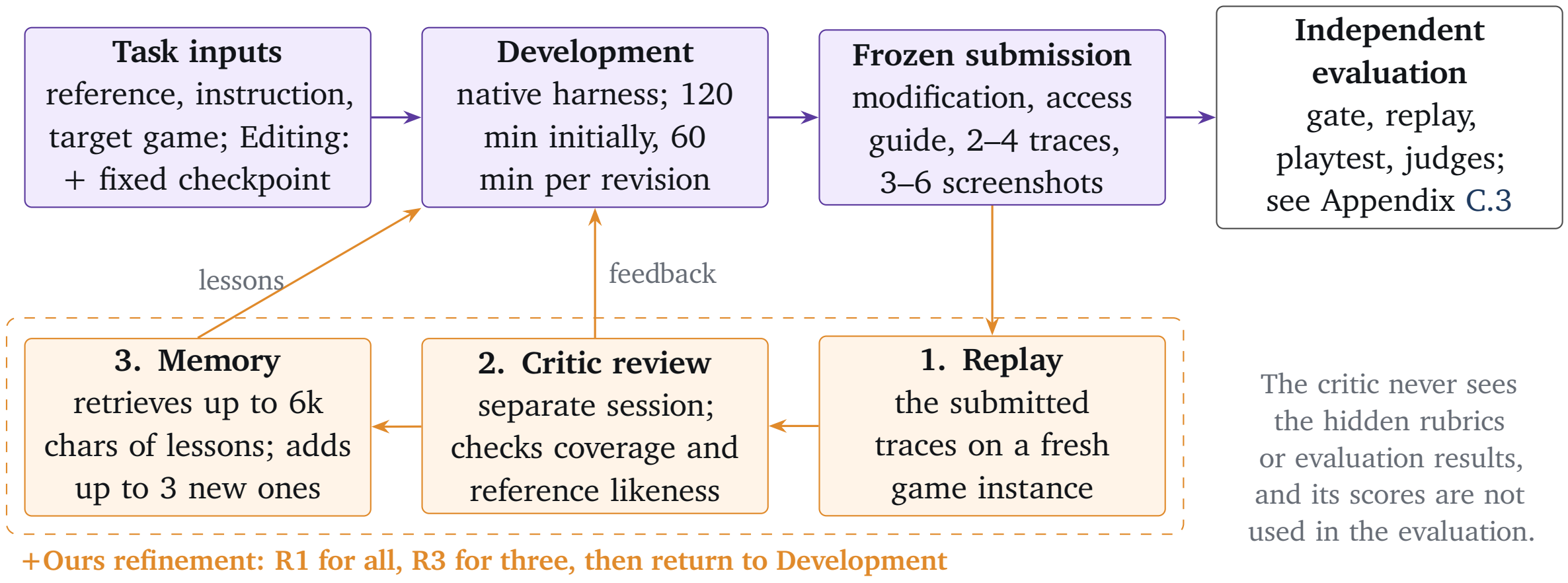

\paragraph{Execution Settings.}
Figure~\ref{fig:app_exec_flow} shows how one task is run. All configurations receive the same inputs and game development interface and retain their native harness tools. Each attempt runs in an isolated container, and its submission is frozen before independent evaluation, as detailed in Table~\ref{tab:app_execution}. The iterative configuration, denoted +Ours, adds refinement rounds before the final evaluation.

\begin{table}[!htbp]
\centering
\caption{Agent execution settings.}
\label{tab:app_execution}
\small
\begin{tabular}{@{}p{0.29\linewidth}p{0.65\linewidth}@{}}
\toprule
\textbf{Setting} & \textbf{Value} \\
\midrule
Attempts & One autonomous attempt per task and configuration \\
Native budget & 120 min of wall clock development time \\
+Ours refinement round & Replay of the submitted traces, critic review using OpenCode and Gemini 3.8 Flash with scores and feedback, then 60 min of revision by the same agent with retrieved lessons; R1 for all configurations, R3 for three. Construction R1 used 40 minutes without lessons on 98 tasks, excluding Cities: Skylines, Flare, Terraria, and Valheim \\
Spend limits & \$20 per development run, \$15 per refinement round, \$3 per critic review; requests above 1M tokens are refused \\
Task inputs & Reference materials, task instruction, target game, and development kit; Editing also receives a fixed checkpoint \\
Environment & Isolated Docker container with a pinned image and a fresh game instance; Ubuntu 24.04 \\
Display & Private virtual X display at $1024\times768$, or a $1280\times720$ game window \\
Rendering & NVIDIA L40S for 3D games, with two tasks per GPU; software OpenGL when required \\
Game interface & \texttt{game} CLI for build, start/reset, player input, screenshot, state queries, trace replay, and logs \\
Completion & Agent marks completion or reaches the time limit; early unmarked stops are resumed \\
Submission & Modification, access guide, 2--4 input traces, and 3--6 screenshots, frozen before evaluation \\
Network and credentials & No direct Internet access; model calls use a host proxy that injects credentials outside the container \\
Environment failures & Infrastructure failures recorded separately from submission failures; transient failures retried up to two times, remaining ones inspected and rerun \\
\bottomrule
\end{tabular}
\end{table}

\subsection{Evaluation and Cost Settings}
\label{app:evaluation_cost}

\paragraph{Evaluation Settings.}
The final benchmark evaluator runs separately from the development agents, including the intermediate playtesting performed by +Ours. We use the same evaluation protocol for all configurations. Table~\ref{tab:app_eval_settings} records the operational settings; the evidence collection and rubric scoring procedures are detailed in Appendix~\ref{app:agentic_eval} and Appendix~\ref{app:eval_details}.

\begin{table}[!htbp]
\centering
\caption{Evaluation and statistical settings.}
\label{tab:app_eval_settings}
\small
\begin{tabular}{@{}p{0.29\linewidth}p{0.65\linewidth}@{}}
\toprule
\textbf{Setting} & \textbf{Value} \\
\midrule
Executability gate & Build, install, and load in a clean instance; submission failure gives zero quality scores \\
Executability gate probes & Three probes per game: a common stability probe and two checks of original game content; see Table~\ref{tab:capability_pools} \\
Host Fidelity & Five-feature host checklist per game, scored 0, 0.5, or 1 by a fixed rule; insufficient evidence excluded \\
Construction Overall & Weighted average of Content, Functions \& Playability, Visual \& Presentation, and Host Fidelity; see Eq.~\ref{eq:overall_task} \\
Editing Overall & Edit Success penalized for failures in Preservation and Host Fidelity; see Eq.~\ref{eq:overall_edit} \\
Evidence & Runtime probes, submitted action replay, independent agent playtesting, and screenshots \\
Independent playtesting & 40 actions and 30 min per submission \\
Judge model & Claude Sonnet 5.5 for replay and playtest verification, adjudication, visual evaluation, and Host Fidelity \\
Rubric scores & 0, 0.5, or 1 \\
Unresolved items & An item testable by only one verifier takes that verifier's score; an item neither verifier could test scores 0 and stays in the denominator \\
Repeated evaluation & One evaluation per submission for the main results; three repeats on 47 submissions for the stability analysis \\
Confidence intervals & 95\% percentile intervals from 4{,}000 task bootstrap resamples paired across configurations \\
\bottomrule
\end{tabular}
\end{table}

\paragraph{Cost Settings.}
\label{app:eval_cost}
Development cost is computed from recorded model usage under a common pricing convention and is averaged over all attempted tasks, including unsuccessful submissions. For +Ours, it includes model calls from all development rounds and intermediate playtesting phases. Construction and Editing costs are reported separately. Calls made by the independent benchmark evaluator are excluded from development cost and reported separately.
A Full evaluation costs about \$0.56 per submission at API prices, with a median of \$0.52, and takes about 13 minutes of wall-clock time. Lite evaluation costs about \$0.20 and takes 6 minutes.
Prices follow the standard non-batch OpenRouter list prices of 2026-10-30, with cache reads priced separately at the cached-input rate. Claude Code reports its own cost; for Codex CLI the reported token counts are priced with this list; for OpenCode we use the cost reported by the provider when available and the list price otherwise.

\FloatBarrier

\section{Per-Game Results}
\label{app:per_game_results}

Table~\ref{tab:per_game} breaks the Construction Overall score of Table~\ref{tab:main_results} down by target game, for the open-source and the commercial games.

\begin{table*}[!htbp]
\centering
\caption{Construction Overall percentages by target game and agent configuration. The open-source track contains 77 tasks across nine games, and the commercial track contains 48 tasks across five games. Avg. is the mean over all tasks in the track. Task counts appear below the game names.}
\label{tab:per_game}
\begingroup
\setlength{\tabcolsep}{3pt}
\renewcommand{\arraystretch}{1.2}
\resizebox{\textwidth}{!}{%
\begin{tabular}{@{}ll*{10}{c}|*{6}{c}@{}}
\toprule
& & \multicolumn{10}{c|}{\textbf{Open-source games}} & \multicolumn{6}{c}{\textbf{Commercial games}} \\
\cmidrule(lr){3-12}\cmidrule(lr){13-18}
\textbf{Harness} & \textbf{Model} & \begin{tabular}[c]{@{}c@{}}VoxeLibre\\$n=18$\end{tabular} & \begin{tabular}[c]{@{}c@{}}Tuxemon\\$n=13$\end{tabular} & \begin{tabular}[c]{@{}c@{}}Godot\\TPS\\$n=11$\end{tabular} & \begin{tabular}[c]{@{}c@{}}Super\\TuxKart\\$n=10$\end{tabular} & \begin{tabular}[c]{@{}c@{}}Xonotic\\$n=7$\end{tabular} & \begin{tabular}[c]{@{}c@{}}Celeste\\64\\$n=6$\end{tabular} & \begin{tabular}[c]{@{}c@{}}Flare\\$n=5$\end{tabular} & \begin{tabular}[c]{@{}c@{}}Tanks of\\Freedom II\\$n=4$\end{tabular} & \begin{tabular}[c]{@{}c@{}}Colobot\\$n=3$\end{tabular} & \textbf{Avg.} & \begin{tabular}[c]{@{}c@{}}Minecraft\\$n=13$\end{tabular} & \begin{tabular}[c]{@{}c@{}}Stardew\\Valley\\$n=13$\end{tabular} & \begin{tabular}[c]{@{}c@{}}Valheim\\$n=8$\end{tabular} & \begin{tabular}[c]{@{}c@{}}Terraria\\$n=7$\end{tabular} & \begin{tabular}[c]{@{}c@{}}Cities:\\Skylines\\$n=7$\end{tabular} & \textbf{Avg.} \\
\midrule
\multirow{4}{*}{Claude Code}
& Claude Fable 5.2
& 88.2 & 90.9 & 87.5 & 80.4 & 85.1 & 88.0 & 79.4 & 93.4 & 91.7 & 87.1 & 85.8 & 82.7 & 62.8 & 89.0 & 70.8 & 79.3 \\

& Claude Sonnet 5.5
& 70.3 & 69.3 & 53.6 & 41.7 & 57.8 & 68.8 & 50.7 & 59.9 & 66.5 & 60.8 & 57.7 & 72.3 & 36.6 & 73.0 & 31.5 & 56.5 \\

& \quad\emph{+Ours R1}
& 69.5 & 71.6 & 57.4 & 46.9 & 54.1 & 73.1 & 53.0 & 71.3 & 66.2 & 63.0 & 59.1 & 78.7 & 54.2 & 71.7 & 51.6 & 64.9 \\

& \quad\emph{+Ours R3}
& 76.9 & 70.2 & 72.9 & 50.5 & 64.7 & 73.2 & 58.4 & 69.4 & 83.6 & 69.0 & 71.1 & 82.4 & 64.1 & 76.7 & 49.8 & 70.7 \\

\midrule

\multirow{2}{*}{Codex CLI}
& GPT-5.6 Sol
& 44.2 & 43.5 & 9.7 & 35.9 & 39.1 & 46.7 & 45.6 & 44.4 & 34.3 & 37.5 & 40.8 & 54.9 & 35.7 & 62.0 & 36.1 & 46.2 \\

& \quad\emph{+Ours R1}
& 47.5 & 43.9 & 12.2 & 37.9 & 45.4 & 50.7 & 42.3 & 50.6 & 48.6 & 40.5 & 49.6 & 69.7 & 44.0 & 65.8 & 53.9 & 57.1 \\

\midrule

\multirow{14}{*}{OpenCode}
& Gemini 3.8 Flash
& 32.6 & 18.7 & 25.5 & 31.6 & 17.0 & 28.2 & 16.6 & 37.5 & 36.7 & 26.7 & 32.9 & 28.7 & 17.2 & 36.9 & 28.4 & 29.1 \\

& \quad\emph{+Ours R1}
& 42.0 & 35.7 & 39.2 & 37.1 & 26.5 & 50.3 & 27.8 & 49.4 & 41.9 & 38.6 & 38.5 & 32.4 & 24.9 & 53.2 & 35.7 & 36.3 \\

& GLM-5.3-Flash
& 42.4 & 27.2 & 25.4 & 32.1 & 17.0 & 27.0 & 29.3 & 55.1 & 32.8 & 32.0 & 33.8 & 24.5 & 2.8 & 39.3 & 14.6 & 24.1 \\

& \quad\emph{+Ours R1}
& 43.4 & 30.0 & 29.6 & 30.9 & 8.6 & 22.6 & 39.5 & 48.1 & 31.5 & 32.3 & 32.7 & 25.2 & 6.6 & 43.3 & 21.7 & 26.2 \\

& Qwen3.8-Max
& 45.5 & 28.2 & 27.4 & 24.9 & 12.4 & 48.9 & 28.9 & 56.7 & 25.0 & 33.3 & 40.1 & 31.1 & 13.4 & 45.7 & 16.5 & 30.6 \\

& \quad\emph{+Ours R1}
& 49.9 & 34.5 & 24.3 & 35.4 & 16.0 & 61.8 & 32.2 & 56.2 & 20.3 & 37.6 & 39.9 & 40.6 & 16.4 & 51.9 & 24.5 & 35.7 \\

& Kimi K3
& 49.7 & 29.3 & 26.6 & 23.3 & 22.8 & 58.7 & 17.1 & 47.6 & 22.0 & 34.5 & 56.6 & 53.8 & 8.9 & 41.5 & 1.4 & 37.6 \\

& \quad\emph{+Ours R1}
& 47.6 & 30.0 & 29.2 & 28.1 & 32.6 & 49.1 & 49.9 & 50.9 & 21.7 & 37.5 & 56.9 & 48.0 & 25.2 & 53.2 & 17.1 & 42.9 \\

& Qwen3.8-Flash
& 39.5 & 24.5 & 44.8 & 32.4 & 18.7 & 56.8 & 38.1 & 51.2 & 39.3 & 36.8 & 44.9 & 36.9 & 20.5 & 46.6 & 2.6 & 32.8 \\

& \quad\emph{+Ours R1}
& 39.6 & 22.8 & 49.8 & 28.1 & 17.9 & 43.1 & 51.5 & 50.6 & 46.9 & 36.7 & 41.2 & 33.4 & 32.2 & 48.6 & 12.6 & 35.0 \\

& \quad\emph{+Ours R3}
& 60.0 & 50.4 & 73.0 & 36.9 & 33.9 & 52.9 & 41.7 & 47.4 & 56.0 & 52.3 & 53.7 & 54.6 & 18.7 & 66.6 & 20.3 & 46.2 \\

& DeepSeek V4.1 Flash
& 65.0 & 70.3 & 37.3 & 38.9 & 45.1 & 67.6 & 45.8 & 64.6 & 39.6 & 54.7 & 56.3 & 65.9 & 21.1 & 50.2 & 15.8 & 46.2 \\

& \quad\emph{+Ours R1}
& 61.5 & 70.3 & 37.9 & 37.4 & 43.8 & 65.5 & 46.8 & 58.9 & 43.3 & 53.4 & 56.4 & 66.3 & 32.4 & 64.4 & 16.0 & 51.1 \\

& \quad\emph{+Ours R3}
& 64.4 & 69.0 & 63.6 & 35.3 & 56.7 & 70.1 & 55.3 & 59.3 & 58.1 & 59.9 & 60.1 & 75.2 & 27.8 & 61.6 & 50.8 & 57.7 \\

\bottomrule
\end{tabular}%
}
\endgroup
\end{table*}


\FloatBarrier

\section{Additional Analysis Results}
\label{app:analysis_tables}

Table~\ref{tab:refinement_results} reports the iterative refinement results of Section~\ref{sec:experiments} round by round, including R2. Tables~\ref{tab:app_capability}, \ref{tab:app_editclass}, and~\ref{tab:app_crossgame} give the numbers behind Section~\ref{sec:analysis}. All values are means of the per-task scores used for Table~\ref{tab:main_results}.

\begin{table*}[!htbp]
\centering
\caption{Construction Overall percentages of native agents across input modalities and task categories. The column headers give the number of tasks.}
\label{tab:app_capability}
\footnotesize
\setlength{\tabcolsep}{5pt}
\begin{tabular}{@{}lc|cccc|ccc@{}}
\toprule
& & \multicolumn{4}{c|}{\textbf{Input modality}} & \multicolumn{3}{c}{\textbf{Task category}} \\
\textbf{Model} & \textbf{Overall} & \begin{tabular}[c]{@{}c@{}}Video\\$n=40$\end{tabular} & \begin{tabular}[c]{@{}c@{}}Image\\$n=36$\end{tabular} & \begin{tabular}[c]{@{}c@{}}Text\\$n=30$\end{tabular} & \begin{tabular}[c]{@{}c@{}}Hybrid\\$n=19$\end{tabular} & \begin{tabular}[c]{@{}c@{}}Scene \&\\World\\$n=46$\end{tabular} & \begin{tabular}[c]{@{}c@{}}Mechanics \&\\Behavior\\$n=46$\end{tabular} & \begin{tabular}[c]{@{}c@{}}Narrative \&\\Event\\$n=33$\end{tabular} \\
\midrule
Claude Fable 5.2 & 84.1 & 83.2 & 89.0 & 80.6 & 82.6 & 82.7 & 85.4 & 84.3 \\
Claude Sonnet 5.5 & 59.2 & 52.9 & 66.5 & 62.5 & 53.2 & 53.1 & 61.6 & 64.1 \\
DeepSeek V4.1 Flash & 51.4 & 45.6 & 60.1 & 53.0 & 44.9 & 45.9 & 51.5 & 59.1 \\
GPT-5.6 Sol & 40.9 & 37.0 & 48.9 & 37.6 & 38.7 & 39.2 & 39.0 & 45.8 \\
Kimi K3 & 35.7 & 30.9 & 38.4 & 34.7 & 42.1 & 35.4 & 38.2 & 32.7 \\
Qwen3.8-Flash & 35.2 & 35.1 & 40.1 & 33.1 & 29.6 & 29.8 & 44.4 & 30.0 \\
Qwen3.8-Max & 32.2 & 40.7 & 28.1 & 26.7 & 31.0 & 26.6 & 41.2 & 27.7 \\
GLM-5.3-Flash & 29.0 & 28.6 & 36.2 & 23.4 & 24.9 & 27.8 & 34.1 & 23.5 \\
Gemini 3.8 Flash & 27.6 & 30.2 & 33.1 & 21.6 & 21.3 & 24.2 & 35.2 & 21.8 \\
\midrule
Mean & 43.9 & 42.7 & 48.9 & 41.5 & 40.9 & 40.5 & 47.8 & 43.2 \\
\bottomrule
\end{tabular}

\end{table*}

\begin{table*}[!htbp]
\centering
\caption{
\textbf{Effect of iterative refinement on GameAnything-Bench, round by round.}
R0 is the native run and R1--R3 are successive +Ours rounds; R2 and R3 were run for the three backbones above the line.
$\Delta$: change in Overall from R0. HC: Headroom Closed, $\Delta/(100-\mathrm{R0})$. \$: cumulative development cost in USD per task.
}
\label{tab:refinement_results}

\begingroup
\footnotesize
\setlength{\tabcolsep}{5pt}
\renewcommand{\arraystretch}{1.0}

\begin{tabular}{@{}ll*{4}{c}*{4}{c}@{}}
\toprule
\multirow{2}{*}{\textbf{Model}} & \multirow{2}{*}{\textbf{Round}}
& \multicolumn{4}{c}{\textbf{Construction}}
& \multicolumn{4}{c}{\textbf{Editing}} \\
\cmidrule(lr){3-6}
\cmidrule(lr){7-10}
& & Overall & $\Delta$ & HC & \$ & Overall & $\Delta$ & HC & \$ \\
\midrule
\multirow{4}{*}{Claude Sonnet 5.5} & R0 & 59.2 & -- & -- & 1.84 & 72.8 & -- & -- & 0.53 \\
 & R1 & 63.7 & +4.5 & 11.1\% & 2.85 & 74.2 & +1.4 & 5.2\% & 1.09 \\
 & R2 & 68.0 & +8.9 & 21.7\% & 4.09 & 76.3 & +3.5 & 13.0\% & 1.66 \\
 & R3 & 69.7 & +10.5 & 25.7\% & 5.16 & 76.4 & +3.6 & 13.4\% & 2.15 \\
\arrayrulecolor{black!30}\midrule\arrayrulecolor{black}
\multirow{4}{*}{DeepSeek V4.1 Flash} & R0 & 51.4 & -- & -- & 0.26 & 69.9 & -- & -- & 0.11 \\
 & R1 & 52.5 & +1.1 & 2.2\% & 0.65 & 76.8 & +7.0 & 23.1\% & 0.53 \\
 & R2 & 55.3 & +3.8 & 7.9\% & 1.18 & 76.8 & +7.0 & 23.2\% & 0.95 \\
 & R3 & 59.1 & +7.6 & 15.7\% & 1.72 & 77.2 & +7.3 & 24.2\% & 1.46 \\
\arrayrulecolor{black!30}\midrule\arrayrulecolor{black}
\multirow{4}{*}{Qwen3.8-Flash} & R0 & 35.2 & -- & -- & 0.35 & 61.7 & -- & -- & 0.15 \\
 & R1 & 36.0 & +0.8 & 1.2\% & 0.76 & 71.9 & +10.2 & 26.6\% & 0.67 \\
 & R2 & 44.4 & +9.2 & 14.2\% & 1.30 & 73.2 & +11.5 & 29.9\% & 1.13 \\
 & R3 & 50.0 & +14.8 & 22.9\% & 1.90 & 78.1 & +16.4 & 42.9\% & 1.61 \\
\midrule
\multirow{2}{*}{GPT-5.6 Sol} & R0 & 40.9 & -- & -- & 2.55 & 72.5 & -- & -- & 2.13 \\
 & R1 & 46.9 & +6.0 & 10.2\% & 6.01 & 73.4 & +0.8 & 3.1\% & 5.87 \\
\arrayrulecolor{black!30}\midrule\arrayrulecolor{black}
\multirow{2}{*}{Kimi K3} & R0 & 35.7 & -- & -- & 5.48 & 67.7 & -- & -- & 2.62 \\
 & R1 & 39.6 & +3.9 & 6.1\% & 7.49 & 73.3 & +5.6 & 17.3\% & 4.52 \\
\arrayrulecolor{black!30}\midrule\arrayrulecolor{black}
\multirow{2}{*}{Qwen3.8-Max} & R0 & 32.2 & -- & -- & 3.33 & 68.5 & -- & -- & 1.65 \\
 & R1 & 36.9 & +4.6 & 6.9\% & 4.93 & 75.1 & +6.6 & 21.0\% & 3.25 \\
\arrayrulecolor{black!30}\midrule\arrayrulecolor{black}
\multirow{2}{*}{GLM-5.3-Flash} & R0 & 29.0 & -- & -- & 0.52 & 63.4 & -- & -- & 0.19 \\
 & R1 & 30.0 & +1.0 & 1.4\% & 0.99 & 72.7 & +9.3 & 25.4\% & 0.70 \\
\arrayrulecolor{black!30}\midrule\arrayrulecolor{black}
\multirow{2}{*}{Gemini 3.8 Flash} & R0 & 27.6 & -- & -- & 2.65 & 61.3 & -- & -- & 2.42 \\
 & R1 & 37.7 & +10.1 & 13.9\% & 5.25 & 66.9 & +5.6 & 14.4\% & 4.34 \\
\bottomrule
\end{tabular}

\endgroup
\end{table*}


\begin{table*}[!htbp]
\centering
\caption{Edit Success and Preservation scores of native agents across edit classes. The headers give task counts, and Overall follows Table~\ref{tab:main_results}.}
\label{tab:app_editclass}
\footnotesize
\begin{tabular}{@{}lc|cc|cc|cc@{}}
\toprule
& & \multicolumn{2}{c|}{\textbf{Appearance} $n=26$} & \multicolumn{2}{c|}{\textbf{Content \& Structure} $n=26$} & \multicolumn{2}{c}{\textbf{Behavior \& Flow} $n=28$} \\
\textbf{Model} & \textbf{Overall} & Succ. & Pres. & Succ. & Pres. & Succ. & Pres. \\
\midrule
Claude Fable 5.2 & 86.5 & 87.3 & 99.0 & 88.4 & 99.2 & 86.6 & 98.0 \\
Claude Sonnet 5.5 & 72.8 & 72.5 & 98.7 & 76.1 & 92.1 & 74.5 & 95.4 \\
GPT-5.6 Sol & 72.5 & 77.0 & 99.3 & 76.3 & 92.4 & 69.8 & 93.5 \\
DeepSeek V4.1 Flash & 69.9 & 73.5 & 98.6 & 77.3 & 94.7 & 64.1 & 93.7 \\
Qwen3.8-Max & 68.5 & 79.5 & 99.0 & 69.0 & 95.9 & 61.5 & 89.7 \\
Kimi K3 & 67.7 & 69.0 & 96.7 & 73.4 & 96.0 & 66.1 & 94.8 \\
GLM-5.3-Flash & 63.4 & 68.2 & 99.7 & 65.9 & 91.7 & 60.4 & 96.8 \\
Qwen3.8-Flash & 61.7 & 69.2 & 99.0 & 64.4 & 95.2 & 55.2 & 89.7 \\
Gemini 3.8 Flash & 61.3 & 63.7 & 98.4 & 66.3 & 93.6 & 58.2 & 94.8 \\
\midrule
Mean & 69.4 & 73.3 & 98.7 & 73.0 & 94.5 & 66.3 & 94.0 \\
\bottomrule
\end{tabular}

\end{table*}

\begin{table}[!htbp]
\centering
\caption{Construction Overall by game track and variation across games. SD is the standard deviation over 14 games; Open and Comm. give the means over 77 open-source tasks and 48 commercial tasks. The last column identifies each model's lowest-scoring game.}
\label{tab:app_crossgame}
\footnotesize
\begin{tabular}{@{}lccccl@{}}
\toprule
\textbf{Model} & \textbf{Overall} & \textbf{SD} & \textbf{Open} & \textbf{Comm.} & \textbf{Lowest game score} \\
\midrule
Claude Fable 5.2 & 84.1 & 8.5 & 87.1 & 79.3 & Valheim, 62.8 \\
Claude Sonnet 5.5 & 59.2 & 13.6 & 60.8 & 56.5 & Cities: Skylines, 31.5 \\
DeepSeek V4.1 Flash & 51.4 & 17.3 & 54.7 & 46.2 & Cities: Skylines, 15.8 \\
GPT-5.6 Sol & 40.9 & 11.8 & 37.5 & 46.2 & Godot TPS, 9.7 \\
Kimi K3 & 35.7 & 18.4 & 34.5 & 37.6 & Cities: Skylines, 1.4 \\
Qwen3.8-Flash & 35.2 & 14.6 & 36.8 & 32.8 & Cities: Skylines, 2.6 \\
Qwen3.8-Max & 32.2 & 13.7 & 33.3 & 30.6 & Xonotic, 12.4 \\
GLM-5.3-Flash & 29.0 & 12.7 & 32.0 & 24.1 & Valheim, 2.8 \\
Gemini 3.8 Flash & 27.6 & 7.7 & 26.7 & 29.1 & Flare, 16.7 \\
\bottomrule
\end{tabular}

\end{table}

\section{Additional Ablations and Robustness Analysis}
\label{app:ablations}

\subsection{Shared Task Subset and Protocol}
\label{app:ablation_protocol}
Both analyses use the same 40-task Construction subset, spanning open-source and commercial games, all four input modalities, and the three task categories. Each task contains rubric criteria that depend on information in the references but not the task instruction. We evaluate submissions with the Lite protocol and compute paired bootstrap confidence intervals over tasks.

\subsection{Reference Ablation}
\label{app:reference_ablation}
Multimodal references provide information that task instructions alone cannot recover. We test this by withholding the references while keeping the task instructions and evaluation unchanged. On the subset of 40 tasks, Claude Sonnet 5.5 loses 7.1 Overall points in a comparison using the same harness version. The decline is concentrated in Content and Visual \& Presentation, while Functions \& Playability shows no clear change. DeepSeek V4.1 Flash, compared against its main run, declines on all dimensions, including Functions \& Playability ($-16.7$). These results support the benchmark's focus on development grounded in external references rather than instruction following alone. Table~\ref{tab:app_reference_ablation} reports the full ablation results.

\begin{table*}[!htbp]
\centering
\caption{Reference ablation on 40 Construction tasks using Lite evaluation. Scores compare full references with instruction-only inputs. $\Delta$ denotes instruction only minus full reference, with paired bootstrap 95\% confidence intervals. Task counts are shown alongside each modality.}
\label{tab:app_reference_ablation}
\footnotesize
\begin{tabular}{@{}lcccc@{}}
\toprule
 & \textbf{Overall} & \textbf{Content} & \textbf{Visual} & \textbf{Functions} \\
\midrule
\multicolumn{5}{@{}l}{\textit{Claude Sonnet 5.5, vs.\ full-reference runs with the same harness version using Runs 2 and 3}} \\
Full reference & 60.5 & 61.3 & 51.5 & 55.4 \\
Instruction only & 53.4 & 51.9 & 36.6 & 56.2 \\
$\Delta$ and 95\% CI & $-$7.1, $-$12.4 to $-$1.9 & $-$9.4, $-$17.0 to $-$1.9 & $-$14.9, $-$21.9 to $-$8.1 & +0.8, $-$7.4 to +9.2 \\
\midrule
\multicolumn{5}{@{}l}{\textit{Claude Sonnet 5.5, vs.\ the main run using Run 1}} \\
$\Delta$ and 95\% CI & $-$11.6, $-$16.5 to $-$6.8 & $-$15.9, $-$23.4 to $-$8.9 & $-$11.9, $-$19.5 to $-$4.3 & $-$10.9, $-$18.2 to $-$3.5 \\
\midrule
\multicolumn{5}{@{}l}{\textit{DeepSeek V4.1 Flash, vs.\ the main run}} \\
Full reference & 59.0 & 59.7 & 47.1 & 58.2 \\
Instruction only & 42.0 & 37.9 & 29.0 & 41.5 \\
$\Delta$ and 95\% CI & $-$17.0, $-$23.4 to $-$10.8 & $-$21.7, $-$30.9 to $-$12.8 & $-$18.1, $-$26.4 to $-$10.0 & $-$16.7, $-$24.4 to $-$9.0 \\
\midrule
\multicolumn{5}{@{}l}{\textit{$\Delta$ Overall by input modality for Sonnet Run 1 and DeepSeek}} \\
\multicolumn{5}{@{}l}{Video, 12 tasks $-$10.1 / $-$6.8;\quad Image, 10 tasks $-$10.0 / $-$22.6;\quad Text, 8 tasks $-$20.3 / $-$25.3;\quad Hybrid, 10 tasks $-$8.0 / $-$16.8} \\
\bottomrule
\end{tabular}

\end{table*}

\subsection{Variance across Development Runs}
\label{app:dev_variance}
Individual development outcomes vary substantially across attempts, while aggregate model comparisons are more stable. We repeat Claude Sonnet 5.5 development on the same 40 tasks with independent seeds and the same 120-minute budget. The two additional runs share a harness version, whereas the original main run used an earlier version, so differences across all three runs cannot be attributed entirely to randomness. Across the two comparable runs, the mean standard deviation per task is 7.6 points. Nevertheless, Sonnet ranks second on the subset in all three runs. Its margin over DeepSeek V4.1 Flash changes considerably, indicating that small differences between agents should be interpreted cautiously. Table~\ref{tab:app_dev_variance} reports the run-level comparisons.

\begin{table}[!htbp]
\centering
\caption{Development variance of Claude Sonnet 5.5 on 40 Construction tasks evaluated with Lite. Run 1 is the original run, while Runs 2 and 3 are independent repeats. The dagger marks results based on 38 tasks with valid scores for all nine agents.}
\label{tab:app_dev_variance}
\footnotesize
\begin{tabular}{@{}lccc@{}}
\toprule
 & \textbf{Run 1} & \textbf{Run 2} & \textbf{Run 3} \\
\midrule
Subset Overall, 40 tasks & 65.0 & 58.9 & 62.1 \\
Rank among the nine agents$^{\dagger}$ & 2 & 2 & 2 \\
Lead over DeepSeek V4.1 Flash$^{\dagger}$ & +5.8 & +0.3 & +3.5 \\
\midrule
\multicolumn{4}{@{}l}{\textit{Across runs}} \\
SD of the subset Overall & \multicolumn{3}{c}{3.1} \\
Per-task SD, mean / median over all runs & \multicolumn{3}{c}{9.7 / 8.3} \\
Per-task SD, mean / median over comparable runs & \multicolumn{3}{c}{7.6 / 7.1} \\
Tasks with a range $\ge$ 20 points & \multicolumn{3}{c}{12 / 40} \\
Tasks where the order vs.\ DeepSeek reverses & \multicolumn{3}{c}{16 / 40} \\
\bottomrule
\end{tabular}

\end{table}

\section{Evaluation Reliability Analysis}
\label{app:reliability_eval}

The following analyses complement Section~\ref{sec:eval_reliability} by examining agreement with human judgments, the value of evaluation evidence, and stability across repeated runs and judge families. We call the replay verifier \emph{Lite} and the protocol combining replay, independent playtesting, and adjudication \emph{Full}. Human comparisons cover Content, Functions \& Playability, and Visual \& Presentation because annotators did not score Host Fidelity. Unless otherwise noted, scores use a 0 to 100 scale and confidence intervals are computed from 4{,}000 bootstrap resamples over submissions.

\subsection{Agreement with Human Judgments}
\label{app:human_agreement}
The Full evaluator closely matches human judgments and preserves comparisons among models. We evaluate 40 frozen submissions from six agent models, with two human annotators using the same rubrics and gameplay evidence. Full achieves a Pearson correlation of $r=0.87$ and a mean absolute error of 8.8 points, and it reproduces the human ranking of all six models exactly. At the rubric level, 71\% of items receive the same score, with few disagreements at opposite extremes. These results support aggregate agent comparisons while leaving room for error on individual submissions. Figure~\ref{fig:eval_human_agreement} and Table~\ref{tab:model_ranking} provide the detailed comparisons.

\begin{table}[!htbp]
\centering
\small
\setlength{\tabcolsep}{3.8pt}
\definecolor{rankdiff}{HTML}{E4DCF5}
\caption{\textbf{Agent-model ranking.} Mean overall score, 0--100, and rank of each agent model on the 40 human-labeled trials under human labels, the Full judge, and three judge families re-scoring the same saved evidence. Shaded ranks differ from the human ranking. Spearman correlation with the human ranking: Full judge 1.00, Claude 0.94, Codex 0.94, Gemini 0.94.}
\label{tab:model_ranking}
\begin{tabular}{@{}llcccccccccc@{}}
\toprule
 & & & & & & \multicolumn{6}{c}{Same-evidence re-judge} \\ \cmidrule(lr){7-12}
 & & \multicolumn{2}{c}{Human} & \multicolumn{2}{c}{Full judge} & \multicolumn{2}{c}{Claude} & \multicolumn{2}{c}{Codex} & \multicolumn{2}{c}{Gemini} \\
\cmidrule(lr){3-4}\cmidrule(lr){5-6}\cmidrule(lr){7-8}\cmidrule(lr){9-10}\cmidrule(lr){11-12}
Model & Agent & Score & Rank & Score & Rank & Score & Rank & Score & Rank & Score & Rank \\
\midrule
GPT-6-Astra & Codex & 90.0 & 1 & 88.1 & 1 & 89.2 & 1 & 92.0 & 1 & 89.2 & 1 \\
Claude Fable 5.2 & Claude Code & 77.6 & 2 & 76.3 & 2 & 75.1 & 2 & 81.9 & \cellcolor{rankdiff}3 & 75.6 & 2 \\
DeepSeek V4.1 Flash & OpenCode & 76.3 & 3 & 66.3 & 3 & 62.9 & \cellcolor{rankdiff}4 & 83.3 & \cellcolor{rankdiff}2 & 71.2 & 3 \\
Claude Sonnet 5.5 & Claude Code & 64.0 & 4 & 63.4 & 4 & 63.4 & \cellcolor{rankdiff}3 & 66.0 & 4 & 63.5 & 4 \\
GLM-5.3-Flash & OpenCode & 53.5 & 5 & 56.1 & 5 & 51.3 & 5 & 57.8 & 5 & 52.0 & \cellcolor{rankdiff}6 \\
GPT-5.6 Sol & Codex & 41.9 & 6 & 43.2 & 6 & 43.4 & 6 & 56.4 & 6 & 56.2 & \cellcolor{rankdiff}5 \\
\bottomrule
\end{tabular}
\end{table}

\subsection{Evidence Ablation}
\label{app:evidence_ablation}
Independent playtesting and adjudication primarily reduce conservative scoring rather than improve correlation with human judgments. We compare runtime states, replay screenshots, their combination in Lite, and the Full protocol on the same 40 human-labeled submissions. Screenshots provide substantially more useful evidence than runtime states alone. Lite and Full achieve the same correlation with human scores, but Full reduces the negative scoring bias from $-4.5$ to $-1.4$ and modestly lowers error. This improvement reflects the combined Full protocol rather than a separately isolated effect of playtesting. Table~\ref{tab:e1_evidence} reports the evidence ablations.

\begin{table}[!htbp]
\centering
\small
\caption{\textbf{Judge evidence ablation.} Agreement with human labels on the overall score, 0--100, over 40 trials. Bias is judge minus human with a 95\% bootstrap CI; negative values mean the judge is stricter. $\kappa$: item-level quadratic-weighted kappa. Best value per column in bold; for bias, closest to 0.}
\label{tab:e1_evidence}
\begin{tabular}{@{}lccccc@{}}
\toprule
Judge & $r\uparrow$ & MAE$\downarrow$ & $\kappa\uparrow$ & Bias & 95\% CI \\
\midrule
State only & 0.74 & 17.3 & 0.48 & $-$14.1 & $-$19.6 to $-$9.0 \\
Screenshots only & 0.85 & 10.7 & 0.62 & $-$7.3 & $-$11.7 to $-$3.5 \\
Lite & \textbf{\boldmath 0.87} & 9.8 & 0.67 & $-$4.5 & $-$8.4 to $-$0.8 \\
\midrule
Full & \textbf{\boldmath 0.87} & \textbf{\boldmath 8.8} & \textbf{\boldmath 0.68} & \textbf{\boldmath $-$1.4} & $-$5.1 to $+$2.1 \\
\bottomrule
\end{tabular}
\end{table}

\begin{figure}[t]
    \centering
    \includegraphics[width=\textwidth]{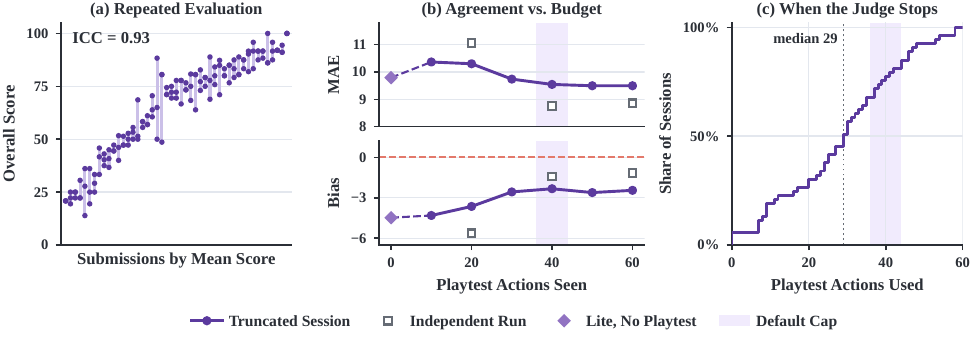}
    \caption{\textbf{Evaluation stability and playtest budget.} Panel a shows Overall scores from three independent evaluations of 47 submissions. Panel b shows scoring error and bias as playtest length increases; $K=0$ denotes Lite, and square markers denote independent runs capped at 20, 40, or 60 actions. Panel c shows the number of actions taken before the judge stopped in 53 sessions with a cap of 60.}
    \label{fig:eval_stability_budget}
\end{figure}

\subsection{Evaluation Stability and Playtest Budget}
\label{app:eval_stability}
Repeated evaluations remain consistent even when independent playtests collect different gameplay evidence. Scoring identical evidence is highly consistent, and three independent Full evaluations of 47 executable submissions achieve an intraclass correlation of 0.93. Variation within each submission is much smaller than differences across submissions, indicating that evaluation can distinguish quality differences despite variation in exploration.

Increasing the playtesting budget beyond 40 actions provides little additional benefit in the tested setting. We compare truncated playtests with independent runs capped at 20, 40, and 60 actions. Scoring error and bias improve until approximately 30 to 40 actions and then level off. A 20-action budget is often insufficient, while increasing the cap to 60 yields no clear improvement. We therefore use 40 actions as the default, as illustrated in Figure~\ref{fig:eval_stability_budget}.

\subsection{Judges from Other Model Families}
\label{app:judge_families}
Judge choice has limited impact on comparisons among models in the human annotated sample. To examine potential preference for a judge's own model family, we rescore the saved evidence of 40 submissions using Claude Sonnet 5.5, GPT-5.6 Sol, and Gemini 3.8 Flash. Each judge closely follows the human ranking, with Spearman $\rho=0.94$ and only one adjacent pair of models reversed. We find no statistically clear preference for a judge's own model family. Since these comparisons reuse saved gameplay evidence, they test scoring robustness rather than differences in independent exploration. Table~\ref{tab:model_ranking} reports the rankings.

\section{GameAnything: User Interface and Workflow}
\label{app:ui}

Section~\ref{sec:method} describes how GameAnything develops a game; this appendix expands Section~\ref{sec:platform} and describes how a user works with it.
GameAnything ships as a skill compatible with Claude Code, Codex, Gemini CLI, OpenCode, and other coding agents, together with a local web application.
The user starts a project with one sentence to the agent; from then on all interaction happens in the browser, and the user never edits code.

\begin{figure}[!htbp]
\centering
\includegraphics[width=\textwidth]{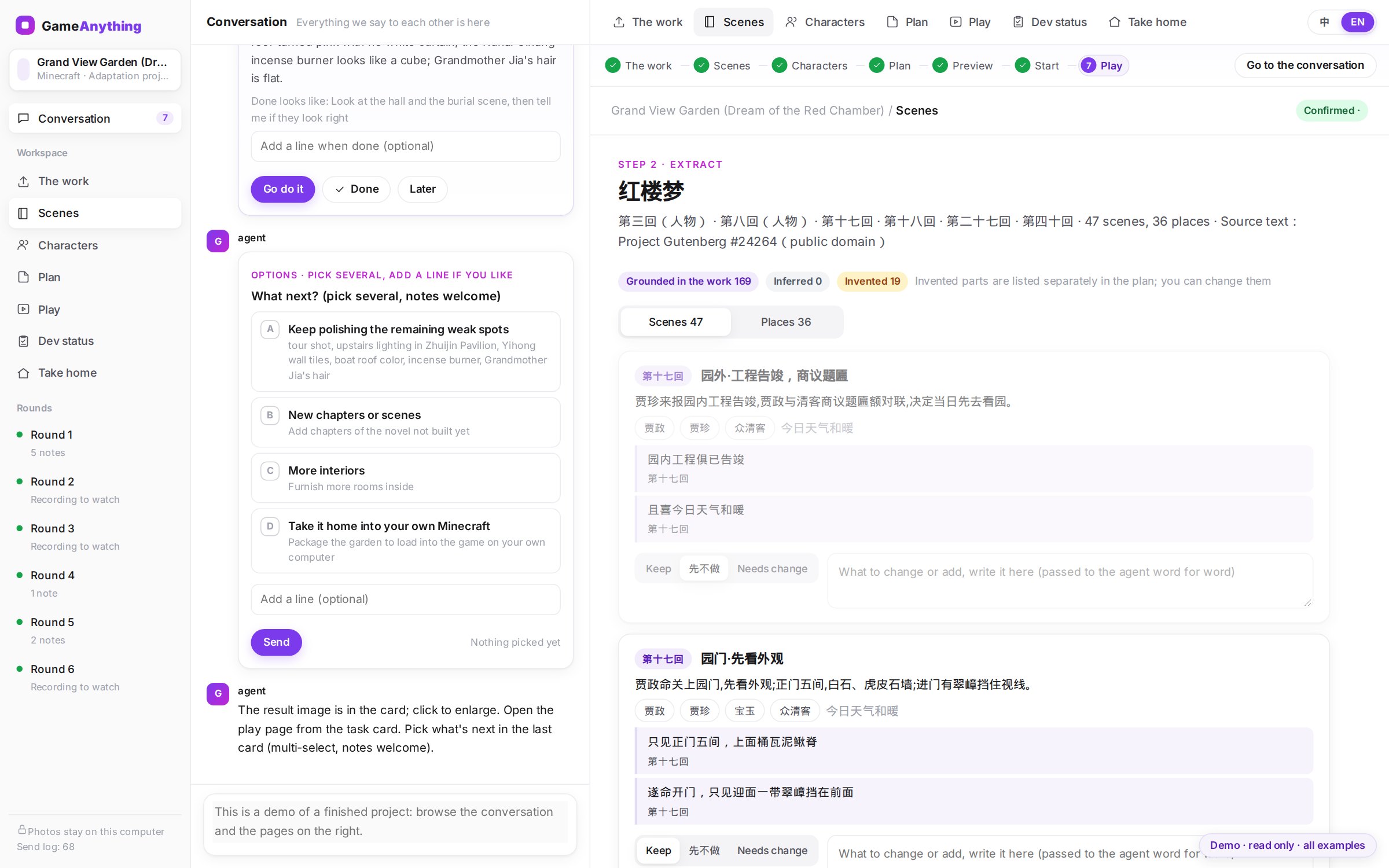}
\caption{The GameAnything web application for \textit{Dream of the Red Chamber} in Minecraft. The left panel lists projects and rounds, the center shows an agent conversation, and the right displays extracted scenes with source evidence, grounding status, and editable review controls. Project steps appear across the top.}
\label{fig:ui_shell}
\end{figure}

\paragraph{Layout.}
The web application runs on the user's own machine and is bound to the local host, as shown in Figure~\ref{fig:ui_shell}.
Each page has three parts: a project sidebar with the list of versions and rounds, a conversation with the agent, and a workspace with tabs for materials, scenes or personal memories, characters, planning, gameplay, development status, and taking home the result.
The agent and the interface share state only through files in the project folder: an append-only conversation log, review files written by the pages, and the plan and status files that the pages render.
Every page reads these files on each request, so a project can be closed, exported, or reopened at any point, and a finished project can be exported as a read-only demo that replays every page.

\begin{figure}[!htbp]
\centering
\includegraphics[width=\textwidth]{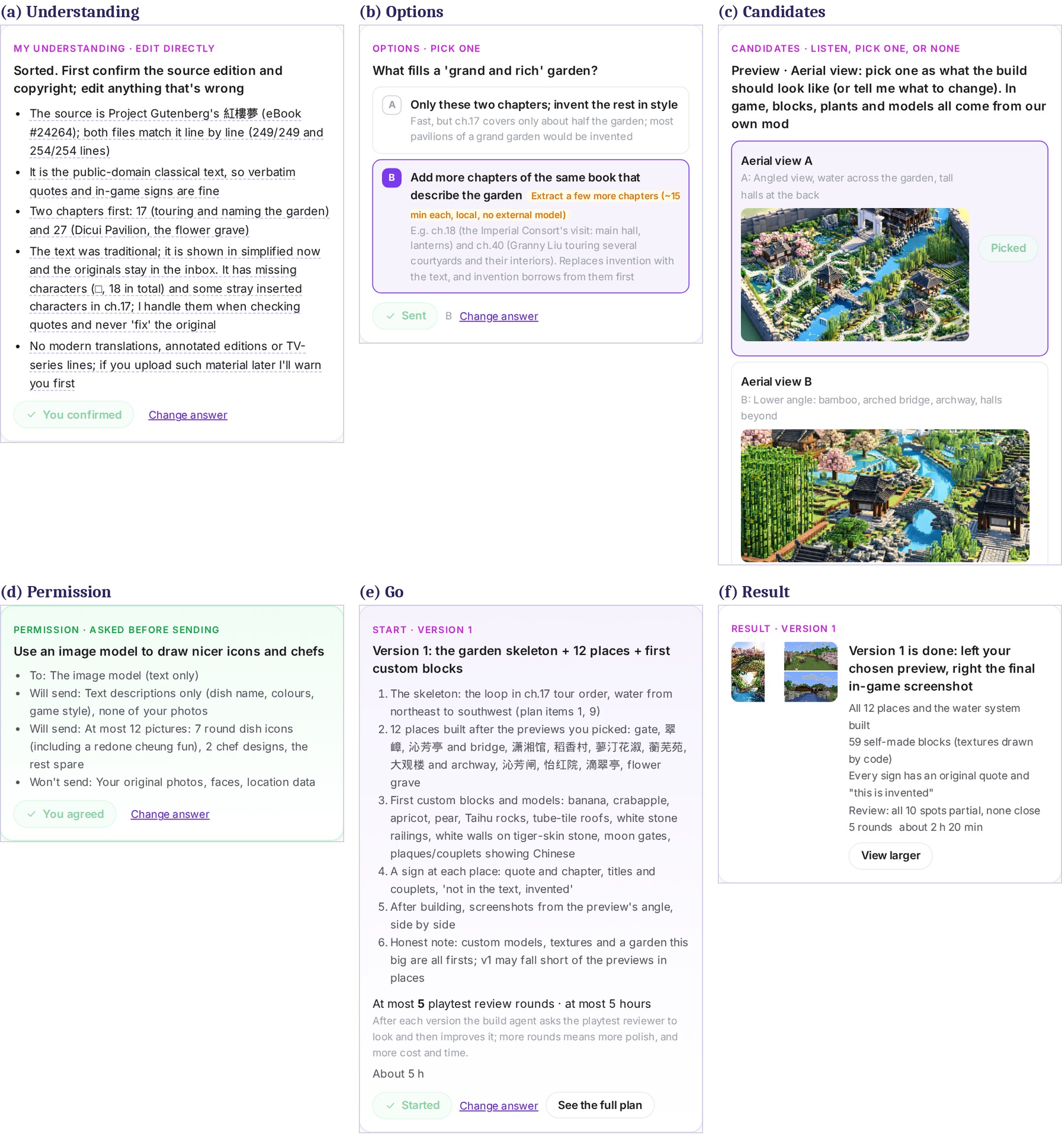}
\caption{Decision cards from \textit{Dream of the Red Chamber} and a Hong Kong food project. Panels a to c show source understanding, design choices, and visual previews. Panel d shows permission controls. Panels e and f show project approval and the final result alongside the selected preview.}
\label{fig:ui_cards}
\end{figure}

\paragraph{Cards instead of open questions.}
The agent asks for decisions through structured cards rather than open-ended questions, as illustrated in Figure~\ref{fig:ui_cards}.
An \textit{understanding} card lists what the agent took from the material and lets the user edit any line before confirming.
An \textit{options} card offers lettered choices with a recommended default and a short note on cost or consequence; a \textit{candidates} card offers generated preview pictures or sounds to pick from, or to reject with a reason; a \textit{questions} card gathers facts that only the user knows.
A \textit{permission} card states what will be sent to which model and what will not, and its estimated cost, before any of the user's material leaves the machine; each request is logged.
A \textit{go} card states the scope of a version, its budget of playtest rounds and hours, and its estimated cost, and nothing is built until the user starts it.
A \textit{result} card reports a delivered version with the user's chosen preview next to the final in-game screenshot.
Only the user's click marks a step as confirmed; the agent cannot confirm on the user's behalf, and every answer can be changed later.

\begin{figure}[!htbp]
\centering
\includegraphics[width=\textwidth]{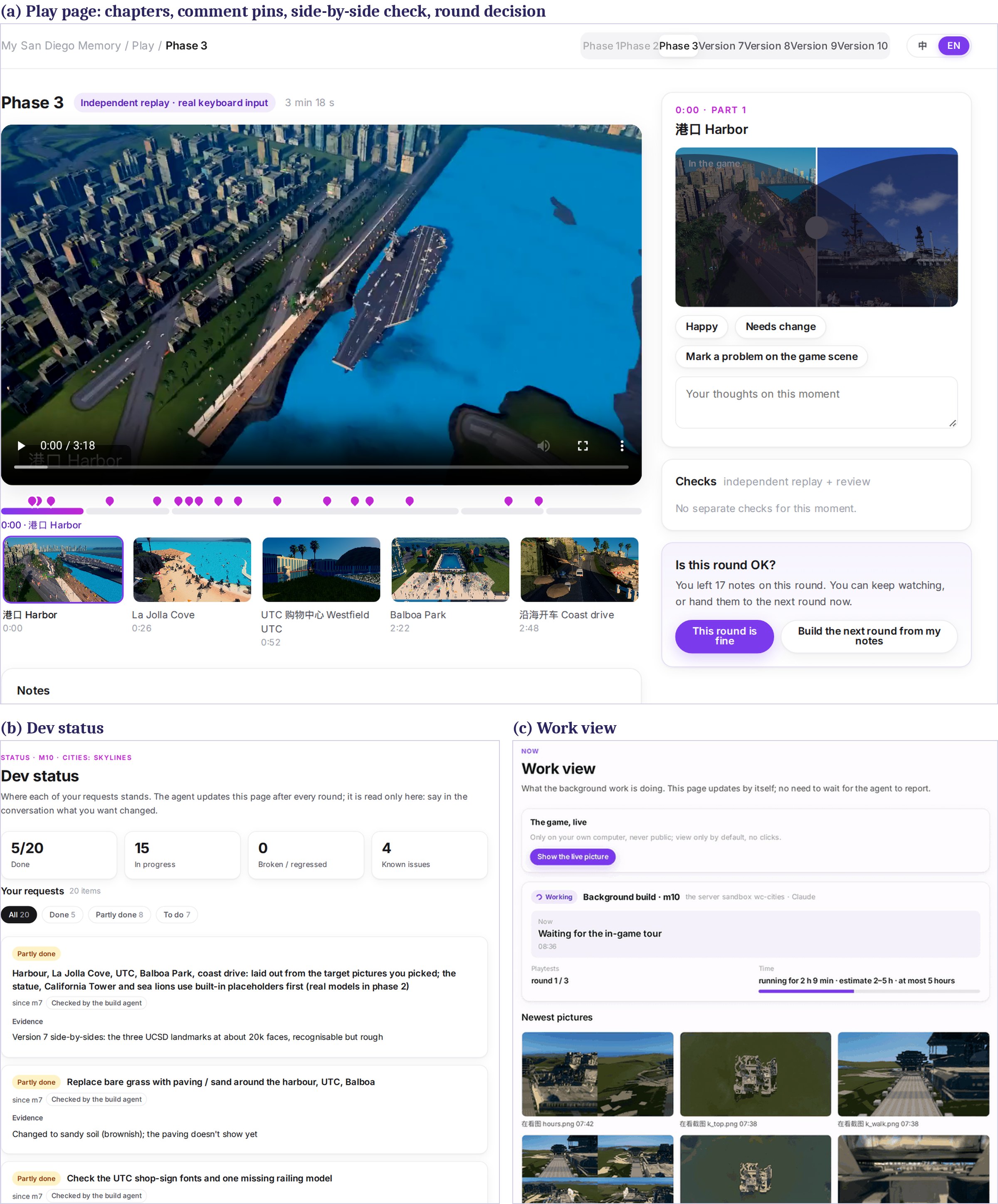}
\caption{Workspace pages from the San Diego project in Cities: Skylines. Panel a shows a playtest video with chapters, time-anchored feedback, reference comparisons, and review controls. Panel b displays the development state and evidence. Panel c shows live development progress.}
\label{fig:ui_pages}
\end{figure}

\paragraph{Four modes.}
Projects follow four modes, each with a defined sequence of steps listed in Table~\ref{tab:ui_modes}.
Steps that need the user's decision are never skipped, and each such step sends its card in the same turn.
The agent prepares extraction, designs, and plans; the user confirms them, approves every permission and paid generation, starts each version, and decides what happens after it.
In the adaptation mode, a copyright step records whether the source is in the public domain: only public-domain text is quoted verbatim in the game, and copyrighted films and modern translations are used as references only.

\begin{table}[!htbp]
\centering
\footnotesize
\setlength{\tabcolsep}{4pt}
\renewcommand{\arraystretch}{1.2}
\caption{The four GameAnything modes. Steps in \textbf{bold} are decided by the user; the others are carried out by the agent, which presents its result in a card or workspace page for the user to confirm.}
\label{tab:ui_modes}
\begin{tabular}{@{}l>{\raggedright\arraybackslash}p{3.2cm}>{\raggedright\arraybackslash}p{9.6cm}@{}}
\toprule
\textbf{Mode} & \textbf{Input $\rightarrow$ output} & \textbf{Steps} \\
\midrule
Memory & Photos and the user's words $\rightarrow$ a world of that day & \textbf{gather} $\cdot$ ingest $\cdot$ perceive $\cdot$ people $\cdot$ board $\cdot$ \textbf{confirm the day} $\cdot$ design $\cdot$ \textbf{confirm the plan} $\cdot$ preview $\cdot$ \textbf{go} $\cdot$ build $\cdot$ \textbf{feedback} $\cdot$ next round $\cdot$ \textbf{keep} \\
Adaptation & A novel or film $\rightarrow$ its places and scenes, playable & \textbf{gather} $\cdot$ ingest $\cdot$ copyright $\cdot$ choose units $\cdot$ extract $\cdot$ registry $\cdot$ \textbf{confirm scenes} $\cdot$ characters $\cdot$ \textbf{confirm characters} $\cdot$ questions $\cdot$ plan $\cdot$ preview $\cdot$ \textbf{go} $\cdot$ build $\cdot$ \textbf{feedback} $\cdot$ next round \\
Feature & One idea $\rightarrow$ one new thing in the game & \textbf{gather} $\cdot$ ingest $\cdot$ questions $\cdot$ spec $\cdot$ \textbf{confirm spec} $\cdot$ plan $\cdot$ \textbf{confirm plan} $\cdot$ \textbf{go} $\cdot$ build $\cdot$ \textbf{feedback} $\cdot$ next round $\cdot$ \textbf{keep} \\
Gameplay & An experience $\rightarrow$ the game's own mechanics & \textbf{gather} $\cdot$ ingest $\cdot$ questions $\cdot$ mechanic mapping $\cdot$ \textbf{confirm mapping} $\cdot$ scope $\cdot$ content $\cdot$ \textbf{confirm content} $\cdot$ \textbf{go} $\cdot$ build with balance runs $\cdot$ \textbf{feedback} $\cdot$ next round \\
\bottomrule
\end{tabular}
\end{table}

\paragraph{Reviewing a version.}
Each delivered version includes a recorded playtest divided into chapters for individual places or scenes, as shown in panel a of Figure~\ref{fig:ui_pages}.
For each chapter the play page shows the in-game frame and the approved target picture side by side, the reviewer's checks, which the user may override, and buttons to accept the moment or request changes.
Feedback is anchored to evidence: a comment is saved with its time and the current video frame, a box can be drawn on the frame to mark a spot, and an image can be attached; a moment can also be sent into the conversation with its frame.
Each piece of feedback becomes a record that keeps the user's own words, the agent's interpretation, and the anchor; when the agent is unsure what the user meant, it asks back with the frame before changing anything.
After a fix, the same anchor is shot again and shown next to the earlier frame, and the record is closed only when the user accepts it.
The user closes the review by marking the version fine or by handing the collected notes to the next version.

\paragraph{Progress and status.}
The development status page renders the persistent state described in Section~\ref{sec:iterative_development}, as shown in panel b of Figure~\ref{fig:ui_pages}. It displays each request's status, verification evidence, unresolved issues, and items to recheck. Statuses include done, partly done, to do, broken, and regressed.
It is read-only; changes are requested in the conversation.
During development, the work view in panel c of Figure~\ref{fig:ui_pages} shows the current activity, round, time budget, and latest screenshots so that users can monitor progress.
A live, view-only picture of the game is available after a separate permission.
If the build stalls, hits a usage limit, or needs a login, a single status line says so in the conversation together with the expected reset time.

\paragraph{Running and installing.}
A project builds either in a sandbox, a container on the user's machine or server with the game, its mod loader, and build tools, or natively on the user's own copy of the game.
In native mode the agent opens the game only after the user agrees or within a time window the user chooses, never takes the window focus, and closes only what it opened.
The take-home page installs a finished version into the user's own game in five steps; all saves are backed up first, only the game's mods folder changes, and every install can be undone.
Preferences learned from user feedback, as described in Section~\ref{sec:experience_memory}, appear on a separate page where each entry can be edited, disabled, or deleted. New entries are used only after the user enables them.

\section{GameAnything: Iterative Development Case Studies}
\label{app:iter_case_studies}
This section illustrates the GameAnything iteration process described in Section~\ref{sec:method} through three longer projects. Hogwarts from the \textit{Harry Potter} novels and the Grand View Garden from \textit{Dream of the Red Chamber} were built as Minecraft mods. Five places on the San Diego coast were built in Cities: Skylines using the user's photos and public images.
Each project went through several versions. Within each version, the builder and an independent critic alternate as described in Section~\ref{sec:iterative_development}. The critic tests the current build against approved targets and identifies concrete issues for the builder to fix. Between versions, the user provides sparse feedback through text and annotated video frames, sometimes extending the scope to new rooms, characters, or landmarks.
In total, Hogwarts took 4 versions with 21 critic rounds, the Grand View Garden 6 versions with 28 critic rounds, and San Diego 3 phases with 12 logged critic playtests.

Tables~\ref{tab:case_iter_hogwarts}--\ref{tab:case_iter_sandiego2} trace these projects from the source material to the final build.
Each column follows one place from the same viewpoint, and each row is one round, so reading a column from top to bottom shows how the place changes; the text above each row summarizes the critic's findings, the builder's changes, and the user's feedback at that point, and frames with an amber border are the user's own annotated feedback.
The critic's findings catch problems that a single development pass leaves behind, such as flat facades, empty rooms, floating blocks, or actors placed below the floor, while the user's feedback adds what the critic cannot judge, such as which places should feel crowded or which landmarks are missing.
Over the rounds, the number of places the critic judges close to their targets rises in both Minecraft projects, and in San Diego the critic's total score for the five places rises from 15 to 27 out of 50 in the first phase.

\providecolor{itCriticBg}{HTML}{ECE6F7}   
\providecolor{itBuilderBg}{HTML}{E4EAF1}  
\providecolor{itUserBg}{HTML}{FBEFD7}     
\providecolor{itUserLine}{HTML}{D9A441}
\providecolor{itSrcBg}{HTML}{F6F2E8}
\ifdefined\itw\else\newlength{\itw}\fi
\setlength{\itw}{0.1615\textwidth}
\providecommand{\itrole}[2]{{\setlength{\fboxsep}{1pt}\colorbox{#1}{\rule[-0.25ex]{0pt}{1.6ex}\textbf{#2}}}}
\providecommand{\itcritic}{\itrole{itCriticBg}{Critic}}
\providecommand{\itbuilder}{\itrole{itBuilderBg}{Builder}}
\providecommand{\ituser}{\itrole{itUserBg}{User}}
\providecommand{\itimg}[1]{\includegraphics[width=\itw]{#1}}
\providecommand{\itnone}{\makebox[\itw]{}}
\providecommand{\ituimg}[1]{{\setlength{\fboxsep}{0pt}\setlength{\fboxrule}{1.2pt}\fcolorbox{itUserLine}{white}{\includegraphics[width=\dimexpr\itw-2.4pt\relax]{#1}}}}
\providecommand{\itcol}[1]{\makebox[\itw][c]{\scriptsize\textbf{#1}}}
\providecommand{\itlab}[1]{\parbox[t]{\itw}{\centering\scriptsize #1}}
\providecommand{\itsrc}[1]{{\setlength{\fboxsep}{2pt}\colorbox{itSrcBg}{\parbox[t][0.95cm][c]{\dimexpr\itw-4pt\relax}{\raggedright\scriptsize\itshape #1}}}}
\providecommand{\itsix}[6]{\noindent#1\hfill#2\hfill#3\hfill#4\hfill#5\hfill#6}
\providecommand{\itround}[1]{\textbf{#1}\quad}
\ifdefined\itwt\else\newlength{\itwt}\fi
\setlength{\itwt}{0.328\textwidth}
\providecommand{\itimgt}[1]{\includegraphics[width=\itwt]{#1}}
\providecommand{\ituimgt}[1]{{\setlength{\fboxsep}{0pt}\setlength{\fboxrule}{1.4pt}\fcolorbox{itUserLine}{white}{\includegraphics[width=\dimexpr\itwt-2.8pt\relax]{#1}}}}
\providecommand{\itcolt}[1]{\makebox[\itwt][c]{\small\textbf{#1}}}
\providecommand{\itlabt}[1]{\parbox[t]{\itwt}{\centering\scriptsize #1}}
\providecommand{\itsrct}[1]{{\setlength{\fboxsep}{2pt}\colorbox{itSrcBg}{\parbox[t][0.75cm][c]{\dimexpr\itwt-4pt\relax}{\centering\footnotesize\itshape #1}}}}
\providecommand{\itthree}[3]{\noindent#1\hfill#2\hfill#3}
\providecommand{\ituimgs}[1]{{\setlength{\fboxsep}{0pt}\setlength{\fboxrule}{1.2pt}\fcolorbox{itUserLine}{white}{\includegraphics[width=0.22\textwidth]{#1}}}}
\providecommand{\itlabs}[1]{\parbox[t]{0.224\textwidth}{\centering\scriptsize #1}}

\begin{table*}[!htbp]
\centering
\footnotesize
\caption{Iterative development of Hogwarts in Minecraft across four versions and 21 critic rounds. Each column tracks the same location from the first round to the final version. The top row shows the agent's interpretation of the source novel. User feedback is provided between versions, and quotations are translated.}
\label{tab:case_iter_hogwarts}
\begin{tabular}{@{}p{\textwidth}@{}}
\toprule
\itthree{\itcolt{Great Hall}}{\itcolt{Portrait door}}{\itcolt{Common room}}\\[2pt]
\itthree{\itsrct{Ch.~7: four long tables, candles floating in mid-air, a ceiling like the night sky}}{\itsrct{Ch.~7: the portrait of the Fat Lady swings open as the entrance}}{\itsrct{Ch.~7: a round, warm red room, a fireplace, soft armchairs}}\\
\midrule
\itround{m1 r1}\itcritic{} Great Hall ``a dark blue-tinted tunnel''; candles ``too large and too uniform \ldots the room is monochrome blue''; the portrait door shows no portrait; the common room is ``dark blue stone'' with a row of sofas.\\[2pt]
\itthree{\itimgt{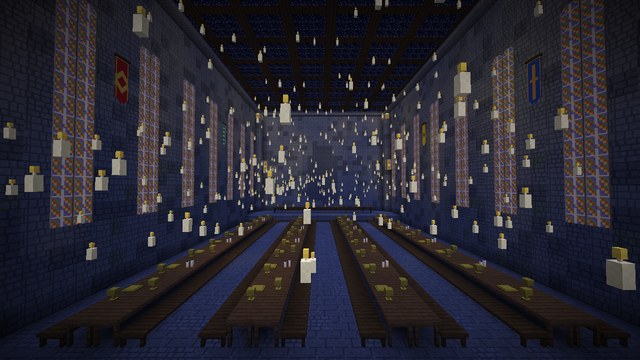}}{\itimgt{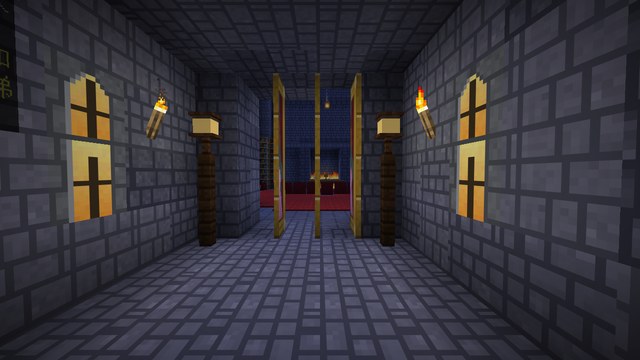}}{\itimgt{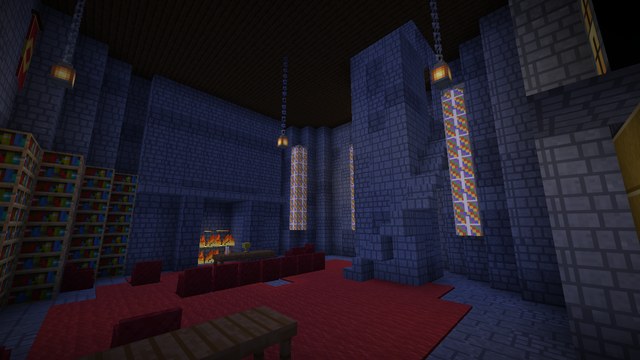}}\\
\midrule
\itround{m1 r5}\itbuilder{} over rounds r2 to r5, smaller candles above the tables, tall house banners, warm torch and lantern light; a gold-framed portrait that opens. \itcritic{} portrait corridor ``close'' to the target; the common room is still ``a wide rectangular room with a flat red carpet''. \ituser{} ``Overall good, but the castle is too empty inside \ldots every room should look full of things, ornate and intricate.''\\[2pt]
\itthree{\itimgt{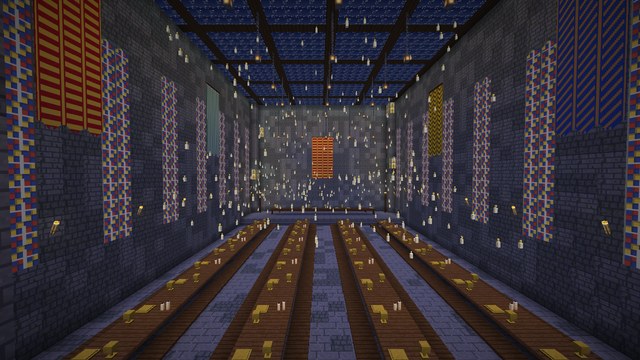}}{\itimgt{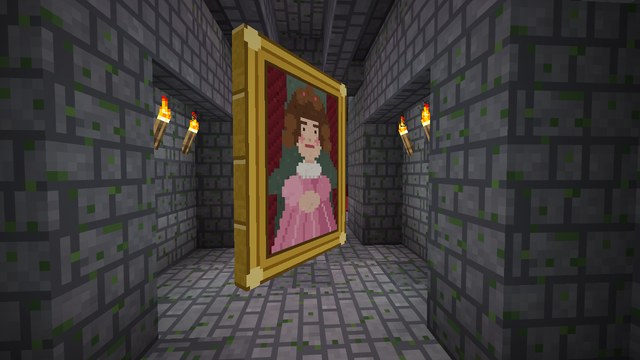}}{\itimgt{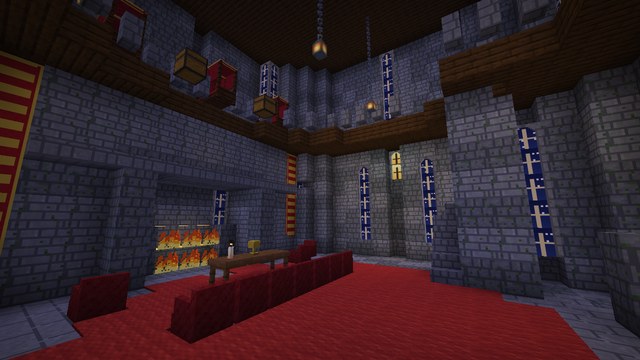}}\\
\midrule
\itround{m2 r5}\itbuilder{} during version m2, 76 new decor blocks in every room: paintings, tapestries, armour, rugs, chess and tea sets, chandeliers; the Great Hall rebuilt with carved piers, pointed stained glass and house carpets. \itcritic{} the interiors are ``genuinely packed with props, rugs, frames and warm light, which answers the user's main request''.\\[2pt]
\itthree{\itimgt{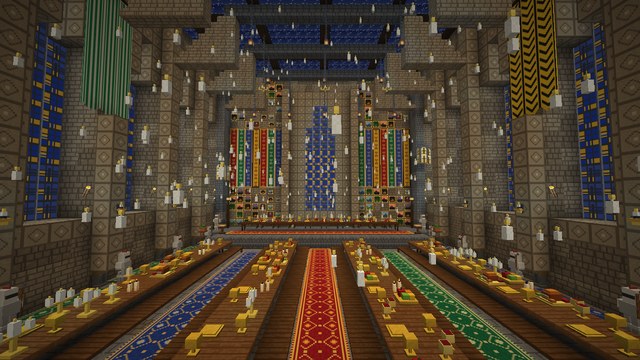}}{\itimgt{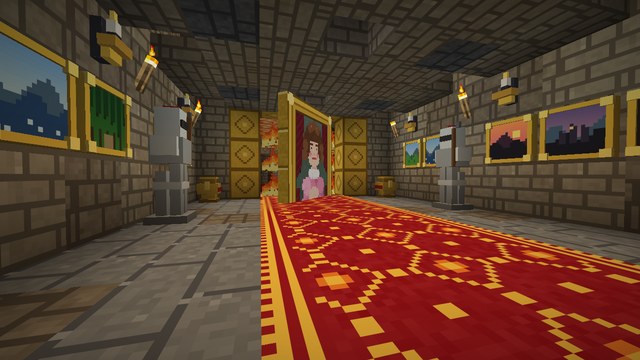}}{\itimgt{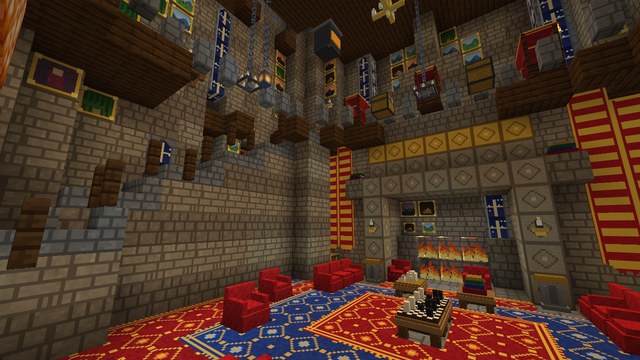}}\\
\midrule
\itround{m4 r5}\ituser{} ``Put the characters in too.'' \itbuilder{} during versions m3 and m4, nine characters and their small scenes: students at the house tables, the trio playing chess by the fire. \itcritic{} Great Hall ``house tables full of students \ldots Fuller than before'' and judged close to the target; portrait door judged close to the target; the common room is warm and full but ``still square''.\\[2pt]
\itthree{\itimgt{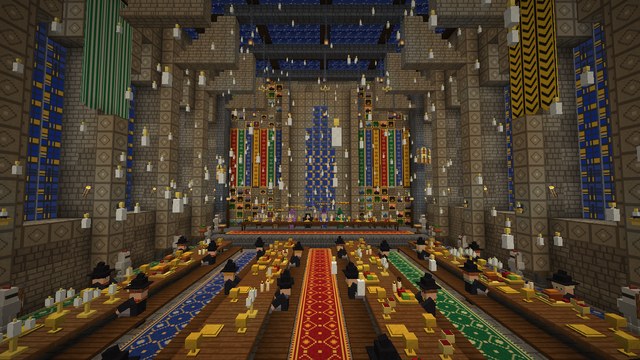}}{\itimgt{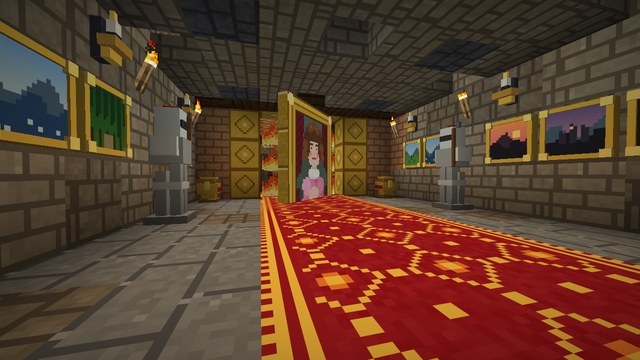}}{\itimgt{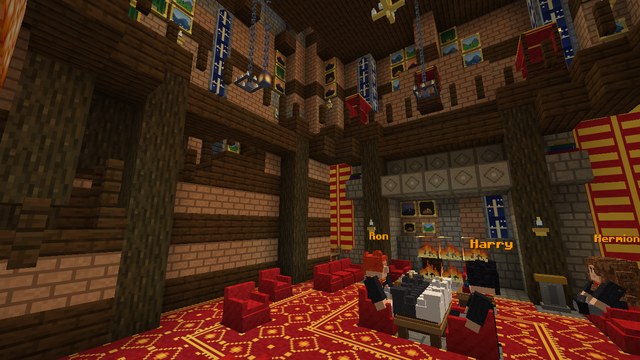}}\\
\bottomrule
\end{tabular}
\end{table*}


\begin{table*}[!htbp]
\centering
\footnotesize
\caption{Iterative development of the Grand View Garden from \textit{Dream of the Red Chamber} in Minecraft across six versions and 28 critic rounds. Columns track the main gate, Daguan archway, and Yihong moon gate. The final row uses the same viewpoints as the earlier critic shots. Amber borders mark user-annotated tour frames. The top row contains the public-domain source text. Quotations are translated.}
\label{tab:case_iter_garden}
\begin{tabular}{@{}p{\textwidth}@{}}
\toprule
\itthree{\itcolt{Main gate}}{\itcolt{Daguan archway}}{\itcolt{Yihong moon gate}}\\[2pt]
\itthree{\includegraphics[width=\itwt]{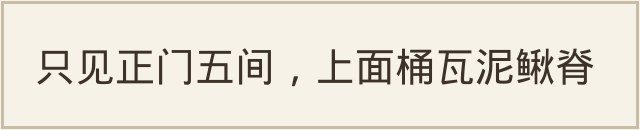}}{\includegraphics[width=\itwt]{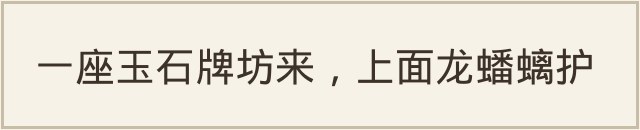}}{\includegraphics[width=\itwt]{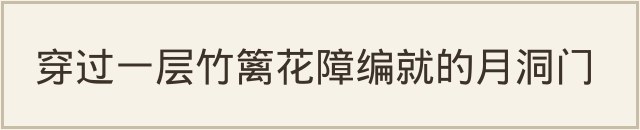}}\\[0pt]
\itthree{\itlabt{Ch.~17: a gate of five bays, tube-tile roof with a loach-back ridge}}{\itlabt{Ch.~17: a jade archway with coiling dragons}}{\itlabt{Ch.~17: a moon gate in a fence of bamboo and flowers}}\\
\midrule
\itround{m1 r1}\itcritic{} reports a gate roof resembling a flat pyramid with two chimney-like posts, a flat wall with a diamond-shaped rather than round opening at Yihong, and a plain grey archway.\\[2pt]
\itthree{\itimgt{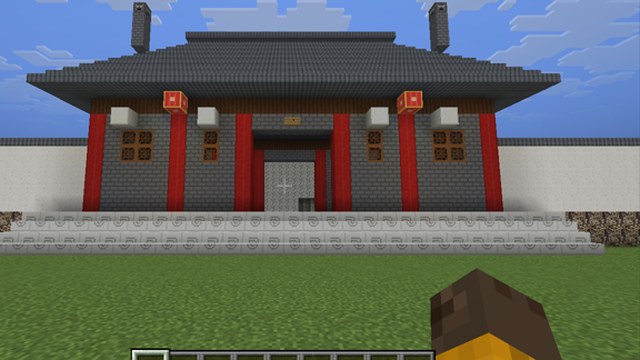}}{\itimgt{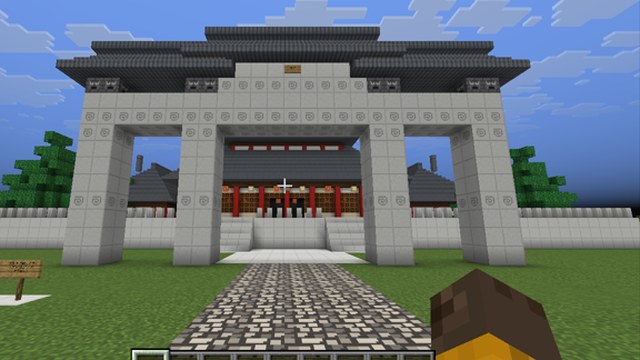}}{\itimgt{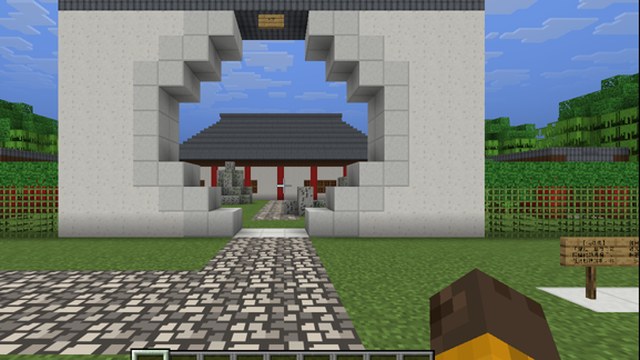}}\\
\midrule
\itround{m1 r2--m2 r5}\ituser{} on the m1 video: ``Roughly right, but still crude \ldots too sparse''; ``Shouldn't every courtyard have a plaque with its name?'' \itbuilder{} from m1 r2 through m2 r5, hip roofs with ridge ornaments; carved dragon pillars and gold roofs at the archway; a round moon gate in a hedge wall; then name plaques for 21 places, a mossy rockery behind the gate, a three-tier Daguan hall, a bamboo-lattice wall at Yihong. \itcritic{} finds that the gate matches the target composition and judges it close to the intended result.\\[2pt]
\itthree{\itimgt{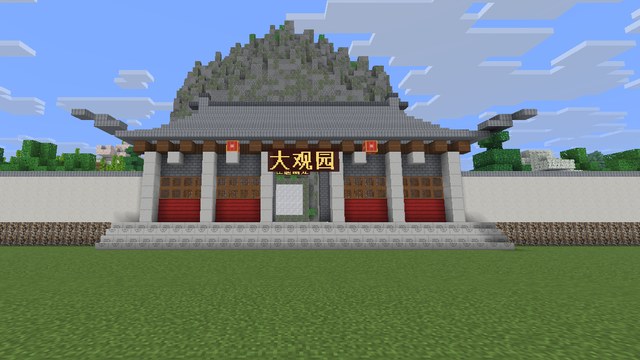}}{\itimgt{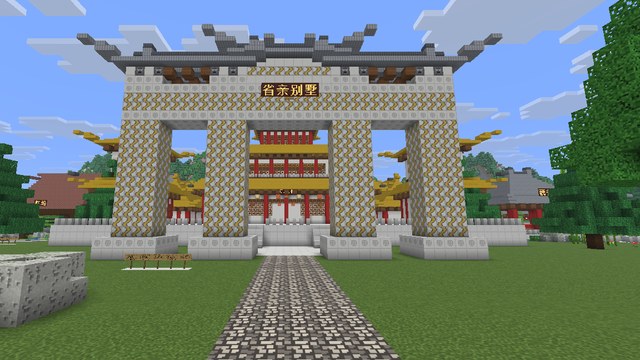}}{\itimgt{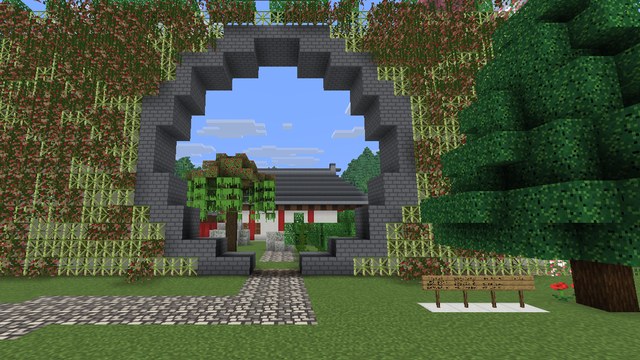}}\\
\midrule
\itround{m4--m5}\ituser{} annotates tour frames after rooms and story scenes are added during m3 to m5. ``Why are these people hanging in the air?''; ``Rough here, these big red and yellow colour blocks''; ``Why is a pink block hanging in the air? Daiyu holds the hoe the wrong way.'' Each is fixed in the next version.\\[2pt]
\noindent\hfill\ituimgs{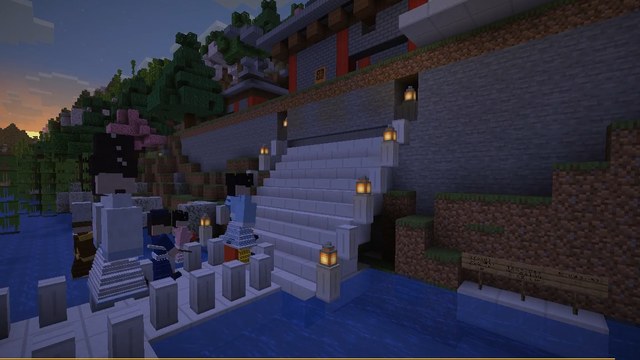}\hfill\ituimgs{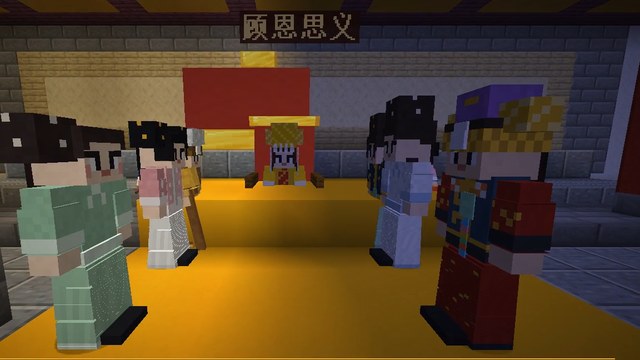}\hfill\ituimgs{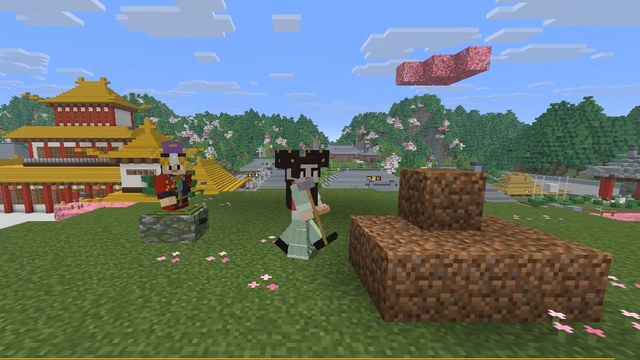}\hfill\null\\[-1pt]
\noindent\hfill\itlabs{User, after m4: temple}\hfill\itlabs{User, after m5: main hall}\hfill\itlabs{User, after m5: flower burial}\hfill\null\\
\midrule
\itround{m6}\itbuilder{} during m3 to m6, characters, story scenes and furnished rooms; the final tour shows each place's name as the player arrives.\\[2pt]
\itthree{\itimgt{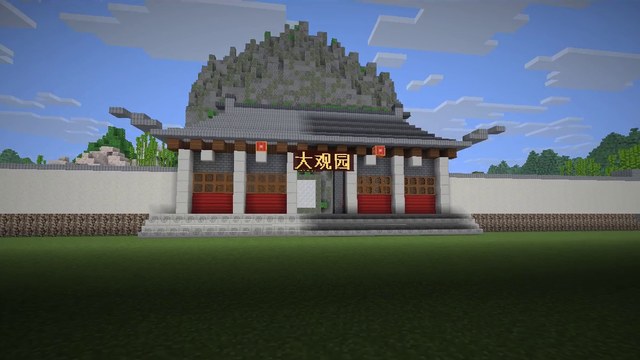}}{\itimgt{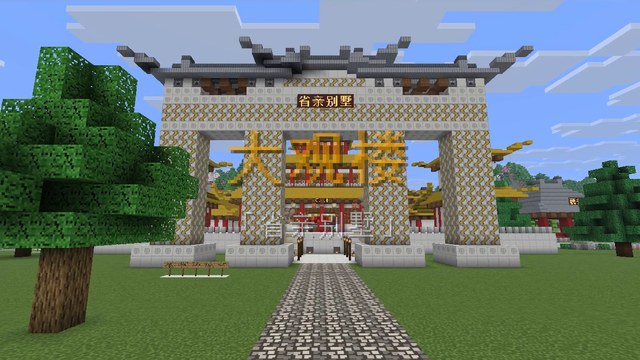}}{\itimgt{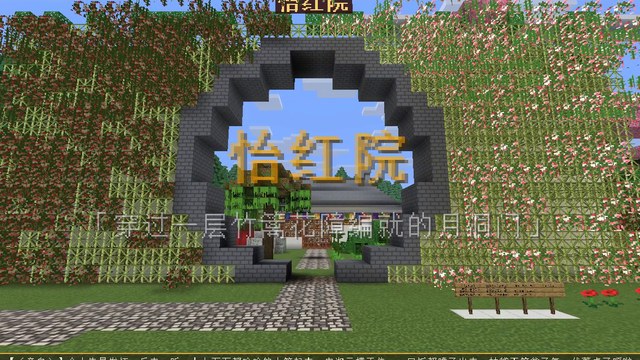}}\\
\bottomrule
\end{tabular}
\end{table*}


\begin{table*}[!htbp]
\centering
\footnotesize
\caption{Iterative development of San Diego locations in Cities: Skylines across three phases and 12 critic playtests. Each row shows the harbor, La Jolla Cove, and UTC from matched camera views, with critic scores shown in the row headings. The top row contains user photos and selected Wikimedia Commons images licensed under CC BY-SA. Amber borders mark user-annotated tour frames. Quotations are translated.}
\label{tab:case_iter_sandiego}
\begin{tabular}{@{}p{\textwidth}@{}}
\toprule
\itsix{\itcol{Harbor}}{\itcol{Harbor, alt.}}{\itcol{La Jolla Cove}}{\itcol{Cove, alt.}}{\itcol{UTC}}{\itcol{UTC, alt.}}\\[1pt]
\itsix{\itimg{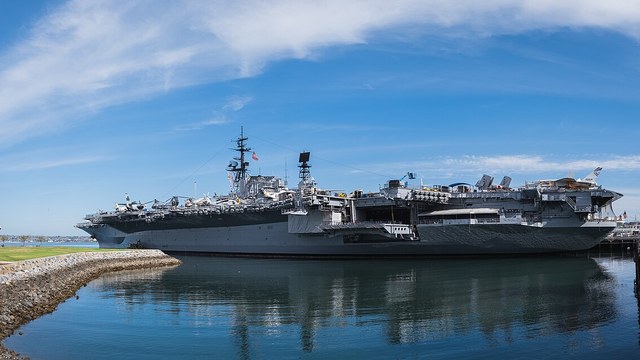}}{\itimg{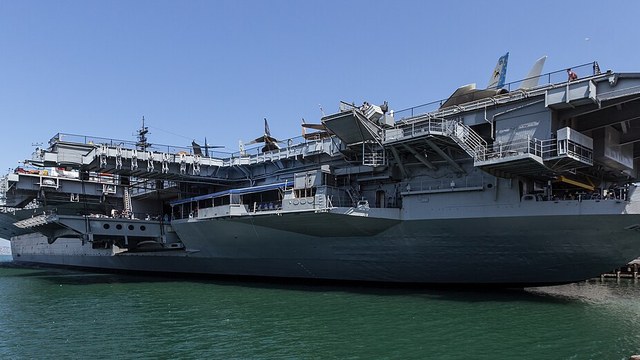}}{\itimg{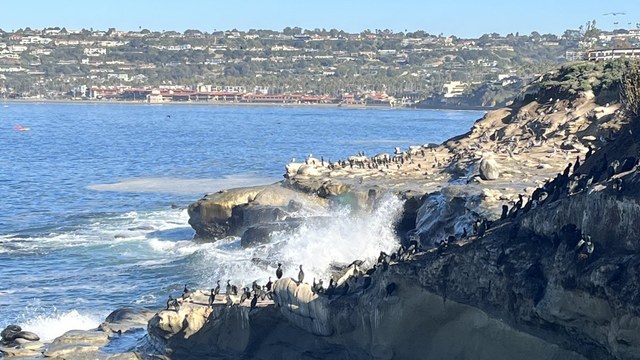}}{\itimg{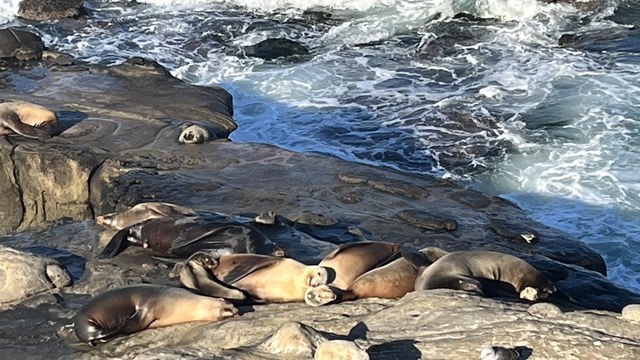}}{\itimg{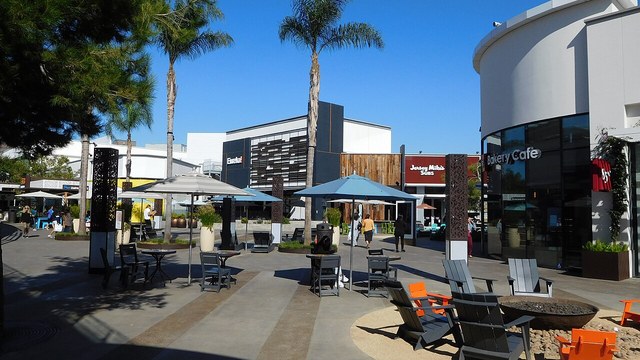}}{\itimg{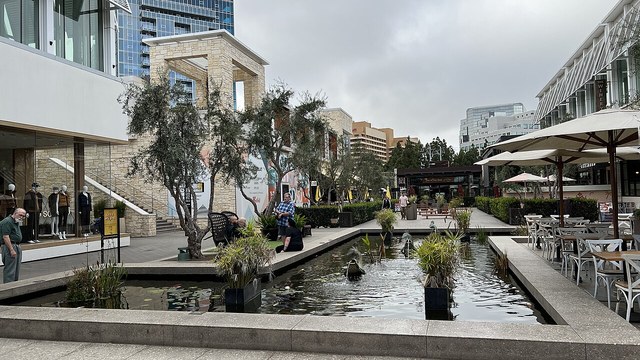}}\\[0pt]
\itsix{\itlab{Commons photo}}{\itlab{Commons photo}}{\itlab{user's photo}}{\itlab{user's photo}}{\itlab{Commons photo}}{\itlab{Commons photo}}\\
\midrule
\itround{P1 r1 \quad\textmd{harbor 3, cove 2, UTC 4}}\itcritic{} harbor: ``a yellow sand desert with a road, not a promenade; no skyline backdrop''; cove: ``a flat green meadow with scattered tiny red houses; the cove has no cliffs, no rocks, no sea lions''; UTC: ``best match in spirit \ldots real shop names in two rows''. \itbuilder{} during rounds r2 and r3, a hard quay and a wide promenade with palms and benches, a skyline cluster behind the carrier; an 18\,m cliff and an enclosed cove filled with water; a continuous shop row with parked cars and umbrellas.\\[2pt]
\itsix{\itimg{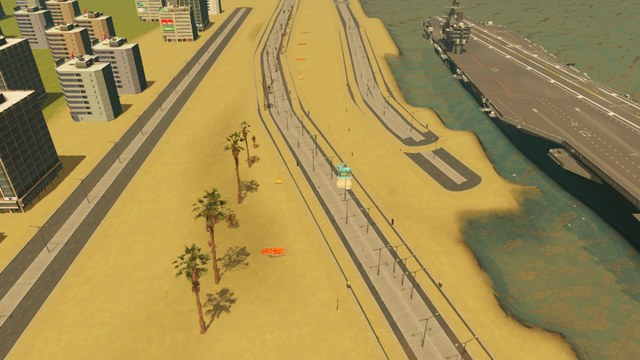}}{\itimg{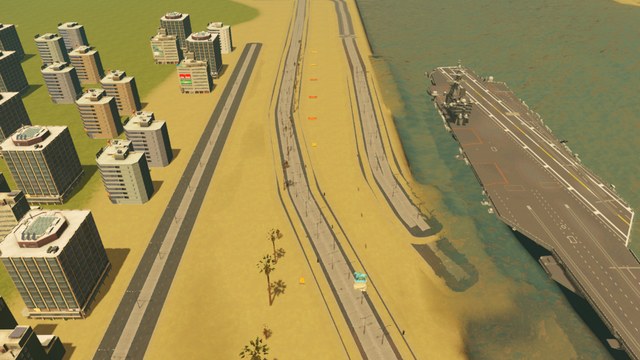}}{\itimg{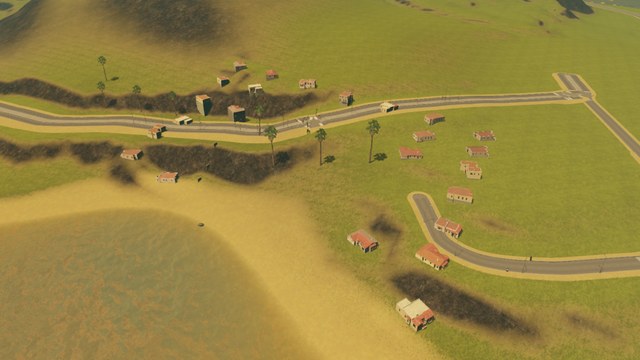}}{\itimg{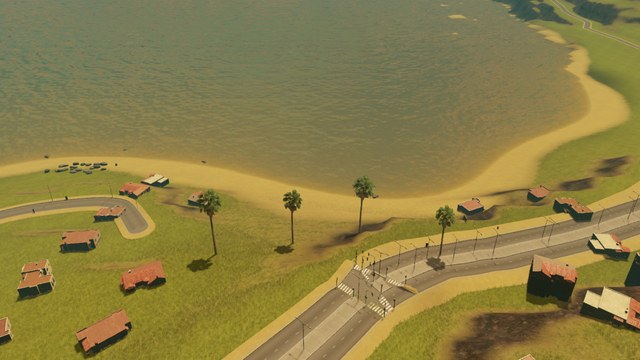}}{\itimg{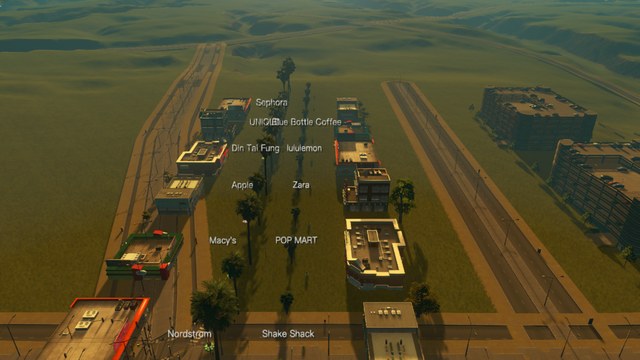}}{\itimg{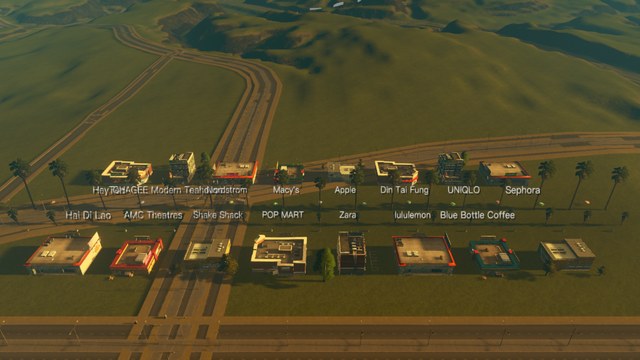}}\\
\midrule
\itround{P1 r3 \quad\textmd{6, 5, 5}}\itcritic{} cove: ``Water now fills the cove \ldots first time it reads as a bay''; UTC: ``Shops now continuous row with names, plaza umbrellas, palm rows; clearly better''; the ground is still yellow sand and the sea brown-green. \itbuilder{} during rounds r4 to r6, sea level fixed at 38.5\,m, terraced coast with a sand crescent, a dense car lot and cream shopfronts at UTC.\\[2pt]
\itsix{\itimg{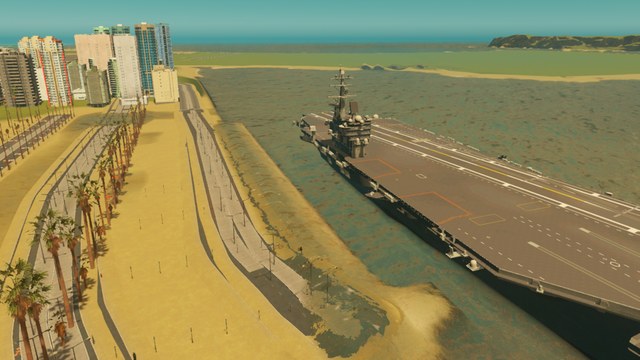}}{\itimg{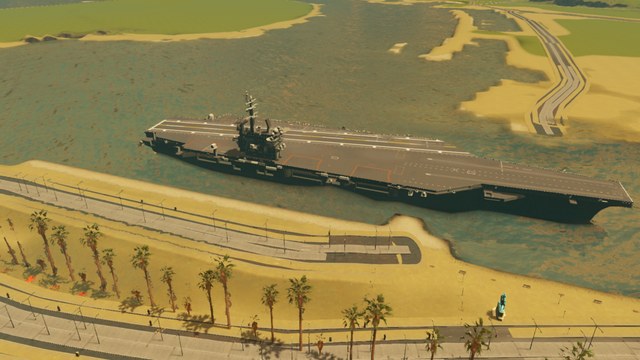}}{\itimg{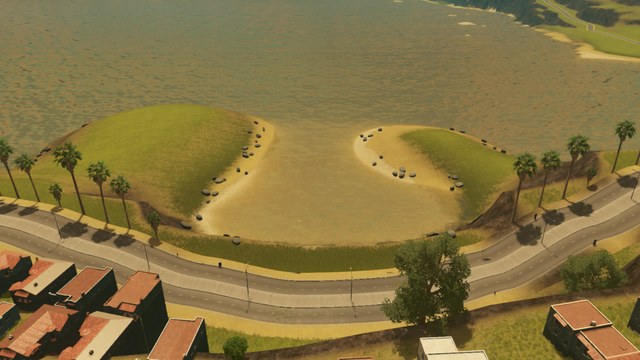}}{\itimg{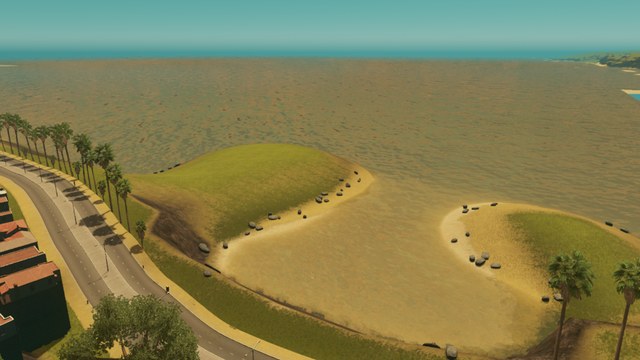}}{\itimg{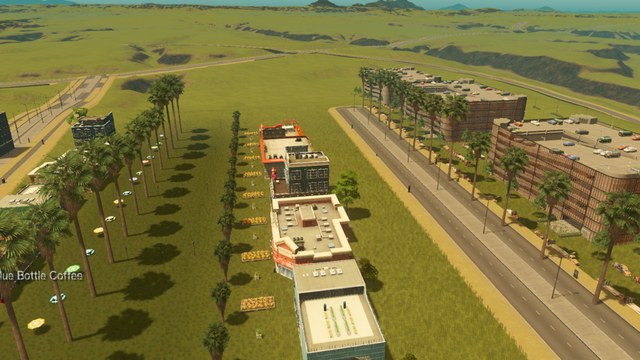}}{\itimg{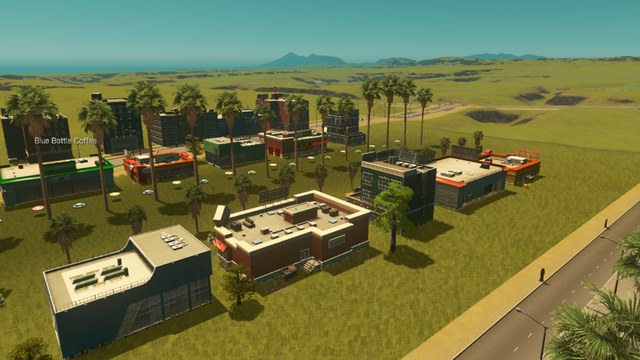}}\\
\midrule
\itround{P1 r6 \quad\textmd{6, 4, 5}}\itcritic{} finds the harbor's quay and ship arrangement recognisable, but the water remains olive-brown; cove: ``a grey pond ringed by rocks in a flat green meadow \ldots Further from the target than r5.'' The critic also reports regressions. \itbuilder{} during phase 2, a colour grade that turns the sea blue, painted pavement, 98 sea lions on the ledge, about 2{,}400 buildings.\\[2pt]
\itsix{\itimg{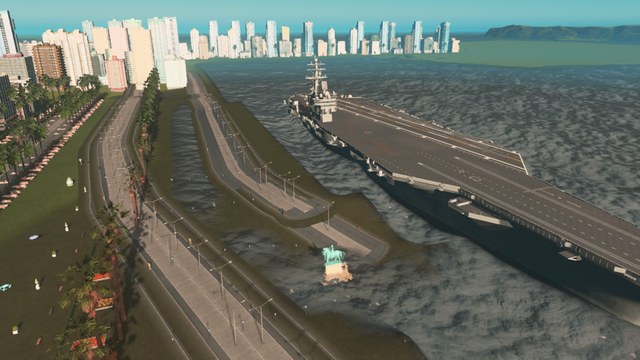}}{\itimg{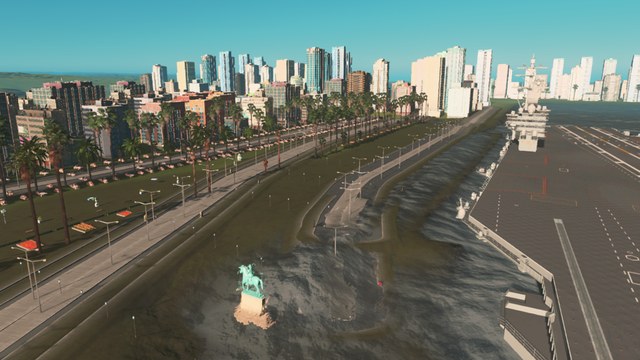}}{\itimg{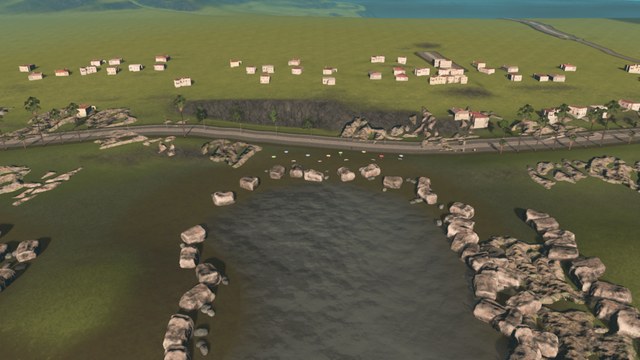}}{\itimg{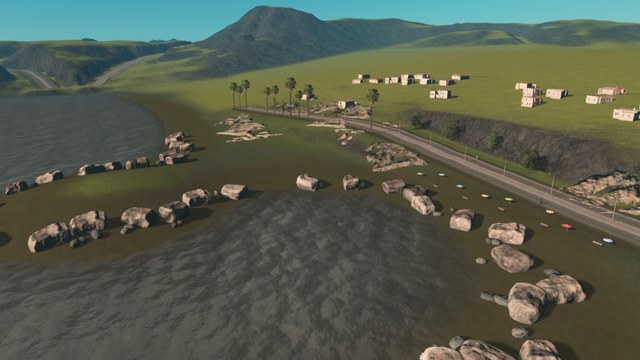}}{\itimg{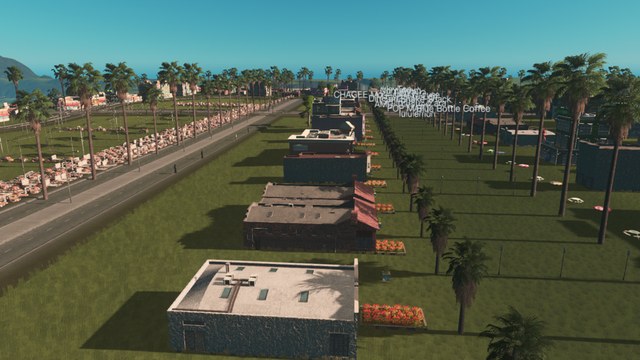}}{\itimg{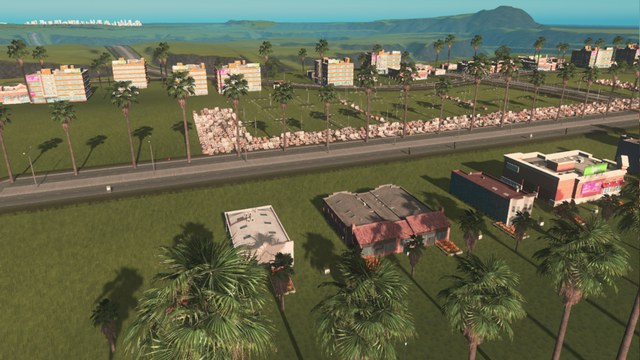}}\\
\bottomrule
\end{tabular}
\end{table*}

\begin{table*}[!htbp]
\centering
\footnotesize
\caption{Continued iterative development of the San Diego coast in Cities: Skylines from Table~\ref{tab:case_iter_sandiego}. This table shows phases 2 and 3 after user feedback. Views remain matched, and amber borders mark user-annotated frames.}
\label{tab:case_iter_sandiego2}
\begin{tabular}{@{}p{\textwidth}@{}}
\toprule
\itsix{\itcol{Harbor}}{\itcol{Harbor, alt.}}{\itcol{La Jolla Cove}}{\itcol{Cove, alt.}}{\itcol{UTC}}{\itcol{UTC, alt.}}\\[1pt]
\midrule
\itround{P2 r1 \quad\textmd{6, 4, 5}}\itcritic{} reports a major improvement from the blue water, carrier, statue, and dense skyline; cove: ``Sea lions are tiny orange specks on grass''; UTC: ``Much better density''. \ituser{} on two tour video frames shown below. ``Why is the seaside so empty? Not enough sea lions''; ``Where are UTC's landmark buildings? Now it is just two low buildings beside a street \ldots the landmark malls and stores are missing.''\\[2pt]
\itsix{\itimg{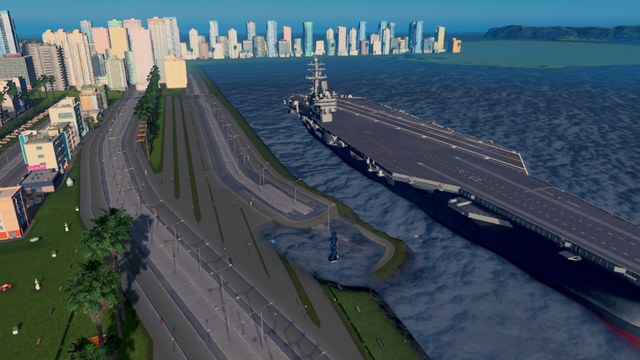}}{\itimg{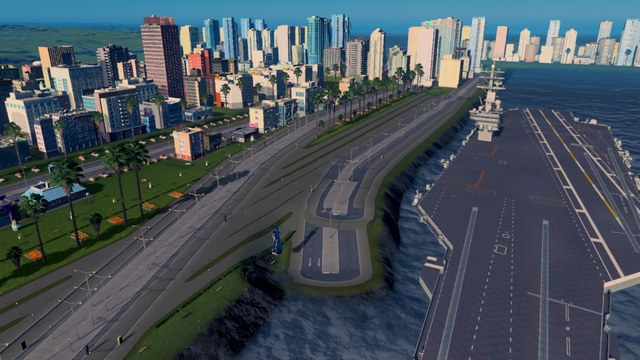}}{\itimg{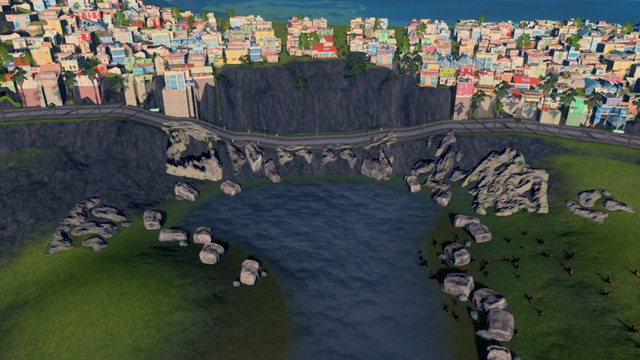}}{\itimg{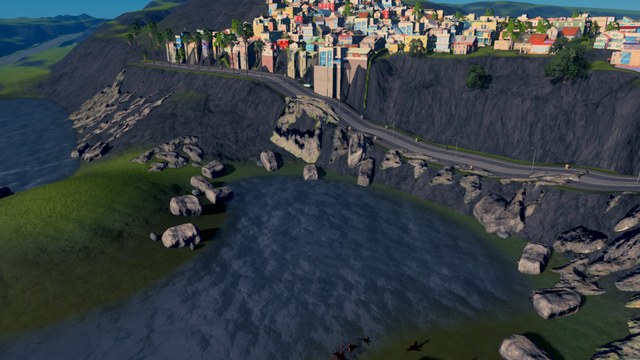}}{\itimg{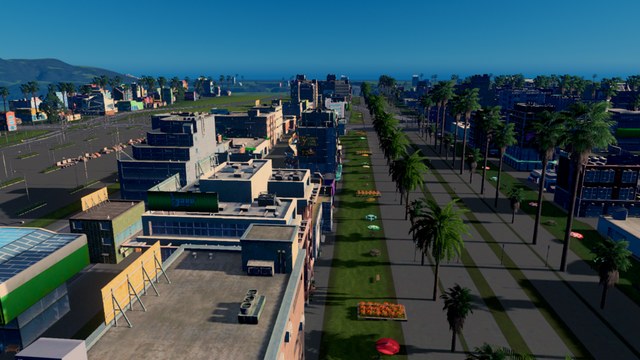}}{\itimg{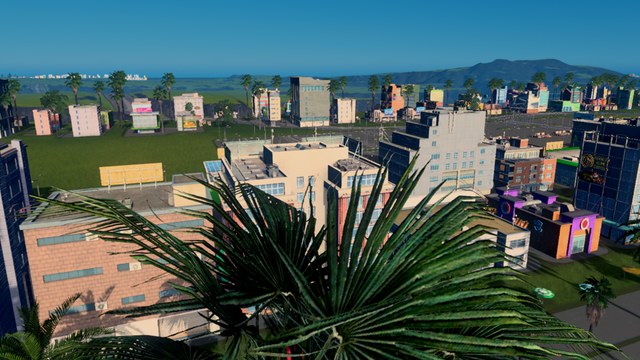}}\\
\itsix{\itnone}{\itnone}{\ituimg{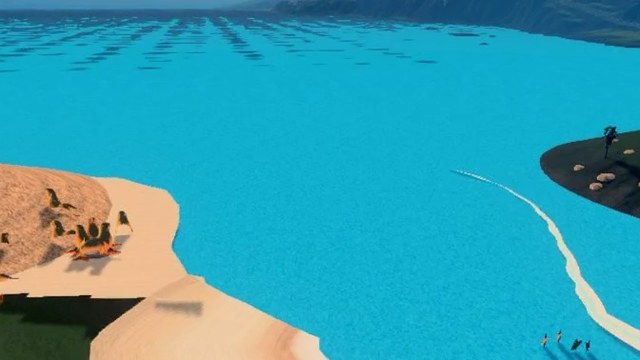}}{\itnone}{\ituimg{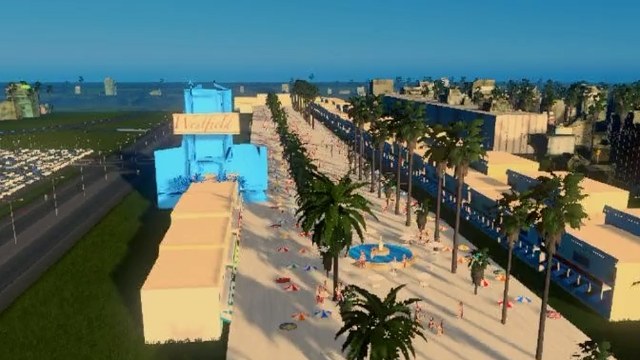}}{\itnone}\\[-1pt]
\itsix{\itnone}{\itnone}{\itlab{User, before P3}}{\itnone}{\itlab{User, before P3}}{\itnone}\\
\midrule
\itround{P3 r2 \quad\textmd{5, 5, 3}}\itbuilder{} during P3 rounds 1 and 2, generated 3D models for the landmarks including the mall tower with its sign and sea lions on sandstone boulders, pale walk and plaza surfaces, people on the beach and walk. \itcritic{} finds that Cove now has many sea lions on boulders and sand, while the improved UTC pylon still has a sign floating above it.\\[2pt]
\itsix{\itimg{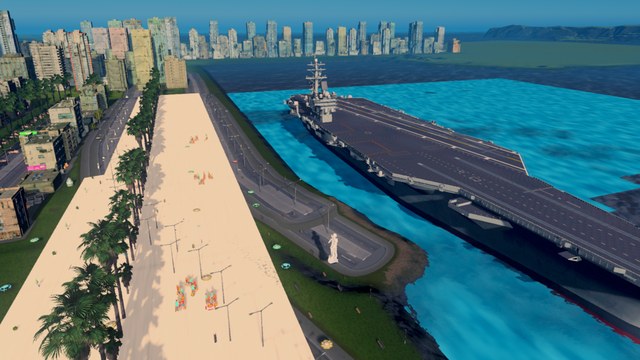}}{\itimg{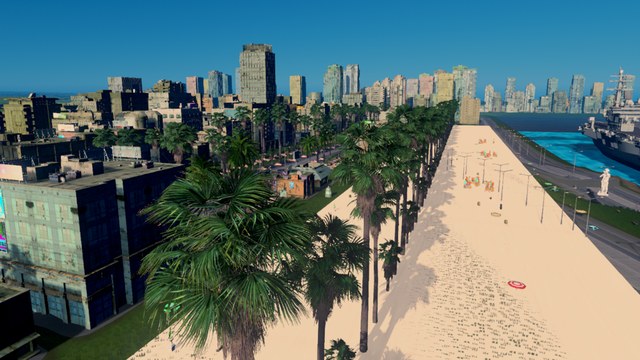}}{\itimg{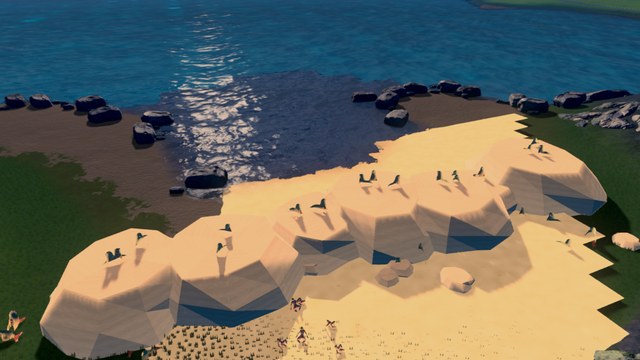}}{\itimg{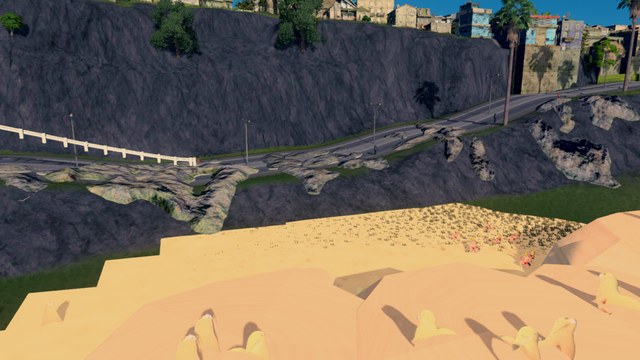}}{\itimg{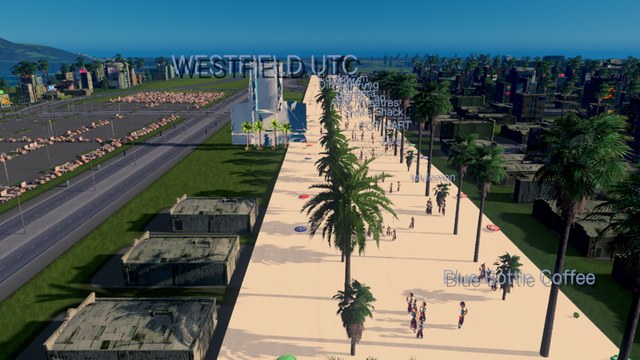}}{\itimg{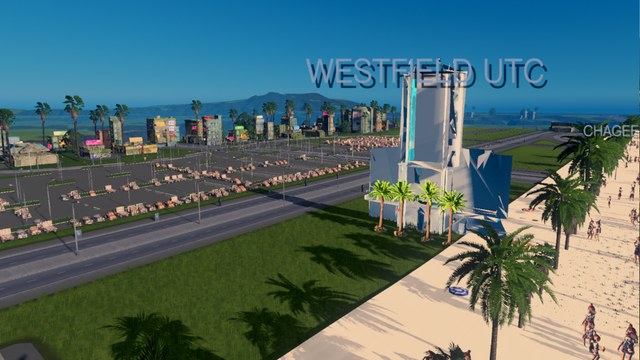}}\\
\midrule
\itround{P3 r4 \quad\textmd{5, 4, 5}}\itbuilder{} during rounds r3 and r4, the tower tiled blue with a red script sign, a round tiled fountain with umbrellas on the walk, a department-store block and a parking structure; palms on the cove headland. \itcritic{} UTC: ``long pale walk, palms, blue tile \ldots tower with red sign, round fountain + umbrellas''; cove: ``no crescent beach \ldots no buff sandstone cliffs''.\\[2pt]
\itsix{\itimg{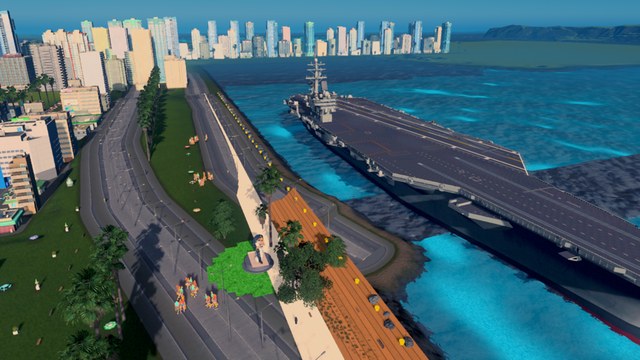}}{\itimg{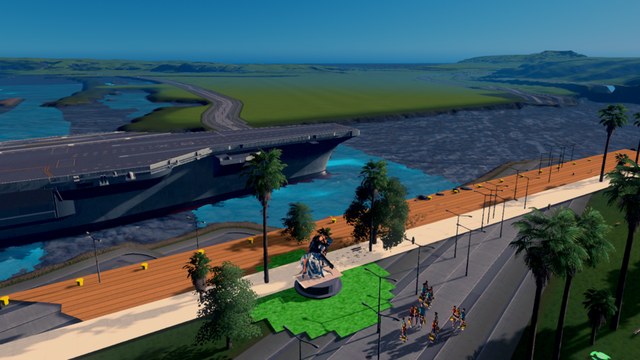}}{\itimg{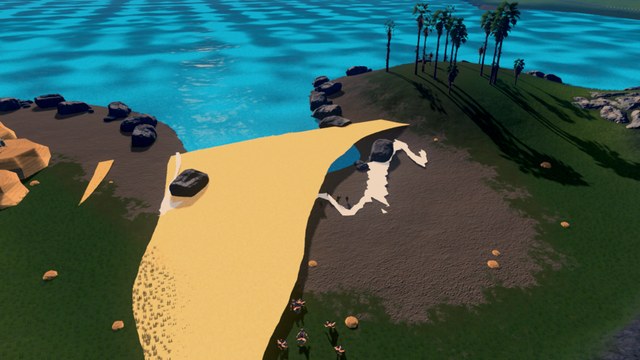}}{\itimg{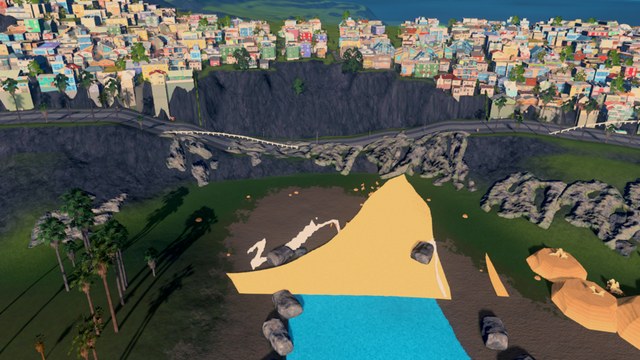}}{\itimg{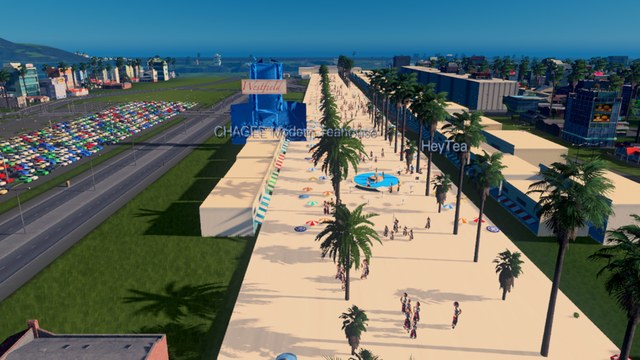}}{\itimg{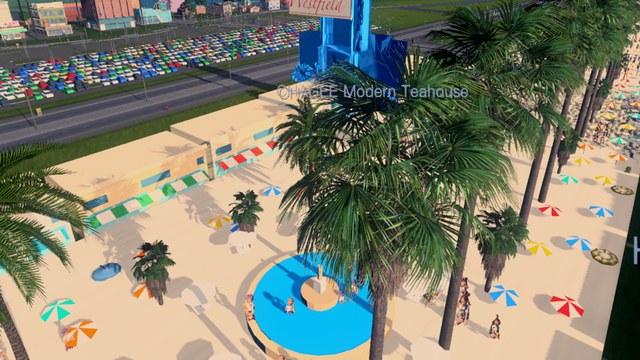}}\\
\bottomrule
\end{tabular}
\end{table*}

\end{document}